\documentclass[11pt]{report} 

\usepackage[papersize={8.5 in, 11 in}, nohead, includeheadfoot, left=1.5 in, right = 1 in, vmargin= 1 in]{geometry}

\usepackage[doublespacing]{setspace} 
\usepackage{calc}

\usepackage{array} 
\usepackage{tabularx}
\usepackage{booktabs}
\usepackage{enumitem}

\usepackage{tocloft}
\renewcommand{\cftchappresnum}{CHAPTER }
\renewcommand{\cftchapaftersnum}{ :}
\newlength{\mylen}   
\renewcommand{\cfttabpresnum}{TABLE }

\newlength{\mylent}   
\usepackage{remreset}

\renewcommand{\cftfigpresnum}{FIGURE }

\newlength{\mylenf}   
\usepackage[compact]{titlesec}

\titleformat{\section}[hang]{\large}{\thesection.}{6 pt}{}
\titleformat{\subsection}[hang]{\normalsize\itshape}{\thesubsection.}{6 pt}{}
\titlespacing*{\section}{0pt}{0pt}{0pt}
\titlespacing*{\subsection}{0pt}{0pt}{0pt}

\newcommand{\bert}{\textsc{BERT}}
\newcommand{\ignore}[1]{}

\newcommand{\datasetname}{\textsc{TacoQA}}

\usepackage{parskip} 
\usepackage{url} 
\usepackage{natbib} 
\bibpunct{(}{)}{;}{a}{,}{,} 

\usepackage{enumitem}
\usepackage{graphicx}
\usepackage{booktabs}
\usepackage{multirow}
\usepackage{sidecap}
\usepackage{amsthm}
\usepackage{amsmath}
\usepackage{amssymb}
\usepackage{amsfonts}
\usepackage{url}
\usepackage{pifont}
\usepackage{eqparbox}
\usepackage{arydshln} 
\usepackage{bigstrut}
\usepackage[linewidth=1pt]{mdframed}
\usepackage{framed}
\usepackage{booktabs}
\usepackage{caption}
\usepackage{colortbl}
\usepackage{multirow}
\usepackage{float}
\restylefloat{table}
\usepackage{algpseudocode}
\usepackage[scaled=0.90]{helvet}

\usepackage{tikz}
\usetikzlibrary{fit,arrows,calc,positioning}
\usepackage{setspace}

\usepackage{epigraph}
\usepackage{framed}
\usepackage[makeroom]{cancel}

\usepackage{etoolbox}

\makeatletter
\pretocmd{\@epitext}{\fontsize{8pt}{6pt} \sffamily \selectfont\em}{}{}
\patchcmd{\epigraph}{\@epitext{#1}}{\itshape\@epitext{#1}}{}{}
\makeatother

\usepackage[ruled,vlined]{algorithm2e}

\usepackage{array}
\newcolumntype{L}[1]{>{\raggedright\let\newline\\\arraybackslash\hspace{0pt}}m{#1}}
\newcolumntype{C}[1]{>{\centering\let\newline\\\arraybackslash\hspace{0pt}}m{#1}}
\newcolumntype{R}[1]{>{\raggedleft\let\newline\\\arraybackslash\hspace{0pt}}m{#1}}

\newcommand{\specialcell}[2][c]{\begin{tabular}[#1]{@{}c@{}}#2\end{tabular}}

\newenvironment{myquote}{\list{}{\leftmargin=3ex\rightmargin=2ex}\item[]}{\endlist}

\usepackage{amssymb}
\usepackage{amsfonts}
\usepackage{graphicx}
\usepackage{longtable}
\usepackage{dcolumn}
\usepackage{tabularx}
\usepackage{wrapfig}
\usepackage{verbatim}
\usepackage{subfiles} 
\usepackage{color}
\usepackage{xcolor}

\definecolor{darkgreen}{RGB}{0,140,0}
\definecolor{darkred}{RGB}{200,0,0}
\definecolor{lightgreen}{RGB}{238,247,233}
\definecolor{lightred}{RGB}{252,231,234}
\definecolor{lightyellow}{RGB}{250,253,191}

\definecolor{purple}{rgb}{0.5,0,1}

\newcommand{\daniel}[1]{\textcolor{blue}{[DK: #1]}}

\newcommand{\todo}[1]{{\color{purple} [TODO: {#1}]}}

\theoremstyle{definition}

\DeclareMathOperator*{\argmax}{arg\,max}

\begin{document}
\makeatletter
\@removefromreset{table}{chapter}
\makeatother
\renewcommand{\thetable}{\arabic{table}}
\makeatletter
\@removefromreset{figure}{chapter}
\makeatother
\renewcommand{\thefigure}{\arabic{figure}}

\newcommand{\tableilp}{\textsc{TableILP}}
\newcommand{\paragram}{\textsc{Paragram}}
\newcommand{\mln}{\textsc{MLNSolver}}
\newcommand{\tupleinf}{\textsc{TupleInf}}
\newcommand{\textilp}{\textsc{SemanticILP}}
\newcommand{\tacoqa}{\textsc{TacoQA}}
\newcommand{\multirc}{\textsc{MultiRC}}
\newcommand{\bidaf}{\textsc{BiDAF}}
\newcommand{\bidafTrained}{\textsc{BiDAF'}}
\newcommand{\proread}{\textsc{Proread}}
\newcommand{\syntprox}{\textsc{SyntProx}}
\newcommand{\textprox}{\textsc{TextProx}}
\newcommand{\cogcompnlp}{\textsc{CogCompNLP}}
\newcommand{\allennlp}{\textsc{AllenNLP}}
\newcommand{\edison}{\textsc{Edison}}
\newcommand\lucene{IR}
\newcommand\arizona{SVM solver}
\newcommand\salience{PMI}
\newcommand\inference{RULE solver}
\newcommand\oldilp{ILP solver}
\newcommand\praline{MLN}
\newcommand\combiner{Combiner}

\newcommand{\processBank}{\textsc{ProcessBank}}
\newcommand{\regentsFourth}{\textsc{Regents 4th}}
\newcommand{\regentsEighth}{\textsc{Regents 8th}}
\newcommand{\publicFourth}{\textsc{AI2Public 4th}}
\newcommand{\publicEighth}{\textsc{AI2Public 8th}}

\newcommand{\tokenLabelView}{\textsf{\small TokenLabelView}}
\newcommand{\spanLabelView}{\textsf{\small SpanLabelView}}
\newcommand{\sequence}{\textsf{\small Sequence}}
\newcommand{\spann}{\textsf{\small Span}}
\newcommand{\predArgView}{\textsf{\small PredArg}}
\newcommand{\semanticGraph}{\textsf{SemanticGraph}}
\newcommand{\clusterView}{\textsf{\small Cluster}}
\newcommand{\treeView}{\textsf{\small Tree}}
\newcommand{\tokens}{\textsf{\small Tokens}}
\newcommand{\shallowP}{\textsf{\small Shallow-Parse}}
\newcommand{\shallowA}{\textsf{\small Shallow}}
\newcommand{\pos}{\textsf{\small POS}}
\newcommand{\lemmaa}{\textsf{\small Lemma}}
\newcommand{\quantities}{\textsf{\small Quantities}}
\newcommand{\coref}{\textsf{\small Coreference}}
\newcommand{\dep}{\textsf{\small Dependency}}
\newcommand{\ner}{\textsf{\small NER}}
\newcommand{\verbSRL}{\textsf{\small Verb-SRL}}
\newcommand{\prepSRL}{\textsf{\small Prep-SRL}}
\newcommand{\frameSRL}{\textsf{\small Frame-SRL}}
\newcommand{\nomSRL}{\textsf{\small Nom-SRL}}
\newcommand{\commaSRL}{\textsf{\small Comma-SRL}}
\newcommand{\simple}{\textsf{\small Shallow}}
\newcommand{\ttitle}[1]{ \hspace{0.1cm} * \textit{#1}}
\newcommand{\know}[1]{\mathcal{K}(#1)}
\newcommand{\cons}[1]{\mathcal{S}(#1)}
\newcommand{\nodes}[1]{\mathbf{v}(#1)}
\newcommand{\edges}[1]{\mathbf{e}(#1)}
\newcommand{\score}[1]{\text{score}(#1)}
\newcommand{\eye}[1]{{\bf1}\!\!\braces{#1}}

\newcommand{\fancyname}{{\textsc{MultiRC}}}

\def\sym#1{\ifmmode^{#1}\else\(^{#1}\)\fi} 

\def\mytitle{
REASONING-DRIVEN QUESTION-ANSWERING \\ FOR NATURAL LANGUAGE UNDERSTANDING
} 
\def\myauthor{Daniel Khashabi}
\def\myauthorfull{Daniel Khashabi}
\def\mysupervisorname{Dan Roth}
\def\mysupervisortitle{Professor of Computer and Information Science}
\newlength{\superlen}   
\settowidth{\superlen}{\mysupervisorname, \mysupervisortitle} 
\def\gradchairname{Rajeev Alur}
\def\gradchairtitle{Professor of Computer and Information Science}
\newlength{\chairlen}   
\settowidth{\chairlen}{\gradchairname, \gradchairtitle} 
\newlength{\maxlen}
\setlength{\maxlen}{\maxof{\superlen}{\chairlen}}
\def\mydepartment{Computer and Information Sciences}
\def\myyear{2019}
\def\signatures{46 pt} 

\newcommand{\isBinary}{\in \{0,1\}}
\newcommand{\tableLen}{|T|}
\newcommand{\tableVar}{T_{i}}
\newcommand{\tableVarPrime}{T_{i'}}
\newcommand{\tableCell}{t_{ijk}}
\newcommand{\tableCellSameRow}{t_{ijk'}} 
\newcommand{\tableCellPrime}{t_{ij'k'}} 
\newcommand{\tableCellPrimePrime}{t_{i'j'k'}} 
\newcommand{\header}{h_{ik}}
\newcommand{\option}{a_{m}}
\newcommand{\optionPrime}{a_{m'}}
\newcommand{\question}{Q}
\newcommand{\qLen}{|\question|}
\newcommand{\qCons}{q_\ell}
\newcommand{\qConsPrime}{q_{\ell'}}
\newcommand{\rowVar}{r_{ij}}
\newcommand{\rowVarPrime}{r_{ij'}}
\newcommand{\columnVar}{\ell_{ik}}
\newcommand{\columnVarPrime}{\ell_{ik'}}

\newcommand{\xOne}[1]{x\left(#1\right)}
\newcommand{\xTwo}[2]{y\left(#1, #2\right)}
\newcommand{\xFour}[4]{r\left(#1, #2, #3, #4\right)}

\newcommand{\setOf}[1]{\left\lbrace #1 \right\rbrace}

\newcommand{\paran}[1]{\left( #1 \right)}
\newcommand{\brac}[1]{\left[ #1 \right]}
\newcommand{\braces}[1]{\left\lbrace #1 \right\rbrace}


\newcommand{\onlyOne}{\exists !}
\newcommand*{\medcap}{\mathbin{\scalebox{0.75}{\ensuremath{\bigcap}}}}%
\newcommand*{\medcup}{\mathbin{\scalebox{1.2}{\ensuremath{\bigcup}}}}%

\newcommand{\essty}{{essentiality}}
\newcommand{\ess}{{essential}}
\newcommand{\Ess}{{Essential}}
\newcommand{\et}{\textsc{ET}} 

\newcommand{\conj}[1]{\textsc{Conj} \big\lbrace #1 \big\rbrace }
\newcommand{\base}[1]{\textsc{Base} \big\lbrace #1 \big\rbrace }
\newcommand{\sur}{\textsc{Surf}}
\newcommand{\tok}{\textsc{Tok}}
\newcommand{\lem}{\textsc{Lem}}
\newcommand{\chk}{\textsc{Chk}}
\newcommand{\ed}[1]{{\tt ed.#1}}
\newcommand{\neighbor}[1]{\textsc{Neighbor}\big(#1\big)}
\newcommand{\pair}[1]{\textsc{Pair}\big(#1\big)}
\newcommand{\scienceTerm}[1]{\textsc{isScienceTerm}\big\lbrace #1 \big\rbrace}
\newcommand{\maxSalience}{\textsc{MaxPMI}~}
\newcommand{\sumSalience}{\textsc{SumPMI}~}
\newcommand{\wordBaseline}{\textsc{PropSurf}~}
\newcommand{\lemmaBaseline}{\textsc{PropLem}~}

\newcommand{\cascades}{\textsc{Cascades}}
\newcommand{\queryFiltering}{IR + ET}
\newcommand{\baseline}{\textsc{Baseline}}

\newcommand{\aiTwoPublic}{\textsc{AI2Public}}
\newcommand{\regentsPerturbed}{\textsc{RegtsPertd}}
\newcommand{\regents}{\textsc{Regents}}

\newcommand{\norm}[1]{\| #1 \|}
\newcommand{\namecite}[1]{\citeauthor{#1}~\shortcite{#1}}
\newcommand{\dist}{\mathrm{dist}}
\newcommand{\npathk}[1]{\overset{#1}{\cancel{\leftrightsquigarrow}}}
\newcommand{\pathk}[1]{\overset{#1}{\leftrightsquigarrow}}
\newcommand{\prob}[1]{\mathbb{P}\left[#1 \right]}
\newcommand{\probTwo}[2]{\mathbb{P}^{(#1)}\left[#2 \right]}
\newcommand{\eone}{\pathk{d}}
\newcommand{\etwo}{\npathk{}}
\newcommand{\ball}{\text{$\mathcal{B}$}}
\newcommand{\floor}[1]{\lfloor #1 \rfloor}
\newcommand{\vol}[1]{\text{Vol}(#1)}
\newcommand{\degA}{\text{deg}}
\newcommand{\ins}{\mathcal{P}}
\newcommand{\renymodel}{\mathsf{G(n, p)}}
\newcommand{\gm}{\text{{$G_M$}}}
\newcommand{\gs}{\text{{$G_S$}}}

\newcommand{\vm}{\text{$V_M$}}
\newcommand{\vs}{\text{$V_S$}}

\newcommand{\emm}{\text{$E_M$}}
\newcommand{\es}{\text{$E_S$}}

\newcommand{\Gm}{\gm(\vm, \emm)\xspace}
\newcommand{\Gs}{\gs(\vs, \es)\xspace}

\newcommand{\oracle}{\text{$\mathcal{O}$}}
\newcommand{\noisy}{\text{$\mathcal{F}$}}
\newcommand{\radius}{\text{$\mathcal{R}$}}
\newcommand{\perf}{\text{$\mathcal{P}$}}
\newcommand{\bernoulli}{\text{Bern}}
\newcommand{\dimm}{|\mathcal{U}|}

\newcommand{\ex}[1]{{\sf #1}}
\newcommand{\tr}{\mathrm{tr}}
\newcommand{\alg}[1]{\mathsf{ALG}(#1)}
\newcommand{\tuple}{\mathcal{V}_S}
\newcommand{\tuplem}{\mathcal{V}_M}

\newcommand{\s}{Section~}
\newcommand{\f}{Figure~}
\newcommand{\ta}{Table~}

\newcommand{\pois}[1]{\text{Pois}(#1)}
\newcommand{\normal}[1]{\mathcal{N}(#1)}
\newcommand{\bin}[1]{\text{Bin}(#1)}
\newcommand{\bern}[1]{\text{Bern}(#1)}
\newcommand*\circled[1]{\tikz[baseline=(char.base)]{
             \node[shape=circle,draw,inner sep=0.2pt] (char) {\tiny #1};}}

\newtheorem{theorem}{Theorem}
\newtheorem{remark}{Remark}
\newtheorem{conjecture}{Conjecture}
\newtheorem{proposition}{Proposition}
\newtheorem{definition}{Definition}
\newtheorem{lemma}{Lemma}
\newtheorem{property}{Property}
\newtheorem{specialc}{Special Case}
\newtheorem{corollary}{Corollary}

\pagenumbering{roman}
\pagestyle{plain}


\begin{titlepage}
\thispagestyle{empty} 
\begin{center}

\onehalfspacing

\mytitle

\myauthor

A DISSERTATION

in 

\mydepartment 

Presented to the Faculties of the University of Pennsylvania

in 

Partial Fulfillment of the Requirements for the

Degree of Doctor of Philosophy 
\vspace{-0.2cm}
\myyear \\ 
\end{center}
\vfill 

\begin{flushleft}

Supervisor of Dissertation\\[\signatures] 
\vspace{-0.2cm}
\renewcommand{\tabcolsep}{0 pt}
\begin{table}[h]
\begin{tabularx}{\maxlen}{l}
\toprule
\mysupervisorname, \mysupervisortitle\\ 
\end{tabularx}
\end{table}

Graduate Group Chairperson\\[\signatures] 
\vspace{-0.2cm}
\begin{table}[h]
\begin{tabularx}{\maxlen}{l}
\toprule
\gradchairname, \gradchairtitle\\ 
\end{tabularx}
\end{table}
\singlespacing
\vspace{-0.2cm}

Dissertation Committee 

Dan Roth, Professor, Computer and Information Science, University of Pennsylvania  

Mitch Marcus, Professor of Computer and Information Science, University of Pennsylvania    

Zachary Ives, Professor of Computer and Information Sciences, University of Pennsylvania  

Chris Callison-Burch, Associate Professor of Computer Science, University of Pennsylvania  

Ashish Sabharwal, Senior Research Scientist, Allen Institute for Artificial Intelligence

\end{flushleft}

\end{titlepage}


\doublespacing

\thispagestyle{empty} 

\vspace*{\fill}

\begin{flushleft}
\mytitle

 \copyright \space COPYRIGHT
 
\myyear

\myauthorfull\\[24 pt] 

This work is licensed under the \\
Creative Commons Attribution \\
NonCommercial-ShareAlike 3.0 \\
License

To view a copy of this license, visit

\url{http://creativecommons.org/licenses/by-nc-sa/3.0/}

\end{flushleft}
\pagebreak 

\newenvironment{preliminary}{}{}
\titleformat{\chapter}[hang]{\large\center}{\thechapter}{0 pt}{}
\titlespacing*{\chapter}{0pt}{-33 pt}{6 pt} 
\begin{preliminary}

\setcounter{page}{3}  
\begin{center}
\textit{
Dedicated to the loving memory of my gramma, An'nah \\ 
Your patience and kindness will forever stay with me. 
}
\end{center}


\clearpage
\chapter*{ACKNOWLEDGEMENT}
\addcontentsline{toc}{chapter}{ACKNOWLEDGEMENT} 

I feel incredibly lucky to have Dan Roth as my advisor. 
I am grateful to Dan for trusting me, especially when 
I had only a basic understanding of many key challenges in natural language. It took me a while to catch up with \emph{what is important} in the field and be able to communicate the challenges effectively. 
During these years, 
Dan's vision has always been the guiding principle to many of my works.   
His insistence on focusing on the long-term progress, rather than ``easy'' 
wins, shaped the foundation of many of the ideas I pursued. 
This perspective 
pushed me to think differently than the popular trends. 
It has been a genuine privilege to work together. 

I want to thank my thesis committee at UPenn, Mitch Marcus, Zach Ives and Chris Callison-Burch  for being a constant source of invaluable feedback and guidance. 
Additionally, I would like to thanks the many professors who have touched parts of my thinking: Jerry DeJong, for encouraging me read the classic literature; 
Chandra Chekuri and Avrim Blum, for their emphasis on intuition, rather than details; 
and my undergraduate advisor Hamid Sheikhzadeh Nadjar, for encouraging me to work on important problems. 

A huge thank you to the Allen Institute for Artificial Intelligence (AI2) for much support during my PhD studies. 
Any time I needed any resources (computing resources, crowdsourcing credits, engineering help, etc), without any hesitation, AI2 has provided me what was needed. 
Special thanks to Ashish Sabhwaral and Tushar Khot for being a constant source of wisdom and guidance, and investing lots of time and effort. They both have always been present to listen to my random thoughts, almost on a weekly basis. 
I am grateful to other members of AI2 for their help throughout my projects: Oren Etzioni, Peter Clark, Oyvind Tafjord, Peter Turney, Ingmar Ellenberger, Dirk Groeneveld,
Michael Schmitz, Chandra Bhagavatula and Scott Yih. 
Moreover, I would like to remember Paul Allen (1953-2018): his vision and constant generous support has tremendously changed our field (and my life, in particular).  


My collaborators, especially past and present CogComp members, have been major contributors and influencers throughout my works. I would like to thank Mark Sammons, Vivek Srikumar, Christos Christodoulopoulos, Erfan Sadeqi Azer, Snigdha Chaturvedi, Kent Quanrud, Amirhossein Taghvaei, Chen-Tse Tsai, and many other CogComp members. 
Furthermore, I thank Eric Horn and Jennifer Sheffield for their tremendous contributions to many of my write-ups. And thank you to all the friends I have made at Penn, UIUC, and elsewhere, for all the happiness you've brought me. Thanks to Whitney, for sharing many happy and sad moments with me, and for helping me become a better version of myself. 

Last, but never least, my family, for their unconditional sacrifice and support. I wouldn't have been 
able to go this far without you.

\clearpage
\chapter*{ABSTRACT}
\addcontentsline{toc}{chapter}{ABSTRACT} 
\vspace{-.2cm}
\begin{center}
\mytitle

\myauthor

\vspace{-.2cm}
\mysupervisorname
\vspace{-.2cm}
\end{center}
\emph{Natural language understanding} (NLU)
of 
text is a fundamental challenge in AI, and it has received significant attention throughout the history
of NLP research. This primary goal
has been studied under different tasks, such as 
Question Answering (QA) and Textual Entailment (TE). 
In this thesis, we investigate the NLU problem through the QA task and focus on the aspects that make it a challenge for the current state-of-the-art technology. 
This thesis is organized into \underline{three} main parts: 

In the first part, we explore multiple formalisms to improve existing machine comprehension systems. We propose a formulation for abductive reasoning in natural language  and show its effectiveness, especially in domains 
with 
limited training data. Additionally, 
to help reasoning systems cope with irrelevant or redundant information,
we create a supervised approach to learn and detect the essential terms in questions. 

In the second part, 
we propose two new challenge datasets. 
In particular, we create two datasets of natural language questions where (i) the first one requires reasoning over multiple sentences; (ii) the second one requires \emph{temporal common sense} reasoning.
We hope that the two proposed datasets will motivate the field to address  more complex problems. 

In the final part, 
we present the first formal framework for multi-step reasoning algorithms, 
in the presence of 
a few important properties 
of 
language use, such as incompleteness, ambiguity, etc. 
We apply this framework to prove fundamental limitations for reasoning algorithms. These theoretical results provide extra intuition into the
existing empirical evidence in the field. 


\clearpage
\tableofcontents


\clearpage
\listoftables
\addcontentsline{toc}{chapter}{LIST OF TABLES}


\clearpage
\listoffigures
\addcontentsline{toc}{chapter}{LIST OF ILLUSTRATIONS}

\clearpage
\chapter*{PUBLICATION NOTES}
\addcontentsline{toc}{chapter}{PUBLICATION NOTES}
{
\begin{enumerate}
    \item Daniel Khashabi, Tushar Khot, Ashish Sabharwal, Peter Clark, Oren Etzioni, and Dan Roth. Question answering via integer programming over semi-structured knowledge. In \emph{Proceedings of the 25th International Joint Conference on Artificial Intelligence (IJCAI), 2016}. URL \url{http://cogcomp.org/page/publication_view/786}.
    \item Daniel Khashabi, Tushar Khot, Ashish Sabharwal, and Dan Roth. Learning what is essential in questions. In \emph{Proceedings of the Conference on Computational Natural Language Learning (CoNLL), 2017}. URL \url{http://cogcomp.org/page/publication_view/813}.
    \item  Daniel Khashabi, Snigdha Chaturvedi, Michael Roth, Shyam Upadhyay, and Dan Roth. Looking beyond the surface: A challenge set for reading comprehension over multiple sentences. In \emph{Proceedings of the Annual Conference of the North American Chapter of the Association for Computational Linguistics (NAACL), 2018a}. URL \url{http://cogcomp.org/page/publication_view/833}.
    \item  Daniel Khashabi, Tushar Khot, Ashish Sabharwal, and Dan Roth. Question answering as global reasoning over semantic abstractions. In \emph{Proceedings of the Fifteenth Conference on Artificial Intelligence (AAAI), 2018b.} URL \url{http://cogcomp.org/page/publication_view/824}.
    \item Daniel Khashabi, Erfan Sadeqi Azer, Tushar Khot, Ashish Sabharwal, and Dan Roth. On the capabilities and limitations of reasoning for natural language understanding, 2019. URL \url{https://arxiv.org/abs/1901.02522}  
    \item Ben Zhou, Daniel Khashabi, Qiang Ning, and Dan Roth. ``Going on a vacation'' takes longer than ``Going for a walk'': A Study of Temporal Commonsense Understanding. In \emph{Proceedings of the Conference on Empirical Methods in Natural Language Processing (EMNLP), 2019.}
\end{enumerate}
}


\end{preliminary}

\newenvironment{mainf}{}{}
\titleformat{\chapter}[hang]{\large\center}{CHAPTER \thechapter}{0 pt}{ : }
\titlespacing*{\chapter}{0pt}{-29 pt}{6 pt} 
\begin{mainf}

\newpage
\pagenumbering{arabic}
\pagestyle{plain} 

\setlength{\parskip}{10 pt} 
\setlength{\parindent}{0pt}

\addtocontents{toc}{\protect\setcounter{tocdepth}{2}} 

\chapter{Introduction}
\epigraph{ 
``To model this language understanding process in a computer, we need a program which combines grammar, semantics, and reasoning in an intimate way, concentrating on their interaction.''
}{--- \textup{T. Winograd}, Understanding Natural Language, 1972}


\section{Motivation}
The purpose of \emph{Natural Language Understanding} (NLU) is to enable systems to interpret a given text, as close as possible to the many ways humans would interpret it.

Improving NLU is increasingly changing the way humans interact with machines. 
The current NLU technology is already making significant impacts. 
For example, we can see it used by 
speech agents, including Alexa, Siri, and Google Assistant. 
In the near future, 
with better NLU systems, we will witness a more active presence of these systems in our daily lives: social media interactions, in financial estimates, during the course of product recommendation, in accelerating of  scientific findings, etc. 

The importance of NLU was understood by many pioneers in Artificial Intelligence (starting in the '60s and '70s). The initial excitement about the field ushered a decade of activity in this area~\citep{McCarthy63,Winograd72,Schank72,Woods73,Zadeh78}. The beginning of these trends was overly positive at times, and it took years (if not decades) to comprehend and appreciate the real difficulty of language understanding. 

\section{Challenges along the way to NLU}
\label{intro:challenges:nlu}
\begin{figure}
    \centering
    \begin{tabular}{L{\textwidth}}
    \toprule
        \justify A 61-year-old furniture salesman was pushed down the shaft of a freight elevator yesterday in his downtown Brooklyn store by two robbers while a third attempted to crush him with the elevator car because they were dissatisfied with the \$1,200 they had forced him to give them.  \\ 
        The buffer springs at the bottom of the shaft prevented the car from crushing the salesman, John J. Hug, after he was pushed from the first floor to the basement. The car stopped about 12 inches above him as he flattened himself at the bottom of the pit.  \\ 
        Mr. Hug was pinned in the shaft for about half an hour until his cries attracted the attention of a porter. The store at 340 Livingston Street is part of the Seaman’s Quality Furniture chain. 
        \\ 
        Mr. Hug was removed by members of the Police Emergency Squad and taken to Long Island College Hospital. He was badly shaken, but after being treated for scrapes of his left arm and for a spinal injury was released and went home. He lives at 62-01 69th Lane, Maspeth, Queens.
        \\ 
        He has worked for seven years at the store, on the corner of Nevins Street, and this was the fourth time he had been held up in the store. The last time was about one year ago, when his right arm was slashed by a knife-wielding robber. \\ 
        \tabularnewline
    \bottomrule
    \end{tabular}
    \caption{A sample story appeared on the New York Times (taken from~\cite{McCarthy76}).  }
    \label{fig:mcarthy:example:paragraph}
\end{figure}

We, humans, are so used to using language that it's almost impossible to see its complexity, without a closer look into instances of this problem. 
As an example, consider the story shown in Figure~\ref{fig:mcarthy:example:paragraph}, which appeared in an issue of the New York Times (taken from~\cite{McCarthy76}). With relatively simple wording, this story is understandable to English speakers. Despite the simplicity, many nuances have to come together to form a coherent understanding of this story. 

We flesh out a few general factors which contribute to the complexity of language understanding in the context of the story given in Figure~\ref{fig:mcarthy:example:paragraph}:
\begin{figure}
    \centering
    \includegraphics[scale=0.4]{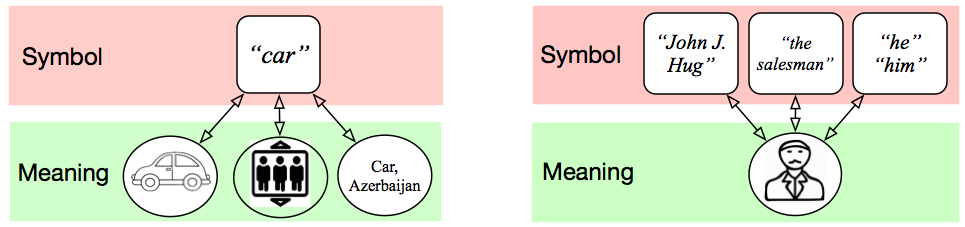}
    \caption{\emph{Ambiguity} (left) appears when mapping a raw string to its actual meaning; \emph{Variability} (right) is having many ways of referring to the same meaning. }
    \label{fig:ambiguity}
\end{figure}
\begin{itemize}
    \item \textit{Ambiguity} comes along when trying to make sense of a given string. While an average human might be good at this, it's incredibly hard for machines to map symbols or characters to their actual meaning. 
    For example, the mention of ``car'' that appears in our story has multiple meanings (see Figure~\ref{fig:ambiguity}; left). In particular, this mention in the story refers to a sense other than its usual meaning (here refers to \emph{the elevator cabin}; the usual meaning is a \emph{road vehicle}).   
    \item \textit{Variability} of language means that a single idea could be phrased in many different ways. For instance, the same character in the story, ``Mr. Hug,'' has been referred to in  different ways: ``the salesman,'' ``he,'' ``him,'' ``himself,'' etc. 
    Beyond lexical level, there is even more variability in bigger constructs of language, such as phrases, sentences, paragraphs, etc. 
    \item Reading and understanding text involves an implicit formation of a mental structure with many elements. 
    Some of these elements are directly described in the given story, but a significant portion of the understanding involves 
    information that is \emph{implied} based on a readers' background knowledge. 
    \textit{Common sense} refers to our (humans) understanding of everyday activities (e.g., sizes of objects, duration of events, etc), usually shared among many individuals. Take the following sentence from the story: 
    \begin{framed}
    \noindent
    The car stopped about 12 inches above him as he flattened himself at the bottom of the pit.
    \end{framed}
    There is a significant amount of imagination hiding in this sentence; each person after reading this sentence has a mental picture of the incident. And based on this mental picture, we have implied meanings: \emph{we know he is lucky to be alive now; if he didn't flatten himself, he would have died; he had nowhere to go at the bottom of the pit; the car is significantly heavier than the man}; etc. 
    Such understanding is common and easy for humans and rarely gets direct mention in text, since they are considered trivial (for humans). Humans are able to form such implicit understanding as a result of our own world model and past shared experiences.   
    \item \textit{Many small bits combine to make a big picture.}  
    We understand that ``downtown Brooklyn'' is probably not a safe neighborhood, since ``this was the fourth time he had been held up here.'' We also understand that despite all that happened to ``Mr. Hug,'' he likely goes back to work after treatment because similar incidents have happened in the past. Machines don't really make these connections (for now!). 
\end{itemize}

Challenges in NLU don't end here; there are many other aspects to language understanding that we skip here since they go beyond the scope of this thesis. 

\section{Measuring the progress towards NLU via Question Answering}
To measure machines' ability to understand a given text, one can create numerous questions about the story. A system that better understands language should have a higher chance of answering these questions.
This approach has been a popular way of measuring NLU since its early days~\citep{McCarthy76,Winograd72,Lehnert77}. 

\begin{table}
    \centering
    \small
    \begin{tabular}{L{0.8 \textwidth}}
        \toprule
        \textbf{Question 1:} Where did the robbers push Mr. Hug?\\ 
        \textbf{Answer 1:} down the shaft of a freight elevator \\ 
        \midrule
        \textbf{Question 2:} How old is Mr. Hug? \\ 
        \textbf{Answer 2:} 61 years old \\ 
        \midrule
        \textbf{Question 3:} On what street is Mr. Hug's store located? \\ 
        \textbf{Answer 3:} 340 Livingston Street, on the corder of Nevins Street \\ 
        \midrule
        \textbf{Question 4:} How far is his house to work? \\ 
        \textbf{Answer 4:} About 30 minutes train ride \\ 
        \midrule
        \textbf{Question 5:} How long did the whole robbery take? \\ 
        \textbf{Answer 5:} Probably a few minutes \\ 
        \midrule
        \textbf{Question 6:} Was he trapped in the elevator car, or under? \\
        \textbf{Answer 6:} under  \\ 
        \midrule
        \textbf{Question 7:} Was Mr. Hug conscious after the robbers left?  \\ 
        \textbf{Answer 7:} Yes, he cried out and his cries were heard.  \\
        \midrule
        \textbf{Question 8:} How many floors does Mr. Hug's store have? \\
        \textbf{Answer 8:} More than one, since he has an elevator  
        \tabularnewline
        \bottomrule
    \end{tabular}
    \caption{Natural language questions about the story in Figure~\ref{fig:mcarthy:example:paragraph}. }
    \label{tab:mr:hug:questions}
\end{table}

Table~\ref{tab:mr:hug:questions} shows examples of such questions. 
Consider \textbf{Question 1}. The answer to this question is directly mentioned in text and the only thing that needs to be done is creating a representation to handle the variability of text. 
For instance, a reoresentation of the meaning that are conveyed by \emph{verb} predicates, since a major portion of meanings are centered around verbs. 
For example, to understand the various elements around a verb ``push,'' one has to figure out \emph{who pushed, who was pushed, pushed where}, etc. The subtask of \textbf{semantic role labeling}~\citep{PunyakanokRoYi04} is dedicated to resolving such inferences (Figure~\ref{fig:annotation:mrhug}; top). The output of this annotation of indicates that \emph{the location pushed to} is ``the shaft of a freight elevator.'' 
In addition, the output of  
the \textbf{coreference} task~\citep{CarbonellBr88,McCarthy95} informs computers about such equivalences between the mentions of the main character of the story (namely, the equivalence between ``Mr. Hug'' and ``A 61-year-old furniture salesman'').  




\begin{figure}
    \centering
    \frame{
        \resizebox{\textwidth}{!}{
            \includegraphics[scale=0.30,trim=0.19cm 3.5cm 17cm 0.2cm, clip=true]{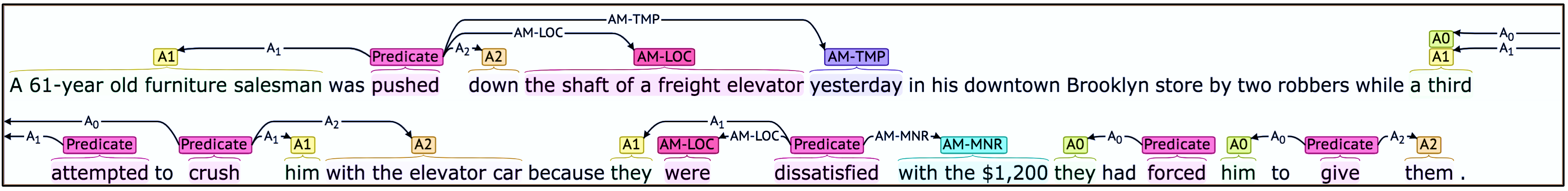}
        }
    }
    \frame{
        \resizebox{\textwidth}{!}{
            \includegraphics[scale=0.24,trim=0cm 0cm 0cm -0.05cm]{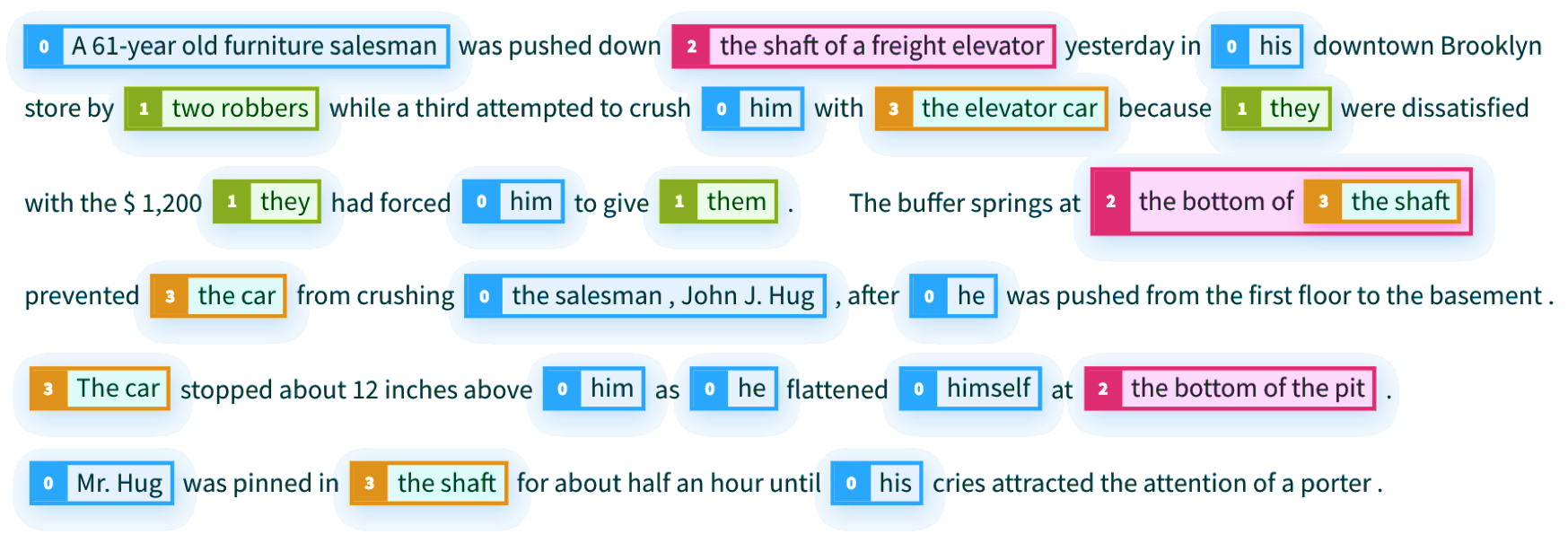}
        }
    }
    \caption{Visualization of two semantic tasks for the given story in Figure~\ref{fig:mcarthy:example:paragraph}. Top figure shows \emph{verb semantic roles}; bottom figure shows clusters of \emph{coreferred} mentions. 
    The visualizations use CogCompNLP~\citep{KSZRCSRRLDTRMFWYSGUANLR18} and AllenNLP~\citep{GGNTDLPSZ18}.
    }
    \label{fig:annotation:mrhug}
\end{figure}

Similarly the answers to \textbf{Question 2 and 3} are directly included in the paragraph, although they both require some intermediate processing like the coreference task. The system we introduce in Chapter 4 uses such representations (coreference, semantic roles, etc) and in principle should be able to answer such questions. 
The dataset introduced in Chapter 6 also motivates addressing questions that require chaining information from multiple pieces of text. In a similar vein, Chapter 8 takes a theoretical perspective on the limits of chaining information. 

The rest of the questions in Table~\ref{tab:mr:hug:questions} are harder for machines, as they require information beyond what is directly mentioned in the paragraph. For example, \textbf{Question 4} requires knowledge of the distance between ``Queens'' and ``Brooklyn,'' which can be looked up on the internet. 
Similarly, \textbf{Question 5} requires information beyond text; however, it is unlikely to be looked up easily on the web. 
Understanding that ``the robbery'' took only a few minutes (and not hours or days) is part of our common sense understanding. The dataset that we introduce in Chapter 7 motivates addressing such understanding (temporal common sense). 
\textbf{Question 6 and 7} require different forms of common sense understanding, beyond the scope of this thesis. 

In this thesis we focus on the task of Question Answering (QA), aiming to progress towards NLU. And for this goal, we study various representations and reasoning algorithms. 
In summary, this thesis is centered around the following statement: 

\paragraph{Thesis Statement.} \emph{
Progress in {\color{blue}automated} question answering could be facilitated by incorporating the ability to {\color{blue}reason} over natural language {\color{blue}abstractions} and {\color{blue}world knowledge}. 
More {\color{blue}challenging, yet realistic} QA datasets pose problems to current technologies; hence, more {\color{blue}opportunities} for {\color{blue}improvement}.
}

\section{Thesis outline}
In the thesis we use QA as a medium to tackle a few important challenges in the context of NLU. We start with an in-depth review of past work and its connections to our work in Chapter 2. The main content of the thesis is organized as follows (see also Figure~\ref{fig:contributions:table}): 

\begin{figure}
    \centering
    \includegraphics[scale=0.89,trim=5.5cm 14.2cm 0cm 1cm, clip=false]{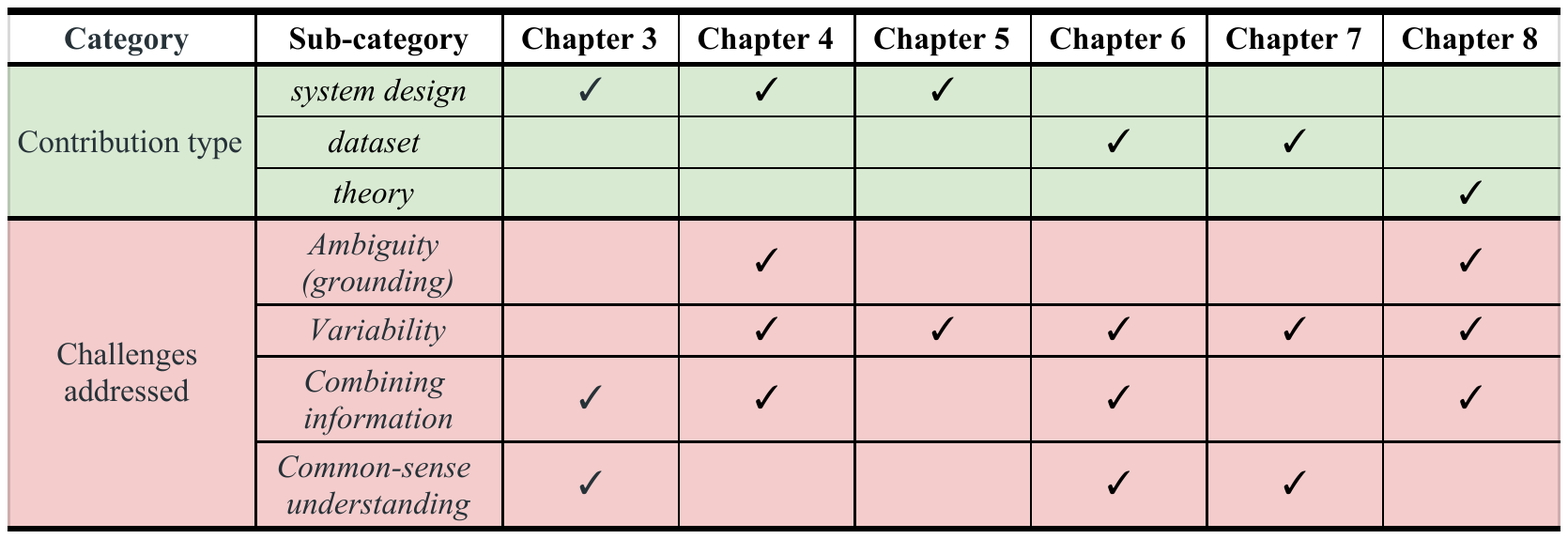}
    \caption{An overview of the contributions and challenges addressed in each chapter of this thesis. 
    }
    \label{fig:contributions:table}
\end{figure}

\begin{itemize}
    \item Part 1: \emph{Reasoning-Driven QA System Design}
    \begin{itemize}
        \item Chapter 3 discusses \tableilp, a model for abductive reasoning over natural language questions, with internal knowledge available in tabular representation. 
        \item Chapter 4 presents \textilp, an extension of the system in the previous chapter to function on raw text knowledge. 

        \item Chapter 5 
        studies 
        the notion of {\em essential question terms} with the goal of making QA solvers more robust to distractions and irrelevant information. 
        
    \end{itemize}
    \item Part 2: \emph{Moving the Peaks Higher: More Challenging QA datasets}
    \begin{itemize}
        \item Chapter 5 presents \multirc, a reading comprehension challenge which requires combining information from \underline{multiple sentences}.


        \item Chapter 6 presents \tacoqa, a reading comprehension challenge which requires  the ability to resolve \underline{temporal common sense}. 
    \end{itemize}
    \item Part 3: \emph{Formal Study of Reasoning in Natural Language}
    \begin{itemize}
        \item Chapter 7 presents a formalism, in an effort to provide \underline{theoretical grounds} to the existing intuitions on \underline{the limits and possibilities} in reasoning, in the context of natural language.
    \end{itemize}
\end{itemize}

\chapter{Background and Related Work}

\epigraph{ 
``Whoever wishes to foresee the future must consult the past.''
}{--- \textup{Nicolo Machiavelli}, 1469-1527}
\label{chapter:background}

\section{Overview}
\label{sec:intro:qa:review}
In this chapter, we review the related literature that addresses different aspects of \emph{natural language understanding.}

Before anything else, we define the terminology (Section~\ref{sec:relate:terminology}). 
We divide the discussion into multiple interrelated axes: 
Section~\ref{sec:related:protocols} discusses 
various evaluation protocols and datasets introduced in the field. 
We then provide an overview of the field from the perspective of \emph{knowledge representation and abstraction} in Section~\ref{sec:related:kr}. Building on the 
discussion of representation, we provide a survey of \emph{reasoning} algorithms, in Section~\ref{sec:related:reasoning}. We end the chapter with 
a short section on the technical background necessary for the forthcoming chapters (Section~\ref{sec:background:notation:technical}). 

To put everything into perspective, we show a summary of the highlights of the field in Figure~\ref{fig:nlu:history}. Each highlight is color-coded to indicate its contribution type. In the following sections, we go over a select few of these works and explain the evolution of the field, especially those directly related to the focus of this work. 

\begin{figure}
    \centering
    \includegraphics[scale=0.5,trim=1cm 0.0cm 1cm 5cm, clip=false]{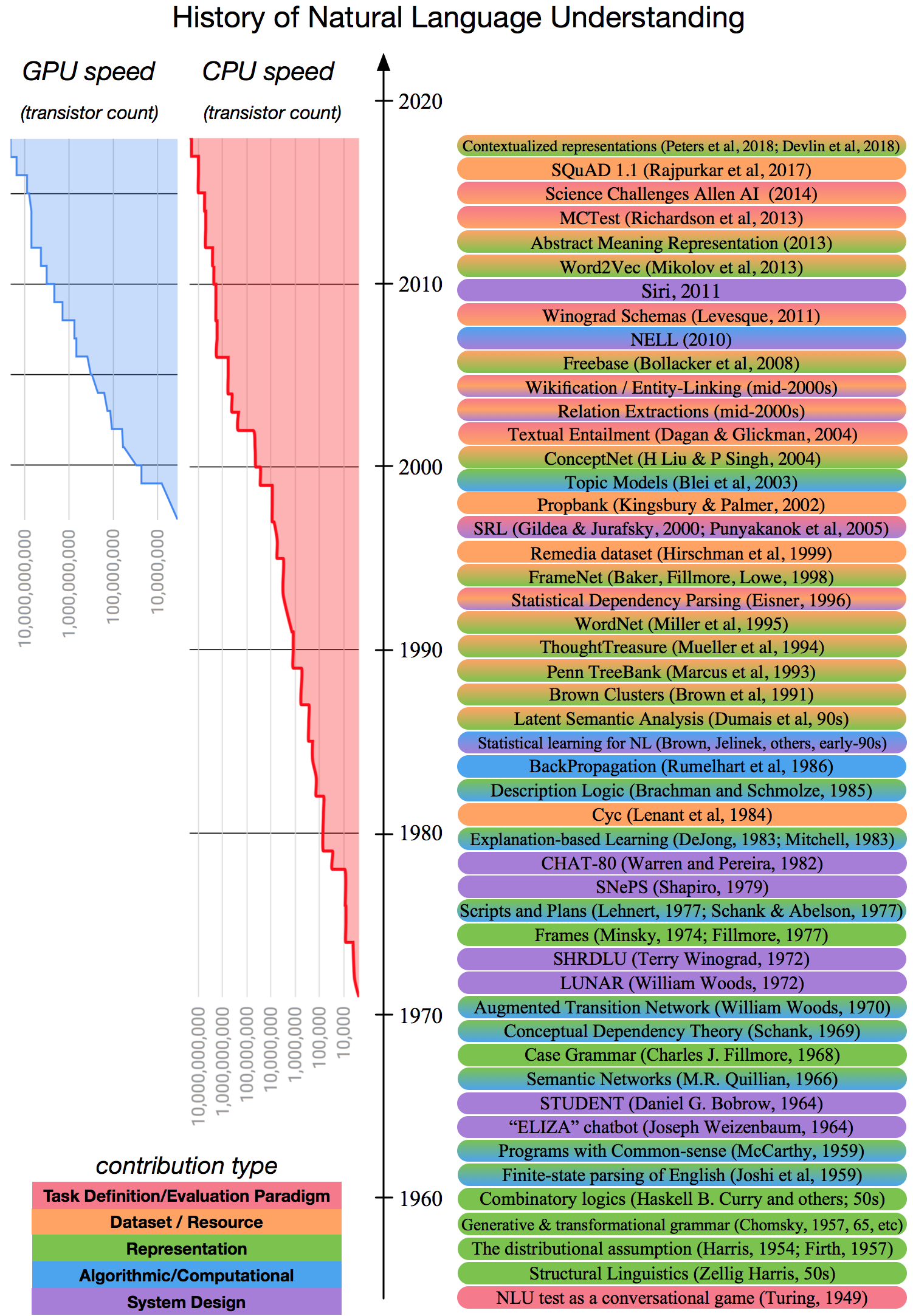}
    \caption{
        Major highlights of NLU in the past 50 years (within the AI community). For each work, its contribution-type is color-coded. To provide perspective about the role of the computational resources available at each period, we 
        show the progress of CPU/GPU hardware over time. 
    }
    \label{fig:nlu:history}
\end{figure}

\section{Terminology}
\label{sec:relate:terminology}
Before starting our main conversation, we define the terminology we will be using throughout this document. 
\begin{itemize}
    \item 
    \underline{Propositions} are \textit{judgments} or \textit{opinions} which can be true or false. A proposition is not necessarily a sentence, although a sentence can express a proposition (e.g., ``cats cannot fly"). 
    \item 
    A \underline{concept} is either a physical entity (like a \textit{tree, bicycle, etc}) or an abstract idea (like \textit{happiness, thought, betrayal, etc}). 
    \item 
    a \underline{belief} is an expression of faith and/or trust in the truthfulness of a proposition. 
    We also use \emph{confidence} or \emph{likelihood} to refer to the same notion. 
    \item 
    \underline{Knowledge} is information, facts, understanding acquired through experience or education. 
    The discussion on the philosophical nature of knowledge 
    and its various forms is studied under \textit{epistemology} \citep{Steup14}. 
    \item 
    \underline{Representation} 
    is a medium through which knowledge is provided to a system. 
    For example, the number 5 could be represented as the string ``5", as bits {\tt 101}, or 
    Roman numeral ``V", etc. 
    \item 
    \underline{Abstraction} 
     defines the level of granularity in a representation. For example, the mentions ``New York City'', ``Buenos Aires'', ``Maragheh'' could all be \emph{abstracted} as {\tt \small city}.  
    \item 
    \underline{Knowledge acquisition} is the process of identifying and acquiring the relevant knowledge, according to the representation. 
    \item 
    \underline{Reasoning} is the process of drawing a conclusion based on the given information. 
    We sometimes refer to this process as \emph{decision-making} or \emph{inference}. 
\end{itemize}
\section{Measuring the progress towards NLU}
\label{sec:related:protocols}
\subsection{Measurement protocols}
Evaluation protocols are critical in incentivizing the field to solve the right problems. 
One of the earliest proposals is due to Alan Turing: if you had a pen-pal for years, you would not know whether you're corresponding to a human or a machine~\citep{Turing50,Harnad92}. A major limitation of this test (and many of its extensions) is that it is ``expensive'' to compute~\citep{Hernandez-Orallo00,French00}. 

The protocol we are focusing on in this work is through answering natural language questions; if an actor (human or computer) understands a given text, it should be able to answer any questions about it. Throughout this thesis, we will refer to this protocol as \textbf{Question Answering} (QA). 
This has been used in the field for many years~\citep{McCarthy76,Winograd72,Lehnert77}. 
There are few other terms popularized in the community to refer 
the same task we are solving here. 
The phrase \textbf{Reading Comprehension} is borrowed from standardized tests (SAT, TOEFL, etc.), usually refers to the scenario where a paragraph is attached to the given question. Another similar phrase is  \textbf{Machine Comprehension}. Throughout this thesis, we use these phrases interchangeably to refer to the same task. 

To make it more formal, for an assumed scenario described by a paragraph $P$, a system $f$ equipped with NLU should be able to answer any questions $Q$ about the given paragraph $P$. One can measure the expected performance of the system on a set of questions $\mathcal{D}$, via some distance measure $d(., .)$ between the predicted answers $f(Q; P)$ and the correct answers $f^*(Q; P)$ (usually a prediction agreed upon by multiple humans): 
$$
\mathfrak{R}(f; \mathcal{D}) = \mathbb{E}_{(Q, P) \sim \mathcal{D}} \Big[ d \Big( f\left(Q; P\right), f^*\left(Q; P\right) \Big) \Big] 
$$

A critical question here is the choice of question set $\mathcal{D}$ so that $\mathfrak{R}(f; \mathcal{D})$ is an effective measure of $f$'s progress towards NLU. 
Denote the set of all the possible English questions as $\mathcal{D}_u$. This is an enormous set and, in practice it is unlikely that we could write them all in one place. Instead, it might be more practical to sample from this set. In practice, this sampling is replaced with \emph{static} datasets. This introduces a problem: datasets are hardly a uniform subset of $\mathcal{D}_u$; instead, they are heavily skewed towards more simplicity.

Figure~\ref{fig:question:manifold} depicts a hypothetical high-dimensional manifold of all the natural language questions in terms of an arbitrary representation (bytes, characters, etc.) 
Unfortunately, datasets are usually biased samples of the universal set $\mathcal{D}_u$. And they are often biased towards simplicity. This issue makes the dataset design of extra importance since performance results on a single set might not be a true representative of our progress. Two chapters of this work are dedicated to the construction of QA datasets. 

\begin{figure}[h]
    \centering
    \includegraphics[scale=0.35,trim=1cm 0cm 0cm 0cm, clip=false]{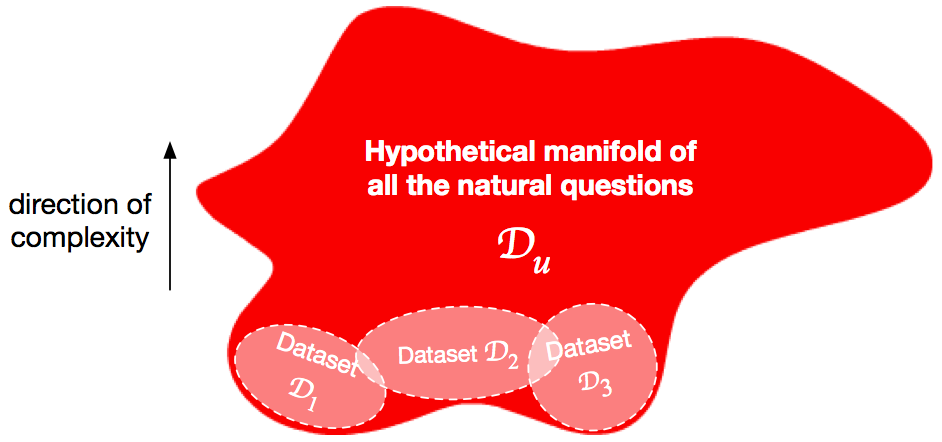}
    \caption{A hypothetical manifold of all the NLU instances. Static datasets make it easy to evaluate our progress but since they usually give a biased estimate, they limit the scope of the challenge.}
    \label{fig:question:manifold}
\end{figure}

There are few flavors of QA in terms of their answer representations (see Table~\ref{tab:sample:questions:background}): 
     (i) questions with multiple candidate-answers, a subset of which are correct; 
    (ii) extractive questions, where the correct answer is a  substring of a given paragraph; 
    (iii) Direct-answer questions; a hypothetical system has to \emph{generate} a string for such questions. 
The choice of answer-representation has direct consequences for the representational richness of the dataset and ease of evaluation. The first two settings (multiple-choice and extractive questions) are easy to evaluate but restrict the richness of the dataset. Direct-answer questions can result in richer datasets but are more expensive to evaluate. 

\begin{table}[]
    \centering
    \small
    \begin{tabular}{c|L{0.93 \textwidth }}
        \toprule 
        \parbox[t]{2mm}{\multirow{3}{*}{\rotatebox[origin=c]{90}{\footnotesize Multiple-choice \hspace{-0.5cm} }}}  & Dirk Diggler was born as Steven Samuel Adams on April 15, 1961 outside of Saint Paul, Minnesota. His parents were a construction worker and a boutique shop owner who attended church every Sunday and
        believed in God. Looking for a career as a male model, Diggler dropped out of school at age 16 and left home. He was discovered at a falafel stand by Jack Horner. Diggler met his friend, Reed Rothchild, through Horner in 1979 while working on a film. \\ 
        & \textbf{Question:} How old was Dirk when he met his friend Reed?\\ 
        & \textbf{Answers:} *(A) 18 \hspace{0.75cm} (B) 16 \hspace{0.75cm} (C) 22  \\ 
        \midrule
        \parbox[t]{2mm}{\multirow{3}{*}{\rotatebox[origin=c]{90}{\footnotesize Extractive}}} & The city developed around the Roman settlement Pons Aelius and was named after the castle built in 1080 by Robert Curthose, William the Conqueror's eldest son. The city grew as an important centre for the wool trade in the 14th century, and later became a major coal mining area. The port developed in the 16th century and, along with the shipyards lower down the River Tyne, was amongst the world's largest shipbuilding and ship-repairing centres. \\ 
        & \textbf{Question:} Who built a castle in Newcastle in 1080? \\ 
        & \textbf{Answers:} ``Robert Curthose'' \\
        \midrule
        \parbox[t]{2mm}{\multirow{3}{*}{\rotatebox[origin=c]{90}{{\footnotesize Direct-answer}}}} & \\ 
         & \textbf{Question:} Some birds fly south in the fall. This seasonal adaptation is known as migration. Explain why these birds migrate.  \\ 
        & \textbf{Answers:} ``A(n) bird can migrate, which helps cope with lack of food resources in harsh cold conditions by getting it to a warmer habitat with more food resources.'' \\ 
        \bottomrule
    \end{tabular}
    \caption{Various answer representation paradigms in QA systems; examples selected from \cite{KCRUR18,RZLL16,CEKSTTK16}. }
    \label{tab:sample:questions:background}
\end{table}



Datasets 
make it possible to automate the evaluation of the progress towards NLU and 
be able to compare systems to each other on fixed problems sets. 
One of the earliest NLU datasets published in the field is the Remedia dataset~\citep{HLBB99} which contains 
short-stories written in simple language for kids provided by Remedia Publications. 
Each story has 5 types of questions (\emph{who, when, why, where, what}). 
Since then, there has been many suggestions as to what kind of question-answering dataset is a better test of NLU. 
\cite{BGBGHF05} suggests SAT exams as a challenge for AI. \cite{Davis14} proposes 
multiple-choice challenge sets that are easy for children but difficult for computers.  In a similar spirit, \cite{ClarkEt16} advocate elementary-school science tests. Many science questions have answers that are not explicitly stated in text and instead, require combining information together. 
In Chapter 2, 3 we use 
elementary-school science tests as our target challenge.

While the field has produced many datasets in the past few years, many of these datasets are either too restricted in terms of their linguistic richness or they contain annotation biases~\citep{GSLSBS18,PNHRD18}. 
For many of these datasets, 
it has been pointed out
that many of the high-performing models neither need to `comprehend' in order to correctly predict an answer, nor learn to `reason' in a way that generalizes across datasets~\citep{ChenBoMa16,JiaLi17,KaushikLi18}. In Section~\ref{subsec:perturbation} we show that 
adversarially-selected candidate-answers 
result in a significant drop in performance of a few state-of-art science QA systems. 
To address these weaknesses, in Chapter 4, 5 we 
propose two new challenge datasets which, we believe, pose better challenges for systems.  


A closely related task is the task of Recognizing Textual Entailment (RTE)~\citep{KSZRCSRRLDTRMFWYSGUANLR18,DRSZ13}, as QA can be cast as entailment (Does {\it $P$} entail {\it $Q+A$}? \citep{BCDG08}). 
While we do not directly address this task, in some cases we use it as a component within out proposed QA systems (in Chapter 3 and 4). 






\section{Knowledge Representation and Abstraction for NLU}
\label{sec:related:kr}
The discussion of knowledge representation has been with AI since its beginning and it is central to the progress of language understanding. 
Since directly dealing with the raw input/output complicates the reasoning stage, historically researchers have preferred to devise 
a  
middleman between the 
raw 
information and the reasoning engine. 
Therefore, the need for an intermediate level seems to be essential. In addition, in many problems, there is a significant amount of knowledge that is not mentioned directly, but rather  implied from the context. Somehow the extra information has to be provided to the reasoning system. 
As a result, the discussion goes beyond just creating formalism for information, and also includes issues like, how to \emph{acquire}, \emph{encode} and \emph{access} it.
The issue of representations applies to both \emph{input} level information and the \emph{internal} knowledge of a reasoning system. 
We refer to some of the relevant debates in the forthcoming sections.  


\subsection{Early Works: ``Neats vs Scruffies''\protect\footnote{
Terms originally made by Roger Schank to characterize two different camps: the first group that represented commonsense knowledge in the form of large amorphous semantic networks, as opposed to another from the camp of whose work was based on logic and formal extensions of logic. 
}}
An early trend emerged as the family of symbolic and logical representations, such as propositional and 1st-order logic~\citep{McCarthy63}.
This approach has deep roots in \emph{philosophy} and \emph{mathematical logic}, where the theories have evolved since Aristotle's time. Logic, provided a general purpose, clean and uniform language, both in terms of representations and reasoning.

The other closely-related school of thought evolved from \emph{linguistics} and \emph{psychology}. This trend was less concerned with mathematical rigor, but more concerned with richer psychological and linguistic motivations. For example, \emph{semantic networks}~\citep{Quillan66}, a network of concepts and links, was based on the idea that memory consists of associations between mental entities. In 
Chapter 8 we study a formalism for reasoning with such graph-like representations. 
\emph{Scripts} and \emph{plans} are representational tools to model frequently encountered activities; e.g., going to a restaurant~\citep{SchankAb75,Lehnert77}. 
Minsky and Fillmore, separately and in parallel, advocated \emph{frame}-based representations~\citep{Minsky74,Fillmore77}. 
The following decades, these approaches have evolved into fine-grained representations and hybrid systems for specific problems. 
One of the first NLU programs was the \textsc{STUDENT} program of~\cite{Bobrow64}, written in \textsc{LISP}~\citep{McCarthyLe65}, which could read and solve high school algebra problems expressed in natural language. 


\begin{figure}
    \centering
    \includegraphics[scale=0.294,trim=1cm 0cm 0cm 0cm 0cm, clip=false]{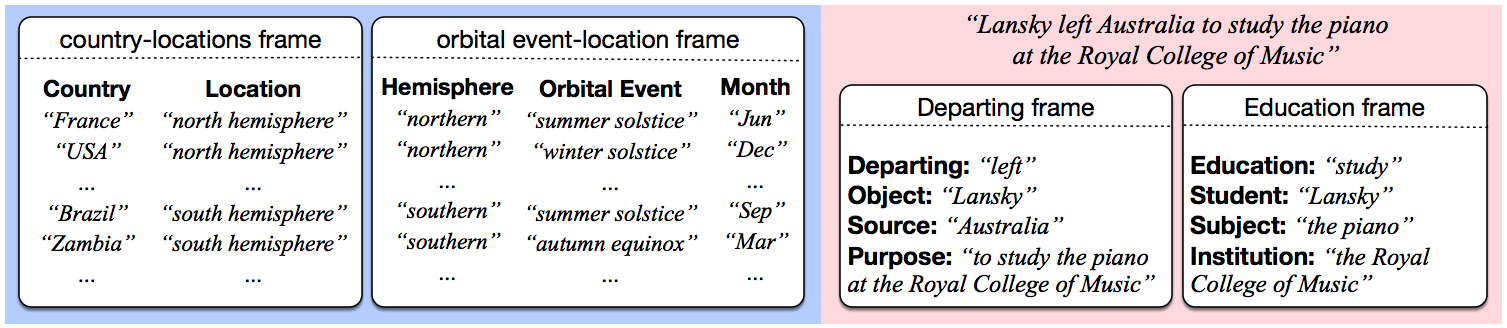}
    \caption{Example frames used in this work. Generic basic science frames (left), used in Chapter~3;  event frames with values filled with the given sentence (right), used in Chapter~4. }
    \label{fig:frame:example}
\end{figure}

Intuitively, a \emph{frame} induces a grouping of concepts and creates abstract hierarchies among them. 
For example, \emph{``Monday'', ``Tuesday'', ...} are distinct concepts, but all members of the same conceptual frame. 
A frame consists of a group of slots and fillers to define a stereotypical object or activity. A slot can contain values such as rules, facts, images, video, procedures or even another frame~\citep{FikesKe85}. 
Frames can be organized hierarchically, where the default values can be inherited the value directly from parent frames. 
This is part of our underlying representation in Chapter 3, where the reasoning is done over tables of information (an example in Figure~\ref{fig:frame:example}, left). 
Decades later after its proposal, the frame-based approach resulted in resources like FrameNet~\citep{BakerFiLo98}, or tasks like Semantic Role Labeling~\citep{GildeaJu02,PalmerGiKi05,PunyakanokRoYi04}. 
This forms the basis for some of the key representations we use in Chapter 4 (see Figure~\ref{fig:frame:example}, right).  



\subsection{Connectionism}
\label{sec:connectionism}
There is another important trend  inspired by the apparent brain function emergent from interconnected networks of neural units~\citep{Rosenblatt58}.  
It lost many of its fans after \cite{MinskyPa69} showed representational limitations of shallow networks in approximating few functions. 
However, a series of events reinvigorated this thread: 
Notably, \cite{RumelhartHiWi88} found a formalized way to train networks with more than one layer (nowadays known as Back-Propagation algorithm). This work emphasized the \textit{parallel} and \textit{distributed} nature of information processing and  gave rise to the ``connectionism" movement.
Around the same time \citep{Funahashi89} showed the universal approximation property for feed-forward networks (any continuous function on the real numbers can be uniformly approximated by neural networks). Over the past decade, this school has enjoyed newfound excitement by effectively exploiting parallel processing power of GPUs and harnessing large datasets to show progress on certain datasets.

\subsection{Unsupervised representations}
\label{sec:representations:intro}
Unsupervised representations are one of the areas that have shown tangible impacts across the board. A pioneering work is \cite{BDMPL92} which creates binary term representations based on co-occurrence information. Over the years, a wide variety of such representations emerged; using Wikipedia concepts~\citep{GabrilovichMa07}, word co-occurrences~\citep{TurneyPa10}, co-occurrence factorization~\citep{MCCD13,LevyGo14,PenningtonSoMa14,LXTJZC15}, and using context-sensitive representation~\citep{PNIGCLZ18,DCLT18}. In particular, the latter two are inspired by the connectionist frameworks in the 80s and have shown to be effective across a wide range of NLU tasks. In this thesis we use unsupervised representations in various ways: In Chapter 3 and 4 we use such representations for phrasal semantic equivalence within reasoning modules. In Chapter 5 we use as features of our supervised system. In Chapter 6, 7 we create NLU systems based on such representations in order to create baselines for the datasets we introduce. 

A more recent highlight along this path is the emergence of new unsupervised representations that have been shown to capture many interesting associations in freely available data~\citep{PNIGCLZ18,DCLT18}. 

\subsection{Grounding of meanings}
A tightly related issue to the abstraction issue is \emph{grounding} natural language surface information to their actual meaning~\citep{Harnad90}, as discussed in Section~\ref{intro:challenges:nlu}.
Practitioners often address this challenging by enriching their representations; 
for example by mapping textual information to Wikipedia entries~\citep{MihalceaCs07}. 
In Chapter 4 we use the disambiguation of semantic actions and their roles~\citep{PunyakanokRoYi04,DangPa05}. 
Chapter 8 of this thesis provides a formalism that incorporates elements of the symbol-grounding problem and shed theoretical light on existing empirical intuitions.

\subsection{Common sense and implied meanings}
A major portion of our language understanding is only implied in language and not explicitly mention (examples in  Section~\ref{intro:challenges:nlu}). 
This difficulty of this challenge has historically been under-estimated. 
Early AI, during the sixties and onward, experienced a lot of interest in modeling common sense knowledge. McCarthy, one of the founders of AI, believed in formal logic as a solution to common sense reasoning~\citep{McCarthyLi90}. 
\cite{Minsky88} estimated that ``... commonsense is knowing maybe 30 or 60 million things about the world and having them represented so that when something happens, you can make analogies with others". There have been decade-long efforts to create knowledge bases of common sense information, such as Cyc~\citep{Lenat95} and ConceptNet~\citep{LiuSi04}, but none of these have yielded any major impact so far. 
A roadblock in progress towards such goals is the lack of \emph{natural end-tasks} that can provide an objective measure of progress in the field. 
To facilitate research in this direction, in Chapter 6 we provide a new natural language QA dataset that performing well on it requires significant progress on multiple \emph{temporal} common sense tasks.


\subsection{Abstractions of the representations}
\emph{Abstraction} of information is one of the key issues in any effort towards an effective representation. 
Having coarser abstraction could result in better generalization. However, too much abstraction could result in losing potentially-useful details. 
In general, there is a trade-off between the expressive level of the representation and the reasoning complexity. 
We also deal with this issue in multiple ways: (i) we use unsupervised representations that have been shown to indirectly capture  abstractions~\citep{MahabalRoMi18}. 
(ii) we use systems pre-trained with annotations that abstract over raw text; for example, in Chapter 4 we use semantic roles representations of sentences, which abstract over low-level words and map the argument into their high-level thematic roles.

For a given problem instance, how does a system internally choose the right level of abstraction?
The human attention structure is extremely good in abstracting concepts~\citep{JohnsonPr04,JanzenVi97}, although automating this is an open question. 
One way of dealing with such issues is 
to use multiple levels of abstraction and let the reasoning algorithm use the right level of abstraction when available~\citep{Rasmussen85,BisantzVi94}. 
In Chapter 4, we take a similar approach by using a collection of different abstractions.

\section{Reasoning/Decision-making Paradigms for NLU}
\label{sec:related:reasoning}
\subsection{Early formalisms of reasoning}
The idea of automated reasoning dates back before AI itself and can be traced to ancient Greece. Aristotle's syllogisms paved the way for \emph{deductive} reasoning formalism. It continued its way with philosophers like Al-Kindi, Al-Farabi, and Avicenna~\citep{Davidson92}, before culminating as the modern mathematics and logic. 

Within AI research, \cite{McCarthy63} pioneered the use of logic for automating reasoning for language problems, which over time branched into other classes of reasoning~\citep{HHNT89,EvansNeBy93}.  

A closely related reasoning to what we study here is \emph{abduction}~\citep{Peirce83,HSAM93}, which is the process of finding \textit{the best minimal explanation} from a set of observations (see Figure~\ref{fig:abduction:example}). 
Unlike in deductive reasoning, in abductive reasoning the premises do not guarantee the conclusion. 
Informally speaking, abduction is inferring cause from effect (reverse direction from deductive reasoning). The two reasoning systems in Chapter 3 and 4 can be interpreted as abductive systems. 

We define the notation to make the exposition slightly more formal. 
Let $\vdash$ denote entailment and $\bot$ denote contradiction. Formally, (logical) abductive reasoning is defined as follows:
\begin{framed}
\noindent Given background knowledge $B$ and observations $O$, find a hypothesis $H$, 
such that $B \cup H \nvdash \bot$ (consistency with the given background) and $B \cup H \vdash O$ (explaining the observations).
\end{framed}

\begin{figure}
    \centering
    \includegraphics[scale=0.5,trim=0cm 0cm 0cm 0cm]{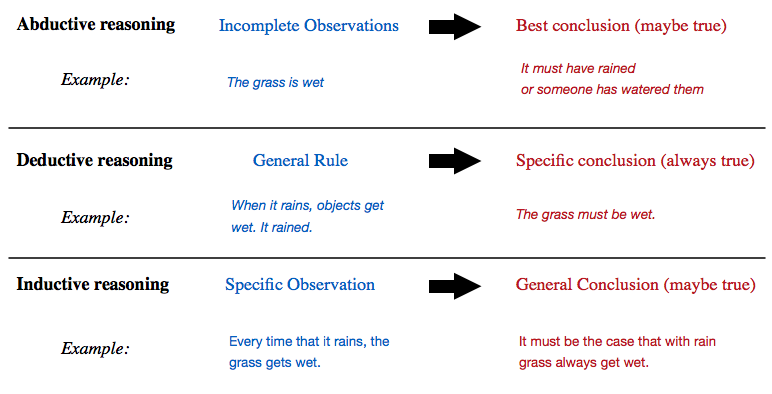}
    \caption{Brief definitions for popular reasoning classes and their examples.}
    \label{fig:abduction:example}
\end{figure}

In practical settings,  
this purely logical definition has many limitations: (a) There could be multiple hypotheses $H$ that explain a particular set of observations given the background knowledge. The best hypothesis has to be selected based on some measure of goodness and the simplicity of the hypothesis (Occam's Razor). (b) Real life has many uncertain elements, i.e. there are degrees of certainties (rather than binary assignments) associated with observations and background knowledge. Hence the decision of consistency and explainability has to be done with respect to this fuzzy measure. (c) The inference problem in its general form is computationally intractable; often assumptions have to be made to have tractable inference (e.g., restricting the representation to Horn clauses).

\subsection{Incorporating ``uncertainty'' in reasoning}
Over the years, a wide variety of \textit{soft} alternatives have emerged for reasoning algorithms, by incorporating uncertainty into symbolic models. 
 This resulted in theories like fuzzy-logic~\citep{Zadeh75}, or probabilistic Bayesian networks~\citep{Pearl88,Dechter13}, soft abduction~\citep{HSME88,SelmanLe90,Poole90}. 
In Bayesian networks, the (uncertain) background knowledge is encoded in a graphical structure and upon receiving observations, the probabilistic explanation is derived by maximizing a posterior probability distribution. These models are essentially based on propositional logic and cannot handle quantifiers~\citep{KateMo09}. 
Weighted abduction combines the weights of relevance/plausibility  with first-order logic rules~\citep{HSME88}. 
However, unlike probability theoretic frameworks, their weighting scheme does not have any solid theoretical basis and does not lend itself to a complete probabilistic analysis. Our framework in Chapter 3,4 is also a way to perform abductive reasoning under uncertainty. Our proposal is different from the previous models in a few ways:  (i) Unlike Bayesian network our framework is not limited to propositional rules; in fact, there are first-order relations used in the design of \tableilp\ (more details in Chapter 3). 
(ii) unlike many other previous works, we do not make representational assumptions to make the inference simpler (like limiting to Horn clauses, or certain independence assumptions). In fact, the inference might be NP-hard, but with the existence of industrial ILP solvers this is not an issue in practice. 
Our work is inspired by a prior line of work on inference on structured representations to reason on (and with) language; see \cite{CRRR08,CGRS10,ChangRaRo12}, among others.

\subsection{Macro-reading vs micro-reading}
With increased availability of  information (especially through the internet) \textbf{macro-reading} systems have emerged with the aim of leveraging 
a large variety of resources and exploiting the redundancy of information~\citep{MBCHW09}. Even if a system does not understand one text, there might be many other texts that convey a similar meaning. Such systems derive significant leverage from relatively shallow statistical methods with surprisingly strong performance~\citep{CEKSTTK16}. 
Today's Internet search engines, for instance, can successfully retrieve \emph{factoid} style answers to many natural language queries by efficiently searching the Web. Information Retrieval (IR) systems work under the assumption that answers to many questions of interest are often explicitly stated somewhere~\citep{KwokEtWe01}, and all one needs, in principle, is access to a sufficiently large corpus. Similarly, statistical correlation based methods, such as those using Pointwise Mutual Information or PMI~\citep{ChurchHa89}, work under the assumption that many questions can be answered by looking for words that tend to co-occur with the question words in a large corpus.
While both of these approaches help identify correct answers, they
are not suitable for questions requiring language understanding and reasoning, such as chaining together multiple facts in order to arrive at a conclusion. 
On the other hand, \textbf{micro-reading} aims at \emph{understanding} a piece of evidence given to the system, without reliance of \emph{redundancy}. The focus of this thesis is \emph{micro-reading} as it directly addresses NLU; that being said, whenever possible, we use macro-reading systems as our baselines.

\subsection{Reasoning on ``structured'' representations}
With increasing knowledge resources and diversity of the available knowledge representations, 
numerous QA systems are developed to operate over large-scale \textit{explicit} knowledge representations. These approaches perform reasoning over \textit{structured} (discrete) abstractions. For instance, \cite{CGRS10} address RTE (and other tasks) via inference on structured representations), 
\cite{BBCGGHKPS13} use AMR annotators~\citep{WXPP15}, 
\cite{UBLNGC12} use RDF knowledge~\citep{YZWYW17}, 
 \cite{ZettlemoyerCo05,CGCR10,GoldwasserRo14,KrishnamurthyTaKe16}  
use semantic parsers to answer a given question, and \cite{DoChRo11,DoLuRo12} employ constrained inference for temporal/causal reasoning. 
The framework we study in Chapter 3 is a reasoning algorithm functioning over  tabular knowledge (frames) of basic science concepts. 



An important limitation of IR-based systems is their inability to connect distant pieces of information together. 
However, many other realistic domains (such as science questions or biology articles) have answers that are not explicitly stated in text, and instead require combining facts together. 
\cite{KhotSaCl17} creates an inference system capable of combining Open IE tuples~\citep{BCSBE07}. \cite{JSSC17} propose reasoning by aggregating sentential information from multiple knowledge bases. \cite{SCMN13,MNDB17} propose frameworks for chaining relations to infer new (unseen) relations. Our work in Chapter 3 creates chaining of information over multiple tables. The reasoning framework in Chapter 4 investigates reasoning over multiple peaces of raw text. The QA dataset in Chapter 5 we propose also encourages the use of information from different segments of the story. Chapter 8 proposes a formalism to study limits of chaining long-range information.

\subsection{Models utilizing massive annotated data}
A  highlight over the past two decades is the advent of statistical techniques into NLP~\citep{HLBB99}. Since then, a wide variety of supervised-learning algorithms have shown strong performances on different datasets.

The increasingly large amount of data available for recent benchmarks make it possible to train neural models (see ``Connectionism''; Section~\ref{sec:connectionism})~\citep{SKFH16,PTDU16,WYW18,LSDG18,HPHXWZ2018}.
Moreover, an additional technical shift was using distributional representation of words (word vectors or embeddings) extracted from large-scale text corpora~\citep{MCCD13,PenningtonSoMa14} (see Section~\ref{sec:representations:intro}). 

Despite all the decade-long excited about supervised-learning algorithms, the main progress, especially in the past few years, has mostly been due to the re-emergence of \emph{unsupervised} representations~\citep{PNIGCLZ18,DCLT18}.\footnote{\emph{Unsupervised} in the sense that they are constructed with freely available data, as opposed to task-specific annotated data. }

\section{Technical background and notation}
\label{sec:background:notation:technical}
In this section, we provide the relevant mathematical background used throughout this thesis. We cover three main areas used widely across this document. 

\subsection{Complexity theory}
We follow the standard notation for asymptotic comparison of functions: 
$O(.), o(.), \Theta(.), \Omega(.),$ and $\omega(.)$~\citep{CLRS09}.

We use \emph{P} and \emph{NP} to refer to the basic complexity classes. We briefly review these classes: \emph{P} consists of all problems that can be solved efficiently (in polynomial time). \emph{NP} (non-deterministic polynomial time) includes all problems that given a solution, one can efficiently \emph{verify} the solution.   
When a problem is called \emph{intractable}, it refers to its complexity class being at least \emph{NP}-hard. 

\subsection{Probability Theory}
\label{background:probability:theory}
$X \sim f(\theta)$ denotes a random variable $X$ distributed according to probability distribution $f(\theta)$, paramterized by $\theta$. The mean and variance of $X$ are denoted as 
$\mathbb{E}_{X \sim f(\theta)}[X]$ and $\mathbb{V}[X]$, resp.
$\bern{p}$ and $\bin{n,p}$ denote the Bernoulli and Binomial distributions, resp.

\subsection{Graph theory}
\label{sec:graph:theory}
We denote an undirected graph with $G(V, E)$ where $V$ and $E$ are the sets of nodes and edges, resp. We use the notations $V_G$ and $E_G$ to  refer to the nodes and edges of a graph $G$, respectively. 

A \emph{subgraph} of a graph $G$ is another graph formed from a subset of the vertices and edges of $G$. The vertex subset must include all endpoints of the edge subset, but may also include additional vertices. 

A \emph{cut} $C = (S,T)$ in $G$ is a partition of the nodes $V$ into subsets $S$ and $T$. The \emph{size} of the cut $C$ is the number of edges in $E$ with one endpoint in $S$ and the other in $T$.


\subsection{Optimization Theory}
\label{sec:optimization:background}

As it is widely known an ILP can be written as the following: 
\begin{align}
& \text{maximize}   \hspace{-3cm} & \mathbf{w}^\mathrm{T} \mathbf{x} &  \label{eq:ilp:obj}\\
& \text{subject to} \hspace{-3cm} & A \mathbf{x} \le \mathbf{b}, & \label{eq:ilp:cons} \\
& \text{and} \hspace{-3cm} & \mathbf{x} \in \mathbb{Z}^n. & \label{eq:ilp:ints}
\end{align}
We first introduce the basic variables, and define the full definition of the ILP program: define the weights in the objective function ($\mathbf{w}$ in Equation \ref{eq:ilp:obj}), and the constraints ($A$ and $\mathbf{b}$ in Equation \ref{eq:ilp:cons}).

This formulation is incredibly powerful and has been used for many  problems. In the context of NLP problems, ILP based discrete optimization was introduced by \citet{RothYi04} and has been successfully  used~\citep{CGRS10,BerantDaGo10,SrikumarRo11,GoldwasserRo14}. 
In Chapter 3 and 4 also, we formalize our desired behavior as an optimization problem. 

This optimization problem with integrality constraint and its general form, is an NP-hard problem. That being said, the industrial solvers (which use cutting-plane and other heuristics) are quite fast across a wide variety of problems. 


\part{Reasoning-Driven System Design}
\label{part:abductive}
\chapter{QA as Subgraph Optimization on Tabular Knowledge}
\label{chapter:tableilp}
\epigraph{ 
``The techniques of artificial intelligence are to the mind what bureaucracy is to human social interaction.''
}{--- \textup{Terry Winograd}, Thinking Machines: Can there be? 1991}

\newif\ifarxiv

\arxivtrue

\section{Overview}
\label{sec:intro}

Consider a question from the NY Regents 4th Grade Science Test:\footnote{This chapter is based on the following publication: \cite{KKSCER16}.}
\begin{myquote}
{In New York State, the longest period of daylight occurs during which month? \ \\ (A) June \ (B) March \ (C) December \ (D) September}
\end{myquote}

\noindent
We would like a QA system that, even if the answer is not explicitly stated in a document, can \emph{combine basic scientific and geographic facts} to answer the question, e.g., New York is in the north hemisphere; the longest day occurs during the summer solstice; and the summer solstice in the north hemisphere occurs in June (hence the answer is June). Figure~\ref{fig:ny-example:tableilp} illustrates how our system approaches this, with the highlighted support graph representing its line of reasoning.


\begin{wrapfigure}{r}{0.69\textwidth}
\centering
\includegraphics[trim=2.2cm 18.4cm 1cm 2.5001cm, clip=true, scale=0.76]{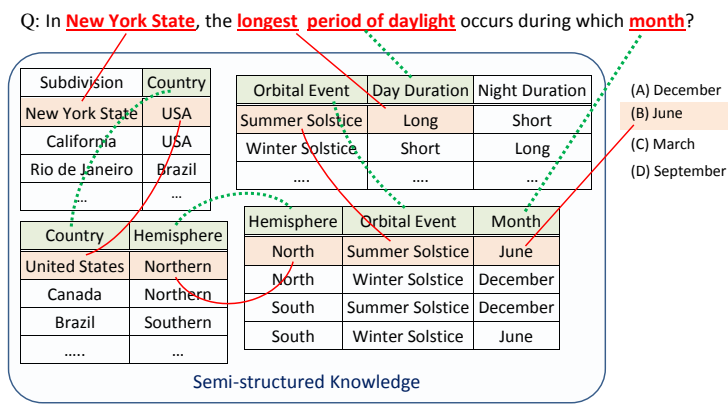}
\caption{
\footnotesize
\tableilp\ searches for the best support graph (chains of reasoning)
connecting the question to an answer, in this case June.
Constraints on the graph define what constitutes valid support and how to score it (Section~\ref{subsec:ilp:tableilp}).} 
\label{fig:ny-example:tableilp}
\end{wrapfigure}
Further, we would like the system to be \emph{robust under simple perturbations}, such as changing New York to New Zealand (in the southern hemisphere) or changing an incorrect answer option to an irrelevant word such as ``last'' that happens to have high co-occurrence with the question text.

To this end, we propose a structured reasoning system, called \tableilp,
that operates over a semi-structured knowledge base derived from text and answers questions by chaining multiple pieces of information and combining parallel evidence.\footnote{{A preliminary version of our ILP model was used in the ensemble solver of \cite{CEKSTTK16}. We build upon this earlier ILP formulation, providing further details and incorporating additional syntactic and semantic constraints that improve the score by 17.7\%.}}
The knowledge base consists of \emph{tables}, each of which is a collection of instances of an $n$-ary relation defined over natural language phrases. E.g., as illustrated in Figure~\ref{fig:ny-example:tableilp}, a simple table with schema \emph{(country, hemisphere)} might contain the instance \emph{(United States, Northern)} while a ternary table with schema \emph{(hemisphere, orbital event, month)} might contain \emph{(North, Summer Solstice, June)}. \tableilp\ treats lexical constituents of the question $Q$, as well as cells of potentially relevant tables $T$, as nodes in a large graph $\mathcal{G}_{Q,T}$, and attempts to find a subgraph $G$ of $\mathcal{G}_{Q,T}$ that ``best'' supports an answer option. The notion of best support is captured via a number of structural and semantic constraints and preferences, which are conveniently expressed in the Integer Linear Programming (ILP) formalism. We then use an off-the-shelf ILP optimization engine called SCIP~\citep{Achterberg09} to determine the best supported answer for $Q$.

Following a recently proposed AI challenge~\citep{Clark15}, {we evaluate \tableilp\ on unseen elementary-school science questions from standardized tests. Specifically, we consider a challenge set \citep{CEKSTTK16} consisting of all non-diagram multiple choice questions from 6 years of NY Regents 4th grade science exams.}
In contrast to a state-of-the-art structured inference method~\citep{KBGSCE15} for this task, which used Markov Logic Networks (MLNs)~\citep{RichardsonDo06}, \tableilp\ achieves a significantly (+14\% absolute) higher test score. 
This suggests that a combination of a rich and fine-grained constraint language, namely ILP,
even with a publicly available solver
is more effective in practice than various MLN formulations of the task. Further, while the scalability of the MLN formulations 
 was limited to very few (typically one or two) selected science rules at a time, our approach easily scales to hundreds of relevant scientific facts.
%
It also complements the kind of questions amenable to IR and PMI techniques, as is evidenced by the fact that a combination (trained using simple Logistic Regression {\citep{CEKSTTK16}}) of \tableilp\ with IR and PMI results in a significant (+10\% absolute) boost in the score compared to IR alone.

Our ablation study suggests that
combining facts from multiple tables or multiple rows within a table plays an important role in \tableilp's performance. 
We also show that \tableilp\ benefits from the table structure, by comparing it with an IR system using the same knowledge (the table rows) but expressed as simple sentences; \tableilp\ scores significantly (+10\%) higher. Finally, we demonstrate that our approach is robust to a simple perturbation of incorrect answer options: while the simple perturbation results in a relative drop of 20\% and 33\% in the performance of IR and PMI methods, respectively, it affects \tableilp's performance by only 12\%.


\section{Related Work}
\label{sec:related-work}
In this section, we provide additional related work, and augment our review related work provided in Section~\ref{sec:intro:qa:review}. 

{\cite{CEKSTTK16} proposed an ensemble approach for the science QA task, demonstrating the effectiveness of a combination of information retrieval, statistical association, rule-based reasoning, and an ILP solver operating on semi-structured knowledge. Our ILP system extends their model with additional constraints and preferences (e.g., semantic relation matching), substantially improving QA performance.}

A number of systems have been developed for answering factoid questions with short answers (e.g., ``What is the capital of France?'') using document collections or databases (e.g., Freebase~\citep{BEPST08}, NELL~\citep{CBKSJM10}), for example~\citep{BrillDuBa02,FaderZeEt14,FBCFGKLMNPo10,KoNySi07,YihHeMe14,YaoDu14,ZHWYHZ14}. However, many science questions have answers that are not explicitly stated in text, and instead require combining information together. Conversely, while there are AI systems for formal scientific reasoning (e.g., \citep{GCCBCG10,Novak77}), they require questions to be posed in logic or restricted English.
Our goal here is a system that operates between these two extremes, able to combine information while still operating with natural language.

There is a relatively rich literature in the databases community, on executing different commands on the tablular content (e.g., searching, joining, etc) via a user commands issued by a semi-novice user~\cite{TJMCIPG08,TalukdarIvPe10}. 
A major distinguishing perspective is that 
in our problem the queries are generated completely independent of the the table content. However, in a database system application, a user is at-least partially  informed of the common keywords, could observe the outputs of the queries and adjust the commands accordingly. 


\section{QA as Subgraph Optimization}

We begin with our knowledge representation formalism, followed by our treatment of QA as an optimal subgraph selection problem over such knowledge, and then briefly describe our ILP model for subgraph selection.

\subsection{Semi-Structured Knowledge as Tables}
\label{subsec:tables}

We use semi-structured knowledge represented in the form of $n$-ary predicates over natural language text {\citep{CEKSTTK16}}. Formally, a $k$-column table in the knowledge base is a predicate $r(x_1, x_2, \ldots, x_k)$ over strings, where each string is a (typically short) natural language phrase. The column headers capture the table schema, akin to a relational database. Each row in the table corresponds to an instance of this predicate. For example, a simple country-hemisphere table represents the binary predicate $r_{\text{ctry-hems}}(c,h)$ with instances such as (Australia, Southern) and (Canada, Northern). Since table content is specified in natural language, the same entity is often represented differently in different tables, posing an additional inference challenge.

thAlthough techniques for constructing this knowledge base are outside the scope of this paper, we briefly mention them. Tables were constructed using a mixture of manual and semi-automatic techniques. First, the table schemas were manually defined based on the syllabus, study guides, and training questions. Tables were then populated both manually and semi-automatically using {IKE~\citep{DalviBhCl16}}, a table-building tool that performs interactive, bootstrapped relation extraction over a corpus of science text. In addition, to augment these tables with the broad knowledge present in study guides that doesn't always fit the manually defined table schemas, we ran an Open IE~\citep{BCSBE07} pattern-based \textit{subject-verb-object} (SVO) extractor from~\cite{CBBHKST14} over several science texts to populate three-column Open IE tables. Methods for further automating table construction are under development.
\subsection{QA as a Search for Desirable Support Graphs}
\label{subsec:qa-as-subgraph-selection}

We treat question answering as the task of pairing the question with an answer such that this pair has the best support in the knowledge base, measured in terms of the strength of a ``support graph'' defined as follows.

Given a multiple choice question $Q$ and tables $T$, we can define a labeled undirected 
graph $\mathcal{G}_{Q,T}$ over nodes $\mathcal{V}$ and edges $\mathcal{E}$ as follows. We first split $Q$ into lexical constituents (e.g., non-stopword tokens, or chunks) $\mathbf{q} = \{\qCons\}$ and answer options $\mathbf{a} = \{\option\}$. For each table $\tableVar$, we consider its cells $\mathbf{t} = \{\tableCell\}$ 
as well as column headers $\mathbf{h} = \{\header\}$. The nodes of $\mathcal{G}_{Q,T}$ are then $\mathcal{V} = \mathbf{q} \cup \mathbf{a} \cup \mathbf{t} \cup \mathbf{h}$.
For presentation purposes, we will equate a graph node with the lexical entity it represents (such as a table cell or a question constituent). The undirected edges of $\mathcal{G}_{Q,T}$ are $\mathcal{E} = ((\mathbf{q} \cup \mathbf{a}) \times (\mathbf{t} \cup \mathbf{h})) \cup (\mathbf{t} \times \mathbf{t}) \cup (\mathbf{h} \times \mathbf{h})$ excluding edges both whose endpoints are within a single table.

Informally, an edge denotes (soft) equality between a question or answer node and a table node, or between two table nodes. To account for lexical variability (e.g., that \emph{tool} and \emph{instrument} are essentially equivalent) and generalization (e.g., that a \emph{dog} is an \emph{animal}), we replace string equality with a phrase-level entailment or similarity function $w : \mathcal{E} \to [0,1]$ that labels each edge $e \in \mathcal{E}$ with an associated score $w(e)$. We use entailment scores (directional) from $\mathbf{q}$ to $\mathbf{t} \cup \mathbf{h}$ and from $\mathbf{t} \cup \mathbf{h}$ to $\mathbf{a}$, and similarity scores (symmetric) between two nodes in $\mathbf{t}$.\footnote{In our evaluations, $w$ for entailment is a simple WordNet-based \citep{Miller95} function that {computes the best word-to-word alignment between phrases, scores these alignments using WordNet's hypernym and synonym relations normalized using relevant word-sense frequency, and returns the weighted sum} of the scores. $w$ for similarity is the {maximum of the entailment score in both directions}. Alternative definitions for these functions may also be used.}
In the special case of column headers across two tables, the score is (manually) set to either 0 or 1, indicating whether this corresponds to a meaningful join. 

Intuitively, we would like the support graph for an answer option to be connected, and to include nodes from the question, the answer option, and at least one table. Since each table row represents a coherent piece of information but cells within a row do not have any edges in $\mathcal{G}_{Q,T}$ (the same holds also for cells and the corresponding column headers), we use the notion of an augmented subgraph to capture the underlying table structure. Let $G = (V,E)$ be a subgraph of $\mathcal{G}_{Q,T}$. The \emph{augmented subgraph} $G^+$ is formed by adding to $G$ edges $(v_1,v_2)$ such that $v_1$ and $v_2$ are in $V$ and they correspond to either the same row (possibly the header row) of a table in $T$ or to a cell and the corresponding column header.

\begin{definition}
\label{def:support-graph:tableilp}
A \emph{support graph} $G = G(Q,T,\option)$ for a question $Q$, tables $T$, and an answer option $\option$ is a subgraph $(V,E)$ of $\mathcal{G}_{Q,T}$ with the following basic properties:
\begin{enumerate}
\item $V \cap \mathbf{a} =  \lbrace \option \rbrace, \ V \cap \mathbf{q} \neq \phi, \ V \cap \mathbf{t} \neq \phi$;
\item $w(e) > 0$ for all $e \in E$;
\item if $e \in E \cap (\mathbf{t} \times \mathbf{t})$ then there exists a corresponding $e' \in E \cap (\mathbf{h} \times \mathbf{h})$ involving the same columns; and
\item the augmented subgraph $G^+$ is connected.
\end{enumerate}
\end{definition}


A support graph thus connects the question constituents to a unique answer option through table cells and (optionally) table headers corresponding to the aligned cells. A given question and tables give rise to a large number of possible support graphs, and the role of the inference process will be to choose the ``best" one under a notion of \emph{desirable} support graphs developed next. We do this through a number of additional structural and semantic properties; the more properties the support graph satisfies, the more desirable it is.



\subsection{ILP Formulation}
\label{subsec:ilp:tableilp}
We model the above support graph search for QA as an ILP optimization problem, i.e., as maximizing a linear objective function over {a finite set of}
variables, subject to a set of linear inequality constraints (see Section~\ref{sec:optimization:background} for a premier on ILP formulation). 
A summary of the model is given below.\footnote{Details of the ILP model may be found in Appendix~\ref{appendix:optimization:details:tableilp}.}	

We note that  the ILP objective and constraints aren't tied to the particular domain of evaluation; they represent general properties that capture what constitutes a well supported answer for a given question.

\begin{wrapfigure}{r}{0.6\textwidth}
\centering
\setlength\tabcolsep{5pt}
\setlength\doublerulesep{\arrayrulewidth}
\small
\setlength\extrarowheight{-7pt}
\begin{tabular}{c|l}
\textsc{Element} & \multicolumn{1}{c}{\textsc{Description}} \\ 
\hline\hline
\bigstrut[t] $\tableVar$ & table $i$ \\ 
$\header $  & header of the $k$-th column of $i$-th table \\
$\tableCell $ & cell in row $j$ and column $k$ of $i$-th table \\ 
$\rowVar$ & row $j$ of $i$-th table \\ 
$\columnVar$ & column $k$ of $i$-th table \\ 
$\qCons$ & $\ell$-th lexical constituent of the question $Q$ \\
$\option$ & $m$-th answer option \\ 
\hline
\end{tabular}
\makeatletter\def\@captype{table}\makeatother
\caption{Notation for the ILP formulation. }
\label{table:summary-notation}
\end{wrapfigure}
Table~\ref{table:summary-notation} summarizes the notation for various elements of the problem, such as $\tableCell$ for cell $(j,k)$ of table $i$. All core variables in the ILP model are binary, i.e., have domain $\{0,1\}$. For each element, the model has a unary
variable capturing whether this element is part of the support graph $G$, i.e., it is ``active''. For instance, row $r_{ij}$ is active if at least one cell in row $j$ of table $i$ is in $G$. The model also has pairwise ``alignment'' variables, capturing edges of $\mathcal{G}_{Q,T}$. The alignment variable for an edge $e$ in $\mathcal{G}_{Q,T}$ is associated with the corresponding weight $w(e)$, and captures whether $e$ is included in $G$. To improve efficiency, we create a pairwise variable for $e$ only if $w(e)$ is larger than a certain threshold. These unary and pairwise variables are then used to define various types of constraints and preferences, as discussed next.

{ 
To make the definitions clear, we introduce the variables used in our optimization, which we will use later to define constraints explicitly. 
We define variables over each element by overloading $\xOne{.}$ or $\xTwo{.}{.}$ notation to refer to a binary variable on a single elements or their pair, respectively. Table~\ref{table:ilp-variables} contains the complete list of the variables, all of which are binary, i.e. they are defined on $\{0,1\}$ domain. 
The unary variables represent presence of a specific element in the support graph as a node. For example $\xOne{\tableVar} = 1$ if and only if the table $\tableVar$ is active. Similarly basic variables are defined between pairs of elements; e.g., $\xTwo{\tableCell}{\qCons}$ is a binary variable that takes value 1 if and only if the corresponding edge is present in the support graph, which can alternatively be referred to as an \emph{alignment} between cell $(j,k)$ of table $i$ and the $\ell$-th constituent of the question. 
\begin{wrapfigure}{r}{0.45\textwidth}
\centering 
\small
\setlength\extrarowheight{-16pt}
\begin{tabular}{r | L{30ex}} 
\multicolumn{2}{c}{\textsc{Basic Pairwise Activity Variables \bigstrut}} \\
\hline \hline 
$\xTwo{\tableCell}{\tableCellPrime} $ & \text{cell to cell} \\
$\xTwo{\tableCell}{\qCons} $ & cell to question constituent \\
$\xTwo{\header}{\qCons} $ & header to question constituent \\
$\xTwo{\tableCell}{\option}$ & cell to answer option \\ 
$\xTwo{\header}{\option}$ & header to answer option \\
$\xTwo{\columnVar}{\option}$ & column to answer option \\
$\xTwo{\tableVar}{\option}$ & table to answer option \\
$\xTwo{\columnVar}{\columnVarPrime}$ & column to column relation \\
\hline
\multicolumn{2}{c}{\textsc{High-level Unary Variables \bigstrut}}   \\
\hline \hline 
$\xOne{\tableVar}$ & active table \\ 
$\xOne{\rowVar}$  & active row \\
$\xOne{\columnVar} $ &  active column \\  
$\xOne{\header} $ & active column header  \\ 
$\xOne{\qCons}$ & active question constituent  \\ 
$ \xOne{\option} $ & active answer option \\ 
\hline
\end{tabular}
\makeatletter\def\@captype{table}\makeatother
\caption{Variables used for defining the optimization problem for \tableilp\ solver. All variables have domain $\{0,1\}$.
}
\label{table:ilp-variables}
\end{wrapfigure}

As previously mentioned, in practice we do not create all possible pairwise variables. Instead we choose the pairs which have the alignment score $w(e)$ exceeding a threshold. For example we create the pairwise variables   $\xTwo{\tableCell}{\tableCellPrimePrime}$ only if the score $w(\tableCell, \tableCellPrimePrime) \geq  \textsc{MinCellCellAlignment} $. 
\ifarxiv 
\footnote{An exhaustive list of the minimum alignment thresholds for creating pairwise variables is in Table~\ref{table:pairwise-thresholds} in the appendix. }   
\else
\footnote{An exhaustive list of the minimum alignment thresholds for creating pairwise variables is in Table~10
of \citep{tableilp2016:arxiv}. }   
\fi 

The objective function is a weighted linear sum of all the variables we instantiate for a given problem. 
\ifarxiv 
\footnote{The complete list of weights for the pairwise and unary variables are included in Table~\ref{table:objective-details} in the appendix.}
\else
\footnote{The complete list of weights for the pairwise and unary variables are included in Table~9
of \citep{tableilp2016:arxiv}.}
\fi 
There is a small set of auxiliary variables defined for linearizing complicated constraints, which will later introduce among constraints.

Constraints are a significant part of our model, which impose our desired behavior on the support graph. 
 However due to lack of space we only show a representative subset here.
\ifarxiv 
\footnote{The complete list of the constraints is explained in Table~\ref{table:constraints} in the appendix.}
\else
\footnote{The complete list of the constraints is explained in Table~13
of \citep{tableilp2016:arxiv}.}
\fi 

Some constraints relate variables to each other. The unary variables are defined through constraints that relate them to the pairwise basic variables. For example, for active row variable $\xOne{\tableVar}$, we ensure that it is active if and only if any cell in row $j$ is active:
$$
\xOne{\rowVar}  \geq \xTwo{\tableCell}{*}, \forall (\tableCell, *) \in \mathcal{R}_{ij}, \forall i, j, k, 
$$
where $\mathcal{R}_{ij}$ is collection of pairwise variables with one end in row  $j$ of table $i$. 



In what follows we outline the some of the important behaviors we expect from our model which come out with different combination of the active variables. 
}

\subsubsection{Basic Lookup}
Consider the following question:
\begin{myquote}
Which characteristic helps a fox find food?
(A) sense of smell (B) thick fur (C) long tail (D) pointed teeth
\end{myquote}
In order to answer such lookup-style questions, we generally seek a row with the highest aggregate alignment to question constituents. We achieve this by 
incorporating the question-table alignment variables with the alignment scores, $w(e)$, as coefficients and the active question constituents variable with a constant coefficient in the objective function. Since any additional question-table edge with a positive entailment score (even to irrelevant tables) in the support graph would result in an increase in the score, we disallow tables with alignments only to the question (or only to a choice) and add a small penalty for every table used in order to reduce noise in the support graph. We also limit the maximum number of alignments of a question constituent and {  table cells to prevent one constituent or cell from having a large influence on the objective function and thereby the solution: 
$$
\sum_{(*, \qCons) \in \mathcal{Q}_l } \xTwo{*}{\qCons} \leq \textsc{MaxAlignmentsPerQCons}, \forall l
$$
where $\mathcal{Q}_l$ is the set of all pairwise variables with one end in question constituent $\ell$. 
}

\subsubsection{Parallel Evidence}
For certain questions, evidence needs to be combined from multiple rows of a table. For example,
\begin{myquote}
Sleet, rain, snow, and hail are forms of (A) erosion (B) evaporation (C) groundwater (D) precipitation
\end{myquote}
To answer this question, we need to combine evidence from multiple table entries from the weather terms table, \textit{(term, type)}, namely (sleet, precipitation), (rain, precipitation), (snow, precipitation), and (hail, precipitation). To achieve this, we allow multiple active rows in the support graph. Similar to the basic constraints, we limit the maximum number of active rows per table and add a penalty for every active row to ensure only relevant rows are considered for reasoning{:
$$
\sum_{j} \xOne{\rowVar} \leq \textsc{MaxRowsPerTable}, \forall i
$$
}
To encourage only coherent parallel evidence within a single table,
we limit our support graph to always use the same columns across multiple rows within a table, i.e., every active row has {the active cells corresponding to the same  set of columns}.

\subsubsection{Evidence Chaining}
Questions requiring chaining of evidence from multiple tables, such as the example in Figure~\ref{fig:ny-example:tableilp}, are typically the most challenging in this domain.
Chaining can be viewed as performing a \emph{join} between two tables. We introduce alignments between cells across columns 
 in pairs of tables to allow for chaining of evidence. 
 To help minimize potential noise introduced by chaining irrelevant facts, we add a penalty for every inter-table alignment and also rely on the 0/1 weights of header-to-header edges
 to ensure only semantically meaningful table joins are considered.

\subsubsection{Semantic Relation Matching}
Our constraints so far have only looked at the content of the table cells, or the structure of the support graph, without explicitly considering the \emph{semantics} of the table schema. By using alignments between the question and column headers (i.e., type information), we exploit the table schema to prefer alignments to columns relevant to the ``topic'' of the question. In particular, for questions of the form ``which X $\ldots$'', we prefer answers that directly entail X or are connected to cells that entail X. However, this is not sufficient for questions such as:
\begin{myquote}
What is one way to change water \underline{from} a liquid \underline{to} a solid? (A) decrease the temperature (B) increase the temperature (C) decrease the mass (D) increase the mass
\end{myquote}
Even if we select the correct table, say $r_\text{change-init-fin}(c, i, f)$ that describes the initial and final states for a phase change event, both choice (A) and choice (B) would have the exact same score in the presence of table rows (increase temperature, solid, liquid) and (decrease temperature, liquid, solid). The table, however, does have the initial vs.\ final state structure. To capture this semantic structure, we annotate pairs of columns within certain tables with the semantic relationship present between them. In this example, we would annotate the phase change table with the relations: changeFrom$(c, i)$, changeTo$(c, f)$, and fromTo$(i, f)$. 

Given such semantic relations for table schemas, we can now impose a preference towards question-table alignments that respect these relations. We associate each semantic relation with a set of linguistic patterns describing how it might be expressed in natural language. \tableilp\ then uses these patterns to spot possible mentions of the relations in the question $Q$. We then add the soft constraint that for every pair of active columns in a table (with an annotated semantic relation) aligned to a pair of question constituents, there should be a valid expression of that relation in $Q$ between those constituents. In our example, we would match the relation fromTo(liquid, solid) in the table to ``liquid \underline{to a} solid'' in the question via the pattern ``X to a Y" associated with fromTo(X,Y), and thereby prefer aligning with the correct row (decrease temperature, liquid, solid).

\section{Evaluation}
\label{sec:experiments:tableilp}

We compare our approach to three existing methods, demonstrating that it outperforms {the best previous} structured approach~\citep{KBGSCE15} and produces a statistically significant improvement when used in combination with IR-based methods~\citep{CEKSTTK16}. For evaluations, we use a 2-core 2.5 GHz Amazon EC2 linux machine with 16 GB RAM.

\vspace{1ex}
\noindent \textbf{Question Set.}
{We use the same question set as \cite{CEKSTTK16}, which consists of all non-diagram multiple-choice questions from 12 years of the NY Regents 4th Grade Science exams.\footnote{These are the only publicly available state-level science exams. http://www.nysedregents.org/Grade4/Science/home.html} The set is split into 108 development questions and 129 hidden test questions based on the year they appeared in (6 years each).} All numbers reported below are for the hidden test set, except for question perturbation experiments which relied on the 108 development questions.

Test scores are reported as percentages. For each question, a solver gets a score of $1$ if it chooses the correct answer and $1/k$ if it reports a $k$-way tie that includes the correct answer. On the 129 test questions, a score difference of 9\% (or 7\%) is statistically significant at the 95\% (or 90\%, resp.) confidence interval based on the binomial exact test~\citep{Howell12}.

%

\vspace{1ex}
\noindent \textbf{Corpora.}
We work with three knowledge corpora:
\begin{enumerate}

\item Web Corpus: This corpus contains $5 \times 10^{10}$ tokens (280 GB of plain text) extracted from Web pages. {It was collected by Charles Clarke at the University of Waterloo, and has been {used previously by~\cite{Turney13}} and \cite{CEKSTTK16}.} We use it here to compute statistical co-occurrence values for the \salience\ solver.

\item Sentence Corpus {\citep{CEKSTTK16}}: This includes sentences from the Web corpus above, as well as around 80,000 sentences from various domain-targeted sources for elementary science:
a Regents study guide, CK12 textbooks (www.ck12.org), and web sentences with similar content as the course material.

\item Table Corpus (cf.~Section~\ref{subsec:tables}): This includes 65 tables totaling around 5,000 rows, designed based on the development set and study guides, as well as 4 Open IE-style~\citep{BCSBE07} automatically generated tables totaling around 2,600 rows.\footnote{{Table Corpus and the ILP model are available at allenai.org}.} 

\end{enumerate}



\subsection{Solvers}
\label{section:baslines:tableilp}

\vspace{1ex}
\noindent \textbf{\tableilp} (our approach). 
Given a question $Q$, we select the top 7 tables from the Table Corpus using the the standard TF-IDF score of $Q$ with tables treated as bag-of-words documents. For each selected table, we choose the 20 rows that overlap with $Q$ the most. This filtering improves efficiency and reduces noise. We then generate an ILP and solve it using the open source SCIP engine~\citep{Achterberg09}, returning the active answer option $\option$ from the optimal solution. To check for ties, we disable $\option$, re-solve the ILP, {and compare the score of the second-best answer, if any, with that of $a_m$}.


\vspace{1ex}
\noindent \textbf{MLN Solver} (structured inference baseline). 
We consider the current state-of-the-art structured reasoning method developed for this specific task by~\cite{KBGSCE15}. We compare against their best performing system, namely Praline, which uses Markov Logic Networks~\citep{RichardsonDo06} to (a) align lexical elements of the question with probabilistic first-order science rules and (b) to control inference. We use {the entire set of} 47,000 science rules from their original work, which were also derived from same domain-targeted sources as the ones used in our Sentence Corpus.

\vspace{1ex}
\noindent \textbf{\lucene\ Solver} (information retrieval baseline).
{We use the IR baseline by \cite{CEKSTTK16},} {which selects the answer option that has} the best matching sentence in a corpus. {Specifically,} for each answer option $a_i$, the \lucene\ solver sends $q + a_i$ as a query to a search engine (we use Lucene) on the Sentence Corpus, and returns the search engine's score for the top retrieved sentence $s$, where $s$ must have at least one non-stopword overlap with $q$, and at least one with $a_i$. The option with the highest Lucene score is returned as the answer.

\vspace{1ex}
\noindent \textbf{\salience\ Solver} (statistical co-occurrence baseline).
{We use the PMI-based approach by \cite{CEKSTTK16}}, {which selects} the answer option that most frequently co-occurs with the question words in a corpus. Specifically, it extracts unigrams, bigrams, trigrams, and skip-bigrams from the question and each answer option.
For a pair $(x,y)$ of $n$-grams, their pointwise mutual information (PMI)~\citep{ChurchHa89} in the corpus is defined as $\log \frac{p(x,y)}{p(x)p(y)}$ where $p(x,y)$ is the co-occurrence frequency of $x$ and $y$ (within some window) in the corpus.
The solver returns the answer option that has the largest average PMI in the Web Corpus, calculated over all pairs of question $n$-grams and answer option $n$-grams.

\subsection{Results}

We first compare the accuracy of our approach against the previous structured (MLN-based) reasoning solver. We also compare against \lucene(tables), an IR solver using table rows expressed as sentences, thus embodying an unstructured approach operating on the same knowledge as \tableilp.


\begin{wrapfigure}{r}{0.35\textwidth}
\centering
\small
\setlength\extrarowheight{-10pt}
\setlength\tabcolsep{10pt}
\setlength\doublerulesep{\arrayrulewidth}
\begin{tabular}{l|c}
Solver & Test Score (\%) \\
\hline\hline \bigstrut[t]
\praline & 47.5 \\
\lucene (tables) & 51.2 \\
\tableilp & {\bf 61.5} \\
\hline
\end{tabular}
\makeatletter\def\@captype{table}\makeatother
\caption{\tableilp~significantly outperforms both the prior MLN reasoner, and IR using identical knowledge as \tableilp}
\label{tab:eval}
\end{wrapfigure}

As Table \ref{tab:eval} shows, among the two structured inference approaches, \tableilp\ outperforms the MLN baseline by 14\%. {The preliminary ILP system reported by \cite{CEKSTTK16} achieves only a score of 43.8\% on this question set.} Further, given the same semi-structured knowledge (i.e., the Table Corpus), \tableilp\ is substantially (+10\%) better at exploiting the structure than the \lucene(tables) baseline, {which, as mentioned above,} uses the same data expressed as sentences.


\subsubsection{Complementary Strengths}
\begin{wrapfigure}{r}{0.5\textwidth}
\centering
\small
\setlength\tabcolsep{10pt}
\setlength\doublerulesep{\arrayrulewidth}
\setlength\extrarowheight{-10pt}
\begin{tabular}{l|c}
Solver & Test Score (\%) \\
\hline\hline \bigstrut[t]
\lucene & 58.5 \\
\salience & 60.7 \\
\tableilp & 61.5  \\
\tableilp\ + \lucene & 66.1 \\
\tableilp\ + \salience & 67.6 \\
\tableilp\ + \lucene + \salience & {\bf 69.0} \\
\hline
\end{tabular}
\makeatletter\def\@captype{table}\makeatother
\caption{Solver combination results}
\label{tab:combination}
\end{wrapfigure}
While their overall score is similar, \tableilp\ and IR-based methods clearly approach QA very differently. To assess whether \tableilp\ {adds} any new capabilities, we considered the 50 (out of 129) questions incorrectly answered by \salience\ solver (ignoring tied scores). On these unseen but arguably more difficult questions, \tableilp\ answered 27 questions correctly, achieving a score of 54\% compared to the random chance of 25\% for 4-way multiple-choice questions. Results with \lucene\ solver were similar: \tableilp\ scored 24.75 on the 52 questions incorrectly answered by \lucene\ (i.e., 47.6\% accuracy).

{This analysis highlights the} complementary strengths of these solvers. {Following \cite{CEKSTTK16}, we create an ensemble of \tableilp, \lucene, and \salience\ solvers, combining their answer predictions} using a simple Logistic Regression model trained on the development set. {This model uses} 4 features derived from each solver's score for each answer option, and 11 features derived from \tableilp's support graphs.
\ifarxiv 
\footnote{Details of the 11 features may be found in the Appendix B.}
\else
\footnote{Details of the 11 features may be found in the extended version~\citep{tableilp2016:arxiv}.}
\fi Table \ref{tab:combination} shows the results,
with the final combination at 69\% representing a significant improvement over individual solvers.

\subsubsection{ILP Solution Properties}
Table~\ref{tab:stats:tableilp} summarizes various ILP and support graph statistics for \tableilp, averaged across all test questions.

The optimization model has around 50 high-level constraints, which result, on average, in around 4000 inequalities over 1000 variables. Model creation, which includes 
computing pairwise entailment scores using WordNet, takes 1.9 seconds on average per question, and the resulting ILP is solved by the SCIP engine in 2.1 seconds (total for all four options), using around 1,300 LP iterations for each option.\footnote{Commercial ILP solvers (e.g., CPLEX, Gurobi) are much faster than the open-source SCIP solver we used for evaluations.} 

\begin{wrapfigure}{r}{0.55\textwidth}
\centering
\small
\setlength\tabcolsep{10pt}
\setlength\doublerulesep{\arrayrulewidth}
\setlength\extrarowheight{-12pt}
\begin{tabular}{llc}
Category & Quantity & Average \\
\hline\hline \bigstrut[t]
\multirow{3}{*}{ILP complexity} & \#variables & 1043.8 \\
& \#constraints & 4417.8 \\
& \#LP iterations & 1348.9 \\
\hline \bigstrut[t]
\multirow{2}{*}{Knowledge use} & \#rows & 2.3 \\
& \#tables & 1.3 \\
\hline \bigstrut[t]
\multirow{2}{*}{Timing stats} & model creation & 1.9 sec \\
& solving the ILP & 2.1 sec \\
\hline
\end{tabular}
\makeatletter\def\@captype{table}\makeatother
\caption{\tableilp\ statistics averaged across questions}
\label{tab:stats:tableilp}
\end{wrapfigure}
Thus, \tableilp\ takes only 4 seconds to answer a question using multiple rows across multiple tables (typically 140 rows in total), as compared to 17 seconds needed by the \praline\ solver for reasoning with four rules (one per answer option).

While the final support graph on this question set relies mostly on a single table to answer the question, it generally combines information from more than two rows (2.3 on average) for reasoning. This suggests parallel evidence is more frequently used on this dataset than evidence chaining.

\subsection{Ablation Study}

\begin{wrapfigure}{r}{0.5\textwidth}
\centering
\small
\setlength\tabcolsep{7pt}
\setlength\extrarowheight{-6pt}
\setlength\doublerulesep{\arrayrulewidth}
\begin{tabular}{ll|cc}
\multicolumn{2}{c|}{Solver} & Test Score (\%) \\
\hline\hline
\multicolumn{2}{l|}{\bigstrut[t] \tableilp} & 61.5  \\
  & No Multiple Row Inference & 51.0 \\
  & No Relation Matching & 55.6 \\
  \cline{2-3} \bigstrut[t] 
  & No Open IE Tables & 52.3  \\
  & No Lexical Entailment & 50.5  \\
\hline 
\end{tabular}
\makeatletter\def\@captype{table}\makeatother
\caption{Ablation results for \tableilp}
\label{tab:ablation:tableilp}
\end{wrapfigure}
To quantify the importance of various components of our system, we performed several ablation experiments, summarized in Table~\ref{tab:ablation:tableilp} and described next.

\vspace{1ex}
\noindent \textbf{No Multiple Row Inference}: We modify the ILP constraints to limit inference to a single row (and hence a single table), thereby disallowing parallel evidence and evidence chaining (Section~\ref{subsec:ilp:tableilp}). This drops the performance by 10.5\%, highlighting the importance of being able to combine evidence from multiple rows (which would correspond to multiple sentences in a corpus) from one or more tables.

\vspace{1ex}
\noindent \textbf{No Relation matching}: To assess the importance of considering the semantics of the table, we remove the requirement of matching the semantic relation present between columns of a table with its lexicalization in the question (Section~\ref{subsec:ilp:tableilp}). The 6\% drop indicates \tableilp\ relies strongly on the table semantics to ensure creating meaningful inferential chains.

\vspace{1ex}
\noindent \textbf{No Open IE tables}: To evaluate the impact of relatively unstructured knowledge from a large corpus, we removed the tables containing Open IE extractions (Section~\ref{subsec:qa-as-subgraph-selection}). The 9\% drop in the score shows that this knowledge is important and \tableilp\ is able to exploit it even though it has a very simple triple structure. This opens up the possibility of extending our approach to triples extracted from larger knowledge bases.

\vspace{1ex}
\noindent \textbf{No Lexical Entailment}: Finally, we test the effect of changing the alignment metric $w$ (Section~\ref{subsec:qa-as-subgraph-selection}) from WordNet based scores to a simple asymmetric word-overlap measured as $\mathit{score}(T, H) = \frac{|T \cap H|}{|H|}$. Relying on just word-matching results in an 11\% drop, which is consistent with our knowledge often being defined in terms of generalities.


\subsection{Question Perturbation}
\label{subsec:perturbation}

One desirable property of QA systems is robustness to simple variations of a question, \emph{especially when a variation would make the question arguably easier for humans}. 
%

To assess this, we {consider a simple, automated way to perturb} each 4-way multiple-choice question: (1) query Microsoft's Bing search engine (www.bing.com) with the question text and obtain the text snippet of the top 2,000 hits; (2) create a list of strings by chunking and tokenizing the results; (3) remove stop words and special characters, as well as any words (or their lemma) appearing in the question; (4) sort the remaining strings based on their frequency; and (5) replace the three incorrect answer options in the question with the most frequently occurring strings, thereby generating a new question. For instance:
\begin{quote}
{In New York State, the longest period of daylight occurs during which month? (A) \emph{eastern} (B) June (C) \emph{history} (D) \emph{years}}
\end{quote}

\begin{wrapfigure}{r}{0.65\textwidth}
\centering
\small
\setlength\tabcolsep{10pt}
\setlength\doublerulesep{\arrayrulewidth}
\setlength\extrarowheight{-10pt}
\begin{tabular}{l|cccc}
 & Original & \multicolumn{2}{c}{\% Drop with Perturbation} \\
 \cline{3-4}
Solver & Score (\%) & \bigstrut[t] absolute & relative \\
\hline\hline \bigstrut[t]
\lucene & 70.7 & 13.8 & 19.5 \\
\salience & 73.6 & 24.4 & 33.2 \\
\tableilp & 85.0 & {\bf 10.5} & {\bf 12.3} \\
\hline
\end{tabular}
\makeatletter\def\@captype{table}\makeatother
\caption{Drop in solver scores (on the development set, rather than the hidden test set) when questions are perturbed}
\label{tab:perturbation}
\end{wrapfigure}

%
As in this example, the perturbations (italicized) are often not even of the correct ``type'', {typically making them much easier for humans. They, however, still remain difficult for solvers.}

For each of the 108 development questions, we generate 10 new perturbed questions, using the 30 most frequently occurring words in step (5) above. {While this approach can introduce new answer options that should be considered correct as well, only 3\% of the questions in a random sample exhibited this behavior.} Table~\ref{tab:perturbation} shows the performance of various solvers on the resulting 1,080 perturbed questions. As one might expect, the \salience\ approach suffers the most at a 33\% relative drop. \tableilp's score drops as well (since answer type matching isn't perfect), but only by 12\%, attesting to its higher resilience to simple question variation.

\section{Summary and Discussion}
This chapter proposed a reasoning system for question answering on elementary-school science exams, using a semi-structured knowledge base. 
We formulate QA as an Integer Linear Program (ILP), that answers natural language questions using a semi-structured knowledge base derived from text, including questions requiring multi-step inference and a combination of multiple facts. On a dataset of real, unseen science questions, our system significantly outperforms (+14\%) the best previous attempt at structured reasoning for this task, which used Markov Logic Networks (MLNs). When combined with unstructured inference methods, the ILP system significantly boosts overall performance (+10\%).  Finally, we show our approach is substantially more robust to {a simple answer} perturbation compared to statistical correlation methods.

There are a few factors that limit the ideas discussed in this chapter. In particular,  the knowledge consumed by this system are in the form of curated tables;
constructing such knowledge is not always easy. In addition, not everything might be representable in that form. 
Another limitation stems from the nature of multi-step reasoning: larger number of reasoning steps could result in more brittle decisions. We study this issue in Chapter 8.

\chapter{QA as Subgraph Optimization over Semantic Abstractions}
\epigraph{ 
``It linked all the perplexed meanings \\ 
Into one perfect peace.''
}{--- \textup{Procter and Sullivan}, The Lost Chord, 1877}
\label{chapter:semantic:ilp}

\section{Overview}
In this chapter, we consider the multiple-choice setting where $Q$ is a question, $A$ is a set of answer candidates, and the knowledge required for answering $Q$ is available in the form of raw text $P$.\footnote{This chapter is based on the following publication: \cite{KKSR18}.} A major difference here with the previous chapter is that, the knowledge given a system is \textit{raw-text}, instead of of being represented in tabular format. 

We demonstrate that we can use existing NLP modules, such as semantic role labeling (SRL) systems with respect to multiple predicate types (verbs, prepositions, nominals, etc.), to derive multiple semantic views of the text and perform reasoning over these views to answer a variety of questions.

As an example, consider the following snippet of sports news text and an associated question:
\begin{framed}
\noindent\footnotesize\emph{\noindent$P$: Teams are under pressure after PSG purchased Neymar this season. \textbf{Chelsea purchased Morata}. The Spaniard looked like he was set for a move to Old Trafford for the majority of the summer only for Manchester United to sign Romelu Lukaku instead, paving the way for Morata to finally move to Chelsea for an initial \pounds 56m.
\\
\noindent
$Q$: Who did Chelsea purchase this season?  \\ 
$A$: \{\checkmark Alvaro Morata, Neymar, Romelu Lukaku \}
}
\end{framed}

Given the bold-faced text $P'$ in $P$, simple word-matching suffices to correctly answer $Q$. However, $P'$ could have stated the same information in many different ways. As paraphrases become more complex, they begin to involve more linguistic constructs such as coreference, punctuation, prepositions, and nominals. This makes understanding the text, and thus the QA task, more challenging.

\begin{figure}[ht]
    \centering
  \includegraphics[trim=-0.00cm 0cm 0cm -0.00cm, scale=0.55]{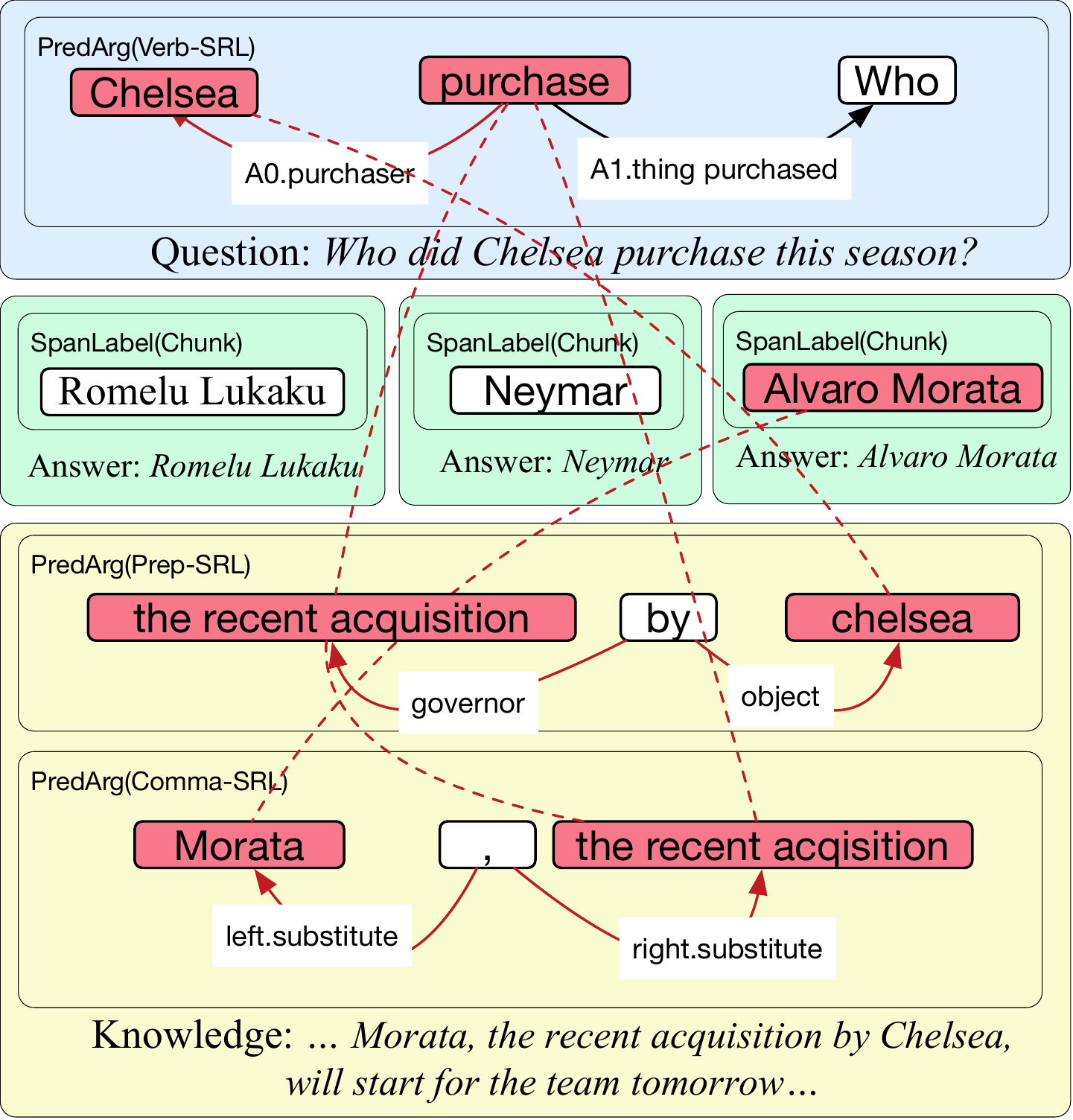}
  \caption{Depiction of \textilp\ reasoning for the example paragraph given in the text.
    Semantic abstractions of the question, answers, knowledge snippet are shown in different colored boxes (blue, green, and yellow, resp.). Red nodes and edges are the elements that are aligned (used) for supporting the correct answer. There are many other unaligned (unused) annotations associated with each piece of text that are omitted for clarity. 
    }
    \label{fig:morata-visualization}
\end{figure}

For instead, $P'$ could instead say \emph{Morata is the recent \textbf{acquisition} by Chelsea}. This simple looking transformation can be surprisingly confusing for highly successful systems such as \bidaf~\citep{SKFH16}, which produces the partially correct phrase \emph{``Neymar this season.\ Morata''}. On the other hand, one can still answer the question confidently by abstracting relevant parts of $Q$ and $P$, and connecting them appropriately. Specifically, a verb SRL frame for $Q$ would indicate that we seek the object of the verb \emph{purchase}, a nominal SRL frame for $P'$ would capture that the \emph{acquisition} was of Morata and was done by Chelsea, and textual similarity would align \emph{purchase} with \emph{acquisition}.


Similarly, suppose $P'$ instead said \emph{Morata\textbf{,} the recent acquisition \textbf{by} Chelsea\textbf{,} will start for the team tomorrow.} \bidaf\ now incorrectly chooses Neymar as the answer, presumably due to its proximity to the words \emph{purchased} and \emph{this season}. However, with the right abstractions, one could still arrive at the correct answer as depicted in Figure~\ref{fig:morata-visualization} for our proposed system, \textilp. This reasoning uses comma SRL to realize that the Morata is referring to the \emph{acquisition}, and a preposition SRL frame to capture that the acquisition was done \emph{by} Chelsea.

One can continue to make $P'$ more complex. For example, $P'$ could introduce the need for coreference resolution by phrasing the information as: \emph{\textbf{Chelsea} is hoping to have a great start this season by actively hunting for new players in the transfer period. Morata, the recent acquisition by the team, will start for \textbf{the team} tomorrow.} Nevertheless, with appropriate semantic abstractions of the text, the underlying reasoning remains relatively simple.

Given sufficiently large QA training data, one could conceivably perform end-to-end training (e.g., using a deep learning method) to address these linguistic challenges. However, existing large scale QA datasets such as SQuAD~\citep{RZLL16} often either have a limited linguistic richness or do not necessarily need reasoning to arrive at the answer~\citep{JiaLi17}. 
Consequently, the resulting models do not transfer easily to other domains.
For instance, the above mentioned BiDAF model trained on the SQuAD dataset performs substantially worse than a simple IR approach on our datasets. On the other hand, many of the QA collections in domains that require some form of reasoning, such as the science questions we use, are small  
(100s to 1000s of questions). This brings into question the viability of the aforementioned paradigm that attempts 
to learn everything from only the QA training data.

Towards the goal of effective structured reasoning in the presence of data sparsity, we propose to use a rich set of general-purpose, pre-trained NLP tools to create various \emph{semantic abstractions} of the raw text\footnote{This applies to all three inputs of the system: $Q$, $A$, and $P$.} in a domain independent fashion, as illustrated for an example in Figure~\ref{fig:morata-visualization}.
We represent these semantic abstractions as \emph{families of graphs}, where the family (e.g., trees, clusters, labeled predicate-argument graphs, etc.) is chosen to match the nature of the abstraction (e.g., parse tree, coreference sets, SRL frames, etc., respectively). The collection of semantic graphs is then augmented with inter- and intra-graph edges capturing lexical similarity (e.g., word-overlap score or word2vec distance).

This semantic graph representation allows us to formulate QA as the search for an optimal \emph{support graph}, a subgraph $G$ of the above augmented graph connecting (the semantic graphs of) $Q$ and $A$ via $P$. The reasoning used to answer the question is captured by a variety of requirements or constraints that $G$ must satisfy,
as well as a number of desired properties, encapsulating the ``correct'' reasoning, that makes $G$ preferable over other valid support graphs. For instance, a simple requirement is that $G$ must be connected and it must touch both $Q$ and $A$. Similarly, if $G$ includes a verb from an SRL frame, it is preferable to also include the corresponding subject. Finally, the resulting constrained optimization problem is formulated as an Integer Linear Program (ILP), and optimized using an off-the-shelf ILP solver (see Section~\ref{sec:optimization:background} for a review of ILP). 

This formalism may be viewed as a generalization of the systems introduced in the previous chapter: 
instead of operating over table rows (which are akin to labeled sequence graphs or predicate-argument graphs), we operate over a much richer class of semantic graphs. It can also be viewed as a generalization of the recent \tupleinf\ system~\citep{KhotSaCl17}, which converts $P$ into a particular kind of semantic abstraction, namely Open IE tuples~\citep{BCSBE07}.

This generalization to multiple semantic abstractions poses two key technical challenges: (a) unlike clean knowledge-bases (e.g., \cite{DWZX15}) used in many QA systems, abstractions generated from NLP tools (e.g., SRL) are noisy; and (b) even if perfect, using their output for QA requires delineating what information in $Q$, $A$, and $P$ is relevant for a given question, and what constitutes valid reasoning. The latter is especially challenging when combining information from diverse abstractions that, even though grounded in the same raw text, may not perfectly align. We address these challenges via our ILP formulation, by using our linguistic knowledge about the abstractions to design requirements and preferences for linking these abstractions.

We present a new QA system, \textilp,\footnote{Code available at: {https://github.com/allenai/semanticilp}} based on these ideas, and evaluate it on multiple-choice questions from two domains involving rich linguistic structure and reasoning: elementary and middle-school level science exams, and early-college level biology reading comprehension. Their data sparsity, as we show, limits the performance of state-of-the-art neural methods such as BiDAF~\citep{SKFH16}. \textilp, on the other hand, is able to successfully capitalize on existing general-purpose NLP tools in order to outperform existing baselines by 2\%-6\% on the science exams, leading to a new state of the art. It also generalizes well, as demonstrated by its strong performance on biology questions in the \processBank\ dataset~\citep{BSCLHHCM14}. Notably, while the best existing system for the latter relies on domain-specific structural annotation and question processing, \textilp\ needs neither.

\subsection{Related Work}
\label{subsec:related-work}
We provide a brief review of the related work, in additional to the discussion provided in Section~\ref{sec:intro:qa:review}.

Our formalism can be seen as an extension of the previous chapter. 
For instance, in our formalism, each table used by \tableilp\ 
can be viewed as a semantic frame and represented as a predicate-argument graph. The table-chaining rules used there are equivalent to the reasoning we define when combining two annotation components. Similarly, Open IE tuples used by~\citep{KhotSaCl17} can also be viewed as a predicate-argument structure.

One key abstraction we use is the predicate-argument structure provided by Semantic Role Labeling (SRL).
 Many SRL systems have been designed \citep{GildeaJu02,PunyakanokRoYi08} using linguistic resources such as FrameNet~\citep{BakerFiLo98}, PropBank~\citep{KingsburyPa02}, and NomBank~\citep{MRMSZYG04}. These systems are meant to convey high-level information about predicates (which can be a verb, a noun, etc.) and related elements in the text. The meaning of each predicate is conveyed by a frame, the schematic representations of a situation. Phrases with similar semantics ideally map to the same frame and roles. Our system also uses other NLP modules, such as for coreference resolution~\citep{LCPCSJ13} and dependency parsing~\citep{CUCSR15}.


While it appears simple to use SRL systems for QA~\citep{PalmerGiKi05}, this has found limited success~\citep{KaisserWe07,PizzatoMo08,MLBP11}. The challenges earlier approaches faced were due to making use of VerbSRL only, while QA depends on richer information, not only verb predicates and their arguments, along with some level of brittleness of all NLP systems. \cite{ShenLa07} have partly addressed the latter challenge with an inference framework that formulates the task as a bipartite matching problem over the assignment of semantic roles, and managed to slightly improve QA. In this work we address both these challenges and go beyond the limitations of using a single predicate SRL system; we make use of SRL abstractions that are based on verbs, nominals, prepositions, and comma predicates, as well as textual similarity. We then develop an inference framework capable of exploiting combinations of these multiple SRL (and other) views, thus operating over a more complete semantic representation of the text.

A key aspect of QA is handling textual variations, on which there has been prior work using dependency-parse transformations~\citep{PunyakanokRoYi04}. 
These approaches often define inference rules, which can generate new trees starting from a base tree. \cite{Bar-HaimDaBe15} and \cite{SSDF12} search over a space of a pre-defined set of text transformations (e.g., coreference substitutions, passive to active). Our work differs in that we consider a much wider range of textual variations by combining multiple abstractions, and make use of a more expressive inference framework.

\section{Knowledge Abstraction and Representation}
\label{subsec:kr}
We begin with our formalism for abstracting knowledge from text and representing it as a family of graphs, followed by specific instantiations of these abstractions using off-the-shelf NLP modules.

\subsection{Semantic Abstractions}
The pivotal ingredient of the abstraction is raw text. This representation is used for question $Q$, each answer option $A_i$ and the knowledge snippet $P$, which potentially contains the answer to the question. The KB for a given raw text, consists of the text itself, embellished with various \semanticGraph s attached to it, as depicted in Figure~\ref{fig:layers}.  
\begin{wrapfigure}{r}{0.4\textwidth}
    \centering\includegraphics[trim=0cm 0cm 0cm 0.0cm, width=0.4 \textwidth]{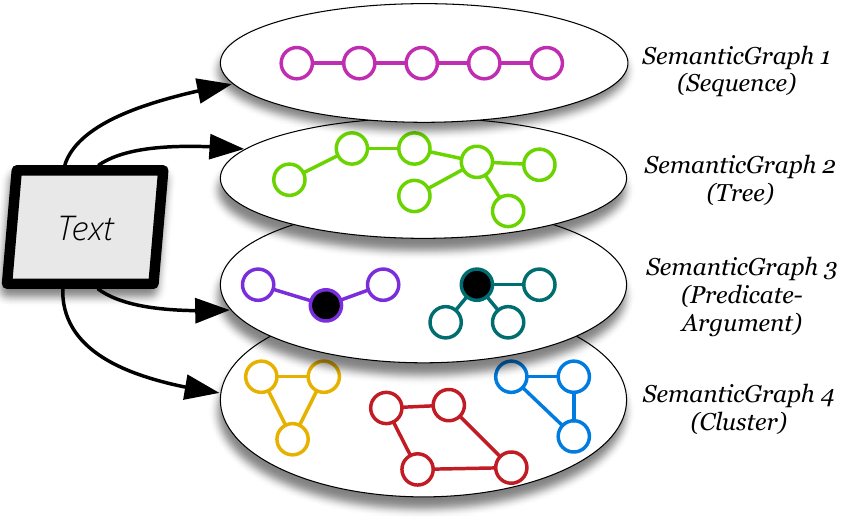}
    \caption{
    \small
    Knowledge Representation used in our formulation. Raw text is associated with a collection of \semanticGraph s, which convey certain information about the text.
    There are implicit similarity edges among the nodes of the connected components of the graphs, and from nodes to the corresponding raw-text spans. }
    \label{fig:layers}
\end{wrapfigure}

Each \semanticGraph\ is representable from a family of graphs. In principle there need not be any constraints on the permitted graph families; however for ease of representation we choose the graphs to belong to one of the 5 following families:
\sequence\ graphs represent labels for each token in the sentence.
\spann\ family represents labels for spans of the text.  
\treeView, is a tree representation of text spans. 
\clusterView\ family, contain spans of text in different groups. 
\predArgView\ family represents predicates and their arguments; in this view edges represent the connections between each single predicates and its arguments. Each \semanticGraph\ belongs to one of the graph families and its content is determined by the semantics of the information it represents and the text itself.

We define the knowledge more formally here. For a given paragraph, $T$, its representation $\know{T}$ consists of a set of semantic graphs  $\know{T} = \setOf{g_1, g_2, \hdots}$. 
We define $\nodes{g} = \setOf{c_i}$ and $\edges{g} = \setOf{(c_{i}, c_{j})} $ to be the set of nodes and edges of a given graph, respectively. 

\subsection{Semantic Graph Generators}

Having introduced a graph-based abstraction for knowledge and categorized it into a family of graphs, we now delineate the instantiations we used for each family. Many of the pre-trained extraction tools we use are available in \cogcompnlp.\footnote{Available at: http://github.com/CogComp/cogcomp-nlp}

\begin{itemize}[leftmargin=0.25cm]
    \item 
         \sequence\, or labels for sequence of tokens; for example \lemmaa\ and \pos\  \citep{RothZe98}. 
    \item
        \spann\ which can contains labels for spans of text; we instantiated \shallowP\ \citep{PunyakanokRo01}, \quantities\ \citep{RoyViRo15}, \ner\  \citep{RatinovRo09,RedmanSaRo16}).   
    \item 
        \treeView, a tree representation connecting spans of text as nodes; for this we used \dep\ of \cite{CUCSR15}. 
    \item 
        \clusterView, or spans of text clustered in groups. An example is \coref\ \citep{LPCCSJ11}. 
    \item 
        \predArgView; for this view we used \verbSRL\ and \nomSRL \citep{PunyakanokRoYi08,RothLa16}, \prepSRL\ \citep{SrikumarRo13}, \commaSRL\ \citep{ArivazhaganChRo16}.
\end{itemize}

Given \semanticGraph\ generators we have the question, answers and paragraph represented as a collection of graphs. Given the instance graphs, creating augmented graph will be done implicitly as an optimization problem in the next step.

\section{QA as Reasoning Over Semantic Graphs}

We introduce our treatment of QA as an optimal subgraph selection problem over knowledge. 
We treat question answering as the task of finding the best support in the knowledge snippet, for a given question and answer pair,
measured in terms of the strength of a ``support graph'' defined as follows.

The inputs to the QA system are, a question $\know{Q}$, the set of answers $\setOf{\know{A_i}}$ and given a knowledge snippet $\know{P}$.\footnote{For simplicity, from now on, we drop ``knowledge"; e.g., instead of saying ``question knowledge", we say ``question". } Given such representations, we will form a reasoning problem, which is formulated as an optimization problem, searching for a ``support graph" that connects the question nodes to a unique answer option through nodes and edges of a snippet. 

Define the instance graph $I = I(Q, \setOf{A_i}, P)$ as the union of knowledge graphs: $I \triangleq \know{Q} \cup \paran{\know{A_i}} \cup \know{P}$. Intuitively, we would like the support graph to be connected, and to include nodes from the question, the answer option, and the knowledge. Since the \semanticGraph\ is composed of many disjoint sub-graph, we define \emph{augmented graph} $I^+$ to 
model a bigger structure over the instance graphs $I$. 
Essentially we \textit{augment} the instance graph and weight the new edges. Define a scoring function $f:(v_1, v_2) $ labels pair of nodes $v_1$ and $v_2$ with an score which represents their phrase-level entailment or similarity. 




\begin{definition}
\label{def:augmented-graph}
An \emph{augmented graph} $I^+$, for a question $Q$, answers $\setOf{A_i}$ and knowledge $P$, is defined with the following properties: 
\begin{enumerate}
    \item Nodes: $\nodes{I^+} = \nodes{I(Q, \setOf{A_i}, P)}$
    \item Edges:\footnote{Define $\know{T_1} \otimes \know{T_2} \triangleq \bigcup_{\substack{ (g_1, g_2) \in \\ \know{T_1} \times \know{T_2}} } \nodes{g_1} \times \nodes{g_2}$, where $\nodes{g_1} \times \nodes{g_2} = \setOf{ (v, w); v \in \nodes{g_1}, w \in \nodes{g_2} }. $} 
    \begin{align*} 
     \edges{I^+} &= \edges{I} \cup  \know{Q} \otimes \know{P} \cup \brac{\cup_i \know{P} \otimes \know{A_i} }
    \end{align*} 
    \item Edge weights: for any $e \in I^+$: 
    \begin{itemize}
      \item If $e \notin I$, the edge connects two nodes in different connected components: 
      $$
      \forall e = (v_1, v_2) \notin I : w(e) = f(v_1, v_2) 
      $$
      \item If $e \in I$, the edge belongs to a connected component, and the edge weight information about the reliability of the \semanticGraph\ and semantics of the two nodes. 
      $$
        \forall g \in I, \forall e \in g: w(e) = f'(e, g)
      $$
    \end{itemize}
\end{enumerate}
\end{definition}

Next, we have to define \textit{support graphs}, the set of graphs that support the reasoning of a question. For this we will apply some structured constraints on the augmented graph. 

\begin{definition}
\label{def:support-graph}
A \emph{support graph} $G = G(Q,\setOf{A_i}, P)$ for a question $Q$, answer-options $\setOf{A_i}$ and paragraph $P$, is a subgraph $(V,E)$ of $I^+$ with the following properties:
\begin{enumerate}
\item $G$ is connected.
\item $G$ has intersection with the question, the knowledge, and exactly one answer candidate:\footnote{$\onlyOne$ here denotes the uniqueness quantifier, meaning ``there exists one and only one".} 
$$
G \cap \know{Q} \neq \emptyset,
\hspace{0.3cm}  G \cap \know{P} \neq \emptyset,
\hspace{0.3cm}  \onlyOne \; i: G \cap \know{A_i} \neq \emptyset 
$$
\item $G$ satisfies structural properties per each connected component, as summarized in Table~\ref{table:structural-properties}. 
\end{enumerate}
\end{definition}

\begin{table}
    \centering
    \small 
    \setlength\tabcolsep{10pt}
    \setlength\doublerulesep{\arrayrulewidth}
    \begin{tabular}{C{2.5cm}|L{5.1cm}}
        Sem. Graph &  Property \\
        \hline\hline \bigstrut[t]
         \predArgView & Use at least (a) a predicate and its argument, or (b) two arguments  \\ 
         \hline 
         \clusterView & Use at least two nodes  \\
         \hline 
         \treeView & Use two nodes with distance less than $k$ \\ 
         \hline 
         \spanLabelView & Use at least $k$ nodes   
    \end{tabular}
    \caption{Minimum requirements for using each family of graphs. Each graph connected component (e.g. a \predArgView\ frame, or a \coref\ chain) cannot be used unless the above-mentioned conditioned is satisfied. }
    \label{table:structural-properties}
\end{table}

Definition~\ref{def:support-graph} characterizes what we call a potential solution to a question. A given question and paragraph give rise to a large number of possible support graphs. We define the space of feasible support graphs as $\mathcal{G}$ (i.e., all the graphs that satisfy Definition~\ref{def:support-graph}, for a given $(Q, \setOf{A_i}, P)$).
To rank various feasible support graphs in such a large space, we define a scoring function $\score{G}$ as:
\begin{equation}
\label{eq:score-function}
\sum_{v \in \nodes{G}} w(v) + \sum_{e \in \edges{G}} w(e) - \sum_{c \in \mathcal{C} } w_c \, \eye{c \text{ is {violated}}}
\end{equation}
for some set of preferences (or soft-constraints) $\mathcal{C}$.
{When $c$ is violated, denoted by the indicator function $\eye{c \text{ is violated}}$ in Eq.~(\ref{eq:score-function}), we penalize the objective value by some fixed amount $w_c$.}
The second term is supposed to bring more sparsity to the desired solutions, just like how regularization terms act in machine learning models \citep{Natarajan95}. The first term is the sum of weights we defined when constructing the augmented-graph, and is supposed to give more weight to the models that have better and more reliable alignments between its nodes. 
The role of the inference process will be to choose the ``best" one under our notion of \emph{desirable} support graphs: 
\begin{equation}
\label{eq:objective}
G^* =\argmax_{G \in \mathcal{G}}
\end{equation}

\subsection{ILP Formulation}
\label{subsec:ilp}

Our QA system, \textilp, models the above support graph search of Eq.~(\ref{eq:objective}) as an ILP optimization problem, i.e., as maximizing a linear objective function over a finite set of variables, subject to a set of linear inequality constraints. A summary of the model is given below. 

The augmented graph is not explicitly created; instead, it is implicitly created. The nodes and edges of the augmented graph are encoded as a set of binary variables. The value of the binary variables reflects whether a node or an edge is used in the optimal graph $G^*$. The properties listed in Table~\ref{table:structural-properties} are implemented as weighted linear constraints using the variables defined for the nodes and edges. 

As mentioned, edge weights in the augmented graph come from a function, $f$, which captures (soft) phrasal entailment between question and paragraph nodes, or paragraph and answer nodes, to account for lexical variability.
In our evaluations, we use two types of $f$. (a) Similar to \cite{KKSCER16}, we use a WordNet-based \citep{Miller95} function to score word-to-word alignments, and use this as a building block to compute a phrase-level alignment score as the weighted sum of word-level alignment scores.
Word-level scores are computed using WordNet's hypernym and synonym relations, and weighted using relevant word-sense frequency.
$f$ for similarity (as opposed to entailment) is taken to be the average of the entailment scores in both directions. (b) For longer phrasal alignments (e.g., when aligning phrasal verbs) we use the Paragram system of \cite{WBGLR15}.

The final optimization is done on Eq.~(\ref{eq:score-function}). The first part of the objective is the sum of the weights of the sub-graph, which is what an ILP does, since the nodes and edges are modeled as variables in the ILP. The second part of Eq.~(\ref{eq:score-function}) contains a set of preferences $\mathcal{C}$, summarized in Table~\ref{tab:preferences}, meant to apply \textit{soft} structural properties that partly dependant on the knowledge instantiation. These preferences are soft in the sense that they are applied with a weight to the overall scoring function (as compare to a hard constraint). For each preference function $c$  there is an associated binary or integer variable with weight $w_c$, and we create appropriate constraints to simulate the corresponding behavior. 

\begin{table}
    \small
    \begin{framed}
    \noindent - Number of sentences used is more than $k$\\ 
    - Active edges connected to each chunk of the answer option, more than $k$ \\ 
    - More than k chunks in the active answer-option \\ 
    - More than k edges to each question constituent  \\ 
    - Number of active question-terms  \\ 
    - If using \predArgView of $\know{Q}$, at least an argument should be used  \\ 
    - If using \predArgView(\verbSRL) of $\know{Q}$, at least one predicate should be used.
    \end{framed}
    \caption{The set of preferences functions in the objective. }
    \label{tab:preferences}
\end{table}

We note that the ILP objective and constraints aren't tied to the particular domain of evaluation; they represent general properties that capture what constitutes a well supported answer for a given question.

\section{Empirical Evaluation}
\label{sec:experiments:semanticilp}

We evaluate on two domains that differ in the nature of the supporting text (concatenated individual sentences vs.\ a coherent paragraph), the underlying reasoning, and the way questions are framed. We show that \textilp\ outperforms a variety of baselines, including retrieval-based methods, neural-networks, structured systems, and the current best system for each domain. These datasets and systems are described next, followed by results.

\subsection{Question Sets}

For the first domain, we have a collection of question sets containing elementary-level science questions from standardized tests \citep{CEKSTTK16,KhotSaCl17}. Specifically, \regentsFourth\ contains all non-diagram multiple choice questions from 6 years of NY Regents 4th grade science exams (127 train questions, 129 test). \regentsEighth\ similarly contains 8th grade questions (147 train, 144 test). The corresponding expanded datasets are \publicFourth\ (432 train, 339 test) and \publicEighth\ (293 train, 282 test).\footnote{AI2 Science Questions V1 at http://data.allenai.org/ai2-science-questions}

For the second domain, we use the \processBank\footnote{https://nlp.stanford.edu/software/bioprocess} dataset for the reading comprehension task proposed by \cite{BSCLHHCM14}. It contains paragraphs about biological processes and two-way multiple choice questions about them. We used a broad subset of this dataset that asks about events or about an argument that depends on another event or argument.\footnote{These are referred to as ``dependency questions" by \cite{BSCLHHCM14}, and cover around 70\% of all questions.}. The resulting dataset has 293 train and 109 test questions, based on 147 biology paragraphs.

Test scores are reported as percentages. For each question, a system gets a score of $1$ if it chooses the correct answer, $1/k$ if it reports a $k$-way tie that includes the correct answer, and $0$ otherwise.

\subsection{Question Answering Systems}
\label{section:baslines}
We consider a variety of baselines, including the best system for each domain.

\textbf{\lucene} (information retrieval baseline).
We use the IR solver from \cite{CEKSTTK16}, which selects the answer option that has the best matching sentence in a corpus. The sentence is forced to have a non-stopword overlap with both $q$ and $a$.

\textbf{\textilp} (our approach). 
Given the input instance (question, answer options, and a paragraph), we invoke various NLP modules to extract semantic graphs. We then generate an ILP and solve it using the open source SCIP engine~\citep{Achterberg09}, returning the active answer option $\option$ from the optimal solution found. To check for ties, we disable $\option$, re-solve the ILP, and compare the score of the second-best answer, if any, with that of the best score.

For the science question sets, where we don't have any paragraphs attached to each question, we create a passage by using the above \lucene\ solver to retrieve scored sentences for each answer option and then combining the top 8 unique sentences (across all answer options) to form a paragraph.

{
While the sub-graph optimization can be done over the entire augmented graph in one shot, 
our current implementation uses multiple simplified solvers, each performing reasoning over augmented graphs for a commonly occurring annotator combination, as listed in Table~\ref{tab:combinations}. For all of these annotator combinations, we let the representation of the answers be $\know{A}=\{\shallowP, \tokens\}$. Importantly, our choice of working with a few annotator combinations is mainly for simplicity of implementation and suffices to demonstrate that reasoning over even just two annotators at a time can be surprisingly powerful. 
There is no fundamental limitation in implementing \textilp\ using one single optimization problem as stated in Eq.~(\ref{eq:objective}). 

\begin{table}
    \small 
  \setlength\tabcolsep{2pt}
  \setlength\doublerulesep{\arrayrulewidth}
    \centering
    \begin{tabular}{c|l}
        Combination & \multicolumn{1}{c}{Representation}   \\
        \hline\hline \bigstrut[t]
         \multirow{2}{*}{
            Comb-1
         } 
         
         & $\know{Q}=\{\shallowP, \tokens\}$  \\ 
           &  $\know{P}=\{\shallowP, \tokens, \dep\}$  \\ 
           \hline 
         \multirow{2}{*}{
            Comb-2
         } &$\know{Q}=\setOf{\verbSRL, \shallowP}$    \\ 
          & $\know{P}=\setOf{ \verbSRL }$ \\ 
         \hline 
          \multirow{2}{*}{
            Comb-3 
          } & $\know{Q}=\setOf{\verbSRL, \shallowP}$    \\  
          &   $\know{P}=\{\verbSRL, \coref \}$  \\  
         \hline 
         \multirow{2}{*}{
            Comb-4
         } & $\know{Q}=\setOf{ \verbSRL,\shallowP}$    \\ 
          &  $\know{P}=\setOf{\commaSRL }$ \\ 
         \hline 
         \multirow{2}{*}{
            Comb-5 
         } & $\know{Q}=\setOf{\verbSRL,\shallowP }$  \\ 
          & $\know{P}=\setOf{ \prepSRL }$ \\ 
    \end{tabular}
    \caption{The semantic annotator combinations used in our implementation of \textilp. 
    }
    \label{tab:combinations}
\end{table}

Each simplified solver associated with an annotator combination in Table~\ref{tab:combinations} produces a confidence score for each answer option. We create an \emph{ensemble} of these solvers as a linear combination of these scores, with weights trained using the union of training data from all questions sets. 
}

\textbf{\bidaf} (neural network baseline).
We use the recent deep learning reading comprehension model of \cite{SKFH16}, which is one of the top performing systems on the SQuAD dataset and has been shown to generalize to another domain as well~\citep{MinSeHa17}. Since \bidaf\ was designed for fill-in-the-blank style questions, we follow the variation used by \cite{KSSCFH17} to apply it to our multiple-choice setting. Specifically, we compare the predicted answer span to each answer candidate and report the one with the highest similarity.

We use two variants: the original system, \bidaf, pre-trained on 100,000+ SQuAD questions, as well as an extended version, \bidafTrained, obtained by performing continuous training to fine-tune the SQuAD-trained parameters using our (smaller) training sets. For the latter, we convert multiple-choice questions into reading comprehension questions by generating all possible text-spans within sentences, with token-length at most \textit{correct answer length + 2}, and choose the ones with the highest similarity score with the correct answer. We use the \allennlp\ re-implementation of \bidaf\footnote{Available at: https://github.com/allenai/allennlp}, train it on SQuAD, followed  by training it on our dataset. We tried different variations (epochs and learning rates) and selected the model which gives the best average score across all the datasets. As we will see, the variant that was further trained on our data often gives better results. 

\textbf{\tupleinf} (semi-structured inference baseline).
Recently proposed by \cite{KhotSaCl17}, this is a state-of-the-art system designed for science questions. It uses Open IE~\citep{BCSBE07} tuples derived from the text as the knowledge representation, and performs reasoning over it via an ILP. It has access to a large knowledge base of Open IE tuples, and exploits redundancy to overcome challenges introduced by noise and linguistic variability.

\textbf{\proread} and \textbf{\syntprox}. 
\proread is a specialized and best performing system on the \processBank\ question set. \cite{BSCLHHCM14} annotated the training data with events and event relations, and trained a system to extract the process structure. Given a question, \proread\ converts it into a query (using regular expression patterns and keywords) and executes it on the process structure as the knowledge base. Its reliance on a question-dependent query generator and on a process structure extractor makes it difficult to apply to other domains.

\syntprox\ is another solver suggested by \citep{BSCLHHCM14}. It aligns content word lemmas in both the question and the answer against the paragraph, and select the answer tokens that are closer to the aligned tokens of the questions. The distance is measured using dependency tree edges. To support multiple sentences they connect roots of adjacent sentences with bidirectional edges.

\subsection{Experimental Results}
We evaluate various QA systems on datasets from the two domains. The results are summarized below, followed by some some insights into \textilp's behavior and an error analysis. 
\paragraph{Science Exams.}
The results of experimenting on different grades' science exams are summarized in Table~\ref{table:science}, which shows the exam scores as a percentage.  
The table demonstrates that \textilp\ consistently outperforms the best baselines in each case by 2\%-6\%. 

Further, there is no absolute winner among the baselines; while \lucene\ is good on the 8th grade questions, \tupleinf\ and \bidafTrained\ are better on 4th grade questions. This highlights the differing nature of questions for different grades. 
\begin{table}[htb]
  \centering
  \small
  \setlength\tabcolsep{2pt}
  \setlength\doublerulesep{\arrayrulewidth}
  \begin{tabular}{l|cccc|c}
  Dataset & \bidaf & \bidafTrained & \lucene & \tupleinf & \textilp \\ 
  \hline\hline \bigstrut[t]
  \regentsFourth  & 56.3 & 53.1 & 59.3 & \emph{61.4} & {\bf 67.6} \\ 
  \publicFourth   & 50.7 & \emph{57.4} & 54.9 & 56.1 & {\bf 59.7} \\ 
  \regentsEighth  & 53.5 & 62.8 & \emph{64.2} & 61.3 & {\bf 66.0} \\ 
  \publicEighth   & 47.7 & 51.9 & \emph{52.8} & 51.6 & {\bf 55.9} \\ 
  \hline
  \end{tabular}
  \caption{Science test scores as a percentage. On elementary level science exams, \textilp\ consistently outperforms baselines. In each row, the best score is in \textbf{bold} and the best baseline is \emph{italicized}.  }
  \label{table:science}
\end{table}

\paragraph{Biology Exam.} 
The results on the \processBank\ dataset are summarized in Table~\ref{table:bio}. While \textilp's performance is substantially better than most baselines and close to that of \proread, it is important to note that this latter baseline enjoys additional supervision of domain-specific event annotations. This, unlike our other relatively general baselines, makes it limited to this dataset,
which is also why we don't include it in Table~\ref{table:science}.

We evaluate \lucene\ on this reading comprehension dataset by creating an ElasticSearch index, containing the sentences of the knowledge paragraphs. 

\begin{table}[htb]
  \centering
  \small
  \setlength\tabcolsep{2pt}
  \setlength\doublerulesep{\arrayrulewidth}
  \begin{tabular}{c|cccc|c}
  \proread &  \syntprox & \lucene &  \bidaf & \bidafTrained & \textilp \\ 
  \hline\hline \bigstrut[t]
  68.1  &  61.9 & 63.8 & 58.7 & 61.3 & {\bf 67.9 }\\ 
  \hline
  \end{tabular}
  \caption{Biology test scores as a percentage. \textilp\ outperforms various baselines on the \processBank\ dataset and roughly matches the specialized best method.}
  \label{table:bio}
\end{table}

\subsection{Error and Timing Analysis}
For some insight into the results, we include a brief analysis of our system's output compared to that of other systems.

We identify a few main reasons for \textilp's errors.
Not surprisingly, some mistakes
(see the appendix figure of~\cite{KKSR18} for an example)
 can be traced back to failures in generating proper annotation (\semanticGraph). Improvement in SRL modules or redundancy can help address this. 
Some mistakes are from the current ILP model not supporting the ideal reasoning, i.e., the requisite knowledge exists in the annotations, but the reasoning fails to exploit it. 
Another group of mistakes is due to the complexity of the sentences, and the system lacking a way to represent the underlying phenomena with our current annotators. 

A weakness (that doesn't seem to be particular to our solver) is reliance on explicit mentions. If there is a meaning indirectly implied by the context and our annotators are not able to capture it, our solver will miss such questions. There will be more room for improvement on such questions with the development of discourse analysis systems.   

When solving the questions that don't have an attached paragraph, relevant sentences need to be fetched from a corpus. A subset of mistakes on this dataset occurs because the extracted knowledge does not contain the correct answer.

\subsubsection{ILP Solution Properties.}


Our system is implemented using many constraints, requires using many linear inequalities which get instantiated on each input instanced, hence there are a different number of variables and inequalities for each input instance. There is an overhead time for pre-processing an input instance, and convert it into an instance graph. Here in the timing analysis we provide we ignore the annotation time, as it is done by black-boxes outside our solver.

Table~\ref{tab:stats} summarizes various ILP and support graph statistics for \textilp, averaged across \processBank\ questions. Next to \textilp\ we have included numbers from \tableilp\ which has similar implementation machinery, but on a very different representation. While the size of the model is a function of the input instance, on average, \textilp\ tends to have a bigger model (number of constraints and variables). The model creation time is significantly time-consuming in \textilp\ as involves many graph traversal operations and jumps between nodes and edges. 
We also providing times statistics for \tupleinf\, which takes roughly half the time of \tableilp, which means that it is faster than \textilp.

\begin{table}[htb]
\centering
\footnotesize
\setlength\tabcolsep{5pt}
\setlength\doublerulesep{\arrayrulewidth}
\begin{tabular}{llccc}
\multirow{2}{*}{Category} & 
\multirow{2}{*}{Quantity} & 
Avg. & Avg. & Avg.  \\
& & { \tiny (\textilp) } & { \tiny (\tableilp) } & { \tiny (\tupleinf) } \\ 
\hline\hline \bigstrut[t]
\multirow{3}{*}{ILP complexity} & \#variables & 2254.9 & 1043.8 & 1691.0 \\
& \#constraints & 4518.7 & 4417.8 & 4440.0 \\
\hline \bigstrut[t]
\multirow{2}{*}{Timing stats} & model creation & 5.3 sec  & 1.9 sec & 1.7 sec \\
& solving the ILP & 1.8 sec   & 2.1 sec &  0.3 sec\\
\hline
\end{tabular}
\caption{\textilp\ statistics averaged across questions, as compared to  \tableilp\ and \tupleinf\ statistics.
}
\label{tab:stats}
\end{table}

\subsection{Ablation Study}
In order to better understand the results, we ablate the contribution of different annotation combinations, where we drop different combination from the ensemble model. We retrain the ensemble, after dropping each combination.

The results are summarized in Table~\ref{tab:ablation}. While Comb-1
seems to be important for science tests, it has limited contribution to the biology tests. On 8th grade exams, the \verbSRL\ and \commaSRL-based alignments provide high value. Structured combinations (e.g., \verbSRL-based alignments) are generally more important for the biology domain. 

\begin{table}[htb]
\centering
\footnotesize
\setlength\tabcolsep{5pt}
\setlength\doublerulesep{\arrayrulewidth}
\begin{tabular}{lC{2.2cm}C{2.2cm}}
 & \publicEighth & \processBank  \\ 
\hline\hline
Full \textilp\ & 
55.9 &	67.9 \\ 
\hline 
no Comb-1  
&
 -3.1 &	-1.8 \\ 
no Comb-2 
& -2.0 &	-4.6 \\ 
no Comb-3  
& -0.6 &	-1.8 \\ 
no Comb-4 
& -3.1 &	-1.8 \\ 
no Comb-5  
& -0.1 &	-5.1 \\ 
\hline
\end{tabular}
\caption{Ablation study of \textilp\ components on various datasets. The first row shows the overall test score of the full system, while other rows report the change in the score as a result of dropping an individual combination. The combinations are listed in Table~\ref{tab:combinations}.}
\label{tab:ablation}
\end{table}

\paragraph{Complementarity to \lucene.}
\begin{wrapfigure}{r}{0pt}
    \centering
    \includegraphics[scale=0.75]{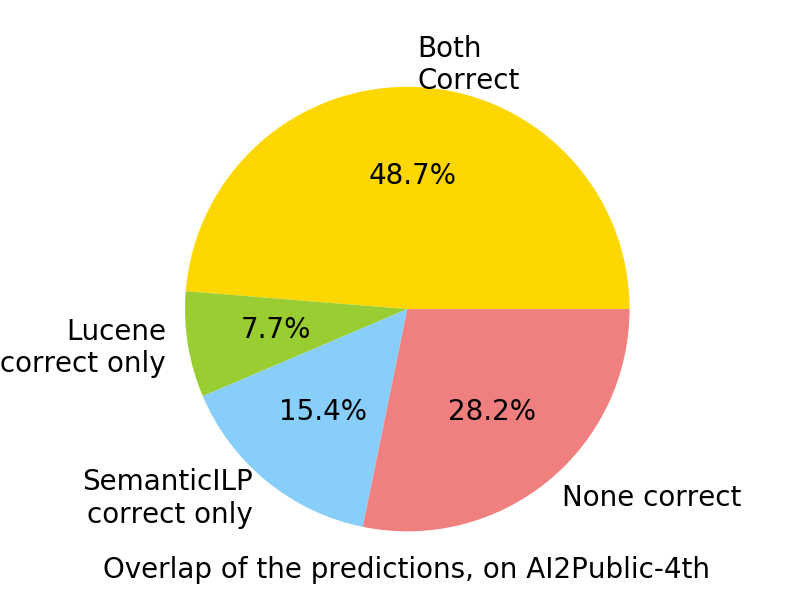}
    \caption{Overlap of the predictions of \textilp\ and \lucene\ on 50 randomly-chosen questions from \publicFourth.
    }
    \label{fig:piechart}
\end{wrapfigure}
Given that in the science domain the input snippets fed to \textilp\ are retrieved through a process similar to the \lucene\ solver, one might naturally expect some similarity in the predictions. 
The pie-chart in Figure~\ref{fig:piechart} shows the overlap between mistakes and correct predictions of \textilp\ and \lucene\ on 50 randomly chosen training questions from \publicFourth. While there is substantial overlap in questions that both answer correctly (the yellow slice) and both miss (the red slice), there is also a significant number of questions solved by \textilp\ but not \lucene\ (the blue slice), almost twice as much as the questions solved by \lucene but not \textilp\ (the green slice).

\begin{figure*}
    \centering
    \hspace{0.8cm}
    \includegraphics[scale=0.33]{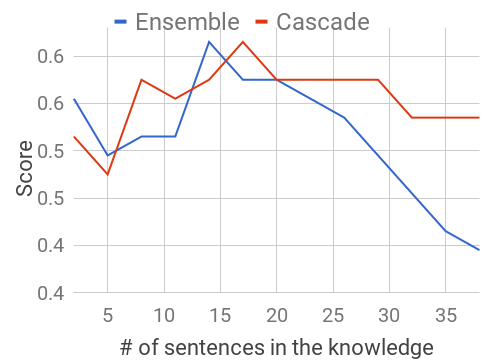}
    \hspace{0.8cm}
    \includegraphics[scale=0.33]{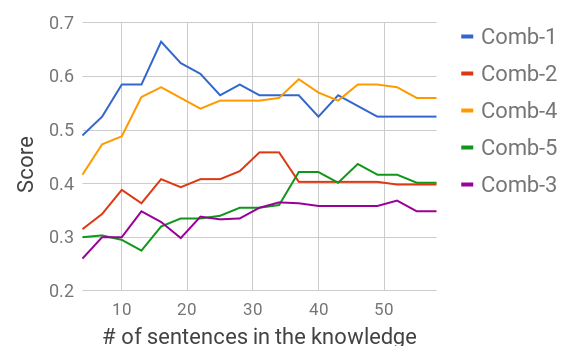}
    \caption{Performance change for varying knowledge length.}
    \label{fig:knowledge-experiment}
\end{figure*}

\subsubsection{Cascade Solvers.}
In Tables~\ref{table:science} and~\ref{table:bio}, we presented one \textit{single} instance of \textilp\ with state-of-art results on multiple datasets, where the solver was an ensemble of semantic combinations (presented in Table~\ref{tab:combinations}). Here we show a simpler approach that achieves stronger results on individual datasets, at the cost of losing a little generalization across domains. 
Specifically, we create two ``cascades'' (i.e., decision lists) of combinations, where the ordering of combinations in the cascade is determined by the training set precision of the simplified solver representing an annotator combination (combinations with higher precision appear earlier). One cascade solver targets science exams and the other the biology dataset.

The results are reported in Table~\ref{table:cascade}. On the 8th grade data, the cascade solver created for science test achieves higher scores than the generic ensemble solver. Similarly, the cascade solver on the biology domain outperforms the ensemble solver on the \processBank\ dataset.

\begin{wrapfigure}{r}{0.6\textwidth}
  \centering
  \small
  \setlength\tabcolsep{5pt}
  \setlength\doublerulesep{\arrayrulewidth}
\setlength\extrarowheight{-8pt}
  \begin{tabular}{cl|ccc}
  \multicolumn{2}{c|}{Dataset} & Ensemble & \specialcell{Cascade \\(Science)} & \specialcell{Cascade \\(Biology)} \\ 
  \hline\hline \bigstrut[t]
   \parbox[t]{2ex}{\multirow{4}{*}{\rotatebox[origin=c]{90}{Science}}} 
   &  \regentsFourth  & {\bf 67.6} & 64.7 & 63.1 \\ 
   & \publicFourth   & {\bf 59.7} & 56.7 & 55.7\\ 
   & \regentsEighth  & 66.0 & {\bf 69.4} & 60.3 \\ 
   & \publicEighth   & 55.9 & {\bf 56.5 } & 54.3 \\ 
  \hline 
    & \processBank   & 67.9 & 59.6 & {\bf 68.8 } \\ 
  \hline
  \end{tabular}
    \makeatletter\def\@captype{table}\makeatother
  \caption{
    Comparison of test scores of \textilp\ using a generic ensemble vs.\ domain-targeted cascades of annotation combinations. 
  }
  \label{table:cascade}
\end{wrapfigure}

\paragraph{Effect of Varying Knowledge Length.}
We analyze the performance of the system as a function of the length of the paragraph fed into \textilp, for 50 randomly selected training questions from the \regentsFourth\ set. Figure~\ref{fig:knowledge-experiment} (left) shows the overall system, for two combinations introduced earlier, as a function of knowledge length, counted as the number of sentences in the paragraph.

As expected, the solver improves with more sentences, until around 12-15 sentences, after which it starts to worsen with the addition of more irrelevant knowledge. While the cascade combinations did not show much generalization across domains, they have the advantage of a smaller drop when adding irrelevant knowledge compared to the ensemble solver. This can be explained by the simplicity of cascading and minimal training compared to the ensemble of annotation combinations. 

Figure~\ref{fig:knowledge-experiment} (right) shows the performance of individual combinations as a function of knowledge length.
It is worth highlighting that while Comb-1
(blue) often achieves higher coverage and good scores in simple paragraphs (e.g., science exams), it is highly sensitive to knowledge length. On the other hand,  highly-constrained combinations have a more consistent performance with increasing knowledge length, at the cost of lower coverage. 

\section{Summary and Discussion}
This chapter extends our abductive reasoning system from Chapter 3 to consume raw text as input knowledge. This is the first system to successfully use a wide range of semantic abstractions to perform a high-level NLP task like Question Answering. 
The approach is especially suitable for domains that require reasoning over a diverse set of linguistic constructs but have limited training data. To address these challenges, we present the first system, to the best of our knowledge, that reasons over a wide range of semantic abstractions of the text, which are derived using off-the-shelf, general-purpose, pre-trained natural language modules such as semantic role labelers. Representing multiple abstractions as a family of graphs, we translate question answering (QA) into a search for an optimal subgraph that satisfies certain global and local properties. This formulation generalizes several prior structured QA systems. Our system, \textilp, demonstrates strong performance on two domains simultaneously. In particular, on a collection of challenging science QA datasets, it outperforms various state-of-the-art approaches, including neural models, broad coverage information retrieval, and specialized techniques using structured knowledge bases, by 2\%-6\%. 

A key limitation of the system here is that its abstractions are mostly extracted from explicit mentions of in a given text. However, a major portion of our understanding come is only implied from text (not directly mention). We propose a challenge dataset for such questions (limited to the \emph{temporal} domain) in Chapter 7. Additionally, the two systems discussed in Chapter 3 and here, lack explicit explicit attention mechanism to the content of the questions. We study this topic in Chapter 5.


\chapter{Learning Essential Terms in Questions}
\epigraph{ 
``The trouble with Artificial Intelligence is that computers don't give a damn-or so I will argue by considering the special case of understanding natural language.''
}{--- \textup{John Haugeland}, 1979}

\newcommand\T{\rule{0pt}{2.6ex}}       
\newcommand\B{\rule[-1.2ex]{0pt}{0pt}} 

\section{Overview}

Many of today's QA systems 
often struggle with seemingly simple questions because they are unable to reliably identify which question words are redundant, irrelevant, or even intentionally distracting.\footnote{This chapter is based on the following publication: \cite{KKSR17}} This reduces the systems' precision and results in questionable ``reasoning'' even when the correct answer is selected among the given alternatives. The variability of subject domain and question style makes identifying essential question words challenging. Further, essentiality is context dependent---a word like `animals' can be critical for one question and distracting for another.
%
%
Consider the following example:
%
\FrameSep2pt
\begin{framed}
\small 
\noindent 
One way animals usually respond to a sudden drop in temperature is by (A) sweating (B) shivering (C) blinking (D) salivating. 
\end{framed}
%
\noindent 
The system we discussed in Chapter 3, 
\tableilp~\citep{KKSCER16}, which performs reasoning by aligning the question to semi-structured knowledge, aligns only the word `animals' when answering this question. Not surprisingly, it chooses an incorrect answer. The issue is that it does not recognize that ``drop in temperature" is an \ess\ aspect of the question. 

\begin{wrapfigure}{r}{0.48\textwidth}
  \centering
  \vspace{-1.1cm}
  \includegraphics[trim={0.1cm 0.1cm 0.1cm 1.6cm}, clip=true, scale=0.7]{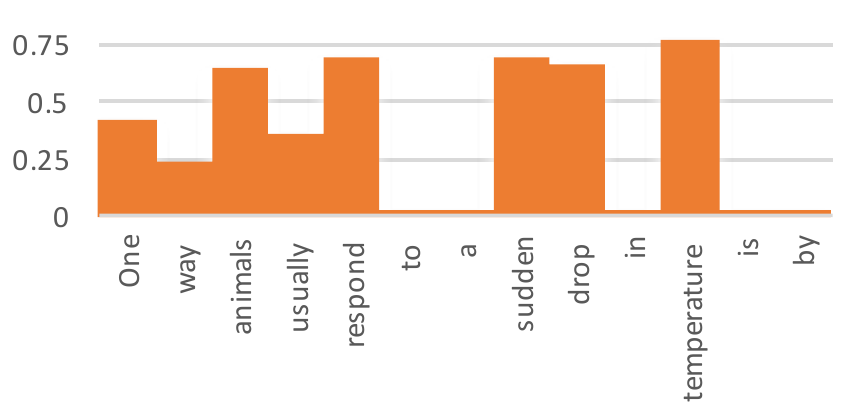}
  \vspace{-.5cm}
  \caption{\label{fig:scores-example} 
  \footnotesize
  Essentiality scores generated by our system, which assigns high essentiality to ``drop" and ``temperature".
  }
\end{wrapfigure}
Towards this goal, we propose a system that can assign an essentiality score to each term in the question. For the above example, our system generates the scores shown in Figure~\ref{fig:scores-example}, where more weight is put on ``temperature" and ``sudden drop". A QA system, when armed with such information, is expected to exhibit a more informed behavior.

We make the following contributions:

(A) We introduce the notion of \emph{question term \essty} and release a new dataset of 2,223 crowd-sourced essential term annotated questions (total 19K annotated terms) that capture this concept.\footnote{\label{footnote1} Annotated dataset and classifier available at
\url{https://github.com/allenai/essential-terms}
} We illustrate the importance of this concept by demonstrating that humans become substantially worse at QA when even a few essential question terms are dropped.

(B) We design a classifier that is effective at predicting question term essentiality. The F1 (0.80) and per-sentence mean average precision (MAP, 0.90) scores of our classifier supercede the closest baselines by 3\%-5\%. Further, our classifier generalizes substantially better to unseen terms.

(C) We show that this classifier can be used to improve a surprisingly effective IR based QA system~\citep{CEKSTTK16} by 4\%-5\% on previously used question sets and by 1.2\% on a larger question set.
We also incorporate the classifier in \tableilp~\citep{KKSCER16}, resulting in fewer errors when sufficient knowledge is present for questions to be meaningfully answerable.


\ignore{
We take a supervised learning approach for identifying \emph{terms of a question that are \ess\ to answering it correctly}. Arguably, whether a particular term is \ess\ or not is somewhat subjective. Instead of attempting to define this concept formally, we use crowdsourcing based on certain broad guidelines in order to obtain an annotated dataset, thereby capturing human intuition about what is essential. 

We make three contributions:
(1) \textbf{Term-Essentiality Dataset}:
We introduce a dataset of over 2,200 questions with annotated \ess\ terms.
    We also validate the notion of term essentiality via a more refined (but expensive to annotate) notion of term-group essentiality.
    (2) \textbf{ET Classifier}: We demonstrate that a new max-margin essential terms (ET) classifier trained on the above dataset significantly outperform supervised and unsupervised baselines.
%
    (3) \textbf{Enhancing QA Systems}: First, we illustrate sensible behavior of the above ET classifier as a sub-component of a model-agnostic ``oracle'' QA system that has humans in the loop. Second, for a state-of-the-art \lucene\ based QA system mentioned earlier, we demonstrate improved performance across multiple datasets. Third, for \tableilp, a constraint-based solver mentioned earlier, we didn't achieve an overall improvement in score but provide a tunable precision-recall tradeoff and a detailed analysis in support of more informed QA.
}


\subsection{Related Work}

Our work can be viewed as the study of an intermediate layer in QA systems.
Some systems implicitly model and learn it, often via indirect signals from end-to-end training data. For instance, Neural Networks based models \citep{WangLiZh16,TymoshenkoBoMo16,YinEbSc16} implicitly compute some kind of \emph{attention}. While this is intuitively meant to weigh key words in the question more heavily,
this aspect hasn't been systematically evaluated, in part due to the lack of ground truth annotations.

There is related work on extracting \emph{question type} information \citep{LiRo02,LZYZ07} and applying it to the design and analysis of end-to-end QA systems \citep{MPHS03}. 
The concept of term \essty\ studied in this work is
different, and so is
our supervised learning approach compared to the typical rule-based systems for question type identification.


Another line of relevant work is sentence  compression \citep{ClarkeLa08}, where the goal is to minimize the content while maintaining grammatical soundness. These approaches typically build an internal importance assignment component to assign significance scores to various terms, which is often done using language models, co-occurrence statistics, or their variants \citep{KnightMa02,HoriSa04}. We compare against unsupervised  baselines inspired by such importance assignment techniques. 

In a similar spirit,~\cite{ParkCr15} use translation models to extract key terms to prevent semantic drift in query expansion.

One key difference from general text summarization literature is that we operate on questions, which tend to have different essentiality characteristics than, say, paragraphs or news articles. As we discuss in Section~\ref{subsec:dataset}, typical indicators of essentiality such as being a proper noun or a verb (for event extraction) are much less informative for questions. Similarly, while the opening sentence of a Wikipedia article is often a good summary, it is the last sentence (in multi-sentence questions) that contains the most pertinent words.

In parallel to our effort, \citet{JSSC17} recently introduced a science QA system that uses the notion of \emph{focus words}. Their rule-based system incorporates grammatical structure, answer types, etc. We take a different approach by learning a supervised model using a new annotated dataset. 


\section{Essential Question Terms}
\label{sec:essential-terms}

In this section, we introduce the notion of \emph{essential question terms}, present a dataset annotated with these terms, and describe two experimental studies that illustrate the importance of this notion---we show that when dropping terms from questions, humans' performance degrades significantly faster if the dropped terms are essential question terms.

Given a question $q$, we consider each non-stopword token in $q$ as a candidate for being an essential question term. Precisely defining what is essential and what isn't is not an easy task and involves some level of inherent subjectivity. We specified  \emph{three broad criteria}: 1) altering an \ess\ term should change the intended meaning of $q$,  2) dropping non-\ess\ terms should not change the correct answer for $q$, and 3) grammatical correctness is not important. We found that given these relatively simple criteria, human annotators had a surprisingly high agreement when annotating elementary-level science questions. Next we discuss the specifics of the crowd-sourcing task and the resulting dataset.

\subsection{Crowd-Sourced Essentiality Dataset}
\label{subsec:dataset}

We collected 2,223 elementary school science exam questions for the annotation of \ess\ terms. This set includes the questions used by \citet{CEKSTTK16}\footnote{These are the only publicly available state-level science exams. http://www.nysedregents.org/Grade4/Science/} and additional ones obtained from other public resources such as the Internet or textbooks. 
For each of these questions, we asked crowd workers\footnote{We use Amazon Mechanical Turk for crowd-sourcing.} to annotate essential question terms based on the above criteria as well as a few examples of essential and non-essential terms. \f\ref{fig:mturk:interface} depicts the annotation interface.

\begin{figure*}[tb]
    \centering
    \begin{framed}
        \includegraphics[trim=0cm 0.6cm 0cm 0.3cm, clip=true, scale=0.25]{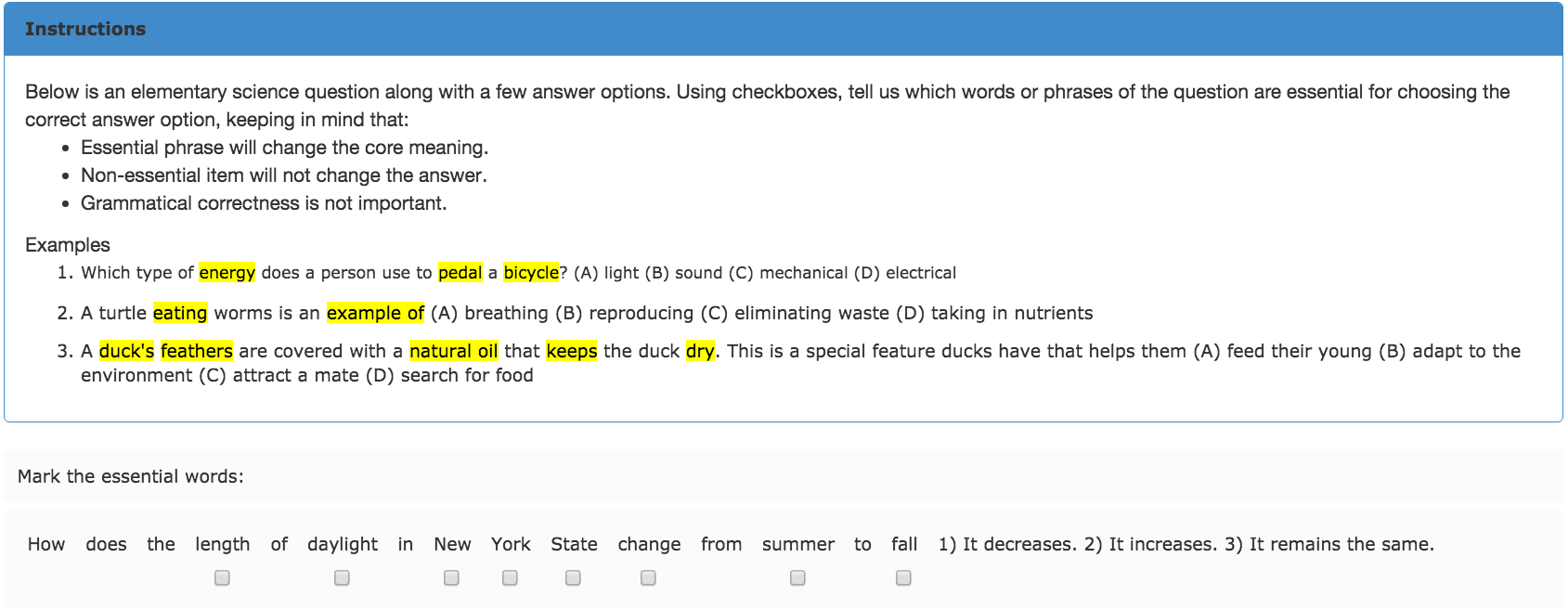}
    \end{framed}
    \caption{Crowd-sourcing interface for annotating \ess\ terms in a question, including the criteria for \essty\ and sample annotations.}
    \label{fig:mturk:interface}
\end{figure*}


\ignore{
\begin{table}[t]
    \setlength\doublerulesep{\arrayrulewidth}
    \small
    \centering
    \begin{tabular}{lc}
        \multicolumn{1}{c}{Statistic} & Value \\ 
        \hline 
        \hline 
        \# of annotated questions &  2,223 \\ 
        \# of annotated question terms & 19,380 \\ 
        \hline
        most frequent essential POS & NN \\
        question terms marked essential (avg.) & 29.9\%\\
        science terms marked essential & 76.6\%\\
        \ashish{add more}
        \\
        \hline 
    \end{tabular}
    \caption{Annotation statistics. 2,199 questions received at least 5 annotations (79 received 10 annotations due to unintended question repetition), 21 received 4 annotations, and 4 received 3 annotations.     \daniel{Ashish: The percentage for NN is, among the essential terms, how often its label is NN. Although it can be misinterpreted as how often a word with POS=NN was essential. } }
    \label{tab:annotationStatistics}
\end{table}

\ta\ref{tab:annotationStatistics} summarizes key statistics of the annotated data.
}

The questions were annotated by 5 crowd workers,\footnote{A few invalid annotations resulted in about 1\% of the questions receiving fewer annotations. 2,199 questions received at least 5 annotations (79 received 10 annotations due to unintended question repetition), 21 received 4 annotations, and 4 received 3 annotations.} and resulted in 19,380 annotated terms. The Fleiss' kappa statistic \citep{Fleiss71} for this task was $\kappa$ = 0.58, indicating a level of inter-annotator agreement very close to `substantial'. In particular, all workers agreed on 36.5\% of the terms and at least 4 agreed on 69.9\% of the terms. We use the proportion of workers that marked a term as essential to be its annotated essentiality score.

On average, less than one-third (29.9\%) of the terms in each question were marked as essential (i.e., score $>$ 0.5). This shows the large proportion of distractors in these science tests (as compared to traditional QA datasets), further showing the importance of this task.
Next we provide some insights into these terms.

We found that part-of-speech (POS) tags are not a reliable predictor of \essty, making it difficult to hand-author POS tag based rules. Among the proper nouns (NNP, NNPS) mentioned in the questions, fewer than half (47.0\%) were marked as essential. This is in contrast with domains such as news articles where proper nouns carry perhaps the most important information.
Nearly two-thirds (65.3\%) of the mentioned comparative adjectives (JJR) were marked as essential, whereas only a quarter of the mentioned superlative adjectives (JJS) were deemed essential. Verbs were marked essential less than a third (32.4\%) of the time. This differs from domains such as math word problems where verbs have been found to play a key role \citep{HHEK14}.

The best single indicator of essential terms, not surprisingly, was being a scientific term\footnote{We use 9,144 science terms from \citet{KKSCER16}.} (such as \emph{precipitation} and \emph{gravity}). 76.6\% of such terms occurring in questions were marked as essential.


In summary, we have a term essentiality annotated dataset of 2,223 questions. We split this into train/development/test subsets in a 70/9/21 ratio, resulting in 483 test sentences
used for per-question evaluation.

We also derive from the above an annotated dataset of 19,380 terms by pooling together all terms across all questions. Each term in this larger dataset is annotated with an essentiality score in the context of the question it appears in. This results in 4,124 test instances (derived from the above 483 test questions)
. We use this dataset for per-term evaluation.



\subsection{The Importance of \Ess\ Terms}

Here we report a second crowd-sourcing experiment that validates our hypothesis that the question terms marked above as essential are, in fact, essential for understanding and answering the questions. Specifically, we ask: \emph{Is the question still answerable by a human if a fraction of the essential question terms are eliminated?}
For instance, the sample question in the introduction is unanswerable when ``drop'' and ``temperature'' are removed from the question: \textit{One way animals usually respond to a sudden * in * is by \_\_\_?}

\begin{figure*}[ht]
    \centering
    \begin{framed}
            \includegraphics[trim=0cm 1.3cm 0cm 0.4cm, clip=true, scale=0.307]{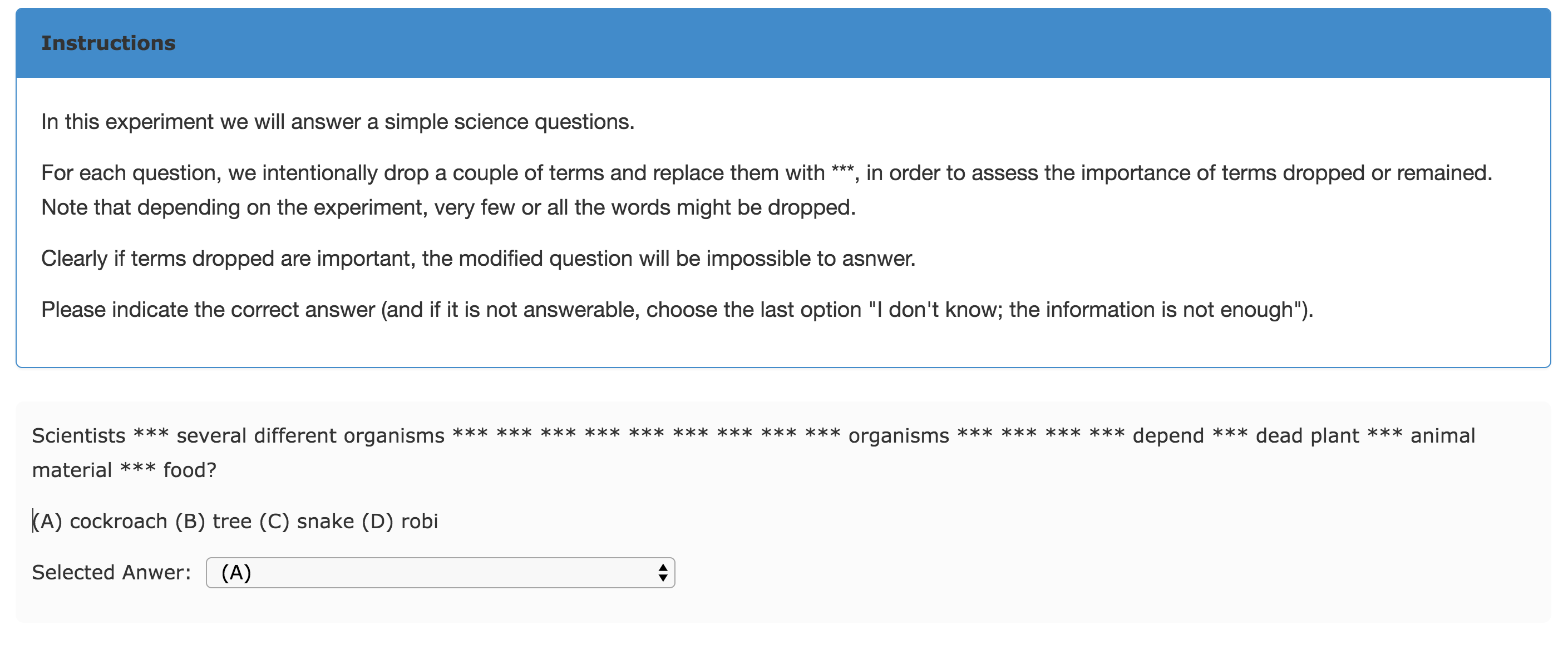}
    \end{framed}
    \caption{Crowd-sourcing interface for verifying the validity of \essty\ annotations generated by the first task. Annotators are asked to answer, if possible, questions with a group of terms dropped.}
    \label{fig:mturk:interface:group}
\end{figure*}

To this end, we consider both the annotated essentiality scores 
 as well as the score produced by our trained classifier (to be presented in Section \ref{sec:ET}).
We first generate candidate sets of terms to eliminate using these \essty\ scores based on a threshold $\xi \in \{0, 0.2, \hdots, 1.0\}$: (a) \textbf{essential set}: terms with score $\geq \xi$; (b) \textbf{non-essential set}: terms with score $< \xi$.
We then ask crowd workers to try to answer a question after replacing each candidate set of terms with ``***". In addition to four original answer options, we now also include ``I don't know. The information is not enough" (cf.\ Figure~\ref{fig:mturk:interface:group} for the user interface).\footnote{It is also possible to directly collect essential term groups using this task. However, collecting such sets of essential terms would be substantially more expensive, as one must iterate over exponentially many subsets rather than the linear number of terms used in our annotation scheme.} For each value of $\xi$, we obtain 
$5 \times 269$ 
annotations for 
269
questions. We measure how often the workers feel there is sufficient information to attempt the question and, when they do attempt, how often do they choose the right answer.

\begin{wrapfigure}{r}{0.5\textwidth}
    \includegraphics[trim={4cm 10.2cm 0cm 9.9cm}, clip=true, scale=0.60]{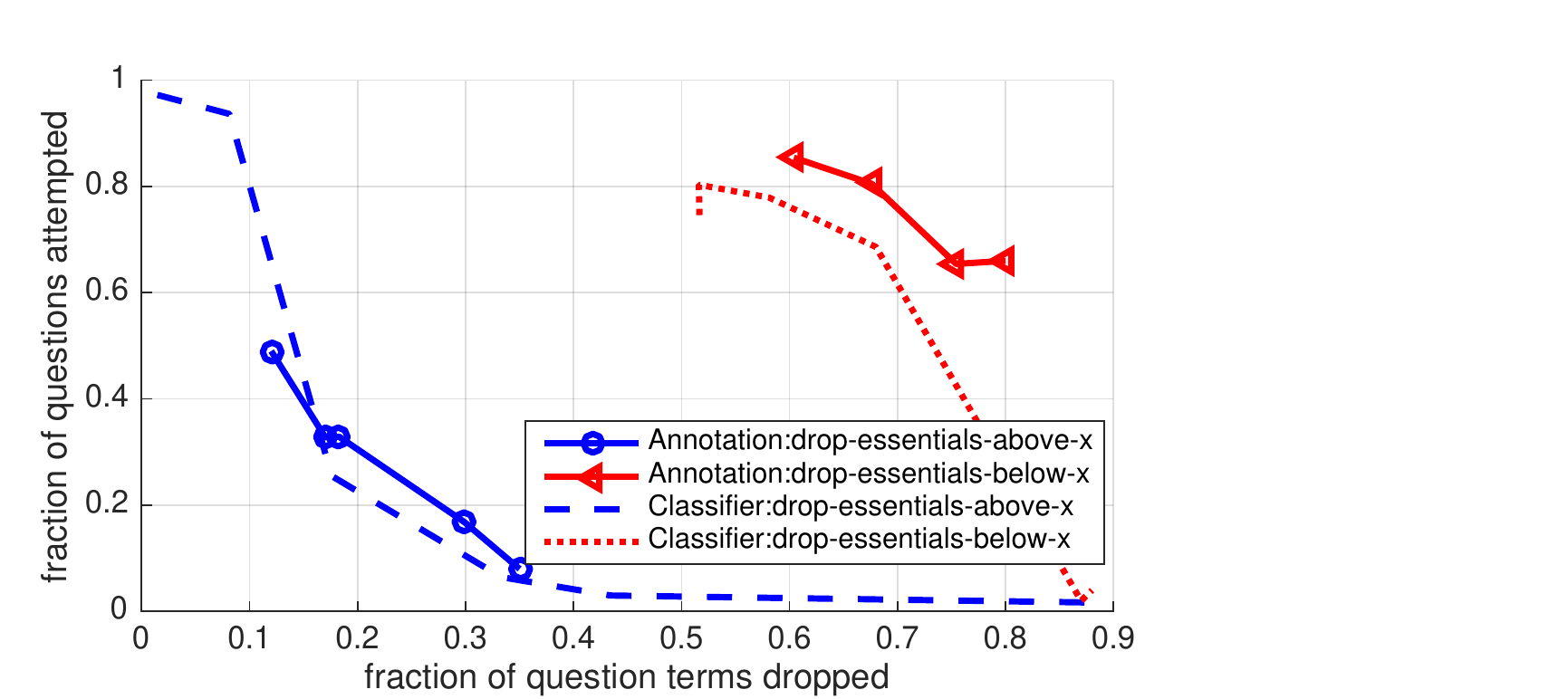}
    \caption{
        \footnotesize
        The relationship between the fraction of question words dropped and the fraction of the questions attempted (fraction of the questions workers felt comfortable answering). Dropping most essential terms (blue lines) results in very few questions remaining answerable, while least essential terms (red lines) allows most questions to still be answerable. Solid lines indicate human annotation scores while dashed lines indicate predicted scores. 
    }
    \label{fig:combinedDropTerms}
\end{wrapfigure}
Each value of $\xi$ results in some fraction of terms to be dropped from a question; the exact number depends on the question and on whether we use annotated scores or our classifier's scores. In Figure~\ref{fig:combinedDropTerms}, we plot the average fraction of terms dropped on the horizontal axis and the corresponding fraction of questions attempted on the vertical axis. Solid lines indicate annotated scores and dashed lines indicate classifier scores. Blue lines (bottom left) illustrate the effect of eliminating essential sets while red lines (top right) reflect eliminating non-essential sets.

We make two observations. First, the solid blue line (bottom-left) demonstrates that dropping even a small fraction of question terms marked as essential dramatically reduces the QA performance of humans. E.g., dropping just 12\% of the terms (with high  essentiality scores) makes 51\% of the questions unanswerable. The solid red line (top-right), on the other hand, shows the opposite trend for terms marked as not-essential: even after dropping 80\% of such terms, 65\% of the questions remained answerable.

Second, the dashed lines reflecting the results when using scores from our \et\ classifier are very close to the solid lines based on human annotation. This indicates that our classifier, to be described next, closely captures human intuition.


\ignore{
\subsection{Validation via Group Essentiality} 

\tushar{Do we need the next sub-section ?}

There is no a priori guarantee that our annotation mechanism will result in well-defined 
scores. It's quite possible that per-token annotations might not capture correlation between terms. In fact, a clearer definition of \ess\ terms is perhaps via term-\textit{groups} (rather than per-token judgments): \emph{a term-group $G$ is essential for a question $Q$ if dropping $G$ from $Q$ leaves insufficient information to answer $Q$}.
Clearly, if $G$ is essential, so are its supersets, but not necessarily its subsets. For instance, here are a few annotated term-groups for the example discussed in the introduction: 

{
\small 
\indent $\set{\text{``animals"}} \rightarrow$ not essential\\
\indent $\set{\text{``animals", ``sudden drop"}} \rightarrow$ essential \\ 
\indent $\set{\text{``animals", ``temperature"}}\rightarrow$ not essential \\
\indent $\set{\text{``animals", ``sudden drop", ``temperature"}}\rightarrow$ essential
}

While such a group-based representation is more natural, its annotation mechanism is substantially more expensive, as one must iterate over exponentially many subsets rather than a linear number of individual terms.

To address this challenge, we use our per-token annotation and a per-token learning model (next section) to create candidates for essential groups. 
We then perform a smaller scale crowdsourcing task to verify that term-groups composed of terms marked essential during per-term annotation are indeed essential for answering the question. This verifies that our per-term annotations provide a cost-effective yet useful \emph{approximation} of term-group annotation.

Specifically, in this second task, we ask annotators to answer a question after replacing some of the terms with ``***". In addition to four original answer options, we include ``I don't know. The information is not enough" (Figure~\ref{fig:mturk:interface:group}). 
Candidate term-groups are generated in two ways: For a threshold $\xi$, we keep terms with per-token \essty\ score (a) at least $\xi$ and (b) less than $\xi$.
For each of these settings and for $\xi \in \{0, 0.2, \hdots, 1.0\}$, we repeat the experiment, keeping track of the number of questions attempted, i.e., where the new ``I don't know'' option wasn't selected.

Intuitively, we expect a much higher number of question attempted in scenario (a) compared to scenario (b). 
Figure~\ref{fig:droppQuestionTerms} shows the relationship between the fraction of question words dropped and the resulting number of questions workers felt comfortable answering. 
Interestingly, when term-groups comprising only 12\% of the terms, marked ``highly essential'', were dropped from the question, the workers felt there wasn't enough information left to answer 51\% of the questions, showing the value of these annotations. On the other hand, even after dropping 80\% of the terms, ranking low on the essentiality score, the workers answered around 65\% of the questions.
}

\section{\Ess\ Terms Classifier}
\label{sec:ET}

Given the dataset of questions and their terms annotated with \ess\ scores,
is it possible to learn the underlying concept? Towards this end,
given a question $q$ , answer options $a$, and a question term $q_l$, we seek a classifier that predicts whether $q_l$ is essential for answering $q$. We also extend it to produce an \essty\ score $et(q_l, q, a) \in
[0,1]$.\footnote{The \essty\ score may alternatively be defined as $et(q_l, q)$, independent of the answer options $a$. This is more suitable for non-multiple choice questions. Our system uses $a$ only to compute PMI-based statistical association features for the classifier. In our experiments, dropping these features resulted in only a small drop in the classifier's performance.}
We use the annotated dataset from \s2, where real-valued \essty\ scores are binarized to 1 if they are at least 0.5, and to 0 otherwise.

We train a linear SVM classifier \citep{Joachims98}, henceforth referred to as \textbf{\et\ classifier}. Given the complex nature of the task, the features of this classifier include syntactic (e.g., dependency parse based) and semantic (e.g., Brown cluster representation of words \citep{BDMPL92}, a list of scientific words) properties of question words, as well as their combinations.
In total, we use 120 types of features (cf.~Appendix of~\cite{KKSR17}).


\paragraph{Baselines.}
To evaluate our approach, we devise a few simple yet relatively powerful baselines.

First, for our supervised baseline, given $(q_l, q, a)$ as before, we ignore $q$ and compute how often is $q_l$ annotated as essential in the entire dataset. In other words, the score for $q_l$ is the proportion of times it was marked as essential in the annotated dataset. If the instance is never observer in training, we choose an arbitrary label as prediction.
We refer to this baseline as \emph{label proportion baseline} and create two variants of it: \wordBaseline based on surface string and \lemmaBaseline based on lemmatizing the surface string. For unseen $q_l$, this baseline makes a random guess with uniform distribution. 

Our unsupervised baseline is inspired by work on sentence compression \citep{ClarkeLa08} 
and the PMI solver of \citet{CEKSTTK16}, which compute word importance based on co-occurrence statistics in a large corpus. 
In a corpus $\mathcal{C}$ of 280 GB of plain text ($5 \times 10^{10}$ tokens) extracted from Web pages,\footnote{Collected by Charles Clarke at the University of Waterloo,
and used previously by \citet{Turney13}.} 
we identify unigrams, bigrams, trigrams, and skip-bigrams from $q$ and each answer option $a_i$. For a pair $(x, y)$ of $n$-grams, their pointwise mutual information (PMI) \citep{ChurchHa89} in $\mathcal{C}$ is defined as $\log \frac{p(x,y)}{p(x)p(y)}$ where $p(x, y)$ is the co-occurrence frequency of $x$ and $y$ (within some window) in $\mathcal{C}$. For a given word $x$, we find all pairs of question $n$-grams and answer option $n$-grams. \maxSalience and \sumSalience
score the importance of a word $x$ by max-ing or summing, resp., PMI scores $p(x,y)$ across all answer options $y$ for $q$. A limitation of this baseline is its dependence on the existence of answer options, while our system makes essentiality predictions independent of the answer options. 

We note that all of the aforementioned baselines produce real-valued confidence scores (for each term in the question), which can be turned into binary labels (\ess\ and non-\ess) by thresholding at a certain confidence value.

\subsection{Evaluation}

We consider two natural evaluation metrics for essentiality detection, first treating it as a binary prediction task at the level of individual terms and then as a task of ranking terms within each question by the degree of essentiality.

\paragraph{Binary Classification of Terms. }
We consider all question terms pooled together as described in Section~\ref{subsec:dataset}, resulting in a dataset of 19,380 terms annotated (in the context of the corresponding question) independently as essential or not. The \et\ classifier is trained on the train subset, and the threshold is tuned using the dev subset.

\begin{table}[ht]
    \setlength\doublerulesep{\arrayrulewidth}
    \centering 
    \small
    \setlength\extrarowheight{-10pt}
    \begin{tabular}{l|cc|ccc}
           &  AUC & Acc & P  & R   & F1   \\ 
        \hline \hline 
        \maxSalience$^\dagger$ & 0.74 & 0.67 & 0.88 & 0.65 & 0.75\\
        \sumSalience$^\dagger$ & 0.74 & 0.67 & 0.88 & 0.65 & 0.75\\
        \wordBaseline & 0.79 & 0.61 & 0.68 & 0.64 & 0.66 \\
        \lemmaBaseline & 0.80 & 0.63 & 0.76 & 0.64 & 0.69 \\
        {\bf \et\ Classifier} & \textbf{0.79} & \textbf{0.75} & \textbf{0.91} & \textbf{0.71} & \textbf{0.80} \\
        \hline
    \end{tabular}
    \makeatletter\def\@captype{table}\makeatother
    \caption{
    \small
    Effectiveness of various methods for identifying essential question terms in the test set, including area under the PR curve (AUC), accuracy (Acc), precision (P), recall (R), and F1 score. \et\ classifier substantially outperforms all supervised and unsupervised (denoted with $^\dagger$) baselines.}
    \label{tab:results:classifier}
\end{table}

For each term in the corresponding test set of 4,124 instances, we use various methods to predict whether the term is essential (for the corresponding question) or not. Table~\ref{tab:results:classifier} summarizes the resulting performance. For the threshold-based scores, each method was tuned to maximize the F1 score based on the dev set. The \et\ classifier achieves an F1 score of 0.80, which is 5\%-14\% higher than the baselines. Its accuracy at 0.75 is statistically significantly better than all baselines based on the Binomial\footnote{Each test term prediction is assumed to be a binomial.} exact test~\citep{Howell12} at $p$-value 0.05.

\begin{figure}[htb]
    \centering
        \includegraphics[trim=4.7cm 9.85cm 4.8cm 10.2cm, clip=true, scale=0.63]{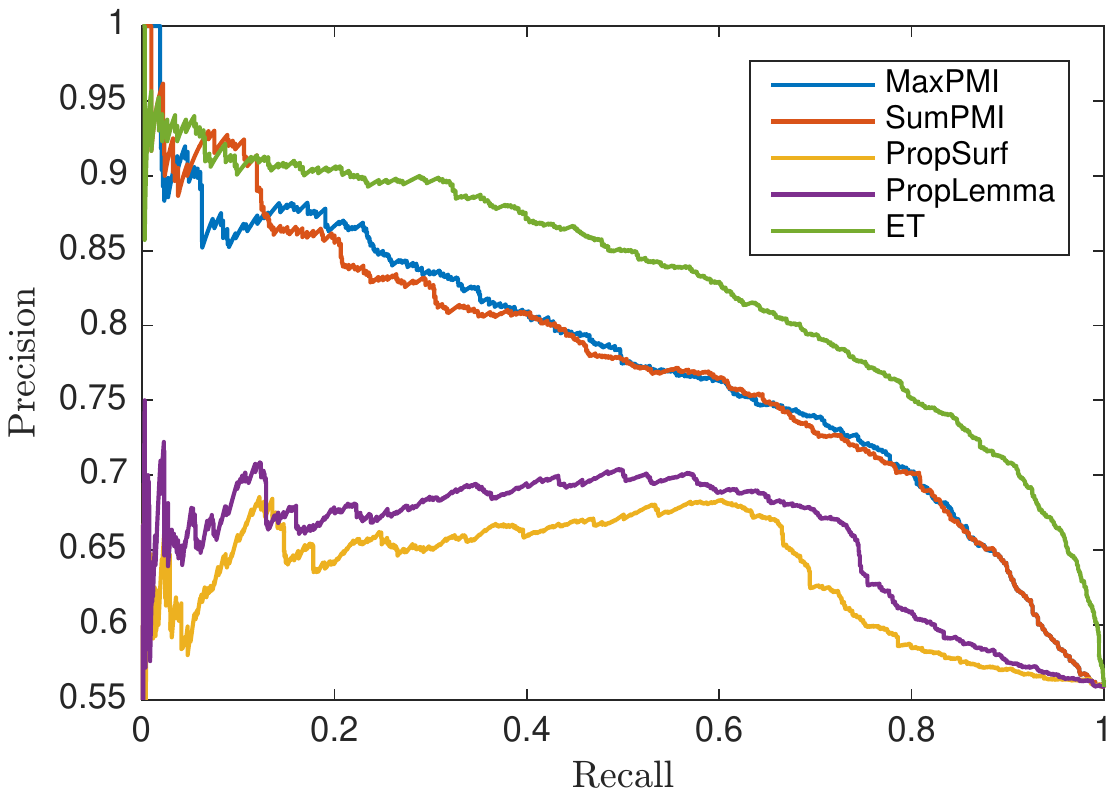}
    \caption{
    \small
    Precision-recall trade-off for various classifiers as the threshold is varied.
    \et\ classifier (green) is significantly better throughout.
    }
    \label{fig:pr:differentSolvers}
\end{figure}
As noted earlier, each of these essentiality identification methods are parameterized by a threshold for balancing precision and recall. This allows them to be tuned for end-to-end performance of the downstream task. We use this feature later when incorporating the \et\ classifier in QA systems. Figure~\ref{fig:pr:differentSolvers} depicts the PR curves for various methods as the threshold is varied, highlighting that the \et\ classifier performs reliably at various recall points. Its precision, when tuned to optimize F1, is 0.91, which is very suitable for high-precision applications. It has a 5\% higher AUC (area under the curve) and outperforms baselines by roughly 5\% throughout the precision-recall spectrum. 

\begin{table}[tb]
    \setlength\doublerulesep{\arrayrulewidth}
    \centering 
    \small
    \setlength\extrarowheight{-10pt}
    \begin{tabular}{l|cc|ccc}
           &  AUC & Acc & P  & R   & F1   \\ 
        \hline \hline 
        \maxSalience$^\dagger$ & 0.75 & 0.63 & 0.81 & 0.65 & 0.72 \\
        \sumSalience$^\dagger$ & 0.75 & 0.63 & 0.80 & 0.66 & 0.72 \\
        \wordBaseline & 0.57 & 0.51 & 0.49 & 0.61 & 0.54 \\
        \lemmaBaseline & 0.58 & 0.49 & 0.50 & 0.59 & 0.54 \\
        {\bf \et\ Classifier} & \textbf{0.78} & \textbf{0.71} & \textbf{0.88} & \textbf{0.71} & \textbf{0.78} \\
        \hline
    \end{tabular}
    \makeatletter\def\@captype{table}\makeatother
    \caption{
    \small
    Generalization to unseen terms: Effectiveness of various methods, using the same metrics as in Table~\ref{tab:results:classifier}. As expected, supervised methods perform poorly, similar to a random baseline. Unsupervised methods generalize well, but the \et\ classifier again substantially outperforms them.}
    \label{tab:precisionRecallUnseenTerms}
\end{table}
As a second study, we assess how well our classifier \textbf{generalizes to unseen terms}. For this, we consider
only the 559 test terms that do not appear in the train set.\footnote{In all our other experiments, test and train questions are always distinct but may have some terms in common.} Table~\ref{tab:precisionRecallUnseenTerms} provides the resulting performance metrics. We see that the frequency based supervised baselines, having never seen the test terms, stay close to the default precision of 0.5.  
The unsupervised baselines, by nature, generalize much better but are substantially dominated by our \et\ classifier, which achieves an F1 score of 78\%. This is only 2\% below its own F1 across all seen and unseen terms, and 6\% higher than the second best baseline.

 \ignore{   
\begin{figure}
    \centering
    \includegraphics[trim=3.4cm 9.3cm 1cm 10.0cm, clip=true, scale=0.54]{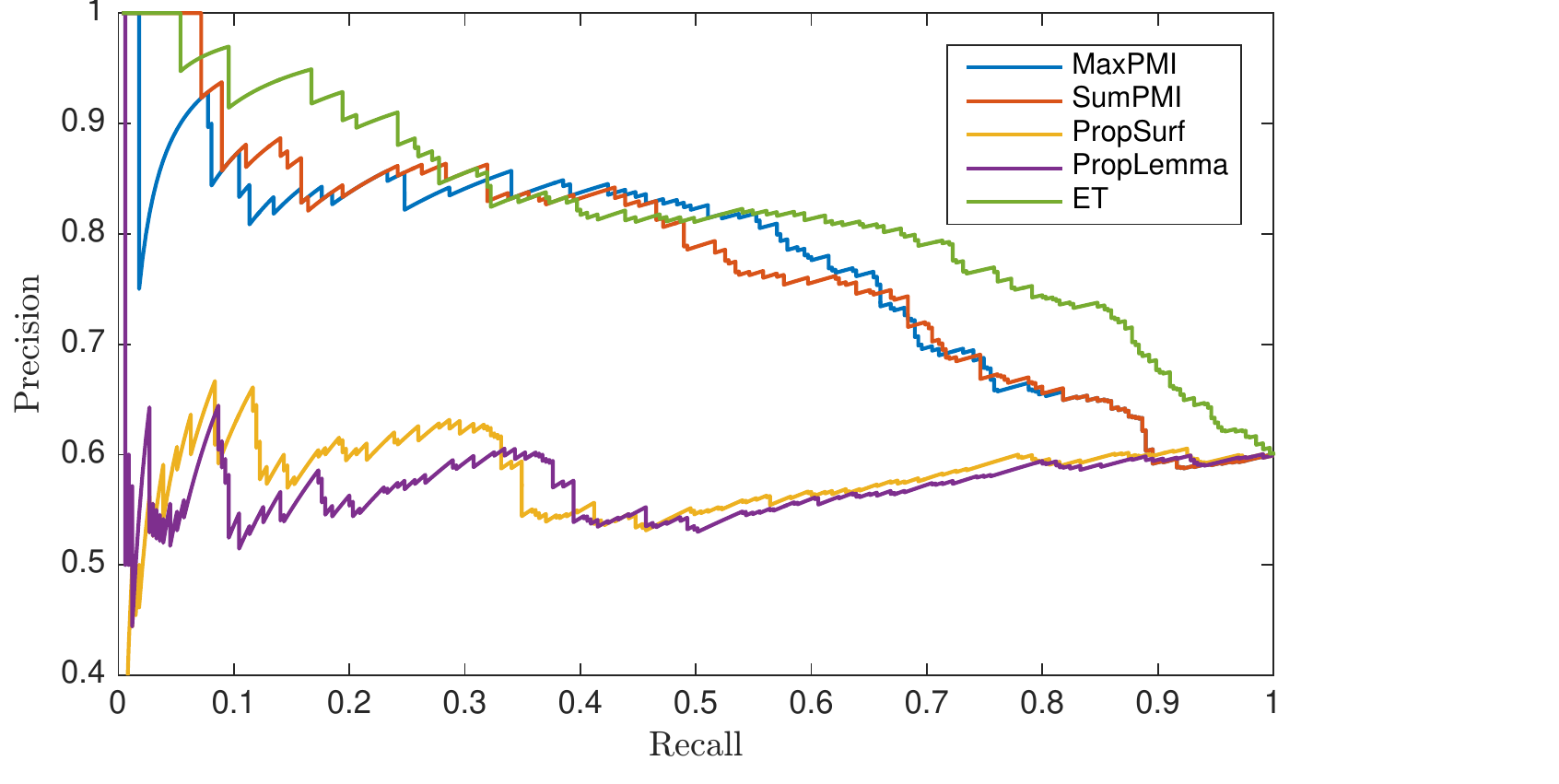}
    \caption{Our classifier generalizes to the unseen data. The supervised baselines stay close to the default precision of 0.5, while the unsupervised baselines generalize much better but are substantially dominated by our \et\ classifier.}
    \label{fig:precisionRecallUnseenTerms}
\end{figure}    
}    
\paragraph{Ranking Question Terms by Essentiality. }

\begin{wrapfigure}{r}{0.4\textwidth}
    \setlength{\tabcolsep}{12pt}
    \setlength\doublerulesep{\arrayrulewidth}
    \centering 
    \small
    \setlength\extrarowheight{-10pt}
    \begin{tabular}{l|c}
        System  &  MAP \\ 
        \hline 
        \hline 
        \maxSalience$^\dagger$ & 0.87 \\
        \sumSalience$^\dagger$ &  0.85 \\
        \wordBaseline & 0.85  \\ 
        \lemmaBaseline & 0.86 \\ 
        {\bf \et\ Classifier} & \textbf{0.90} \\
        \hline
    \end{tabular}
    \makeatletter\def\@captype{table}\makeatother
    \caption{
    \small
    Effectiveness of various methods for ranking the terms in a question by essentiality. $^\dagger$ indicates unsupervised method. Mean-Average Precision (MAP) numbers reflect the mean (across all test set questions) of the average precision of the term ranking for each question. \et\ classifier again substantially outperforms all baselines.}
    \label{tab:results:et:classifier}
\end{wrapfigure}
Next, we investigate the performance of the \et\ classifier as a system that ranks all terms within a question in the order of essentiality. Thus, unlike the previous evaluation that pools terms together across questions, we now consider each question as a unit. For the ranked list produced by each classifier for each question, we compute the average precision (AP).\footnote{We rank all terms within a question based on their essentiality scores. For any true positive instance at rank $k$, the precision at $k$ is defined to be the number of positive instances with rank no more than $k$, divided by $k$.
The average of all these precision values for the ranked list for the question is the \emph{average precision}.} We then take the mean of these AP values across questions to obtain the mean average precision (MAP) score for the classifier.

The results for the test set (483 questions) are shown in Table~\ref{tab:results:et:classifier}. Our \et\ classifier achieves a MAP of 90.2\%, which is 3\%-5\% higher than the baselines, 
and demonstrates that one can learn to reliably identify essential question terms.

\section{Using \et\ Classifier in QA Solvers}
\label{sec:using:in:classifier}

In order to assess the utility of our \et\ classifier, we investigate its impact on two end-to-end QA systems. We start with a brief description of the question sets.

\paragraph{Question Sets.} We use three question sets of 4-way multiple choice questions.\footnote{Available at \url{http://allenai.org/data.html}}
\regents\ and \aiTwoPublic\ are two publicly available elementary school science question set. \regents\ comes with 127 training and 129 test questions; \aiTwoPublic\ contains 432 training and 339 test questions that subsume the smaller question sets used previously \citep{CEKSTTK16,KKSCER16}.
\regentsPerturbed\ set, introduced by \citet{KKSCER16}, has 1,080 questions obtained by automatically perturbing incorrect answer choices for 108 New York Regents 4th grade science questions. We split this into 700 train and 380 test questions. 

For each question, a solver gets a score of 1 if it chooses the correct answer and $1/k$ if it reports a $k$-way tie that includes the correct answer. 

\paragraph{QA Systems.} We investigate the impact of adding the \et\ classifier to two state-of-the-art QA systems for elementary level science questions. Let $q$ be a multiple choice question with answer options $\{a_i\}$. The \emph{\lucene\ Solver} from \citet{CEKSTTK16} searches, for each $a_i$, a large corpus for a sentence that best matches the $(q,a_i)$ pair. It then selects the answer option for which the match score is the highest. The inference based \emph{\tableilp\ Solver} from \citet{KKSCER16}, on the other hand, performs QA by treating it as an optimization problem over a semi-structured knowledge base derived from text. It is designed to answer questions requiring multi-step inference and a combination of multiple facts.

For each multiple-choice question $(q, a)$, we use the \et\ classifier to obtain \ess\ term scores $s_l$ for each token $q_l$ in $q$; $s_l = et(q_l, q, a)$.
We will be interested in the subset $\omega$ of all terms $T_q$ in $q$ with \essty\ score above a threshold $\xi$: 
$\omega(\xi; q) =  \{l \in T_q \mid s_l > \xi \}$.
Let $\overline{\omega}(\xi; q) = T_q \setminus \omega(\xi; q)$. For brevity, we will write $\omega(\xi)$ when $q$ is implicit.

\subsection{\lucene\  solver + \et}

To incorporate the \et\ classifier, we create a parameterized \lucene\ system called \queryFiltering($\xi$) where, instead of querying a $(q,a_i)$ pair, we query 
$(\omega(\xi; q),a_i)$.

While IR solvers are generally easy to implement and are used in popular QA systems with surprisingly good performance, they are often also sensitive to the nature of the questions they receive. \citet{KKSCER16} demonstrated that a minor perturbation of the questions, as embodied in the \regentsPerturbed\ question set, dramatically reduces the performance of IR solvers. Since the perturbation involved the introduction of distracting incorrect answer options, we hypothesize that a system with better knowledge of what's important in the question will demonstrate increased robustness to such perturbation.

\begin{wrapfigure}{r}{0.5\textwidth}
\setlength{\tabcolsep}{8pt}
\setlength{\doublerulesep}{\arrayrulewidth}
\centering 
\small
\setlength\extrarowheight{-10pt}
\begin{tabular}{l|cc}
Dataset & Basic IR & \queryFiltering\  \\ 
\hline 
\hline
\T \regents &  59.11 & {\bf 60.85}  \\
\aiTwoPublic & 57.90 & {\bf 59.10} \\
\regentsPerturbed &  61.84 & {\bf 66.84}  \\
\hline
\end{tabular}
\vspace{-0.1cm}
\makeatletter\def\@captype{table}\makeatother
\caption{Performance of the \lucene\ solver without (Basic IR) and with (\queryFiltering) essential terms. The numbers are solver scores (\%) on the test sets of the three datasets. 
}
\vspace{-0.1cm}
\label{table:public}
\end{wrapfigure}

Table \ref{table:public} validates this hypothesis, showing the result of incorporating \et\ in \lucene, as \queryFiltering($\xi = 0.36$), where $\xi$ was selected by optimizing end-to-end performance on the training set. 
We observe a 5\% boost in the score on \regentsPerturbed, showing that incorporating the notion of \essty\ makes the system more robust to perturbations.

Adding ET to \lucene\ also improves its performance on standard test sets.  On the larger \aiTwoPublic\ question set, we see an improvement of 1.2\%.
On the smaller \regents\ set, introducing \et\ improves \lucene solver’s score by 1.74\%, bringing it close to the state-of-the-art solver, \tableilp, which achieves a score of 61.5\%.
This demonstrates that the notion of essential terms can be fruitfully exploited to improve QA systems.

\subsection{\tableilp\  solver + \et}

Our essentiality guided query filtering helped the \lucene\ solver find sentences that are more relevant to the question. However, for \tableilp\ an added focus on essential terms is expected to help only when the requisite knowledge is present in its relatively small knowledge base. To remove confounding factors, we focus on questions that are, in fact, answerable.

To this end, we consider three (implicit) requirements for \tableilp\ to demonstrate reliable behavior:
(1) the existence of relevant knowledge, (2) correct alignment between the question and the knowledge, and (3) a valid reasoning chain connecting the facts together. Judging this for a question, however, requires a significant manual effort and can only be done at a small scale.

\paragraph{Question Set.} We consider questions for which the \tableilp\ solver does have access to the requisite knowledge and, as judged by a human, a reasoning chain to arrive at the correct answer. To reduce manual effort, we collect such questions by starting with the correct reasoning chains (`support graphs') provided by \tableilp.
A human annotator is then asked to paraphrase the corresponding questions or add distracting terms, while maintaining the general meaning of the question. Note that this is done independent of essentiality scores. For instance, the modified question below changes two words in the question without affecting its core intent:
\vspace{-0.1cm}
\begin{framed}
\small 
\noindent 
\textbf{Original question: } A {\color{blue}fox} grows thicker {\color{blue}fur} as a season changes. This adaptation helps the fox to (A) find food(B) keep warmer(C) grow stronger(D) escape from predators \\ 
\textbf{Generated question: } An {\color{blue} animal} grows thicker {\color{blue} hair} as a season changes. This adaptation helps to (A) find food(B) keep warmer(C) grow stronger(D) escape from predators
\end{framed}
\vspace{-0.1cm}

While these generated questions should arguably remain correctly answerable by \tableilp, we found that this is often not the case. To investigate this, we curate a small dataset $Q_R$ with 12 questions 
(see the Appendix)
on each of which,  despite having the required knowledge and a plausible reasoning chain, \tableilp\ fails. 

\paragraph{Modified Solver.} To incorporate question term essentiality in the \tableilp\ solver while maintaining high recall, we employ a \emph{cascade} system that starts with a strong essentiality requirement and progressively weakens it.

Following the notation of Chapter 3, let $x(q_l)$ be a binary variable that denotes whether or not the $l$-th term of the question is used in the final reasoning graph. We enforce that terms with \essty\ score above a threshold $\xi$ must be used:  $x(q_l) = 1$, $\forall l \in \omega(\xi) $. Let \tableilp+\et($\xi$) denote the resulting system which can now be used in a cascading architecture: 

\tikzstyle{b} = [rectangle, draw, fill=blue!20, text width=10em, text centered, rounded corners, minimum height=2.1em, thick]

\vspace{0.1cm}
\begin{center}
\begin{tikzpicture}[auto]
    \node [b] (sys1)  {{\scriptsize \tableilp+\et($\xi_1$)}};
    \node [right=of sys1, xshift=-0.8cm](dots1) {$\rightarrow$};
    \node [b, right=of dots1, xshift=-0.8cm] (sys2) {{\scriptsize \tableilp+\et($\xi_2$)}};
    \node [right=of sys2, xshift=-0.8cm](dots2) {$\rightarrow$ \hspace{0.2cm} ... };
\end{tikzpicture}
\end{center}
\vspace{0.1cm}
where 
$\xi_1 < \xi_2 < \hdots < \xi_k$ is a sequence of thresholds. Questions unanswered by the first system are delegated to the second,
and so on. 
The cascade has the same recall as \tableilp, as long as the last system 
is the vanilla \tableilp. We refer to this configuration as \cascades$(\xi_1, \xi_2, \hdots, \xi_k)$.

This can be implemented via repeated calls to \tableilp+\et($\xi_j$) with $j$ increasing from $1$ to $k$, stopping if a solution is found. Alternatively, one can simulate the cascade via a single extended ILP using $k$ new binary variables $z_j$ with constraints:
$|\omega(\xi_j)| * z_j  \leq \sum_{l \in \omega(\xi_j)} x(q_l)$
for $j \in \{1, \hdots, k\}$,
and adding $M * \sum_{j = 1}^k z_j$ to the objective function, for a sufficiently large constant $M$.

We evaluate \cascades($0.4, 0.6, 0.8, 1.0$) on our question set, $Q_R$. By employing essentiality information provided by the \et\ classifier, \cascades\ corrects 41.7\% of the mistakes made by vanilla \tableilp. This error-reduction illustrates that the extra attention mechanism added to \tableilp\ via the concept of essential question terms helps it cope with distracting terms.

\ignore {
\subsection{\tableilp\  solver + \et}

We now consider incorporating the notion of question term essentiality in the \tableilp\ solver. This can be done in many ways. We describe a \emph{cascade} system that starts with a strong essentiality requirement and progressively weakens it.

Following the notation of \citet{KKSCER16}, let $x(q_l)$ be a binary variable that denotes whether or not the $l$-th term of the question is used in the final reasoning graph. We enforce that terms with \essty\ score above a threshold $\xi$ must be used:  $x(q_l) = 1$, $\forall l \in \omega(\xi) $. Let \tableilp+\et($\xi$) denote the resulting system.
To maintain high recall for the solver in the presence of incomplete knowledge, we create a cascading architecture: 

\tikzstyle{b} = [rectangle, draw, fill=blue!20, text width=5em, text centered, rounded corners, minimum height=2.1em, thick]

\vspace{0.1cm}
\begin{tikzpicture}[auto]
    \node [b] (sys1)  {{\scriptsize \tableilp+\et($\xi_1$)}};
    \node [right=of sys1, xshift=-0.8cm](dots1) {$\rightarrow$};
    \node [b, right=of dots1, xshift=-0.8cm] (sys2) {{\scriptsize \tableilp+\et($\xi_2$)}};
    \node [right=of sys2, xshift=-0.8cm](dots2) {$\rightarrow$ \hspace{0.2cm} ... };
\end{tikzpicture}
\vspace{0.1cm}
where 
$\xi_1 < \xi_2 < \hdots < \xi_k$ is a sequence of thresholds. Questions unanswered by the first system are delegated to the second,
and so on. 
The cascade has the same recall as \tableilp, as long as the last system 
is the vanilla \tableilp. We refer to this configuration as \cascades$(\xi_1, \xi_2, \hdots, \xi_k)$.
%
This can be implemented via repeated calls to \tableilp+\et($\xi_j$) with $j$ increasing from $1$ to $k$, stopping if a solution is found. Alternatively, one can simulate the cascade via a single extended ILP using $k$ new binary variables $z_j$ with constraints:
$|\omega(\xi_j)| * z_j & \leq \sum_{l \in \omega(\xi_j)} x(q_l)$
for $j \in \{1, \hdots, k\}$,
and adding $M * \sum_{j = 1}^k z_j$ to the objective function, for a sufficiently large constant $M$.
 We used \cascades($0.4, 0.6, 0.8, 1.0$) on our question set, $Q_R$.

While essentiality guided query filtering helps the \lucene\ solver find sentences that are more relevant to the question, an added focus on essential terms is expected to help \tableilp\ only when the requisite knowledge is in fact present in its relatively small knowledge base. To remove confounding factors, we specifically focus on questions that are answerable.

To this end, we consider three (implicit) requirements for \tableilp\ to demonstrate reliable behavior:
(1) the existence of relevant knowledge, (2) correct alignment between the question and the knowledge, and (3) a valid reasoning chain connecting the facts together. Judging this for a question, however, requires a significant manual effort and can only be done at a small scale.

We hypothesize that the \cascades\ solver improves over \tableilp\ when the above requirements are met.\footnote{Note that when these requirements aren't met for a question, \tableilp\ may still choose the correct answer by luck.}
To validate this hypothesis, we create a question set, $Q_R$, where each question does have the requisite knowledge and, as judged by a human, a reasoning chain to arrive at the correct answer. To reduce manual effort, we collect such questions by starting with the correct reasoning chains (`support graphs') provided by the \tableilp\ solver.
A human annotator is then asked to paraphrase the corresponding questions or add distracting terms, while maintaining the general meaning of the question. Note that this is done independent of essentiality scores. For example, the modified question below changes two words in the question without affecting its core intent:
\vspace{-0.1cm}
\begin{framed}
\small 
\noindent 
\textbf{Original question: } A {\color{blue}fox} grows thicker {\color{blue}fur} as a season changes. This adaptation helps the fox to (A) find food(B) keep warmer(C) grow stronger(D) escape from predators \\ 
\textbf{Generated question: } An {\color{blue} animal} grows thicker {\color{blue} hair} as a season changes. This adaptation helps to (A) find food(B) keep warmer(C) grow stronger(D) escape from predators
\end{framed}
\vspace{-0.1cm}

While these generated questions should arguably remain correctly answerable by \tableilp, we found that this is often not the case. To investigate this, we curate a small dataset $Q_R$ with 12 questions 
(see the Appendix)
where,  despite having the required knowledge and a plausible reasoning chain, \tableilp\ fails. Interestingly, simply adding \et\ classifier via the \cascades\ system corrects 41.7\% of these mistakes. This error-reduction illustrates that the extra attention mechanism added to \tableilp\ via the concept of essential question terms helps it cope with distracting terms.
}

\section{Summary}
This chapter introduces and studies the notion of {\em essential question terms} with the goal of improving such QA solvers.
We illustrate the importance of essential question terms by showing that humans' ability to answer questions drops significantly when essential terms are eliminated from questions. We then develop a classifier that reliably (90\% mean average precision) identifies and ranks essential terms in questions. Finally, we use the classifier to demonstrate that the notion of question term essentiality allows state-of-the-art QA solvers for elementary-level science questions to make better and more informed decisions, improving performance by up to 5\%.




\part{Moving the Peaks Higher:  Designing More Challenging Datasets}
\label{part2:datasets}

\chapter{A Challenge Set for Reasoning on Multiple Sentences}
\label{chapter:multirc}
\epigraph{ 
``Human beings, viewed as behaving systems, are quite simple. The apparent complexity of our behavior over time is largely a reflection of the complexity of the environment in which we find ourselves.''
}{--- \textup{Herbert A. Simon}, The Sciences of the Artificial, 1968}

\section{Overview}
In this chapter we 
develop 
a reading comprehension challenge in which answering each of the questions requires reasoning over multiple sentences.\footnote{This chapter is based on the following publication: \cite{KCRUR18}.}

There is evidence that answering `single-sentence questions', i.e. questions that can be answered from a single sentence of the given paragraph, is easier than answering multi-sentence questions', which require multiple sentences to answer a given question. For example, \citep{RichardsonBuRe13} released a reading comprehension dataset that contained both single-sentence and multi-sentence questions; models proposed for this task yielded considerably better performance on the single-sentence questions than on the multi-sentence questions (according to \citep{NarasimhanBa15} accuracy of about $83\%$ and $60\%$ on these two types of questions, respectively).


There could be multiple reasons for this. First, multi-sentence reasoning seems to be inherently a difficult task. Research has shown that while complete-sentence construction emerges as early as first grade for many children, their ability to integrate sentences emerges only in fourth grade~\citep{BerningerNaBe11}. Answering multi-sentence questions might be more challenging for an automated system because it involves more than just processing individual sentences but rather combining linguistic, semantic and background knowledge across sentences---a computational challenges in itself. Despite these challenges, multi-sentence questions can be answered by humans and hence present an interesting yet reasonable goal for AI systems \citep{Davis14}.

\begin{wrapfigure}{r}{0.55\textwidth}
\includegraphics[scale=0.25]{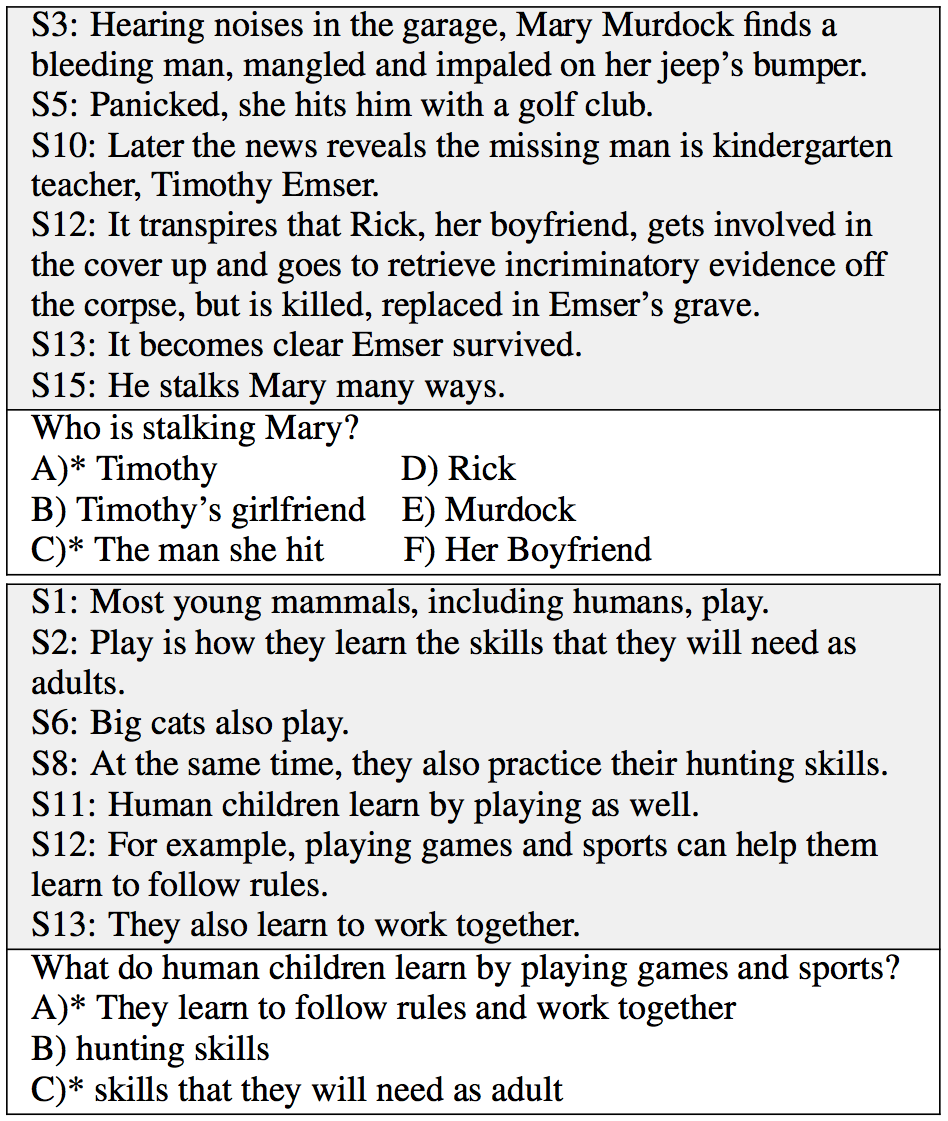}
\vspace{-0.3cm}
\caption{
\small
Examples from our \fancyname corpus. Each example shows relevant excerpts from a paragraph; multi-sentence question that can be answered by combining information from multiple sentences of the paragraph; and corresponding answer-options. The correct answer(s) is indicated by a *. Note that there can be multiple correct answers per question.}
\vspace{-0.2cm}
\label{figure:intro}
\end{wrapfigure}
In this work, we propose a
multi-sentence QA challenge in which questions can be answered only using information from multiple sentences. Specifically, we present \fancyname\ (Multi-Sentence Reading Comprehension)\footnote{\url{http://cogcomp.org/multirc/}}---a dataset of short paragraphs and multi-sentence questions that can be answered from the content of the paragraph. Each question is associated with several choices for answer-options, out of which {\em one or more} correctly answer the question. Figure~\ref{figure:intro} shows two examples from our dataset. Each instance consists of a multi-sentence paragraph, a question, and answer-options. All instances were constructed such that it is not possible to answer a question correctly without gathering information from multiple sentences. Due to space constraints, the figure shows only the relevant sentences from the original paragraph. The entire corpus consists of $871$ paragraphs and about $\sim6$k multi-sentence questions. 

The goal of this dataset is to encourage the research community to explore approaches that can do more than sophisticated lexical-level matching. To accomplish this, we designed the dataset with three key challenges in mind. (i) The number of correct answer-options for each question is not pre-specified. This removes the over-reliance of current approaches on answer-options and forces them to decide on the correctness of each candidate answer independently of others. In other words, unlike previous work, the task here is not to simply identify the best answer-option, but to evaluate the correctness of each answer-option individually. For example, the first question in Figure~\ref{figure:intro} can be answered by combining information from sentences 3, 5, 10, 13 and 15. It requires not only understanding that the stalker's name is Timothy but also that he is the man who Mary had hit. (ii) The correct answer(s) is not required to be a span in the text. For example, the correct answer, A, of the second question in Figure~\ref{figure:intro} is not present in the paragraph verbatim. It is instead a combination of two spans from 2 sentences: 12 and 13. Such answer-options force models to process and understand not only the paragraph and the question but also the answer-options. (iii) The paragraphs in our dataset have diverse provenance by being extracted from 7 different domains such as news, fiction, historical text etc., and hence are expected to be more diverse in their contents as compared to single-domain datasets. We also expect this to lead to  diversity in the types of questions that can be constructed from the passage.

Overall, we introduce a reading comprehension dataset that significantly differs from most other datasets available today in the following ways:
\begin{itemize}[topsep=0pt,itemsep=-3ex,partopsep=1ex,parsep=1ex,nolistsep]
    \item $\sim$6$k$ high-quality multiple-choice RC questions that are generated (and manually verified via crowdsourcing) to require integrating information from multiple sentences.\\[-1em]
    \item The questions are not constrained to have a {\em single} correct answer, 
    generalizing existing paradigms for representing answer-options.\\[-1em]
    \item Our dataset is constructed using 7 different sources, allowing more diversity in content, style, and possible question types.\\[-1em]
    \item We show a significant performance gap between current solvers and human performance, indicating an opportunity for developing sophistical reasoning systems. 
\end{itemize}

\newif\ifshort
\shorttrue
\newif\iftable
\tablefalse 

\section{Relevant Work}
\label{sec:related}

Some recent datasets proposed for machine comprehension pay attention to type of questions and reasoning required. For example, RACE~\citep{LXLYH17} attempts to incorporate different types of reasoning phenomena, and MCTest~\citep{RichardsonBuRe13} attempted to contain at least $50\%$ multi-sentence reasoning questions. However, since the crowdsourced workers who created the dataset were only encouraged, and not required, to write such questions, it is not clear how many of these questions actually require multi-sentence reasoning (see Sec.~\ref{subsec:multi-other-data}).
Similarly, only about 25\% of question in the RACE dataset require multi-sentence reasoning as reported in their paper. Remedia~\citep{HLBB99} also contains $5$ different types of questions (based on question words) but is a much smaller dataset. Other datasets which do not deliberately attempt to include multi-sentence reasoning, like SQuAD \citep{RZLL16} and the CNN/Daily Mail dataset \citep{HKGEKSB15}, suffer from even lower percentage of such questions (12\% and 2\% respectively~\citep{LXLYH17}). There are several other corpora which do not guarantee specific reasoning types, including MS MARCO~\citep{NRSGTMD16}, WikiQA \citep{YangYiMe15}, and TriviaQA \citep{JCWZ17}.

The complexity of reasoning required for a reading comprehension dataset would depend on several factors such as the source of questions or paragraphs; the way they are generated; and the order in which they are generated (i.e. questions from paragraphs, or the reverse). Specifically, paragraphs' source could influence the complexity and diversity of the language of the paragraphs and questions, and hence  the required level of reasoning capabilities. Unlike most current datasets which rely on only one or two sources for their paragraphs (e.g. CNN/Daily Mail and SQuAD rely only on news and Wikipedia articles respectively) our dataset uses $7$ different domains. 

Another factor that distinguishes our dataset from previously proposed corpora is the way answers are represented. Several datasets represent answers as multiple-choices with a single correct answer. While multiple-choice questions are easy to grade, coming up with non-trivial correct and incorrect answers can be challenging. 
Also, assuming exactly one correct answer (e.g., as in MCTest and RACE) inadvertently changes the task from choosing the correct answer to choosing the most likely answer.
Other datasets (e.g MS-MARCO and SQuAD) represent answers as a contiguous substring within the passage. This assumption of the answer being a span of the paragraph, limits the questions to those whose answer is contained verbatim in the paragraph. Unfortunately, it rules out more complicated questions whose answers are only implied by the text and hence require a deeper understanding. Because of these limitations, we designed our dataset to use multiple-choice representations, but without specifying the number of correct answers for each question.

\section{Construction of \fancyname}
\label{sec:construction}

In this section we describe our principles and methodology of dataset collection. This includes automatically collecting paragraphs, composing questions and answer-options through crowd-sourcing platform, and manually curating the collected data. We also summarize a pilot study that helped us design this process, and end with a summary of statistics of the collected corpus.

\subsection{Principles of design}
Questions and answers in our dataset are designed based on the following key principles:

\paragraph{Multi-sentenceness.} Questions in our challenge require models to use information from multiple sentences of a paragraph. This is ensured through explicit validation. We exclude any question that can be answered based on a single sentence from a paragraph.

\paragraph{Open-endedness.} Our dataset is not restricted to questions whose answer can be found verbatim in a paragraph. Instead, we provide a set of hand-crafted answer-options for each question. Notably, they can represent information that is not explicitly stated in the text but is only inferable from it (e.g. implied counts, sentiments, and relationships).

\paragraph{Answers to be judged independently.} The total number of answer options per question is variable in our data and we explicitly allow multiple correct and incorrect answer options (e.g. 2 correct and 1 incorrect options). As a consequence, correct answers cannot be guessed solely by a process of elimination or by simply choosing the best candidates out of the given options.\\

Through these principles, we encourage users to explicitly model the semantics of text beyond individual words and sentences, to incorporate extra-linguistic reasoning mechanisms, and to handle answer options independently of one another. 

\paragraph{Variability.} We encourage variability on different levels. Our dataset is based on paragraphs from multiple domains, leading to  linguistically diverse questions and answers. Also, we do not impose any restrictions on the questions, to encourage different forms of reasoning.

\subsection{Sources of documents}
The paragraphs used in our dataset are extracted from various sources. Here is the complete list of the text types and sources used in our dataset, and the number of paragraphs extracted from each category (indicated in square brackets on the right):  
\begin{enumerate}[topsep=0pt,itemsep=-3ex,partopsep=1ex,parsep=1ex,nolistsep]
    \item News: \hfill [121] 
    \begin{itemize}[nolistsep]
        \item CNN  \citep{HKGEKSB15} 
        \item WSJ  \citep{IBFFP08}
        \item NYT  \citep{IBFFP08}
    \end{itemize}
    \item Wikipedia articles \hfill [92] 
    \item Articles on society, law and justice  \citep{IdeSu06} \hfill [91]
    \item Articles on history and anthropology \citep{IBFFP08} \hfill [65]
    \item Elementary school science textbooks~\footnote{https://www.ck12.org} \hfill [153]
    \item 9/11 reports~\citep{IdeSu06} \hfill [72]
    \item Fiction: \hfill [277]
    \begin{itemize}[nolistsep]
        \item Stories from the Gutenberg project  
        \item Children stories from MCTest~\citep{RichardsonBuRe13} 
        \item Movie plots from CMU Movie Summary corpus~\citep{BammanO'Sm13}
    \end{itemize}
\end{enumerate}

From each of the above-mentioned sources we extracted paragraphs that had enough content. To ensure this we followed a $3$-step process. In the first step we selected top few sentences from paragraphs such that they contained $1$k-$1.5$k characters. To ensure coherence, all sentences were contiguous and extracted from the same paragraph. In this process we also discarded paragraphs that seemed to deviate too much from third person narrative style. For example, while processing Gutenberg corpus we considered files that had at least $5$k lines because we found that most of them were short poetic texts. In the second step, we annotated~\citep{KSZRCSRRLDTRMFWYSGUANLR18} the paragraphs and automatically filtered texts using conditions such as the average number of words per sentence; number of named entities; number of discourse connectives in the paragraph. These were designed by the authors of this paper after reviewing a small sample of paragraphs. A complete set of conditions is listed in Table~\ref{tab:paragraph-selection-condition}. Finally in the last step, we manually verified each paragraph and filtered out the ones that had formatting issues or other concerns that seemed to compromise their usability.


\begin{table}
    \centering
    \footnotesize{
    \centering
    \setlength\tabcolsep{10pt}
    \setlength\doublerulesep{\arrayrulewidth}
    \begin{tabular}{@{}ll@{}}
        \toprule
        Condition  & bound  \\
        \midrule
        Number of sentences & $\geq 6$ \& $\leq 18$ \\
        Number of NER(CoNLL) mentions & $\geq 2$  \\
        Avg. number of NER(CoNLL) mentions  & $\geq 0.2$ \\
        Number of NER(Ontonotes) mentions & $\geq 4$ \\
        Avg. number of NER(Ontonotes) mentions & $\geq 0.25$ \\
        Avg. number of words per sentence  & $\geq 5$ \\
        Number of coreference mentions & $\geq 3$  \\
        Avg. number of coreference mentions & $\geq 0.1$ \\
        Number of coreference relations  & $\geq 3$ \\
        Avg. number of coreference relations  & $\geq 0.08$ \\
        Number of coreference chains & $\geq 2$ \\
        Avg. number of coreference chains & $\geq0.1$ \\
        Number of discourse markers & $\geq2$ \\
\bottomrule
    \end{tabular}
    }
    \vspace{-0.1cm}
    \makeatletter\def\@captype{table}\makeatother
    \caption{Bounds used to select paragraphs for dataset creation.}
    \vspace{-0.1cm}
    \label{tab:paragraph-selection-condition}
\end{table}

\subsection{Pipeline of question extraction}
In this section, we delineate details of the process for collecting questions and answers. Figure~\ref{fig:steps} gives a high-level idea of the process. The first two steps deal with creating multi-sentence questions, followed by two steps for construction of candidate answers.

\begin{figure*}[ht]
    \centering
    \includegraphics[trim=0cm 10.7cm 0cm 0.55cm, clip=true,scale=0.48]{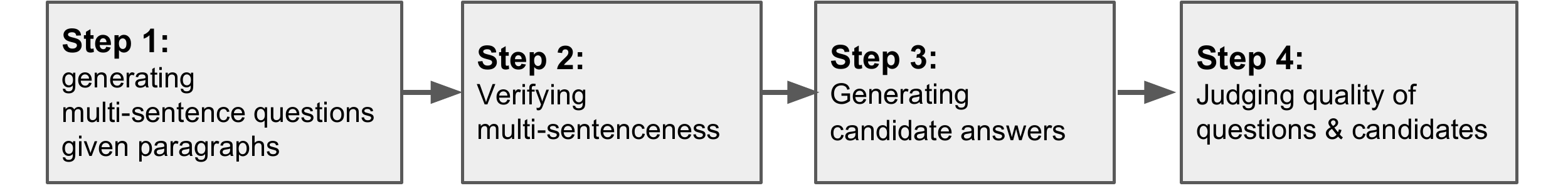}
    \vspace{-0.4cm}
    \caption{Pipeline of our dataset construction.}
    \label{fig:steps}
    \vspace{-0.2cm}
\end{figure*}

\paragraph{Step 1: Generating questions.}

The goal of the first step of our pipeline is to collect multi-sentence questions. We show each paragraph to $5$ turkers and ask them to write $3$-$5$ questions such that:  
(1) the question is answerable from the passage, and
(2) only those questions are allowed whose answer cannot be determined from a single sentence. 
We clarify this point by providing example paragraphs and questions. In order to encourage turkers to write meaningful questions that fit our criteria, we additionally ask them for a correct answer and for the sentence indices required to answer the question. To ensure the grammatical quality of the questions collected in this step, we limit the turkers to the countries with English as their major language. 
After the acquisition of questions in this step, we filter out questions which required less than 2 or more than 4 sentences to be answered; we also run them through an automatic spell-checker\footnote{Grammarly: www.grammarly.com} and manually correct questions regarding typos and unusual wordings.

\paragraph{Step 2: Verifying multi-sentenceness of questions. }
In a second step, we verify that each question can only be answered using more than one sentence. For each question collected in the previous step, we create question-sentence pairs by pairing it with each of the sentences necessary for answering it as indicated in the previous step. For a given question-sentence pair, we then ask turkers to annotate if they could answer the question from the sentence it is paired with (binary annotation). The underlying idea of this step is that a multi-sentence question would not be answerable from a single sentence, hence turkers should not be able to give a correct answer for any of the question-sentence pair.
Accordingly, we determine a question as requiring multiple sentences only if the correct answer cannot be guessed from any single question-sentence pair.
We collected at least 3 annotations per pair, and to avoid sharing of information across sentences, no two pairs shown to a turker came from the same paragraph. We aggregate the above annotations for each question-answer pair and
retain only those questions for which no pair was judged as answerable by a majority of turkers.

\paragraph{Step 3: Generating answer-options.}
In this step, we collect answer-options that will be shown with each question. Specifically, for each verified question from the previous steps, we ask 3 turkers to write as many correct and incorrect answer options as they can think of. In order to not curb creativity, we do not place a restriction on the number of options they have to write. We explicitly ask turkers to design difficult and non-trivial incorrect answer-options (e.g. if the question is about a person, a non-trivial incorrect answer-option would be other people mentioned in the paragraph). 

After this step, we perform a light clean up of the candidate answers by manually correcting minor errors (such as typos), completing 
incomplete sentences and rephrasing any ambiguous sentences.
We further make sure there is not much repetition in the answer-options, to prevent potential exploitation of correlation between some candidate answers in order to find the correct answer. For example, we drop obviously duplicate answer-options (i.e.\ identical options after lower-casing, lemmatization, and removing stop-words).   

\paragraph{Step 4: Verifying quality of the dataset.}
This step serves as the final quality check for both questions and the answer-options generated in the previous steps. We show each paragraph, its questions, and the corresponding answer-options to $3$ turkers, and ask them to indicate if they find any errors (grammatical or otherwise), in the questions and/or answer-options. We then manually review, and correct if needed, all erroneous questions and answer-options. This ensures that we have meaningful questions and answer-options. In this step, we also want to verify that the correct (or incorrect) options obtained from Step 3 were indeed correct (or incorrect). For this, we additionally ask the annotators to select all correct answer-options for the question. If their annotations did not agree with the ones we had after Step 3 (e.g. if they unanimously selected an `incorrect' option as the answer), we manually reviewed and corrected (if needed) the annotation. 



\subsection{Pilot experiments}
The $4$-step process described above was a result of detailed analysis and substantial refinement after two small pilot studies. 

In the first pilot study, we ran a set of $10$ paragraphs extracted from the CMU Movie Summary Corpus through our pipeline. Our then pipeline looked considerably different from the one described above. We found the steps that required turkers to write questions and answer-options to often have grammatical errors, possibly because a large majority of turkers were non-native speakers of English. This probslem was more prominent in questions than in answer-options. Because of this, we decided to limit the task to native speakers. Also, based on the results of this pilot, we overhauled the instructions of these steps by including examples of grammatically correct---but undesirable (not multi-sentence)---questions and answer-options, in addition to several minor changes. 

Thereafter, we decided to perform a manual validation of the verification steps (current Steps 2 and 4). For this, we (the authors of this paper) performed additional annotations ourselves on the data shown to turkers, and compared our results with those provided by the turkers. We found that in the verification of answer-options, our annotations were in high agreement ($98\%$) with those obtained from mechanical turk. However, that was not the case for the verification of multi-sentence questions. We made several further changes to the first two steps. Among other things, we clarified in the instructions that turkers should not use their background knowledge when writing and verifying questions, and also included negative examples of such questions. Additionally, when turkers judged a question to be answerable using a single sentence, we decided to encourage (but not require) them to guess the answer to the question. This improved our results considerably, possibly because it forced  annotators to think more carefully about what the answer might be, and whether they \textit{actually} knew the answer or they just \emph{thought} that they knew it (possibly because of background knowledge or because the sentence contained a lot of information relevant to the question). Guessed answers in this step were only used to verify the validity of multi-sentence questions. They were not used in the dataset or subsequent steps.

After revision, we ran a second pilot study in which we processed a set of $50$ paragraphs through our updated pipeline. This second pilot confirmed that our revisions were helpful, but thanks to its larger size,  also allowed us to identify a couple of borderline cases for which additional clarifications were required. Based on the results of the second pilot, we made some additional minor changes and then decided to apply the pipeline for creating the final dataset. 


\subsection{Verifying multi-sentenceness}
\label{subsec:multi-other-data}
While collecting our dataset, we found that, even though Step 1 instructed turkers to write multi-sentence questions, not all generated questions indeed required multi-sentence reasoning. This happened even after clarifications and revisions to the corresponding instructions, and we attribute it to honest mistakes. Therefore, we designed the subsequent verification step (Step 2). 

There are other datasets which aim to include multi-sentence reasoning questions, especially MCTest. 
Using our verification step, we 
systematically verify their multi-sentenceness. For this, we conducted a small pilot study on about $60$ multi-sentence questions from MCTest. As for our own verification, we created question-sentence pairs for each question and asked annotators to judge whether they can answer a question from the single sentence shown. Because we did not know which sentences contain information relevant to a question, we created question-sentence pairs using all sentences from a paragraph. After aggregation of turker annotations, we found that about half of the questions annotated as multi-sentence could be answered from a single sentence of the paragraph. This study, though performed on a subset of the data, underscores the necessity of rigorous verification step for multi-sentence reasoning when studying this phenomenon.

\subsection{Statistics on the dataset}
We now provide a brief summary of \fancyname. 
Overall, it contains roughly $\sim$~$6k$ multi-sentence questions collected for about $+800$ paragraphs.\footnote{We will also release the $3.7k$ questions that did not pass Step 2. Though not multi-sentence questions, they could be a valuable resource on their own.}  The median number of correct and total answer options for each question is $2$ and $5$, respectively. Additional statistics are given in Table~\ref{tab:statistics:multirc}.

\begin{wrapfigure}{r}{0.5\textwidth}
    \footnotesize
    \centering
    \resizebox{210px}{!}{
    \setlength\extrarowheight{-3pt}
    \begin{tabular}{l|c}
        \toprule
        Parameter & Value \\ 
        \hline 
         \# of paragraphs & 871\\ 
        \# of questions & 9,872 \\ 
        \# of multi-sentence questions & 5,825  \\ 
        avg \# of candidates (per question) &  5.44 \\ 
         avg \# of correct answers (per question) & 2.58  \\ 
        \hline 
         avg paragraph length (in sentences) & 14.3 (4.1) \\  
         avg paragraph length (in tokens) & 263.1 (92.4) \\  
         avg question length (in tokens) & 10.9 (4.8) \\ 
         avg answer length (in tokens) & 4.7 (5.5) \\ 
        \hline         
        \% of yes/no/true/false questions & 27.57\% \\ 
        \hline 
        avg \# of sent. used for questions & 2.37 (0.63) \\ 
        avg distance between the sent.'s used & 2.4 (2.58) \\ 
        \hline
        \% of correct answers verbatim in paragraph & 34.96\%  \\ 
        \% of incorrect answers verbatim in paragraph & 25.84\%  \\ 
        \bottomrule
    \end{tabular}
    }
    \vspace{-0.1cm}
    \makeatletter\def\@captype{table}\makeatother
    \caption{Various statistics of our dataset. Figures in parentheses represent standard deviation.}
    \label{tab:statistics:multirc}
\end{wrapfigure}

In Step 1, we also asked annotators to 
identify sentences required to answer a given question. We found that answering each question required $2.4$ sentences on average. Also, required sentences are often not contiguous, and the average distance between sentences is $2.4$. Next, we analyze the types of questions in our dataset. Figure~\ref{fig:first-chunk-distribution} shows the count of first word(s) for our questions. We can see that while the popular question words (\emph{What}, \emph{Who}, etc.) are very common, there is a wide variety in the first word(s) indicating a diversity in question types. About $28\%$ of our questions require binary decisions (true/false or yes/no). 

\begin{figure*}
    \includegraphics[scale=0.41]{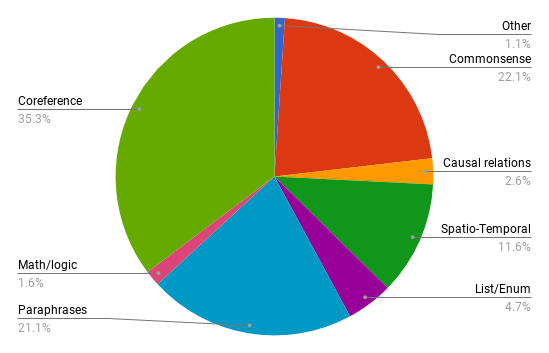}
    \includegraphics[scale=0.41]{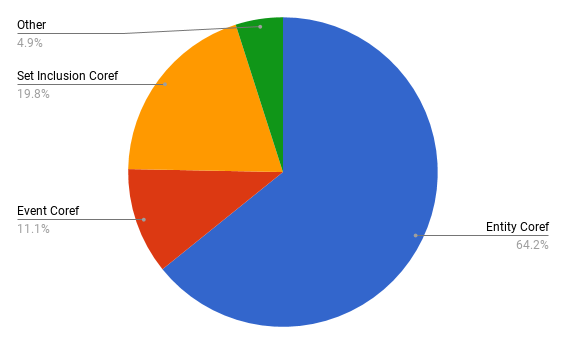}
    \caption{Distribution of (left) general phenomena; (right) variations of the ``coreference" phenomena.  }
    \label{fig:phenomena}
\end{figure*}

We randomly selected 60 multi-sentence questions from our corpus and asked two independent annotators to label them with the type of reasoning phenomenon required to answer them.\footnote{The annotations were  adjudicated by two authors of this paper.} During this process, the annotators were shown a list of common reasoning phenomena (shown below), and they had to identify one or more of the phenomena relevant to a given question. The list of phenomena shown to the annotators included the following categories: mathematical and logical reasoning, spatio-temporal reasoning, list/enumeration, coreference resolution (including implicit references, abstract pronouns, event coreference, etc.), causal relations, paraphrases and contrasts (including lexical relations such as synonyms, antonyms), commonsense knowledge, and `other'. The categories were selected after a manual inspection of a subset of questions by two of the authors. The annotation process revealed that answering questions in our corpus requires a broad variety of reasoning phenomena. The left plot in Figure~\ref{fig:phenomena} provides detailed results. 

The figure shows that a large fraction of questions require coreference resolution, and a more careful inspection revealed that there were different types of coreference phenomena at play here. To investigate these further, we conducted a follow-up experiment in which manually annotated all questions that required coreference resolution into finer categories. Specifically, each question was shown to two annotators who were asked to select one or more of the following categories:  entity coreference (between two entities), event coreference	(between two events), set inclusion coreference (one item is part of or included in the other) and `other'. Figure~\ref{fig:phenomena} (right) shows the results of this experiment. We can see that, as expected, entity coreference is the most common type of coreference resolution needed in our corpus. However, a significant number of questions also require other types of coreference resolution.
\begin{figure}
    \centering
    \includegraphics[trim=1.3cm 1.1cm 0.0cm 0.0cm, scale=0.36]{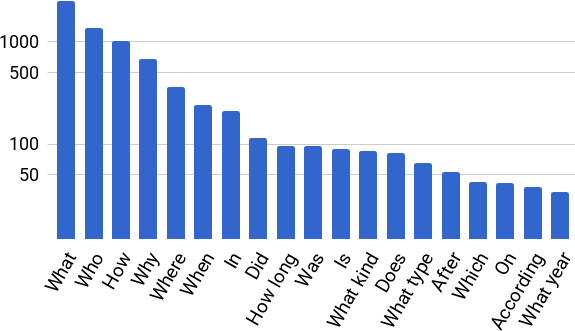}
    \caption{Most frequent first chunks of the questions (counts in log scale). }
    \label{fig:first-chunk-distribution}
\end{figure}

\section{Analysis}
\label{sec:analysis}

In this section, we provide a quantitative analysis of several baselines for our challenge. 

\paragraph{Evaluation Metrics.}
We define precision and recall for a question $q$ as:
$\text{Pre}(q) = \frac{|A(q) \cap \hat{A}(q)|}{|\hat{A}(q)|}$ and $\text{Rec}(q) = \frac{|A(q) \cap \hat{A}(q)|}{|A(q)|}$, where $A(q)$ and $\hat{A}(q)$ are the sets of correct and selected answer-options. 
We define (macro-average) F1$_m$ as the harmonic mean of average-precision $avg_{q \in Q}(\text{Pre}(q))$ and average-recall $avg_{q \in Q}(\text{Rec}(q))$ with $Q$ as the set of all questions. 

Since by design, each answer-option can be judged independently, we consider another metric, F1$_a$, evaluating binary decisions on all the answer-options in the dataset. We define F1$_a$ to be the harmonic mean of $\text{Pre}(Q)$ and $\text{Rec}(Q)$, with $\text{Pre}(Q) = \frac{|A(Q) \cap \hat{A}(Q)|}{|\hat{A}(Q)|}$; $A(Q)=\bigcup_{q \in Q} A(q)$; and similar definitions for $\hat{A}(Q)$ and $\text{Rec}(Q)$.

\subsection{Baselines}

\paragraph{Human.} Human performance provides us with an estimate of the best achievable results on datasets. Using mechanical turk, we ask 4 people (limited to native speakers) to solve our data. We evaluate score of each label by averaging the decision of the individuals. 

\paragraph{Random.}
To get an estimate on the lower-bound we consider a random baseline, where each answer option is selected as correct with a probability of 50\% (an unbiased coin toss). The numbers reported for this baseline represent the expected outcome (statistical expectation). 

\paragraph{\lucene}(information retrieval baseline). This baseline selects answer-options that best match sentences in a text corpus~\citep{CEKSTTK16}. Specifically, for each question $q$ and answer option $a_i$, the \lucene\ solver sends $q + a_i$ as a query to a search engine (we use Lucene) on a corpus, and returns the search engine's score for the top retrieved sentence $s$, where $s$ must have at least one non-stopword overlap with $q$, and at least one with $a_i$. 

We create two versions of this system. In the first variation IR(paragraphs) we create a corpus of sentences extracted from all the paragraphs in the dataset. In the second variation, IR(web) in addition to the knowledge of the paragraphs, we use extensive external knowledge extracted from the web (Wikipedia, science textbooks and study guidelines, and other webpages), with $5 \times 10^{10}$ tokens (280GB of plain text). 

\paragraph{SurfaceLR}(logistic regression baseline). As a simple baseline that makes use of our small training set, we reimplemented and trained a logistic regression model using word-based overlap features. As described in \citep{MHBSZ18}, this baseline takes into account the lengths of a text, question and each answer candidate, as well as indicator features regarding the (co-)occurrences of any words in them.

\paragraph{SemanticILP}(semi-structured baseline). This state-of-the-art solver, originally proposed for science questions and biology tests, uses a semi-structured representation to formalize the scoring problem as a subgraph optimization problem over multiple layers of semantic abstractions~\citep{KKSR18}. Since the solver is designed for multiple-choice with single-correct answer, we adapt it to our setting by running it for each answer-option. Specifically for each answer-option, we create a single-candidate question, and retrieve a real-valued score from the solver.  

\begin{wrapfigure}{r}{0.5\textwidth}
\centering
\small {
\setlength\extrarowheight{-6pt}
\begin{tabular}{@{}lcc|cc@{}}
\toprule
                            & \multicolumn{2}{c}{Dev} & \multicolumn{2}{c}{Test} \\ 
\cmidrule(lr){2-3}
\cmidrule(lr){4-5}
                            & F1$_m$  & F1$_a$ & F1$_m$  & F1$_a$  \\
\midrule
Random                      & 44.3 & 43.8 & 47.1 & 47.6  \\
\lucene (paragraphs)	    & 64.3 & 60.0 &	54.8 &	53.9 \\
SurfaceLR            	    &  66.1 & 63.7  & 66.7 &	63.5 \\
\hline
Human 	                    & 86.4 & 83.8 & 84.3 & 81.8 \\ 
\bottomrule
\end{tabular}
}
\makeatletter\def\@captype{table}\makeatother
\caption{
\small
Performance comparison for different baselines tested on a subset of our dataset (in percentage). There is a significant gap between the human performance and current statistical methods. }
\label{table:mainResults}
\end{wrapfigure}
\paragraph{BiDAF}(neural network baseline). As a neural baseline, we apply this solver by \cite{SKFH16}, which was originally proposed for SQuAD but has been shown to generalize well to another domain~\citep{MinSeHa17}. Since BiDAF was designed for cloze style questions, we apply it to our multiple-choice setting following the procedure by \cite{KSSCFH17}: Specifically, we score each answer-option by computing the similarity value of it's output span with each of the candidate answers, computed by phrasal similarity tool of \cite{WBGLR15}.

\subsection{Results} 
\begin{wrapfigure}{r}{0.5\textwidth}
    \centering
    \includegraphics[trim=0.45cm 0.55cm 0cm 0.45cm, scale=0.63]{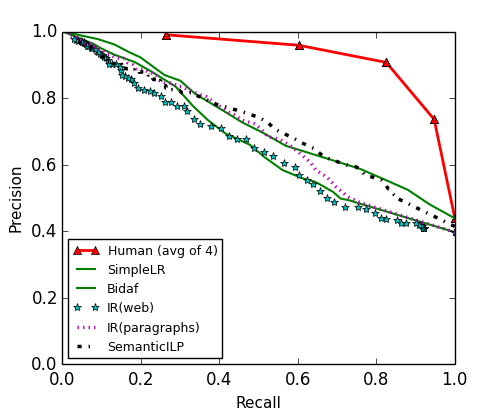}
    \vspace{-0.1cm}
    \caption{PR curve for each of the baselines. There is a considerable gap with the baselines and human. }
    \label{fig:pr-curve}
\end{wrapfigure}
To get a sense of our dataset's hardness, we evaluate both human performance and multiple computational baselines. Each baseline scores an answer-option with a real-valued score, which we threshold to decide whether an answer option is selected or not, where the threshold is tuned on the development set. Table~\ref{table:mainResults} shows performance results for different baselines. The significantly high human performance shows that humans do not have much difficulties in answering the questions. Similar observations can be made in Figure~\ref{fig:pr-curve} where we plot $avg_{q \in Q}(\text{Pre}(q))$ vs. $avg_{q \in Q}(\text{Rec}(q))$, for different threshold values.

\section{Summary}
To motivate the community to work on more challenging forms of natural language comprehension, in this chapter we discussed  a dataset that requires reasoning over multiple sentences. 
We solicit and verify questions and answers for this challenge through a 4-step crowdsourcing experiment.
Our challenge dataset contains $\sim$6$k$ questions for $+800$ paragraphs across 7 different domains (elementary school science, news, travel guides, fiction stories, etc) bringing in linguistic diversity to the texts and to the questions wordings. On a subset of our dataset, we found human solvers to achieve an F1-score of $86.4\%$.
We analyze a range of baselines, including a recent state-of-art reading comprehension system, and demonstrate the difficulty of this challenge,
The dataset is the first to study multi-sentence inference at scale, with an open-ended set of question types that requires reasoning skills.


An additional important aspect of this work is that we challenged the community to change the harmful ``fixed'' test set methodology, and committed to updating the test set every few months.  

\chapter{A Question Answering Benchmark for Temporal Common-sense}
\label{chapter:dataset:tacoqa:commonsense}
\epigraph{ 
``Everything changes and nothing stands still.''
}{--- \textup{Heraclitus}, 535 BC - 475 BC}

\section{Overview}

Automating natural language understanding requires models that are informed
by commonsense knowledge and the ability to reason with it in both common and unexpected
situations. The NLP community has started in the last few year to investigate how to acquire such knowledge \cite{ForbesCh17,ZRDD17,YBWD18,RSASC18,BauerWaBa18,TDGYBC18,ZBSC18}.

This work studies a specific type of commonsense, {\em temporal commonsense}.\footnote{This chapter is based on the following publication: \cite{ZKNR19}.} For instance, given two events \emph{``going on a vacation''} and \emph{``going for a walk,''} most humans would know that a vacation is typically longer and occurs less often than a walk, but our programs currently do not know that. 

\begin{figure}
    \centering
    \includegraphics[width=0.8\textwidth]{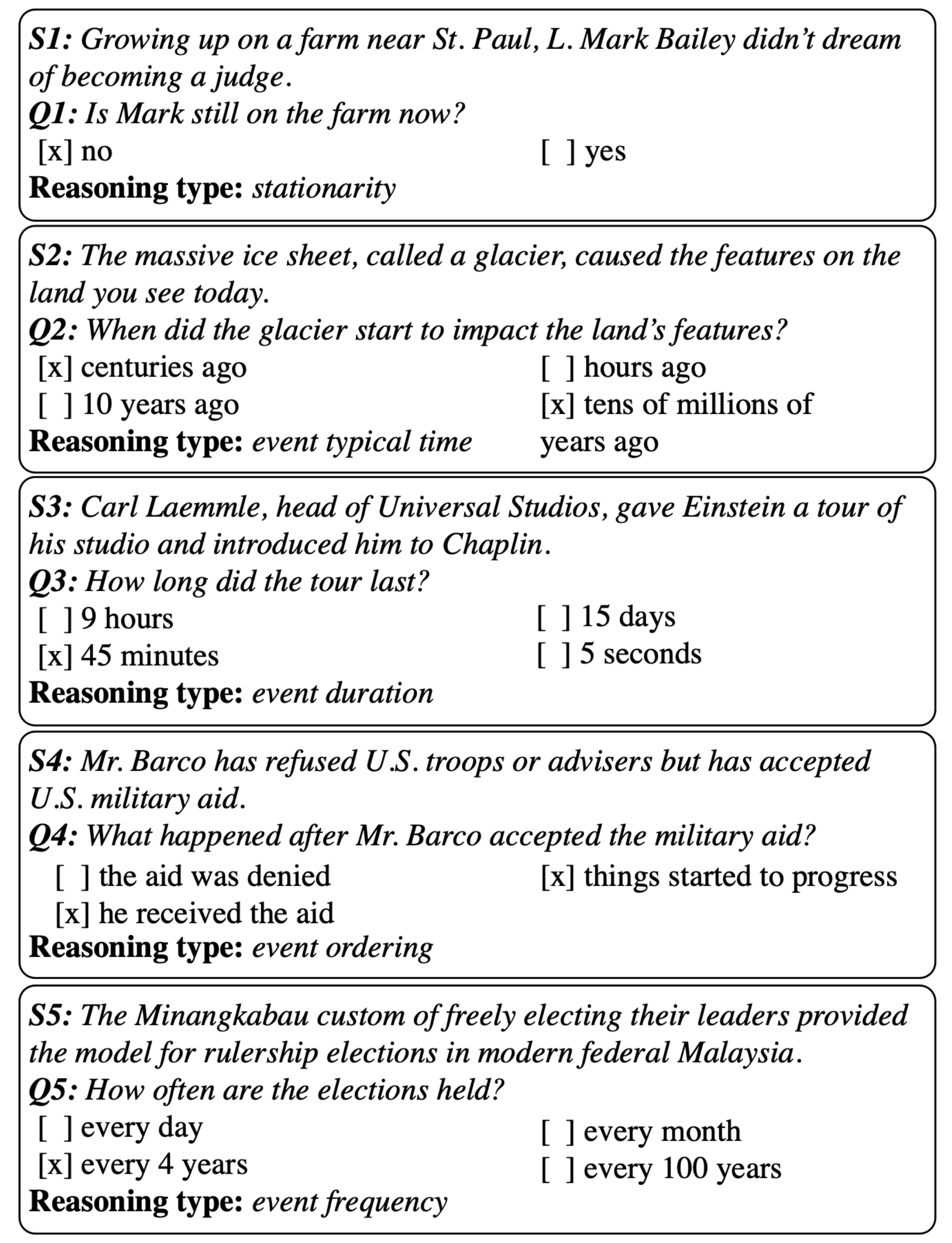}
    \caption{\small Five types of temporal commonsense in \datasetname. Note that a question may have multiple answers. \ignore{Correct answers marked by `x'.}}
    \label{fig:intro:example}
\end{figure}

Temporal commonsense has received limited attention so far. \textbf{Our first contribution} is that, to the best of our knowledge, we are the first to systematically study and quantify performance on a range of temporal commonsense phenomena. Specifically, we consider five temporal properties: \emph{duration} (how long an event takes),  \emph{temporal ordering} (typical order of events), \emph{typical time} (when an event happens), \emph{frequency} (how often an event occurs), and \emph{stationarity} (whether a state holds for a very long time). Previous works have investigated some of them, either explicitly or implicitly (e.g., duration \cite{DivyeKhilnaniJu11,Williams12} and ordering \cite{ChklovskiPa04,NWPR18}), but none of them have defined or studied all aspects of temporal commonsense in a unified framework. \cite{KozarevaHo11} came close, when they defined a few temporal aspects to be investigated, but failed short of distinguishing in text and quantifying performances on these.

\ignore{
In this work, we examine the existing systems in their ability to handle \emph{temporal commonsense} or such world knowledge. 
In particular, we consider five temporal properties: \emph{duration} (typical event duration), \emph{stationarity} (whether an event lasts long-enough), \emph{temporal ordering} (typical order of events), \emph{typical time} (when do events usually happen), and \emph{frequency} (\qiang{how often an event occurs}). Our goal is to study how existing methods could understand all these attributes.
}


Given the lack of an evaluation standards and datasets for temporal commonsense, \textbf{our second contribution} is the collection of a new dataset dedicated for it, \datasetname{} (short for \textbf{t}empor\textbf{a}l \textbf{co}mmon-sense \textbf{q}uestion \textbf{a}nswering).\footnote{The dataset and code will be released upon publication.}
\datasetname{} is constructed via crowdsourcing with three meticulously-designed stages to guarantee its quality. 
An entry in \datasetname{} contains a \emph{sentence} providing context information, a \emph{question} requiring temporal commonsense, and \emph{candidate answers} with or without correct ones (see Fig.~\ref{fig:intro:example}).
More details about \datasetname{} are  in Sec.~\ref{sec:tacoqa}.

\ignore{
To have a natural evaluation to such properties, this paper introduces \datasetname, \textbf{t}empor\textbf{a}l \textbf{co}mmon-sense \textbf{q}uestion \textbf{a}nswering dataset, to serve as a diagnostic environment for further investigations dedicated to temporal commonsense. 
Our dataset is constructed with three meticulously-designed crowdsoursing stages. 
Figure~\ref{fig:intro:example} shows some examples for each class in \datasetname. 
The input contains a \emph{sentence} providing context information, a \emph{question} asking for temporal information, and \emph{candidate answers} with or without the correct ones in it.
To ensure the answers requires commonsense, no questions can be answered with a phrase explicitly contained in the context. 
Moreover, we made sure that the question can still be answered with high agreement by human.
For example, \emph{Q1} asks whether that Mark is still on the farm. Given somebody being at some place's transient property, in addition to Mark is now a judge, humans can agree that he is no longer on the farm, even though the answer is not explicitly mentioned in \emph{S1}.
Similarly, \emph{Q5} asks for the frequency of large mammals' rest, and humans know that they usually rest every day, not too frequent (``once a minute'') or too infrequent (``once every three days'').
\qiangcomment{I feel that this paragraph can be shortened. It's too detailed right now.}
}

\textbf{Our third contribution} is that we propose multiple systems, including \emph{ESIM}, \bert{} and their variants, for this task. \datasetname{} allows us to investigate how state-of-the-art NLP techniques do on temporal commonsense tasks. Results in Sec.~\ref{sec:exp} show that, despite a significant improvement over random-guess baselines, \bert{} is still far behind human performance on temporal commonsense reasoning, indicating that existing NLP techniques still have limited capability of capturing high-level semantics like time.

\ignore{
The temporal problems we are interested here are different from what has been investigated in the field. 
While there are many works on understanding \emph{time}, a vast majority of efforts focus on problems that mostly involve semi-direct mentions in text: 
for instance, extraction and normalization of \emph{temporal mentions}, \emph{events}, and \emph{relations}, 
\cite{Str{\"o}tgenGe10,chang2012sutime,ning2017structured}. 
However, many instances we consider here contains involve a  variety of temporal phenomena that appear implicitly, without direct temporal references. 
\qiangcomment{I feel that this discussion is not in the right position.}
}

\ignore{
\textbf{The contributions of this work can be summarized as follows.}\qiangcomment{I'm going to move the bullets below to different places in-text. Will work on it in the afternoon.}
\begin{itemize}[noitemsep,leftmargin=0.4cm]
    \item A high-quality dataset on \emph{temporal commonsense}, \datasetname, is collected via crowdsourcing, which can significantly facilitate further research along this direction.\footnote{The dataset and codes will be released upon publication.}
    \ignore{To facilitate this research towards the problem of \emph{temporal commonsense}, we create a high-quality dataset for this task using crowdsourcing.}
    
    \item Multiple systems based on the state-of-the-art (including \emph{ESIM} and \bert) are proposed and evaluated on \datasetname. 
    \ignore{We  develop
    multiple competitive systems using state-of-the-art machinery available in the field. Additionally we show that humans are able to achieve significantly higher score on this task. }
    
    \item We show that there remains a significant performance gap between the proposed systems and humans, indicating the limitations of current NLP techniques and pointing out potential research directions.
    \ignore{We provide extra analysis on limitations of current techniques and remaining open challenges.} 
    
\end{itemize}
}

\section{Related Work}
commonsense has been a very popular topic in recent years and existing NLP works have mainly investigated the acquisition and evaluation of commonsense in the physical world, including but not limited to, size, weight, and strength \cite{ForbesCh17}, roundness and deliciousness \cite{YBWD18}, and intensity \cite{CWPAC18}. In terms of commonsense on ``events'', \citet{RSASC18} investigated the intent and reaction of participants of an event, and \citet{ZBSC18} tried to select the most likely subsequent event. As far as we know, no existing work has focused on temporal commonsense yet.

\ignore{
\paragraph{commonsense understanding.}
The majority of the work in this area has focused on formalizing and acquisition of commonsense \emph{knowledge}, either in the form of manually curated knowledge-bases~\cite{LiuSi04}, exploiting weak signals from sensors like vision~\cite{lin2015don,ZBSC18} or  
a combination of weak signals incorporated within relational frameworks ~\cite{ForbesCh17}. Within this literature, to the best of our knowledge, this 
is the first work along the commonsense thread which systematically focuses on the commonsense understanding of \emph{time}.
}

There have also been many works trying to understand time in natural language but not necessarily with respect to commonsense, such as the extraction and normalization of temporal expressions~\cite{LADZ14}, temporal relation extraction \cite{NZFPR18}, and timeline construction \cite{LeeuwenbergMo18}. Among these, some works are implicitly on temporal commonsense, such as event durations \cite{Williams12,VempalaBlPa18}, typical temporal ordering \cite{ChklovskiPa04,NWPR18}, and script learning (i.e., what happens next after certain events) \cite{Granroth-WildingCl16,LiDiLi18}. 
However, either in terms of datasets or approaches, existing works did not study all five types of temporal commonsense in a unified framework as we do here.

\ignore{
\paragraph{Temporal reasoning.}
The field has already studied some tasks related to \emph{time}:
the extraction and normalization of temporal expressions~\cite{LADZ14}, temporal ordering~\cite{NZFPR18}, and duration expressed in text~\cite{VempalaBlPa18}.
There have been some works conceptually relevant to commonsense, independently spreading over many subareas, including \citet{ChklovskiPa04,NWPR18} on temporal ordering of verbs, \citet{Williams12} on duration, and \citet{Granroth-WildingCl16,LiDiLi18} on script learning (i.e., what happens next after certain events). 
All these works were intended to provide prior knowledge for certain aspects of temporal reasoning, but none of them systematically studies a broad range of aspects to the temporal commonsense, which is what this paper is aiming for.
}


\ignore{
This area has advanced in multiple interrelated directions. With respect to harnessing temporal patterns in the wild, there are many proposals~\cite{williams2012extracting,DivyeKhilnaniJu11}. 
In parallel, there is a line of work on
creating richer temporal reasoning formalism~\cite{NFWR18,leeuwenberg2018temporal}. 
Unfortunately, majority the work in this area has not explored 
the extent to which these ideas apply to scenarios which involve commonsense understanding/reasoning with respect to \emph{time}. Our dataset is an opportunity for this area to do such explorations. }



Instead of working on each individual aspect of temporal commonsense, we formulate the problem as a machine reading comprehension task in the format of question-answering (QA). The past few years have also seen significant progress on QA \cite{CCEKSST18,ORMTP18,MHBSZ18}, but mainly on {\em general} natural language comprehension tasks without tailoring it to test {\em specific} reasoning capabilities such as temporal commonsense. Therefore, a new dataset like \datasetname{} is strongly desired.

\ignore{
\paragraph{Machine Reading Comprehension.}
This area has witnessed many recent datasets in the past few years. 
Many of the proposals provide general natural language comprehension tasks, without tailoring it to any specific reasoning ability.   
To name a few, 
\namecite{CCEKSST18,ORMTP18,MHBSZ18} are among such works. 
While these are 
helpful to measure general language 
understanding,
they are not very useful when evaluating narrow 
aspects of language. 
Our reading comprehension dataset, while being naturally generated by 
crowdsourcers,
it is 
focused on class of language understanding problems (various properties of time)
and contains 
granular annotations for the required skills. 
}



\begin{table}[]
\centering
\footnotesize
\resizebox{0.65\textwidth}{!}{
\begin{tabular}{lcc}
\toprule
\multicolumn{2}{l}{Measure}                            & Value         \\
\cmidrule(r){1-2}  \cmidrule(r){3-3}
\multicolumn{2}{l}{\# of unique questions}             & 1893              \\
\multicolumn{2}{l}{\# of unique question-answer pairs}             & 13,225              \\
\multicolumn{2}{l}{avg. sentence length}            & 17.8              \\
\multicolumn{2}{l}{avg. question length}               & 8.2              \\
\multicolumn{2}{l}{avg. answer length}                 & 3.3              \\
\midrule
Category                           & \# questions & avg \# of candidate \\
\cmidrule(r){1-1} \cmidrule(r){2-2} \cmidrule(r){3-3}
\emph{event frequency}                    & 433 & 8.5 \\
\emph{event duration}                     & 440 & 9.4 \\
\emph{event stationarity}                 & 279 & 3.1 \\
\emph{event ordering}                     & 370 & 5.4 \\
\emph{event typical time}                 & 371 & 6.8 \\
\bottomrule
\end{tabular}
}
\caption{\small Statistics of \datasetname. }
    \label{tab:statistics}
\end{table}

\section{Construction of \datasetname}
\label{sec:tacoqa}
\ignore{
In the next steps we describe  a
multi-step crowdsourcing scheme, 
resulting from 
 detailed analysis and substantial refinements after multiple pilots studies. 
}
We describe our crowdsourcing scheme for \datasetname{} that is designed after extensive pilot studies. The multi-step scheme asks annotators to generate questions, validate questions, and then label candidate answers.
We use Amazon Mechanical Turk 
and restrict our tasks to English-speakers only. Before working on our task, annotators need to read through our guidelines and pass a qualification test designed to ensure their understandings.\footnote{
Our dataset and some related details (such as, our annotation interfaces, guidelines and qualification tests) are available at the following link: { \url{https://bit.ly/2tZ1mkd}}
}



\paragraph{Step 1: Question generation.}
In the first step, we ask crowdsourcers to generate questions given a sentence.
We randomly select 630 sentences from MultiRC~\cite{KCRUR18} (70 from each of the 9 domains) as input sentences. 
To make sure that the questions indeed require temporal commonsense knowledge, we instruct annotators to follow two requirements when generating questions:
(a) ``temporal'' questions, from one of our five categories
(see Fig.~\ref{fig:intro:example}); (b) not having direct answers mentioned in the given sentence.  
We also ask annotators to provide a correct answer for each of their questions to make sure that the questions are answerable at least by themselves.

\paragraph{Step 2: Question verification.}
To improve the quality of the questions generated in Step~1, we further ask two different annotators to check (a) whether the two requirements above are satisfied and (b) whether there exist grammatical or logical errors.
We keep a question if and only if both annotators agree on its quality; since the annotator who provided the question in Step~1 also agrees on it, this leads to a [3/3] agreement for each question. For the questions that we keep, we continue to ask annotators to give one correct answer and one incorrect answer,
which serve as a seed set for automatic answer expansion in the next step.

\paragraph{Step 3: Candidate answer expansion.}
In the previous steps, we have collected 3 positive and 2 negative answers for each question.\footnote{One positive answer from Step~1; one positive and one negative answer from each of the two annotators in Step~2.} Step~3 aims to automatically expand this set of candidate answers by three approaches. First, we use a set of rules to extract temporal terms (e.g. ``a.m.", ``1990", ``afternoon", ``day"), or numbers and quantities (``2", ``once"), which are replaced by terms randomly selected from a list of temporal units (``second''), adjectives (``early''), points ( ``a.m.'') and adverbs (``always'').
Examples are ``2 a.m." $\rightarrow$ ``3 p.m.", ``1 day" $\rightarrow$ ``10 days", ``once a week"$\rightarrow$ ``twice a month". 
Second, we mask each individual token in a candidate answer and use \bert{}~\cite{DCLT18} to predict them; we rank those predictions by the confidence level of \bert{} and keep the top ones.
Third, for those candidates representing events, typically there are no temporal terms in them.
We then create a pool of $60k$ event phrases using PropBank~\cite{KingsburyPa02},
and retrieve the most similar ones to a given candidate answer using an information retrieval (\emph{IR}) system.\footnote{\url{www.elastic.co}} 
We use the three approaches sequentially to expand the candidate answer set to 20 candidates per question.



\paragraph{Step 4: Answer labeling.}
In this step, we ask annotators to label each answer with three options: ``likely'', ``unlikely'', or ``garbage'' (incomplete or meaningless phrases). 
We keep a candidate answer if and only if all 4 annotators agree on ``likely'' or ``unlikely'', and ``garbage'' is not marked by any annotator.
We also discard any questions that end up with no valid candidate answers. 
\ignore{
\qiang{
Before asking annotators to label the candidates, we allow them to filter out illegitimate ones (incomplete or meaningless phrases), and discard a candidate answer if any annotator thinks so. This procedure is important since the set of candidate answers augmented by Step~3 is noisy.
}
We \qiang{only keep those} candidate answers with complete agreement (\qiang{[4/4]}),
\qiang{and we also discard any questions that end up with no valid candidate answers.}
}
Finally, the statistics of \datasetname{} is in Table~\ref{tab:statistics}.

\ignore{
\begin{table}[]
    \centering
    \footnotesize
    \resizebox{0.6\textwidth}{!}{
    \begin{tabular}{lc}
        \toprule 
         Measure & Value \\ 
         \cmidrule(lr){1-1} \cmidrule(lr){2-2}
          \# of unique questions & 1,893 \\
          \# of unique question-answer pairs & 13,225 \\
          avg. sentence length & 17.8 \\
          avg. question length & 8.2 \\
          avg. answer length & 3.3 \\
          \midrule 
          \# of \emph{event frequency} questions & 433 \\
          \# of \emph{event duration} questions & 440 \\
          \# of \emph{event stationarity} questions & 279 \\
          \# of \emph{event ordering} questions & 370 \\
          \# of \emph{event typical time} questions & 371 \\
          \midrule
          avg candidate size for \emph{event frequency} questions & 8.5 \\
          avg candidate size for \emph{event duration} questions & 9.4 \\
          avg candidate size for \emph{event stationarity} questions & 3.1\\
          avg candidate size for \emph{event ordering} questions & 5.4  \\
          avg candidate size for \emph{event typical time} questions & 6.8 \\
        \bottomrule 
    \end{tabular}
    }
    \caption{\small Statistics of \datasetname. }
    \label{tab:statistics}
\end{table}
}

\section{Experiments}
\label{sec:exp}
We assess the quality of our dataset using a couple of baseline systems. We create a uniform split of 30\%/70\% of the data to dev/test.
The rationale behind this split is that, a successful system has to bring in a huge amount of world knowledge and derive commonsense understandings {\em prior} to the current task evaluation. We therefore believe that it make no sense to expect a system to {\em train} solely on this data, and we think of the development data as only providing a {\em definition} of the task. Indeed, the gains from our development data are marginal after a certain number of observations.
This intuition has been studied and verified in the appendix of~\cite{ZKNR19}.

\paragraph{Evaluation metrics.}
Two question-level metrics are adopted in this work: exact match (\emph{EM}) and {\em F1}. EM measures in how many questions a system is able to correctly label all candidate answers, while {\em F1} measures the average overlap between one's predictions and the ground truth (see the appendix of~\cite{ZKNR19} for full definition).

\paragraph{Human performance.}
An expert annotator also worked on \datasetname{} to gain a better understanding of the human performance on it.
The expert specifically answered 100 questions randomly sampled from the test set, and could only see a single answer at a time, with its corresponding question and sentence.

\paragraph{Systems.} 
We propose to use two state-of-the-art systems in machine reading comprehension that are suitable for our task.
\emph{ESIM}~\cite{CZLWJI17} is a neural model effective on natural language inference. We initialize the word embeddings in {\em ESIM} via either \emph{GloVe}~\cite{PenningtonSoMa14} or \emph{ELMo}~\cite{PNIGCLZ18} to demonstrate the effect of pre-training in this task.
\bert{} is a recent state-of-the-art contextualized representation used in a broad range of high-level tasks~\cite{DCLT18}. 
We also add unit normalization to \bert{}, which extracts and converts temporal expressions in candidate answers to their most proper units. For example, ``30 months'' will be converted to ``2.5 years".

\paragraph{Experimental setting.}
In both \emph{ESIM} baselines, we model the process as a sentence-pair labeling task, following the \emph{SNLI} setting provided in AllenNLP.\footnote{{https://github.com/allenai/allennlp}}
In both versions of \emph{BERT}, we use the same sequence pair classification model and the same parameters as in \emph{BERT}'s \emph{GLUE} experiment.\footnote{{github.com/huggingface/pytorch-pretrained-BERT}}
A system receives two phrases at a time: (a) the concatenation of the sentence and question, and (b) the answer. The system makes a binary prediction on each instance, positive or negative. 



\begin{table}[]
    \centering
    \small 
    \begin{tabular}{ccc}
        \toprule 
         System & \emph{F1} & \emph{EM}  \\ 
         \cmidrule(lr){1-1} \cmidrule(lr){2-2} \cmidrule(lr){3-3} 
         Random  & 36.2 & 8.1 \\ 
         Always Positive & 49.8 & 12.1  \\
         Always Negative & 17.4 & 17.4 \\
         \midrule
         \emph{ESIM + GloVe} & 50.3 & 20.9 \\
         \emph{ESIM + ELMo} & 54.9 & 26.4\\
         \emph{BERT} & 66.1 & 39.6  \\
         \emph{BERT + unit normalization} & \textbf{69.9} & \textbf{42.7}  \\
         \midrule 
         Single Human & 87.1 & 75.8  \\ 
        \bottomrule 
    \end{tabular}
    \caption{\small Summary of the performances for different baselines. All numbers are in percentages. 
    }
    \label{tab:performance}
\end{table}

\paragraph{Results and discussion.}
Table~\ref{tab:performance} provides a summary of the results on \datasetname{}, where we compare the {\em ESIM} and \bert{} baselines, along with a few naive baselines (always-positive, always-negative, uniformly random), to the human performance.
The significant improvement brought by contextualized pre-training such as \bert{} and \emph{ELMo} indicates that a significant portion of commonsense knowledge is actually acquired via pre-training.
We can also see that human annotators achieved a very high performance under both metrics, indicating the high agreement level humans are on for this task. 
Our baselines, including \emph{BERT}, still fall behind the human performance with a significantly margin. 

\begin{figure}
    \centering
    \includegraphics[scale=0.4,trim=1.6cm 0cm 0cm 0cm, clip=false]{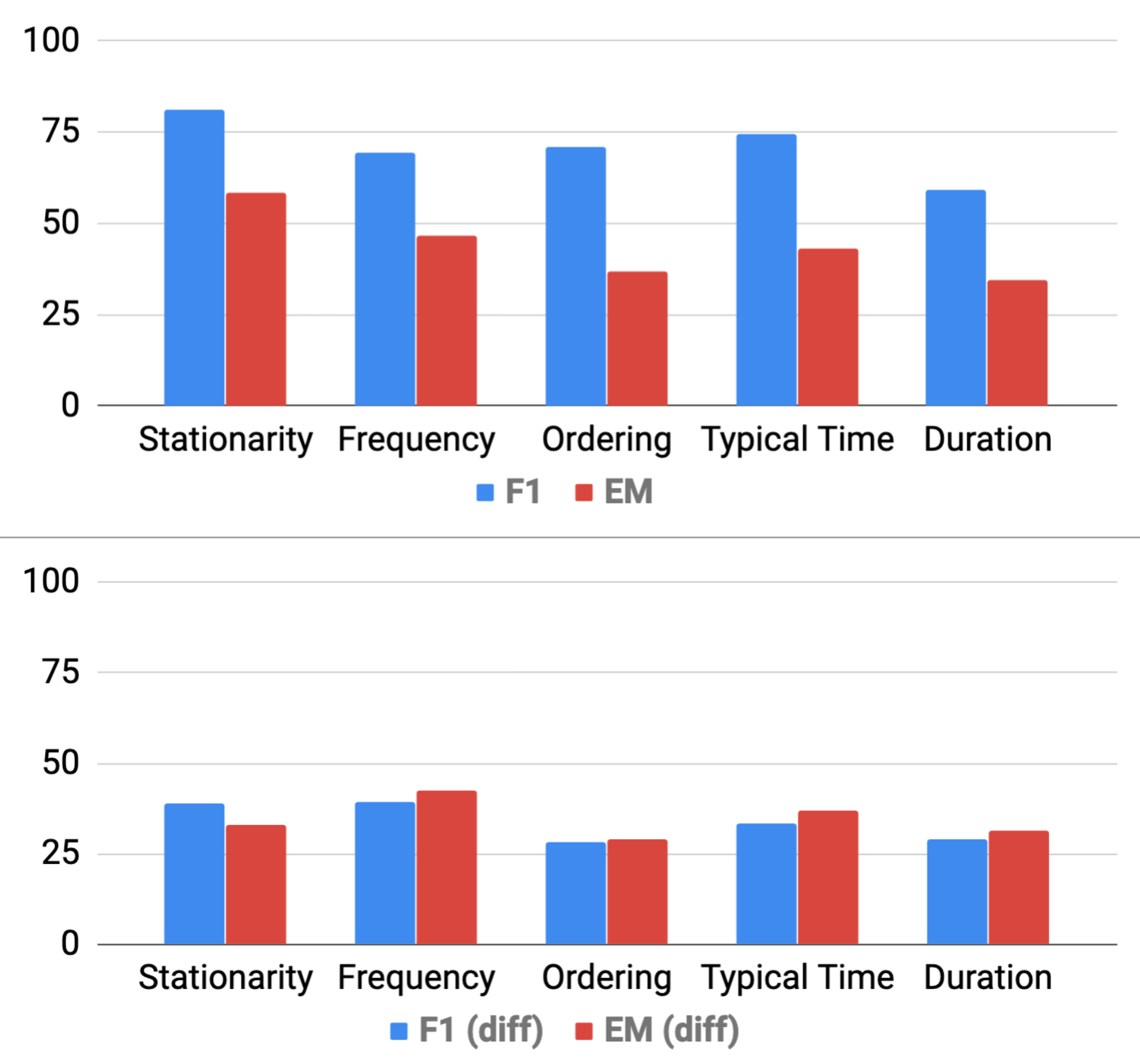}
    \caption{\small \emph{BERT + unit normalization} performance per temporal reasoning category (top), performance gain over random baseline per category (bottom)}
    \label{fig:bert-per-category}
\end{figure}
\ignore{
In this section we provide further 
intuition into our results. 
A major takeaway from the experiments, as expected, is the importance of \emph{contextual pre-training} for a task like ours, which involves much commonsense understanding (unlike tasks where answers are part of the input) \qiangcomment{not sure about the ``unlike'' part}. The best system in our results strongly relies on BERT. Additionally, the combination of \emph{ESIM} with \emph{ELMo} (a contextual pre-trained representation), is much stronger than using \emph{GloVe} as the underlying representations. This indicates the importance of using large amount of world knowledge in this task. \qiangcomment{a bit wordy; can just say ``the significant improvement brought by contextualized pre-training such as \emph{BERT} and \emph{ELMo} indicates that a significant portion of commonsense knowledge is actually acquired via pre-training.''}
}

Further analysis shows that \emph{BERT}, as a language model, is good at associating surface-forms (e.g. associating ``sunrise'' and ``morning'' since they often co-occur), which is highly sensitive to \emph{units} (\emph{days, years, etc}). To address the high sensitivity, we added unit normalization on top of \bert{}, but even with normalization, \emph{BERT+unit normalization} is still far behind the human performance. This implies that the information acquired by \bert{} is still not sufficient to solve this task. Moreover, the low EM scores show that the current systems do not truly understand time in those questions.

\ignore{
However, the system can be brittle when a candidate answer is not mentioned with the typical units (and hence fail to generalize to deeper temporal understanding) \qiangcomment{``deeper'' not clear}.   
}

Figure~\ref{fig:bert-per-category} reveals that the performance of \emph{BERT} is not uniform across different categories, which could stem from the nature of those different types of temporal commonsense, quality of the candidate answers, etc. For example, the number of candidates for stationarity questions are much smaller than those for other questions, leading to a relatively easy task, but the performance gain from a random baseline to \bert{}+normalization is not large, indicating that further improvement on stationarity is still difficult.

\ignore{
\paragraph{Beyond this work.}\qiangcomment{keep it for some time and I'll try to go over it again}
While this work takes a positive step towards the temporal commonsense problem, 
there are aspects of this problem that are not covered by our crowdsourced dataset. 
To demonstrate this, one of the authors created a toy dataset of 55 question-answer pairs (20 questions) that require \emph{composing} temporal commonsense of potentially multiple events, rather than just making inferences about isolated events (examples in the appendix of~\cite{ZKNR19}).
The fine-tuned \emph{BERT} (previous section) achieves \emph{F1}/\emph{EM} of 38.1/0.0; while a human annotator achieves \emph{F1}/\emph{EM} of 69.0/57.1. 
This highlights the necessity of, not only acquiring temporal commonsense, but also being able to \emph{compose} them and make inferences. 
}



\section{Summary}
This chapter has focused on the challenge of temporal commonsense.
Specifically, we framed it as a QA task, defined five categories of questions that capture such ability, and developed a novel crowdsourcing scheme to generate a high-quality dataset for temporal commonsense.
We then showed that systems equipped with state-of-the-art language models such as {\em ELMo} and \bert{} are still far behind humans, thus motivating future research in this area. Our analysis sheds light on the capabilities as well as limitations of current models. 
We hope that this study will inspire further research on temporal commonsense. 


\part{Formal Study of Reasoning in Natural Language}
\label{part3:theory}
\chapter{Capabilities and Limitations of Reasoning in Natural Language}
\label{chapter:limits}
\epigraph{ 
``Language is froth on the surface of thought.''
}{--- \textup{John McCarthy} }


\newif\iftacl

\section{Introduction}

Reasoning can be defined as the process 
of combining facts and beliefs, in order to make decisions~\citep{Johnson-Laird80}. 
In particular, in natural language processing (NLP), it has been 
studied under various settings, such as question-answering (QA)~\citep{HLBB99}.

\begin{wrapfigure}{r}{0.5\textwidth}
    \centering
    \includegraphics[trim=0.31cm 0.05cm 0cm 1.3cm, clip=false, scale=0.31]{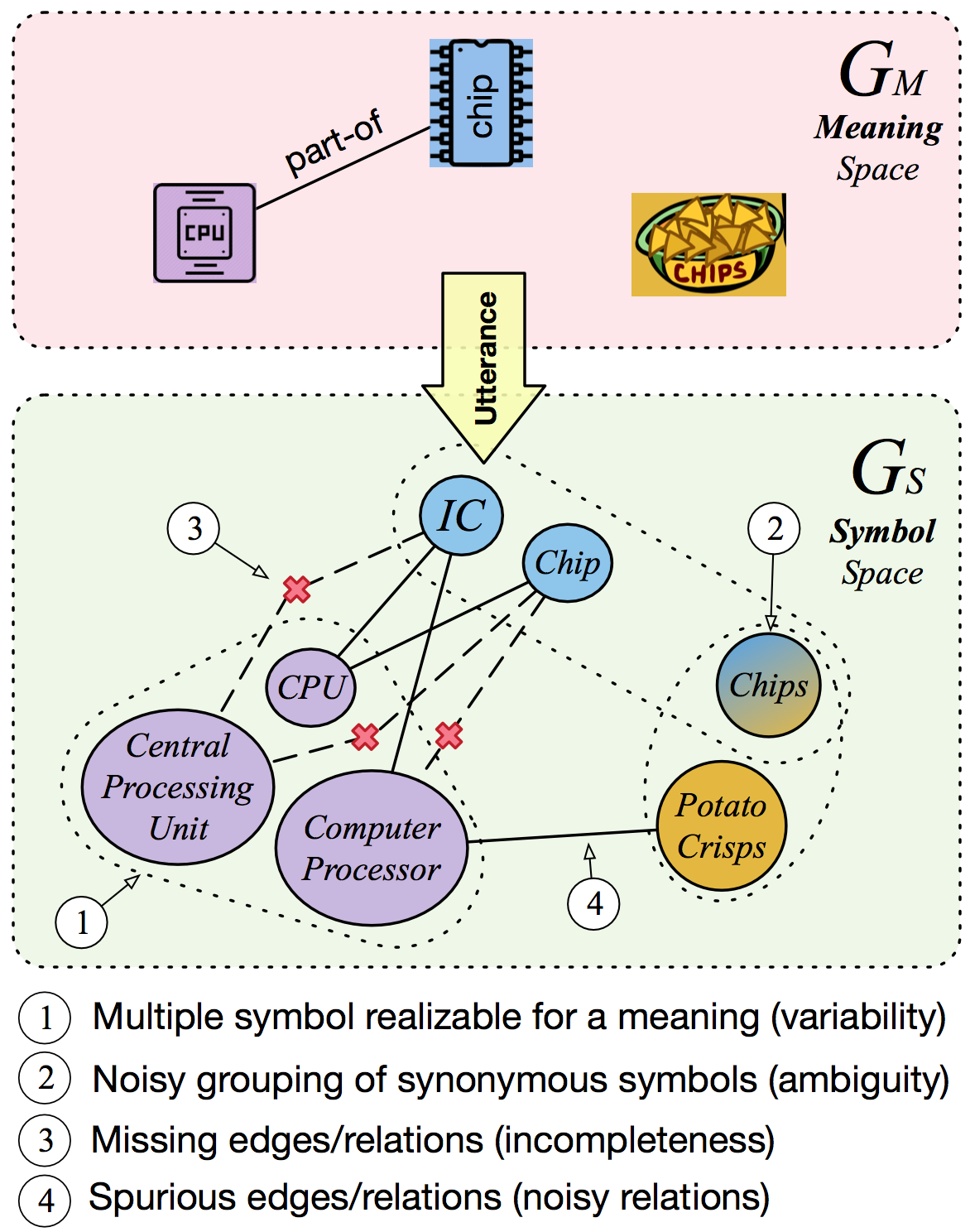}
    \vspace{-0.2cm}
    \caption{
    The interface between meanings and symbols: each \colorbox{lightred}{meaning} (top) can be \colorbox{lightyellow}{\emph{uttered}} in many ways into \colorbox{lightgreen}{symbolic} forms (bottom). 
    }
    \label{fig:intro-figure}
\end{wrapfigure}
While there is a rich literature on reasoning, 
there is little understanding of the 
nature of the problem and its limitations, especially in the context of natural language. 
In particular, there remains a sizable gap between empirical understanding of reasoning algorithms for language and the theoretical guarantees for their quality, often due to the complexity of the reality they operate on.
An important challenge in many language understanding problems 
is 
the \emph{symbol grounding problem}~\citep{Harnad90}, 
the problem of accurately mapping symbols into its underlying meaning representation.
Practitioners often address this challenging by enriching their representations; 
for example by mapping textual information to Wikipedia entries~\citep{MihalceaCs07,RRDA11}, or grounding text to executable rules via semantic parsing~\citep{RTPSL17}. 
{
Building upon such representations, 
has produced various reasoning systems that essentially work by combining local information. 
}

This work introduces a formalism that incorporates elements of the symbol-grounding problem, via the two spaces illustrated in Figure~\ref{fig:intro-figure}, and sheds theoretical light on existing 
intuitions.\footnote{This chapter is based on the following publication: \cite{KSKSR19}.} The formalism consists of (A) an abstract model of linguistic knowledge, and (B) a reasoning model.

\noindent
\textbf{(A) Linguistically-inspired abstract model:}
We propose a theoretical framework to model and study the capabilities/limitations of reasoning, especially when taking into account key difficulties that arise when formalizing linguistic reasoning. Our model uses two spaces; cf.~Figure~\ref{fig:intro-figure}. We refer to the internal conceptualization in the human mind as the \colorbox{lightred}{\emph{meaning space}}. We assume the information in this space is free of noise and uncertainty.
In contrast to human thinking in this space, \emph{human expression} of thought via the \colorbox{lightyellow}{utterance} of language introduces many imperfections. 
The information in this linguistic space---which we refer to as the \colorbox{lightgreen}{\emph{symbol space}}---has many language-specific properties. 
The symbolic space is often redundant (e.g., multiple symbols ``CPU'' and ``computer processor'' express the same meaning), ambiguous (e.g., a symbol like ``chips'' could refer to multiple meanings ), incomplete (relations between some symbolic nodes might be missing), and inaccurate (there might be incorrect edges).
Importantly, this noisy symbol space is also what a machine reasoning algorithm operates in.


\noindent
\textbf{(B) Reasoning model:}
We define reasoning as the ability to infer the existence of properties of interest  
 in the \colorbox{lightred}{\emph{meaning} space}, by observing only its 
 representation in the \colorbox{lightgreen}{\emph{symbol} space}. 
The target property 
in the meaning graph is what characterizes the nature of the reasoning algorithm, e.g., are two nodes connected. 
While there are many flavors of reasoning (including \emph{multi-hop reasoning}), 
in this first study, we explore a
common primitive shared among various reasoning formalisms; namely, the \emph{connectivity problem} between a pair of nodes in an undirected graph in the \colorbox{lightred}{meaning space}, while observing its noisy version in the \colorbox{lightgreen}{symbol space}.
This simplification clarifies 
the exposition and the analysis,and we expect similar results to hold for a broader class of reasoning algorithms
that rely on connectivity. 

\begin{figure*}
    \centering
    \includegraphics[trim=0cm 1.1cm 0cm 0cm, clip=false, scale=0.287]{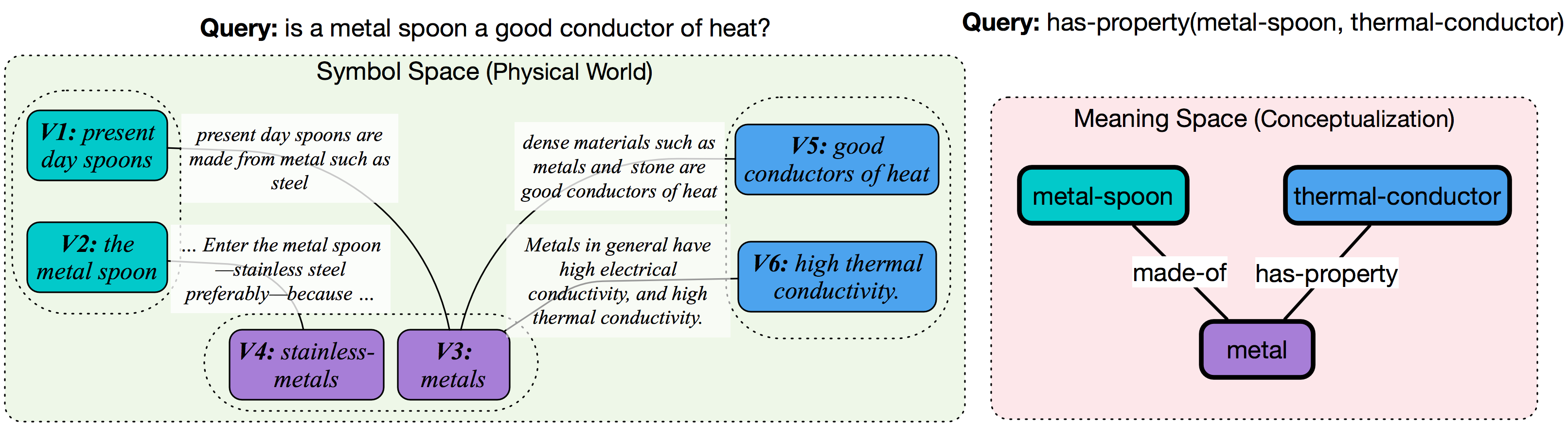}
    \caption{
    \small
    The \colorbox{lightred}{meaning space} contains [clean and unique] symbolic representation and the facts, while the \colorbox{lightgreen}{symbol space} contains [noisy, incomplete and variable] representation of the facts. We show sample meaning and symbol space nodes to answer the question: \textit{Is a metal spoon a good conductor of heat?}.
    }
    \label{fig:ny-example}
\end{figure*}

Figure~\ref{fig:ny-example} illustrates a reasoning setting where the semantics of the edges is included. Most humans understand that \emph{\textbf{V1}:``present day spoons''} and \emph{\textbf{V2}:``the metal spoons''} are equivalent nodes (have the same meaning). However, a machine has to infer this understanding. 
The semantics of the connection between nodes are expressed through natural language sentences. 
For example, connectivity could express the semantic relation between two nodes: {\tt \small \textbf{has-property(metal,thermal-conductor)}}. However a machine may find it difficult to infer this fact from, say, reading text over the Internet as it may be expressed in many different ways, e.g., 
can be found in a sentence like \emph{``dense materials such as [\textbf{V3:}]metals and stones are [\textbf{V5:}]good conductors of heat''}. 

\begin{sloppypar}
To ground this in existing efforts, consider \emph{multi-hop} reasoning for QA systems~\citep{KKSCER16,JWMM18}. Here the reasoning task is to connect local information, via multiple local ``hops'', in order to arrive at a conclusion. In the meaning graph, one can trace a path of locally connected nodes to verify the correctness of a query; for example the query {\tt \small \textbf{has-property(metal-spoon, thermal-conductor)}}
can be verified by tracing a sequence of nodes, as shown in Figure~\ref{fig:ny-example}. 
In other words, answering queries can be cast as inferring the existence of {a 
path} connecting two nodes $m$ and $m'$.
\footnote{This particular grounding is meant to help relate our graph-based formalism to existing applications, and is 
not the only way of realizing reasoning on graphs.}
While doing so on the \colorbox{lightred}{meaning graph} is straightforward, doing so on the noisy \colorbox{lightgreen}{symbol graph} is not. Intuitively, each local ``hop'' introduces more noise, allowing reliable inference to be performed only when it does not require too many steps in the underlying \colorbox{lightred}{meaning space}. To study this issue, one must quantify the effect of noise accumulation for long-range reasoning.
\end{sloppypar}


\textbf{Contributions.}
We believe that this is the first work to provide a mathematical study of the 
challenges
and limitations of reasoning algorithms in the presence of the symbol-meaning mapping challenge.
We make three main contributions.

First, we establish a novel, linguistically motivated formal framework for analyzing the problem of reasoning about the ground truth (the meaning space) while operating over a noisy and incomplete linguistic representation (the symbol space). This framework allows one to derive rigorous intuitions about what various classes of reasoning algorithms \emph{can} and \emph{cannot} achieve.

Second, we study in detail the \emph{connectivity reasoning} problem, in particular the interplay between the noise level in the symbol space (due to ambiguity, variability, and missing information) and the distance (in terms of inference steps, or hops) between two elements in the meaning space. We prove that under low noise levels, it is indeed possible to perform reliable connectivity reasoning up to a few hops (Theorem~\ref{the:reasoning:possiblity}). On the flip side, even a moderate increase in the noise level makes it difficult to assess the connectivity of elements if they are logarithmic distance apart in the meaning space (Theorems~\ref{th:multi-hop:impossiblity} and~\ref{theorem:impossiblity:general:reasoning}). This finding is aligned with empirical observations of 
``semantic drift'', i.e., substantial drop in performance beyond a few (usually 2-3) hops~\citep{FJHSC15,Jansen16}.

Third, we apply the framework to a subset of a real-world knowledge-base, FB15k237, treated as the meaning graph, illustrating how key noise parameters influence the possibility (or not) of accurately solving the connectivity problem.

\section{Related Work}

\paragraph{Classical views on reasoning.}
Philosophers, all the way from Aristotle and Avicenna, were the first ones to notice reasoning and rationalism
~\citep{KirkRaSc83,Davidson92}. 
In modern philosophy, the earlier notions were mixed with mathematical logic, resulting in formal theories of reasoning, such as deductive,
inductive,
and abductive reasoning~\citep{Peirce83}.
Our treatment of reasoning applies to all these, that can be 
modeled and executed using graphical representations. 

\paragraph{Reasoning in AI literature.}
The AI literature has seen a  variety of formalisms for automated reasoning. 
These include, reasoning with logical representations~\citep{McCarthy63}, semantic networks~\citep{Quillan66}, frame-semantic based systems~\citep{Fillmore77}, Bayesian networks~\citep{Pearl88}, among others.

It is widely believed that a key obstacle to progress has been the \emph{symbol grounding problem}~\citep{Harnad90,TaddeoFl05}. 
Our formalism is directly relevant to this issue. 
We assume that symbols available to reasoning systems are results of communication meaning n natural language. This results in ambiguity since a given symbol could be mapped to multiple actual meanings but also in variablity (redundancy).  

\paragraph{Reasoning for natural language comprehension.}
In the context of natural language applications (such as QA) flavors of linguistic theories are blended with the foundation provided by AI. 
A major roadblock has been 
the problem of \emph{symbol grounding}, or grounding free-form texts to a higher-level meaning. Example proposals to deal with this issue are, extracting semantic parses~\citep{KaplanBrot82,SteedmanBa11,BBCGGHKPS13}, linking to the knowledge bases~\citep{MihalceaCs07}, mapping to semantic frames~\citep{PunyakanokRoYi04}, etc. These methods can be thought of as approximate solutions for grounding symbolic information to some \emph{meaning}. \citep{RothYi04} suggested a general abductive framework that addresses it by connecting reasoning to models learned from data; it has been used in multiple NLP reasoning problems~\citep{KKSR18}.


On the execution of reasoning with the disambiguated inputs there are varieties of proposals, e.g., using executable formulas~\citep{RTPSL17,AngeliMa14}, chaining relations to infer new relations~\citep{SCMN13,MNDB17,KhotSaCl17}, 
and possible combinations of the aforementioned paradigms~\citep{GardnerTaMi15,CEKSTTK16}. 
Our analysis covers any algorithm for inferring patterns that can be formulated in graph-based knowledge, e.g.,  
chaining local information, often referred to as \emph{multi-hop} reasoning~\citep{JBSC16,JWMM18,LinSoXi18}. 
For example, \cite{JSSC17} propose a structured multi-hop reasoning by aggregating sentential information from multiple knowledge bases. The work shows that while this strategy improves over baselines with no reasoning (showing the effectiveness of reasoning), with aggregation of more than 2-3 sentences the quality declines (showing a limitation for reasoning). Similar observations were also made in~\citep{KKSCER16}. These empirical observations support the theoretical intuition proven in this work. 




\section{Background and Notation}
\label{sec:background}
We start with basic definitions and notation. 

\paragraph{Graph Theory.}
We denote an undirected graph with $G(V, E)$ where $V$ and $E$ are the sets of nodes and edges, resp. We use the notations $V_G$ and $E_G$ to  refer to the nodes and edges of a graph $G$, respectively. 
Let $\dist(v_i, v_j)$ be the distance between nodes $v_i$ and $v_j$ in $G$. 
A \emph{simple path} (henceforth referred to as just a \emph{path}) is a sequence of adjacent nodes that does not have repeating nodes. Let $v_i \pathk{d} v_j$ denote the existence of a path of length $d$ between $v_i$ and $v_j$. Similarly, $v_i \npathk{} v_j$ denotes that there is no path between $v_i$ and $v_j$. We define the notion of $d$-\emph{neighborhood} in order to analyze local properties of the graphs:  
\begin{definition}
For a graph $G = (V,E)$, $s \in V$, and $d \in \mathbb{N}$, the \emph{$d$-neighbourhood} of $s$ is $\{v \mid \dist(s,v) \leq d\}$, i.e., the `ball' of radius $d$ around $s$. $\ball(s,d)$ denotes the number of nodes in this $d$-neighborhood, and $\ball(d) = \max_{s \in V} \ball(s,d)$.
\end{definition}

Finally, a \emph{cut} $C = (S,T)$ in $G$ is a partition of the nodes $V$ into subsets $S$ and $T$. The \emph{size} of the cut $C$ is the number of edges in $E$ with one endpoint in $S$ and the other in $T$.

\paragraph{Probability Theory.}
$X \sim f(\theta)$ denotes a random variable $X$ distributed according to probability distribution $f(\theta)$, paramterized by $\theta$. 

Given random variables $X\sim\bern{p}$ and $Y\sim\bern{q}$, their disjunction $X \vee Y$ is another Bernoulli $\bern{p \oplus q }$, where  
$p \oplus q \triangleq 1 - (1-p)(1-q) = p + q - pq$. We will make extensive use of this notation throughout this work. 

\section{The Meaning-Symbol Interface}
\label{sec:framework}

We introduce two notions of knowledge spaces: 
\begin{itemize}
\itemsep0em 
    \item The \emph{meaning space}, $M$, is a conceptual hidden space where all the facts are accurate and complete. 
    We assume the knowledge in this space can be represented as an undirected graph, denoted \Gm. 
    This 
    knowledge is 
    hidden, and representative of the information that exists within human minds. 

    \item The \emph{symbol space}, $S$, 
    is the space of written sentences, curated knowledge-based, etc., in which knowledge is represented for human and machine consumption. We assume access to a knowledge graph \Gs\ in this space that is an incomplete, noisy, redundant, and ambiguous approximation of $G_M$.
\end{itemize}

There are interactions between the two spaces: when we read a sentence, we are reading from the \emph{symbol space} and interpreting it 
in the \emph{meaning space}. 
When writing out our thoughts, we symbolize our thought process, by moving them from \emph{meaning space} to the \emph{symbol space}. 
Figure~\ref{fig:intro-figure} provides a high-level view of the framework. A reasoning system is not aware of the exact structure and information encoded in the meaning graph.

The only information given is the ball-assumption, i.e., we assume that each node $m$ is connected to at most $\ball(m, d)$ many nodes, within distance at most $d$. If this bound holds for all the nodes in a graph, we'd simply write it as $\ball(d)$. The ball assumption is a simple understanding of the maximum-connectivity in the meaning-graph, without knowing the details of the connections.

\paragraph{Meaning-Symbol mapping.}
We define an oracle function $\oracle: M \rightarrow  2^S$ that map nodes in the meaning space to those in the symbol space. When $s\in \oracle(m)$, with some abuse of notation, we write $\oracle^{-1}(s)=m$. 


\paragraph{Generative Modeling of Symbol Graphs.}
We now explain a generative process for constructing symbol graphs. Starting with \gm, we sample a symbol graph $G_S \leftarrow \alg{G_M}$ using a stochastic process, detailed in Algorithm~\ref{al:sampling}. Informally, the algorithm simulates the process of transforming conceptual information into linguistic utterances (web-pages, conversations, knowledge-bases).

Our stochastic process has three main parameters: (a) the distribution $r(\lambda)$ of the number of replicated symbols to be created for each node in the meaning space; (b) the edge retention probability $p_+$; and (c) the noisy edge creation probability $p_-$. We will discuss later the regimes under which Algorithm~\ref{al:sampling} generates interesting symbol graphs.

{
\begin{algorithm}[t]
\footnotesize
\KwIn{Meaning graph \Gm, discrete distribution $r(\lambda)$, edge retention probability $p_+$, edge creation probability $p_-$}
\KwOut{Symbol graph \Gs}
 \ForEach{$v \in \vm$}{
     sample $k \sim r(\lambda)$ \\
     construct a collection of new nodes $U$ s.t.\ $|U| = k$ \\
     $\vs \leftarrow \vs  \cup U$  \\ 
    \oracle$(v) \leftarrow U$
 }
 
 \ForEach{$(m_1,  m_2) \in (\vm \times \vm), m_1 \neq m_2$}{
    $S_1 \leftarrow \oracle(m_1)$, $S_2 \leftarrow \oracle(m_2)$  \\ 
    \ForEach{$e \in S_1 \times S_2$}{
        \eIf{$(m_1,  m_2) \in \emm$}{
            with probability $p_+$: $\es \leftarrow \es  \cup \{ e \} $   
        }{ 
            with probability $p_-$: $\es \leftarrow \es  \cup \{ e \} $
        }
    }
 }
 \caption{
 Generative construction of knowledge graphs; sampling a symbol knowledge graph \gs\ given a meaning graph \gm. 
 }
 \label{al:sampling}
\end{algorithm}
}

This construction models a few key properties of linguistic representation of meaning. Each node in the meaning space is potentially mapped to multiple nodes in the symbol space, which models redundancy. 
Incompleteness of knowledge is modeled by the fact that not all meaning space edges appear in the symbol space {(controlled by parameter $p_+$ in Algorithm~\ref{al:sampling})}.  {There are also edges in the symbol space that do not correspond to any edges in the meaning space and account for the noise (controlled by parameter $p_-$ in Algorithm~\ref{al:sampling}).}

Next, we introduce a linguistic similarity based connection to model ambiguity, i.e., a single node in the symbol graph mapping to multiple nodes in the meaning graph.
{
The ambiguity phenomena is modelled indirectly via the linguistic similarity based connections (discussed next). 
We view ambiguity as treating (or confusing) two symbol nodes as the same even when they originate from different nodes in the meaning space.
}

\paragraph{Noisy Similarity Metric.}
Similarity metrics are typically used to judge the equivalence of symbolic assertions. Let $\rho: V_S \times V_S \rightarrow \{0, 1\}$ be such a metric, where $\rho(s, s')=1$ denotes the equivalence of two nodes in the symbol graph. Specifically, we define the similarity to be a noisy version of the true node similarity between node pairs:
\vspace{-0.05cm}
\begin{align*}
\rho(s, s') \triangleq &
  \begin{cases}
  1 - \text{Bern}(\varepsilon_+) &  \text{if\ } \oracle^{-1}(s) = \oracle^{-1}(s') \\ 
  \text{Bern}(\varepsilon_-) &  \text{otherwise}
  \end{cases}, 
  \vspace{-0.05cm}
\end{align*}
where $\varepsilon_+, \varepsilon_- \in (0, 1)$ are the noise parameters of the similarity function, both typically close to zero. 
{
Intuitively, the  similarity function is a \emph{perturbed} version of ground-truth similarities, with small random noise (parameterized with $\varepsilon_+$ and $\varepsilon_-$). Specifically with a high probability $1-\varepsilon_{+/-}$, it returns the correct similarity decision (i.e., whether two symbols have the same meaning); and with a low probability $\varepsilon_{+/-}$ it returns an incorrect similarity decision. In particular, $\varepsilon_+ = \varepsilon_- = 0$ models the perfect similarity metric. In practice, even the best entailment/similarity systems have some noise (modeled as $\varepsilon_{+/-} > 0$). 
}

We assume algorithms have access to the symbol graph $\gs$ and the similarity function $\rho$, and that they use the following procedure to verify the existence of a connection  
between two nodes: 
\FrameSep0pt
\begin{framed}
  \begin{algorithmic}
        \Function{NodePairConnectivity}{$s, s'$}
         \State \textbf{return} $(s, s') \in E_S$ or $\rho(s, s') = 1$ 
        \EndFunction
    \end{algorithmic}
\end{framed}

There are many corner cases that result in uninteresting meaning or symbol graphs. 
Below we define the regime of realistic instances:

\begin{definition}[Nontrivial Graph Instances]
\label{defn:nontrivial}
    A pair ($\gm, \gs$) of a meaning graph and a symbol graph sampled from it is \emph{non-trivial} if it satisfies: 
    \begin{enumerate}
        \itemsep0em 
        \item non-zero noise, i.e., $p_-, \varepsilon_-, \varepsilon_+ > 0$; 
        
        \item incomplete information, i.e., $p_+ < 1$;
        
        \item noise content does not dominate the actual information, i.e., $p_- \ll p_+, \varepsilon_+ < 0.5$ and $p_+ > 0.5$; 
        
        \item \gm\ is not overly-connected, i.e., $\ball(d) \in o(n)$, where $n$ is the number of nodes in $\gm$;
        
        \item \gm\ is not overly-sparse, i.e., $|E_{G_M}| \in \omega(1)$. 
    \end{enumerate}
\end{definition}
Henceforth, we will only consider sampling parameters satisfying the above conditions.

\paragraph{Reasoning About Meaning, through Symbols.}
While the reasoning engine only sees the symbol graph $G_S$, it must make inferences about the potential latent meaning graph. Given a pair of nodes $\tuple:= \{s, s'\} \subset V_S$ in the symbol graph, the reasoning algorithm must then predict properties about the corresponding nodes $\tuplem =\{m, m'\} = \{ \oracle^{-1}(s), \oracle^{-1}(s') \}$ in the meaning graph. 

We use a hypothesis testing setup to assess the likelihood of two disjoint hypotheses defined over these meaning nodes:
$H^{\circled{1}}_M(\tuplem)$ and $H^{\circled{2}}_M(\tuplem)$. Given observations about the symbol nodes, defined as $X_S(\tuple)$, the goal of a reasoning algorithm is to identify which of the two hypotheses about the meaning graph has a higher likelihood of resulting in these observations under the sampling process of Algorithm~\ref{al:sampling}. Formally, we are interested in:
\vspace{-0.05cm}
\begin{equation}
\label{eq:decision}
\underset{h \in \{ H^{\circled{1}}_M(\tuplem), 
H^{\circled{2}}_M(\tuplem)\}}{\text{argmax}}  \probTwo{h}{X_S(\tuple)}
\end{equation}
\vspace{-0.03cm}
where $\probTwo{h}{x}$ denotes the probability of an event $x$ in the sample space induced by Algorithm~\ref{al:sampling} on the latent meaning graph $\gm$ when it satisfies hypothesis $h$.

Since we start with two disjoint hypotheses on $\gm$, the resulting probability spaces are generally different, making it plausible to identify the correct hypothesis with high confidence. At the same time, with sufficient noise in the sampling process, it can also become difficult for an algorithm to distinguish the two resulting probability spaces (corresponding to the two hypotheses)  especially depending on the observations $X_S(\tuple)$ used by the algorithm. For example, the distance between the symbolic nodes can often be an insufficient indicator for distinguishing these hypotheses. We will explore these two contrasting behaviors in the next section.

\begin{definition}[Reasoning Problem]
    The input for an instance $\ins$ of the \emph{reasoning problem} is a collection of parameters that characterize how a symbol graph $\gs$ is generated from a (latent) meaning graph $\gm$, two hypotheses $H^{\circled{1}}_M(\tuplem), H^{\circled{2}}_M(\tuplem)$ about $\gm$, and available observations $X_S(\tuple)$ in $\gs$.
    The reasoning problem, $\ins(p_+$, $p_-$, $\varepsilon_+$, $\varepsilon_-$, $\ball(d)$, n, $\lambda$, $H^{\circled{1}}_M(\tuplem)$, $H^{\circled{2}}_M(\tuplem)$, $X_S(\tuple))$, is to map the input to the hypothesis $h$ as per Eq.~(\ref{eq:decision}).
\end{definition}

We use the following notion to measure the effectiveness of the observation $X_S$ in distinguishing between the two hypotheses as in Eq.~(\ref{eq:decision}):
\vspace{-0.05cm}
\begin{definition}[$\gamma$-Separation]
\label{defn:gamma-separation}
    For $\gamma \in [0,1]$ and a problem instance $\ins$ with two hypotheses $h_1 = H^{\circled{1}}_M(\tuplem)$ and $h_2 = H^{\circled{2}}_M(\tuplem)$, we say an observation $X_S(\tuple)$ in the symbol space $\gamma$-separates $h_1$ from $h_2$ if: 
    $$
        \probTwo{ h_1 }{ X_S(\tuple) } - \probTwo{ h_2 }{ X_S(\tuple) } \, \geq \, \gamma.
    $$
\end{definition}
\vspace{-0.05cm}
We can view $\gamma$ as the \emph{gap} between the likelihoods of the observation $X_S(\tuple)$ having originated from a meaning graph satisfying hypothesis $h_1$ vs.\ one satisfying hypothesis $h_2$. When $\gamma = 1$, $X_S(\tuple)$ is a perfect discriminator for distinguishing $h_1$ and $h_2$. In general, any positive 
$\gamma$ bounded away from $1$ yields a valuable observation.\footnote{If the above probability gap is negative, one can instead use the complement of $X_S(\tuple)$ for $\gamma$-separation.}

Given an observation $X_S$ that $\gamma$-separates $h_1$ and $h_2$, there is a simple algorithm that distinguishes $h_1$ from $h_2$:
\FrameSep0pt
\begin{framed}
  \small 
  \begin{algorithmic}
        \Function{Separator$_{X_S}$}{$\gs, \tuple=\{s, s'\}$}
         \State \textbf{if} $X_S(\tuple) = 1$
            {\textbf{then} return $h_1$}
            {\textbf{else} return $h_2$}
        \EndFunction
    \end{algorithmic}
\end{framed}

Importantly, this algorithm does \emph{not} compute the probabilities in Definition~\ref{defn:gamma-separation}. Rather, it works with a particular instantiation \gs\ of the symbol graph. We refer to such an algorithm $\mathcal{A}$ as \textbf{$\gamma$-accurate} for $h_1$ and $h_2$ if, under the sampling choices of Algorithm~\ref{al:sampling}, it outputs the `correct' hypothesis with probability at least $\gamma$; that is, for both $i \in \{1,2\}$: 
$\probTwo{h_i}{\mathcal{A} \text{\ outputs\ } h_i} \geq \gamma.$

\begin{proposition}
If observation $X_S$ $\gamma$-separates $h_1$ and $h_2$, then algorithm \textsc{Separator}$_{X_S}$ is $\gamma$-accurate for $h_1$ and $h_2$.
\end{proposition}
\begin{proof}
Let $\mathcal{A}$ denote \textsc{Separator}$_{X_S}$ for brevity. Combining $\gamma$-separation of $X_S$ with how $\mathcal{A}$ operates, we obtain:
{
\small
\begin{align*}
    \probTwo{h_1}{\mathcal{A} \text{\ outputs\ } h_1} - \probTwo{h_2}{\mathcal{A} \text{\ outputs\ } h_1} & \geq \gamma \\
    \Rightarrow \probTwo{h_1}{\mathcal{A} \text{\ outputs\ } h_1} + \probTwo{h_2}{\mathcal{A} \text{\ outputs\ } h_2} & \geq 1 + \gamma
\end{align*}
}
Since each term on the left is bounded above by $1$, each of them must also be at least $\gamma$. 
\end{proof}

In the rest of work, we will analyze when one can obtain a $\gamma$-accurate algorithm, using $\gamma$-separation of the underlying observation as a tool for the analysis.

We will assume that the replication factor (i.e., the number of symbol nodes corresponding to each meaning node) is a constant, i.e., $r$ is  such that $\prob{|U| = \lambda} = 1$.

\section{Connectivity Reasoning Algorithm}
One simple but often effective approach for reasoning is to focus on connectivity (as described in Figure~\ref{fig:ny-example}). Specifically, we consider reasoning chains as valid if they correspond to a short path in the meaning space, and invalid if they correspond to disconnected nodes. Given nodes $m, m' \in \gm$, this corresponds to two possible hypotheses: 
$$h_1 = m \eone m'\text{, and}\; h_2 = m \etwo m'$$
We refer to distinguishing between these two worlds as the \textbf{$d$-connectivity reasoning problem}. While we consider two extreme hypotheses for our analysis, we find that with a small amount of noise, even these extreme hypotheses can be difficult to distinguish. 

For the reasoning algorithm, one natural observation that can be used is the connectivity of the symbol nodes in $\gs$. Existing models of multi-hop reasoning~\citep{KhotSaCl17} use similar features 
to identify valid reasoning chains. Specifically, we consider the observation that there is a path of length at most $\tilde{d}$ between $s$ and $s'$: 
\vspace{-0.05cm}
$$
 X^{\tilde{d}}_S(s, s') = \; s \pathk{\leq \tilde{d}} s'
 \vspace{-0.05cm}
$$
The corresponding \textbf{connectivity algorithm} is \textsc{Separator}$_{X^{\tilde{d}}_S}$, which we would like to be $\gamma$-accurate for the two hypotheses under consideration.
Next, we derive bounds on $\gamma$ for these specific hypotheses and observation. Note that while the space of possible hypotheses and observations is large, the above natural and simple choices still allow us to derive valuable intuitions for the limits of reasoning.

\subsection{Possibility of accurate connectivity}

We begin by defining the following accuracy threshold, $\gamma^*$, as a function of the parameters for sampling a symbol graph:
\begin{definition}
Given $n, d \in \mathbb{N}$ and symbol graph sampling parameters $p_+, \varepsilon_+, \lambda$, define $ \gamma^*(n, d, p_+, \varepsilon_+, \varepsilon_-, \lambda) $ as
{
\small
\begin{align*}
       \left( 1 - \left( 1- (p_+ \oplus \varepsilon_-) \right)^{\lambda^2} \right)^{d} 
       \cdot \left( 1 - 2 e^3 \varepsilon_+ ^{\lambda/2} \right)^{d+1}  
        - 2en (\lambda\ball(d))^2 p_-.
\end{align*}
}
\vspace{-0.05cm}
\end{definition}
This expression is somewhat difficult to follow. Nevertheless,
as one might expect, the accuracy threshold $\gamma^*$ increases (higher accuracy) as $p_+$ increases (higher edge retention) or $\varepsilon_+$ decreases (fewer dropped connections between replicas). As $\lambda$ increases (higher replication), the impact of the noise on edges between node cluster
or $d$ decreases (shorter paths), the accuracy threshold will also increase.

	The following theorem (see Appendix for a proof) establishes the possibility of a $\gamma$-accurate algorithm for the connectivity problem:

%
\begin{theorem}
\label{the:reasoning:possiblity:detailed}
Let $p_+, p_-, \varepsilon_+, \varepsilon_-, \lambda$ be parameters of the sampling process in Algorithm~\ref{al:sampling} on a meaning graph with $n$ nodes. Let $d \in \mathbb{N}$ and $\tilde{d} = d (1+\lambda)$. If $p_-$ and $d$ satisfy
\vspace{-0.05cm}
$$
    { (p_- \oplus \varepsilon_-) } \cdot \ball^2(d) < \frac{1}{{2e}\lambda^2n},
    \vspace{-0.05cm}
$$
and $\gamma = \max \{ 0, \gamma^*(n, d, p_+, \varepsilon_+, \varepsilon_-, \lambda) \}$, then the connectivity algorithm \textsc{Separator}$_{X^{\tilde{d}}_S}$ is $\gamma$-accurate for the $d$-connectivity problem.
\end{theorem}
\begin{proof}[Proof idea]
\small
The proof consists of two steps: first show that for the assumed choice of parameters, connectivity in the meaning space is recoverable in the symbol space, with high-probability. Then show that 
spurious connectivity in the symbol space (with no meaning space counterparts) has low probability. 
\end{proof}

\begin{corollary}
\label{the:reasoning:possiblity}
(Informal) If $p_-, \varepsilon_-, d,$ and $\gamma$ are small enough, then the connectivity algorithm \textsc{Separator}$_{X^{\tilde{d}}_S}$ with $\tilde{d} = d (1+\lambda)$ is $\gamma$-accurate for the $d$-connectivity problem.
\end{corollary}

\subsection{Limits of connectivity algorithm}
We show that as $d$, the distance between two nodes in the meaning space, increases, it is unlikely that we will be able to make any inference about their connectivity by assessing connectivity of the corresponding symbol-graph nodes. More specifically, if $d$ is at least logarithmic in the number of nodes in the graph, then, even for relatively small amounts of noise, the algorithm will see all node-pairs as connected within distance $d$;
hence any informative inference will be unlikely.

\begin{theorem}
\label{th:multi-hop:impossiblity}
Let $c > 1$ be a constant and $p_-, \varepsilon_-, \lambda$ be parameters of the sampling process in Algorithm~\ref{al:sampling} on a meaning graph $\gm$ with $n$ nodes. Let $d \in \mathbb{N}$ and $\tilde{d} = \lambda d$. If
$$
    \vspace{-0.04cm}
    {p_- \oplus \varepsilon_-} \geq \frac{c}{\lambda n} \ \ \ \text{and} \ \ \ d \in \Omega (\log n),
    \vspace{-0.05cm}
$$
then the connectivity algorithm \textsc{Separator}$_{X^{\tilde{d}}_S}$ almost-surely infers any node-pair in $\gm$ as connected, and is thus not $\gamma$-accurate for any $\gamma > 0$ for the $d$-connectivity problem.
\end{theorem}
\begin{proof}[Proof idea]
\small
One can show that, for the given choice of parameters, noisy edges would dominate over informative ones and the symbol-graph would be a densely connected graph (i.e., one cannot distinguish actual connectivities from the spurious ones). 
\end{proof}

This result exposes an inherent limitation to multi-hop reasoning: even for small values of noise, the diameter of the symbol graph becomes very small, namely, logarithmic in $n$. 
This has a resemblance to similar observations in various contexts, commonly known as the \emph{small-world phenomenon}. This principle states that in many real-world graphs, nodes are all linked by short chains of acquaintances, such as ``six degrees of separation"~\citep{Milgram67,WattsSt98}. Our result affirms that if NLP reasoning algorithms are not designed carefully, such macro behaviors 
will necessarily become bottlenecks. 

We note that the preconditions of Theorems~\ref{the:reasoning:possiblity:detailed} and~\ref{th:multi-hop:impossiblity} are disjoint, that is, both results do not apply simultaneously. Since $\ball(.) \geq 1$ and $\lambda \geq 1$, Theorem~\ref{the:reasoning:possiblity:detailed} requires ${p_- \oplus \varepsilon_-} 
\leq { \frac{1}{{2e}\lambda^2 n} < \frac{1}{\lambda^2 n}}$, whereas Theorem~\ref{th:multi-hop:impossiblity} applies when ${p_- \oplus \varepsilon_-} \geq \frac{c}{\lambda n} > \frac{1}{\lambda^2 n}$.

\section{Limits of General Algorithms}

While in the previous section we showed limitations of multi-hop reasoning in inferring long-range relations, here we extend the argument to prove the difficulty for \emph{any} reasoning algorithm. 

Our exposition is algorithm independent; in other words, we do not make any assumption on the choice of $E_S(s, s')$ in Equation~\ref{eq:decision}.
In our analysis we use the spectral properties of the graph to quantify local information within graphs. 

\begin{wrapfigure}{r}{0.42\textwidth}
    \centering
    \includegraphics[scale=0.19, trim=0cm 1.0cm 0cm 0cm]{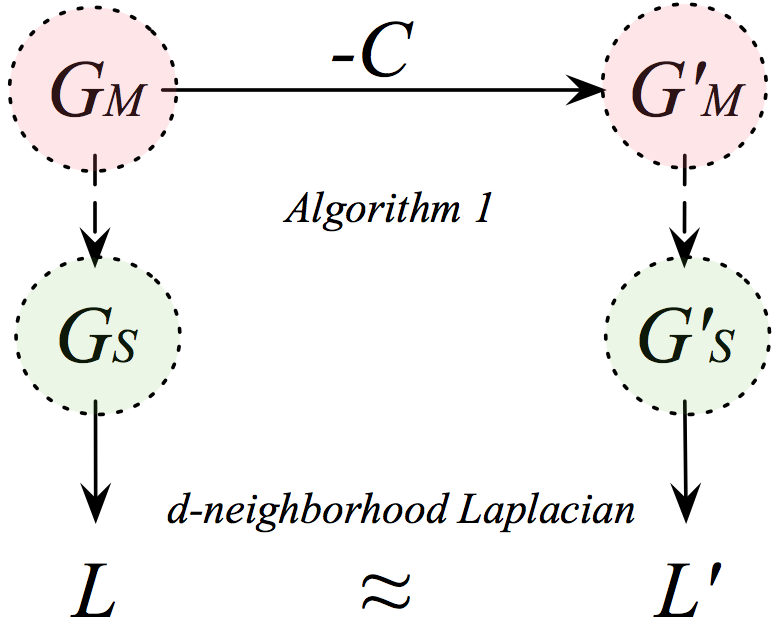}
    \caption{
    \small
    The construction considered in Definition~\ref{cut:construction}. 
    The node-pair $m$-$m'$ is connected with distance $d$ in $G_M$, and disconnected in $G_M'$, after dropping the edges of a cut $C$. For each symbol graph, we consider it ``local'' Laplacian. 
    }
    \label{fig:laplacians}
\end{wrapfigure}
Consider a meaning graph $G_M$ in which two nodes $m$ and $m'$ are connected. 
We drop edges in a min-cut $C$ to make the two nodes disconnected and get $G_M'$ (Figure~\ref{fig:laplacians}).

\begin{definition}
\label{cut:construction}
Define a pair of meaning-graphs $G$ and $G'$, both with size $n$ and satisfying the ball assumption $\ball(d)$, with the following properties: (1) $m \pathk{d} m'$ in $G$, (2) $m \npathk{} m'$ in $G'$, (3) $E_{G'} \subset E_G$, (4) $C = E_G \setminus E_{G'}$, an $(m, m')$ min-cut of $G$. 
\end{definition}
We define a uniform distribution over all the instances that satisfy the construction explained in Definition~\ref{cut:construction}: 
\begin{definition}
\label{def:distribution}
We define a distribution $\mathcal{G}$ over pairs of possible meaning graphs $G, G'$ and pairs of nodes $m,m'$ which satisfies the requirements of Definition \ref{cut:construction}. Formally, \(\mathcal{G}\) is a uniform distribution over the following set:
\vspace{-0.05cm}
{
\small
$$ 
\{(G,G',m,m') \mid G,G',m,m' \text{satisfy Definition \ref{cut:construction}}\}.
$$
}
\end{definition}

For the meaning graphs, we sample a symbol graph $G_S$ and $G'_S$, as denoted in Figure~\ref{fig:laplacians}. 
In the sampling of $G_S$ and $G'_S$, all the edges share the randomization, except for the ones that correspond to $C$ (i.e., the difference between the $G_M$ and $G'_M$).  
Let $\mathcal{U}$ be the union of the nodes involved in $\tilde{d}$-neighborhood of $s, s'$, in $G_S$ and $G'_S$. 
Define $L, L'$ to be the Laplacian matrices corresponding to the nodes of $\mathcal{U}$. As $n$ grows, the two Laplacians become less distinguishable whenever {$p_- \oplus  \varepsilon_-$} and $d$ are large enough: 

\begin{lemma}
\label{lem:normLaplacianClose}
Let $c > 0$ be a constant and $p_-, \lambda$ be parameters of the sampling process in Algorithm~\ref{al:sampling} on a pair of meaning graphs $G$ and $G'$ on $n$ nodes constructed according to Definition~\ref{cut:construction}. Let $d \in \mathbb{N}, \tilde{d} \geq \lambda d,$ and $L, L'$ be the Laplacian matrices for the $\tilde{d}$-neighborhoods of the corresponding nodes in the sampled symbol graphs $\gs$ and $\gs'$. If
${ p_- \oplus \varepsilon_-} \geq \frac{c \log n}{n}$ and 
$d > \log n,$
then, with a high probability, the two Laplacians are close:
$$
\vspace{-0.05cm}
\norm{\tilde{L}-\tilde{L'}} \leq \frac{ \sqrt{2\lambda}\, \ball(1)}{ \sqrt{n \log (n \lambda)} } 
\vspace{-0.05cm}
$$
\end{lemma}

This can be used to show 
that, for such large enough $p_-$ and $d$, the two symbol graphs, $G_S$ and $G'_S$ sampled as above, are indistinguishable by any 
function operating over a $\lambda d$-neighborhood of $s, s'$ in \gs, with a high probability.

A reasoning function can be thought of a mapping defined on normalized Laplacians, 
since they encode all the information in a graph.  
For a reasoning function $f$ with limited precision, the input space can be partitioned into regions where the function is constant; and for large enough values of $n$ both $\tilde{L}, \tilde{L}'$ (with a high probability) fall into regions where $f$ is constant.

Note that a reasoning algorithm is oblivious to the the details of $C$, i.e. it does not know where $C$ is, or where it has to look for the changes. Therefore a realistic algorithm ought to use the neighborhood information collectively.

In the next lemma, we define a function $f$ to characterize the reasoning function, which uses Laplacian information and maps it to binary decisions. We then prove that for any such functions, there are regimes that the function won't be able to distinguish $\tilde{L}$ and $\tilde{L}'$: 

\begin{lemma}
\label{lemma:improbable:recovery}
Let meaning and symbol graphs be constructed under the conditions of Lemma~\ref{lem:normLaplacianClose}. Let $\beta > 0$ and $f: \mathbb{R}^{\dimm \times \dimm} \rightarrow \{0, 1\}$ be the indicator function of an open set.
Then there exists $n_0 \in \mathbb{N}$ such that for all $n \geq n_0$: 
$$
\mathbb{P}_{ \substack{ (G,G',m,m') \sim \mathcal{G} \\ G_S \leftarrow \alg{G}, \\G'_S \leftarrow \alg{G'}  } }\left[ f(\tilde{L}) = f(\tilde{L}') \right] \geq 1 - \beta.
$$
\end{lemma}

This yields the following result:

\begin{theorem}
\label{theorem:impossiblity:general:reasoning}
Let $c > 0$ be a constant and $p_-, \varepsilon_-, \lambda$ be parameters of the sampling process in Algorithm~\ref{al:sampling} on a meaning graph $\gm$ with $n$ nodes. Let $d \in \mathbb{N}$. If
$$
    {p_- \oplus \varepsilon_-} > \frac{c \log n}{\lambda n} \ \ \ \text{and} \ \ \ d > \log n,
$$
then there exists $n_0 \in \mathbb{N}$ such that for all $n \geq n_0$, \emph{any} algorithm cannot distinguish, with a high probability, between two nodes in \gm\ having a $d$-path vs.\ being disconnected, and is thus not $\gamma$-accurate for any $\gamma > 0$ for the $d$-connectivity problem.
\end{theorem}
\begin{proof}[Proof idea]
\small
The proof uses Lemma~\ref{lemma:improbable:recovery} to show that for the given choice of parameters, the informative paths are indistinguishable from the spurious ones, with high probability. 
\end{proof}

This reveals a fundamental limitation: under noisy conditions, our ability to infer interesting phenomena in the meaning space is limited to a small, logarithmic neighborhood.

\section{Empirical Analysis}

Our formal analysis thus far provides worst-case bounds for two regions in the rather large spectrum of noisy sampling parameters for the symbol space, namely, when $p_- \oplus \varepsilon_-$ and $d$ are either both small (Theorem~\ref{the:reasoning:possiblity}), or both large (Theorem~\ref{th:multi-hop:impossiblity}).

This section complements the theoretical findings in two ways: (a) by grounding the formalism empirically into a real-world knowledge graph, and (b) by quantifying the impact of noisy sampling parameters on the success of the connectivity algorithm. 
    We use $\varepsilon_- = 0$ for this experiments, but the effect turns out to be identical as long as {$p_- \oplus \varepsilon_-$} stays unchanged (see Remark~\ref{remark:folding} in Appendix).	

Specifically, we consider FB15k237~\citep{ToutanovaCh15} containing a set of $\langle$head, relation, target$\rangle$ triples from a curated knowledge base, FreeBase~\citep{BEPST08}. For scalability, we use a subset that relates to the movies domain,\footnote{Specifically, relations beginning with {\tt /film/}.} resulting in 2855 distinct entity nodes and 4682 relation edges. 
We treat this as the meaning graph and sample a symbol graph as per Algorithm~\ref{al:sampling} to simulate the observed graph derived from text.

We sample symbol graphs for various values of $p_-$ and plot the resulting symbol and meaning graph distances in Figure~\ref{fig:emp:distance:pm}. For every value of $p_-$ ($y$-axis), we sample points in the meaning graph separated by distance $d$ ($x$-axis). For these points, we compute the average distance between the corresponding symbol nodes, and indicate that in the heat map using color shades.

We make two observations from this simulation. First, for lower values of $p_-$, disconnected nodes in the meaning graph (rightmost column)
are clearly distinguishable from meaning nodes with short paths (small $d$) as predicted by Theorem~\ref{the:reasoning:possiblity}, but harder to distinguish from nodes at large distances (large $d$). Second, and in contrast, for higher values of $p_-$, almost every pair of symbol nodes is connected with a very short path (dark color), making it impossible for a distance-based reasoning algorithm to confidently assess $d$-connectivity in the meaning graph. This simulation also confirms our finding in Theorem~\ref{th:multi-hop:impossiblity}: any graph with $p_- \geq 1/\lambda n$, which is $\sim 0.0001$ in this case, cannot distinguish disconnected meaning nodes from nodes with paths of short (logarithmic) length (top rows). 

\begin{figure}
    \centering
    \includegraphics[scale=0.27,trim=1.9cm 1.4cm 0.5cm 0cm, clip=false]{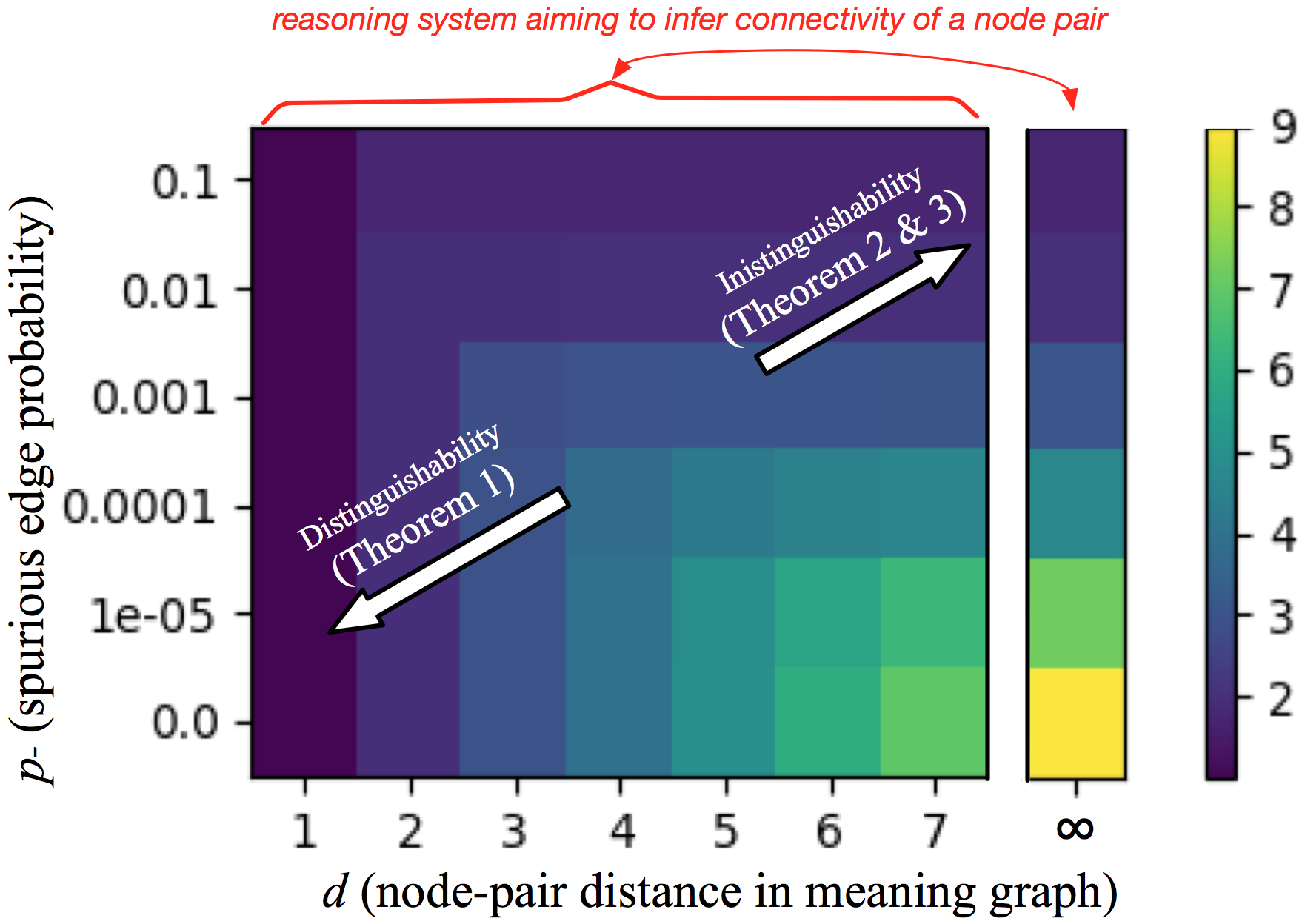}
    \caption{ 
    Various colors in the figure depict the average distance between node-pairs in the symbol graph, for each true meaning-graph distance $d$ (x-axis), as the noise parameter $p_-$ (y-axis) is varied. {The goal is to distinguish squares in the column for a particular $d$ with the corresponding squares in the right-most column, which corresponds to node-pairs being disconnected. This is easy in the bottom-left regime and becomes progressively harder as we move upward (more noise) or rightward (higher meaning-graph distance).} ($\varepsilon_+=0.7, \lambda = 3
    $)}
    \label{fig:emp:distance:pm}
\end{figure}

{
\section{Summary, Discussion and Practical Lessons}
Our work is inspired by empirical observations of ``semantic drift'' of reasoning algorithms, as the number of hops is increased. There are series of works sharing this empirical observation; for example, \cite{FJHSC15} show modest benefits up to 2-3 hops, and then decreasing performance; \cite{Jansen16,JWMM18} made similar observations in graphs built out of larger structures such as sentences, where the performance drops off around 2 hops. This pattern has interestingly been observed in a number of results with a variety of representations, including word-level representations, graphs, and traversal methods. 
The question we are after in this work is whether the field might be hitting a fundamental limit on multi-hop information aggregation using existing methods and noisy knowledge sources.

Our ``impossibility'' results are reaffirmations of the empirical intuition in the field. This means that multi-hop inference (and any algorithm that can be cast in that form), as we've been approaching it, is exceptionally unlikely to breach the few-hop barrier predicted in our analysis. 


There are at least two practical lessons: 
\begin{enumerate}[leftmargin=.55cm]
    \item 
    There are several efforts in the field pursuing ``very long'' multi-hop reasoning. Our results suggest that such efforts, especially without a careful understanding of the limitations, are unlikely to succeed, unless some fundamental building blocks are altered.
    
    \item 
    A corollary of this observation suggests that, due to the limited number of hops, practitioners must focus on richer representations that allow reasoning with only a ``few'' hops. This, in part, requires higher-quality abstraction and grounding mechanisms. It also points to alternatives, such as offline KB completion/expansion, which indirectly reduce the number of steps needed at inference time. It basically suggests that ambiguity and variability must be handled well to reduce the number of hops needed.
\end{enumerate}

Finally, we note that our proposed framework applies to any machine comprehension task over natural text that requires multi-step decision making, such as multi-hop QA or textual entailment.
}

\chapter{Summary and Future Work}
\label{chapter:conclusion:summary:future}
\epigraph{ 
\includegraphics[scale=0.55,trim=4.4cm 22.3cm 2.2cm 3cm, clip=true]{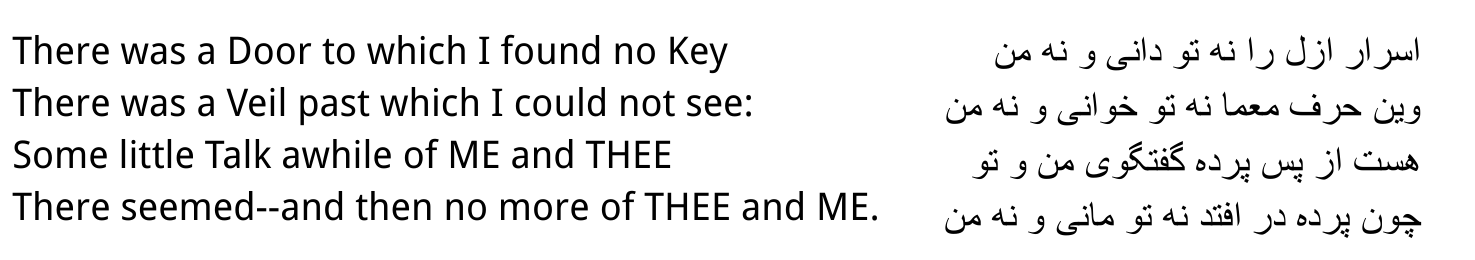}
}{--- \textup{Omar Khayyam}, Rubaiyat, 1120 CE}

This thesis aims at progressing towards natural language understanding, by means of the task of question answering. 
This chapter, gives a summary of our contributions across this document and provides a few angles along which we would like to extend this work. 


\section{Summary of Contributions}
We start the discussion in Chapter 2 by providing a thorough review of the past literature concerning NLU, highlighting the ones that are related to the works in this thesis. 

Chapter 3 studies reasoning systems for question answering on elementary-school science exams, using a semi-structured knowledge base. We treat QA as a subgraph selection problem and then formulate this as an ILP optimization. 
Most importantly, this formulation allows multiple, semi-formally expressed facts to be combined to answer questions, a capability outside the scope of IR-based QA systems.
In our experiments, this approach significantly outperforms both the previous best attempt at structured reasoning for this task,  and an IR engine provided with the same knowledge. 
Our effort has had great impacts since publication. Our work has inspired others to to build systems based on our design and to improve the state of the art in other domains; for instance, \cite{KhotSaCl17} uses similar ideas to reasoning with OpenIE tuples~\citep{EBSW08}.
In addition, the system has been incorporated into Allen Institute's reading-comprehension project\footnote{\url{https://allenai.org/aristo/}} and is shown to give a significant boost to their performance~\citep{CEKSTTK16}. 
Even after a couple of years, the system has been shown to be 
among the best systems on a recently-proposed reading comprehension task~\citep{CCEKSST18}.  


Chapter 4 extends our abductive reasoning system (from Chapter 3) to consume raw text as input knowledge. This is the first system to successfully use a wide range of semantic abstractions to perform a high-level NLP task like Question Answering. 
Departing from the currently popular paradigm of generating a very large dataset and learning ``everything'' from it in an end-to-end fashion, we  demonstrate that one can successfully leverage pre-trained NLP modules to extract a sufficiently complete linguistic abstraction of the text that allows answering interesting questions about it. This approach is particularly valuable in settings where there is a small amount of data. Instead of exploiting peculiarities of a large but homogeneous dataset, as many state-of-the-art QA systems end up doing, we focus on confidently performing certain kinds of reasoning, as captured by our semantic graphs and the ILP formulation of support graph search over them. 
    

Chapter 5 introduces the concept of essential question terms and demonstrates its importance for question answering. 
We introduce a dataset for this task and show that our classifier trained on this dataset substantially outperforms several baselines in identifying and ranking question terms by the degree of essentiality.
Since the publication, this work has been picked up by others to significantly improve their systems~\citep{NZCM18}. 



Chapter 6 presents a reading comprehension dataset in which questions require \emph{reasoning over multiple sentences}. This dataset contains $\sim6k$ questions from different domains and wide variety of complexities.
We have shown a significant performance different between human and state-of-the-art systems and we hope that this performance gap will encourage the community to work towards more sophisticated reasoning systems. It is encouraging to see that 
the work is already been used in a couple of works   
\citep{SYYC19,TKKSB19,WYSCYRM19}.


Chapter 7 offers a question answering dataset dedicated to \emph{temporal common sense understanding}. 
We show that systems equipped with the state-of-the-art techniques are still far behind human performance. 
We hope that the dataset will bring more attention to the study of common sense (especially in the context of understanding of \emph{time}).

In Chapter 8, we develop 
a theoretical formalism to investigate fundamental limitations pertaining to multi-step reasoning in the context of natural language problems.
We present the first analysis of reasoning 
in the context of properties like 
ambiguity, variability, incompleteness, and inaccuracy. 
We show that a multi-hop inference (and any algorithm that can be cast in that
form), as we've been approaching it, is exceptionally unlikely to breach the few-hop barrier predicted in our analysis.
Our results suggest that such efforts, especially without
a careful understanding of the limitations, are unlikely to succeed, unless some fundamental building blocks are altered.
A corollary of this observation suggests that, practitioners must focus on richer representations that allow reasoning
with only a ``few'' hops. 
This, in part, requires higher-quality abstraction and grounding mechanisms.
In other words, ambiguity and variability must be handled well to reduce the number of hops needed.


\section{Discussion and Future Directions}
This thesis has taken a noticeably distinct approach towards a few important problems in the field and has shown progress on multiple ends. 
For example, the formalism of Chapter 3 and 4 are novel and provide general ways to formalize and implement reasoning algorithms. The datasets of Chapter 6 and 7 are distinct from the many QA datasets in the field. The theoretical analysis of Chapter 8 takes a uniquely distinct formal analysis of reasoning in the context of natural language.  

All these said, 
there are many issues that 
are not addressed 
as extensively as we could have (or should have), or there are aspects that turned out slightly differently from  what we initially expected. 

Looking back at the reasoning formalism of Chapter 4, we underestimated the hardness of extracting the underlying semantic representations. Even though the field has made significant progress in low-level NLP tasks (like SRL or Coreference), such tasks still suffer from brittleness and lack of transfer across domains.  And brittleness in the extraction of such annotations, result in exponentially bigger errors when 
reasoning with them (as also justified by the theoretical observations of Chapter 8); in practice, it worked well only for short-ranged chains (1, 2, and sometimes 3 hops). 
With more recent progress in unsupervised representations and improvement of semantic extraction systems, my hope is to redo these ideas in the coming years and revisit the remaining challenges.

A vision that I would like to pursue 
(influenced by discussions with my advisor)
is reasoning with \emph{minimal} data. We (humans) are able to perform the same  reasoning on many high-level concepts and are able to transfer them in all sorts of domains: for instance, an average human uses the same inductive reasoning to conclude \emph{the sky is blue} and inferring that  \emph{there is another number after every number}.  
Effective (unsupervised) representation could potentially need a huge amount of data (and many parameters), but successful reasoning systems will likely need very minimal data (and very simple, but general definitions).

Over the past years, the field has witnessed a wave of activity on unsupervised language models~\citep{PNIGCLZ18,DCLT18}. There are many questions with respect  to the success of such models on several datasets: for instance, what kinds of reasoning are they capable of? what is it that they are missing? And how we can address them by possibly creating hybrid systems.  
What is clear is that these systems will offer increasingly richer representations of meaning; we need better ways to effectively understand what these systems are capable of and what are the scenarios they are used to represent. And in conjunction to understanding their capabilities and limitations, we have to build reasoning algorithms on top of them. It's unlikely that these tools will ever be enough to solve all of our challenges; one has to equip these representations with the ability to reason, especially when they face an unusual/unseen scenario.

In Chapter 5 (essential terms) an initial motivation was to model \emph{knowing what we don't know}~\citep{RajpurkarJiLi18}; basically, systems should be able to infer whether they have enough confidence about the answer to a given query before acting. 
In hindsight, I think our supervised system 
ended up using too many shallow features, which didn't end up generalizing to tricky instances. 
Additionally, it would have been better if the decision of essentiality was more involved within reasoning systems (rather than an independently supervised classifier, which limited its domain transfer). 

The datasets of Chapter 6 and 7 are critical 
parts of this thesis which, I suspect, are likely to be remembered longer than the rest of the chapters. In general, the construction of datasets (including the ones we described) is a menial task. 
It's unfortunate that many small empirical details are usually left out. 
It is not clear to me whether using static datasets is the best way for the road ahead. 
In the future, I hope that the field discovers more effective ways of measuring the progress towards NLU.

A key issue contributing to the complexity of NLU (and Question Answering) is the set of implied information  (common sense). 
We touch upon a class of such understanding in Chapter 7, where we introduce a dataset for such problems. A natural next step is addressing such questions and exploring the many ways we can incorporate such understanding in the models. 

The analysis of Chapter 8 is uniquely distinct within the field. That said, there are many issues that make me feel unsatisfied about our current attempt. In particular, there are many assumptions that may or may not stand the test of time (e.g., the generative construction of symbol graph from the meaning graph or the connectivity reasoning as a proxy for the actual reasoning in language). 
And there are some important reasoning phenomena missing from this formalism: 
conditional reasoning, transitivity and directionality, inductive reasoning, just to name a few. 
In general, our (the field's) understanding of ``reasoning'' (and its formalisms) is very limited. 
And the existing formalisms are not easily applicable, since 
those who formalized reasoning were not intimately aware of the complexity of NLU; they were philosophers and mathematicians. In practice, it's really hard to make the existing theories of reasoning work in the existence of many of the properties of language. 
In the coming years, 
I would like to see more efforts on reconciling the issues in the interface of ``language'' and ``reasoning''.

\end{mainf}

\appendix

\newenvironment{appendixf}{}{}
\titleformat{\chapter}[hang]{\large\center}{APPENDIX}{0 pt}{} 
\titlespacing*{\chapter}{0pt}{-33 pt}{6 pt} 
\begin{appendixf}

\addtocontents{toc}{\protect\setcounter{tocdepth}{-1}} 
\clearpage
\chapter{}
\addtocontents{toc}{\protect\setcounter{tocdepth}{1}} 
\addcontentsline{toc}{chapter}{APPENDIX} 

\section{Supplementary Details for Chapter 3}

\subsection{The ILP Model for \tableilp}
\label{appendix:optimization:details:tableilp}
\textbf{Variables: }
We start with a brief overview of the basic variables and how they are combined into high level variables.

\begin{wrapfigure}{r}{0.6\textwidth}
\centering
\small
  \setlength\extrarowheight{-12pt}
\begin{tabular}{|c|L{40ex}|} 
\hline
 \bigstrut[t] \textsc{Reference} & \bigstrut[t] \textsc{Description} \\ 
\hline
\bigstrut[t] $i$ & index over tables \\ 
$j$ & index over table rows \\ 
$k$ & index over table columns \\ 
$l$ & index over lexical constituents of question \\
$m$ & index over answer options \\
 \hline
\bigstrut[t] $\xOne{.} $ & a unary variable \\
$\xTwo{.}{.} $ & a pairwise variable \\
\hline
\end{tabular}
\caption{Notation for the ILP formulation. }
\label{table:variables-and-description}
\end{wrapfigure}
Table~\ref{table:variables-and-description} summarizes our notation to refer to various elements of the problem, such as $\tableCell$ for cell $(j,k)$ of table $i$, as defined in Section 3. We define variables over each element by overloading $\xOne{.}$ or $\xTwo{.}{.}$ notation which refer to a binary variable on elements or their pair, respectively. Table~\ref{table:ilp-variables} contains the complete list of basic variables in the model, all of which are binary. The pairwise variables are defined between pairs of elements; e.g., $\xTwo{\tableCell}{\qCons}$ takes value 1 if and only if the corresponding edge is present in the support graph. Similarly, if a node corresponding to an element of the problem is present in the support graph, we will refer to that element as being \emph{active}.


In practice we do not create pairwise variables for all possible pairs of elements; instead we create pairwise variables {for edges that have an entailment score exceeding a threshold}. For example we create the pairwise variables $\xTwo{\tableCell}{\tableCellPrimePrime}$ only if $w(\tableCell, \tableCellPrimePrime) \geq  \textsc{MinCellCellAlignment}$. An exhaustive list of the minimum alignment thresholds for creating pairwise variables is in Table~\ref{table:pairwise-thresholds}.   

Table~\ref{table:ilp-variables} also includes some high level unary variables, which help conveniently impose structural constraints on the support graph $G$ we seek. An example is the \emph{active row} variable $\xOne{\tableVar}$ which should take value 1 if and only if at least a cell in row $j$ of table $i$.

\begin{table*} 
\centering
\small 
\renewcommand{\sfdefault}{phv}
\resizebox{\textwidth}{!}
{
\begin{tabular}{|l|lc|lc|lc|lc|}	
    \hline 
	\multirow{2}{*}{Pairwise Variables} &
 	$\xTwo{\tableCell}{\tableCellPrimePrime} $ & 1 &
    $\xTwo{\tableCell}{\tableCellPrime} $ & $w(\tableCell, \tableCellPrime) - 0.1$ &
    $\xTwo{\tableCell}{\qCons} $ &  $w(\qCons, \tableCell)$ &  
    $\xTwo{\header}{\qCons} $ &  $w(\qCons, \header)$  \\   
    & $\xTwo{\tableCell}{\option}$ & $w(\tableCell, \option)$ & 
    $\xTwo{\header}{\option}$ & $w(\header, \option)$ & 
    &
    & & \\   
    \hline 
 	\multirow{2}{*}{Unary Variables}
    & $\xOne{\tableVar}$ & 1.0  &  
      $\xOne{\rowVar}$   & -1.0 & 
      $\xOne{\columnVar} $  & 1.0 &   
      $\xOne{\header} $ & 0.3 \\ 
      & $\xOne{\tableCell} $  & 0.0 &    
      $\xOne{\qCons}$ & 0.3 &   
      & & & \\
	\hline 
\end{tabular}
}
\caption{The weights of the variables in our objective function. In each column, the weight of the variable is mentioned on its right side. The variables that are not mentioned here are set to have zero weight.}
\label{table:objective-details}
\end{table*}

\textbf{Objective function: }
Any of the binary variables defined in our problem are included in the final weighted linear objective function. The weights of the variables in the objective function (i.e. the vector $\textbf{w}$ in Equation~\ref{eq:ilp:obj}) are {set} according to Table~\ref{table:objective-details}. In addition to the current set of variables, {we introduce auxiliary variables for certain constraints}. Defining auxiliary variables is a common trick for linearizing more intricate constraints at the cost of having more variables. 

\textbf{Constraints: }
Constraints are significant part of our model in imposing the desirable behaviors for the \textit{support graph} (cf. Section 3.1). 

The complete list of the constraints is explained in Table~\ref{table:constraints}. While groups of constraints are defined for different purposes, it is hard to partition them into disjoint sets of constraints. Here we give examples of some important constraint groups. 

\textbf{Active variable constraints:} An important group of constraints relate variables to each other. The unary variables are defined through constraints that relate them to the basic pairwise variables. For example, active row variable $\xOne{\tableVar}$ should be active if and only if any cell in row $j$ is active.  (constraint~\ref{cons:rowIsActiveIfAnyCellInRowIsActive}, Table~\ref{table:constraints}).

\textbf{Correctness Constraints:} A simple, but important set of constraints force the basic correctness principles on the final answer. For example $G$ should contain exactly one answer option which is expressed by constraint~\ref{cons:OnlyASingleOption}, Table~\ref{table:constraints}. Another example is that, $G$ should contain at least a certain number of constituents in the question, which is modeled by constraint~\ref{eq:MinActiveQCons}, Table~\ref{table:constraints}.

\textbf{Sparsity Constraints:} Another group of constraint induce simplicity (sparsity) in the output. For example $G$ should use at most a certain number of knowledge base tables (constraint~\ref{eq:MAXTABLESTOCHAIN}, Table~\ref{table:constraints}), since letting the inference use any table could lead to unreasonably long, and likely error-prone, answer chains.



\subsection {Features used in Solver Combination}

{To combine the predictions from all the solvers, we learn a Logistic Regression model~\citep{CEKSTTK16} that returns a probability for an answer option, $a_i$, being correct based on the following features.}

\textbf{Solver-independent features:}
Given the solver scores $s_j$ for all the answer options $j$, we generate the following set of features for the answer option $a_i$, for each of the solvers:
\begin{enumerate}
\item Score = $s_i$
\item Normalized score = $\frac{s_i}{\sum_j s_j}$
\item Softmax score = $\frac{\exp(s_i)}{\sum_j \exp(s_j)}$
\item Best Option, set to $1$ if this is the top-scoring option = $\mathbb{I}(s_i = \max s_j)$ 
\end{enumerate}

\textbf{TableILP-specific features:}
Given the proof graph returned for an option, we generate the following 11 features apart from the solver-independent features:
\begin{enumerate}
\item Average alignment score for question constituents
\item Minimum alignment score for question constituents
\item Number of active question constituents
\item Fraction of active question constituents
\item Average alignment scores for question choice
\item Sum of alignment scores for question choice
\item Number of active table cells
\item Average alignment scores across all the edges
\item Minimum alignment scores across all the edges
\item Log of number of variables in the ILP
\item Log of number of constraints in the ILP
\end{enumerate}

\begin{table*}
\centering
\small 
\renewcommand{\sfdefault}{phv}
\resizebox{\textwidth}{!}{
\begin{tabular}{|lc|lc|lc|}	
    \hline 
    \textsc{MinCellCellAlignment} &  0.6 & \textsc{MinCellQConsAlignment} & 0.1  & \textsc{MinTitleQConsAlignment} & 0.1 \\ 
    \textsc{MinTitleTitleAlignment} &  0.0 & \textsc{MinCellQChoiceAlignment} & 0.2 & \textsc{MinTitleQChoiceAlignment} & 0.2\\ 
    \textsc{MinCellQChoiceConsAlignment} &  0.4 & \textsc{MinCellQChoiceConsAlignment} & 0.4 & \textsc{MinTitleQChoiceConsAlignment} & 0.4 \\ 
    \textsc{MinActiveCellAggrAlignment} & 0.1 & \textsc{MinActiveTitleAggrAlignment} & 0.1&
 &   \\ 
\hline 
\end{tabular}
}
\caption{Minimum thresholds used in creating pairwise variables.  }
\label{table:pairwise-thresholds}
\end{table*}

\begin{table*}
\centering
\small 
\renewcommand{\sfdefault}{phv}
\resizebox{\textwidth}{!}{
\begin{tabular}{|lc|lc|lc|}	
    \hline 
    \textsc{MaxTablesToChain} &  4 & \textsc{qConsCoalignMaxDist} & 4 & \textsc{WhichTermSpan} &  2 \\ 
    \textsc{WhichTermMulBoost} &  1 & \textsc{MinAlignmentWhichTerm} & 0.6 & \textsc{TableUsagePenalty} & 3 \\ 
    \textsc{RowUsagePenalty} &  1 & \textsc{InterTableAlignmentPenalty} & 0.1 & \textsc{MaxAlignmentsPerQCons} & 2 \\ 
    \textsc{MaxAlignmentsPerCell} &  2 & \textsc{RelationMatchCoeff} & 0.2 & \textsc{RelationMatchCoeff} & 0.2 \\ 
     \textsc{EmptyRelationMatchCoeff} & 0.0 & \textsc{NoRelationMatchCoeff} & -5 & \textsc{MaxRowsPerTable}  & 4 \\ 
     \textsc{MinActiveQCons} & 1 & \textsc{MaxActiveColumnChoiceAlignments} & 1 & \textsc{MaxActiveChoiceColumnVars} & 2 \\
\textsc{MinActiveCellsPerRow} & 2 & & & & \\ 
\hline 
\end{tabular}
}
\caption{Some of the important constants and their values in our model.  }
\label{table:coefficients}
\end{table*}

\newpage 
\newcommand{\linespacingInTable}{0.6}

\begin{table*}
	\small 
	\renewcommand{\arraystretch}{0.0}
    \linespread{\linespacingInTable}\selectfont\centering
    \resizebox{\textwidth}{!}{
	\begin{tabular}{|L{7.5cm}L{9.5cm}|}
      \hline 
      Collection of basic variables connected to header column  $k$ of table  $i$: & 
      \begin{equation}
          \mathcal{H}_{ik} = \setOf{(\header, \qCons); \forall l } \cup \setOf{(\header, \option); \forall m} 
      \end{equation} \\
            Collection of basic variables connected to cell $j, k$ of table $i$: & 
       \begin{equation}
              \mathcal{E}_{ijk} = \setOf{(\tableCell, \tableCellPrime); \forall i', j', k' } \cup \setOf{(\tableCell, \option); \forall m} \cup \setOf{(\tableCell, \qCons); \forall l}  
      \end{equation} \\
            Collection of basic variables connected to column  $k$  of table  $i$ & 
       \begin{equation}
               \mathcal{C}_{ik} = \mathcal{H}_{ik} \cup \left(  \bigcup_{j} \mathcal{E}_{ijk} \right)
      \end{equation} \\
            Collection of basic variables connected to row  $j$  of table  $i$: & 
       \begin{equation}
          \mathcal{R}_{ij} =  \bigcup_{k} \mathcal{E}_{ijk} 
      \end{equation} \\
            Collection of non-choice basic variables connected to row  $j$  of table  $i$:  & 
       \begin{equation}
          \mathcal{L}_{ij} = \setOf{(\tableCell,\tableCellPrime); \forall k, i', j', k'} \cup \setOf{ (\tableCell, \qCons); \forall k, l} 
      \end{equation} \\
            Collection of non-question basic variables connected to row  $j$  of table  $i$:   & 
       \begin{equation}
          \mathcal{K}_{ij} = \setOf{(\tableCell,\tableCellPrime); \forall k, i', j', k'} \cup \setOf{ (\tableCell, \option); \forall k, m}  
      \end{equation} \\
            Collection of basic variables connected to table  $i$:  & 
       \begin{equation}
              \mathcal{T}_i = \bigcup_{k} \mathcal{C}_{ik}
      \end{equation} \\
            Collection of non-choice basic variables connected to table  $i$:  & 
       \begin{equation}
              \mathcal{N}_{i} = \setOf{(\header, \qCons); \forall l } \cup \setOf{(\tableCell,\tableCellPrime); \forall j, k, i', j', k'}  \cup \setOf{ (\tableCell, \qCons); \forall j, k, l}  
      \end{equation} \\
            Collection of basic variables connected to question constituent $\qCons$:    & 
       \begin{equation}
              \mathcal{Q}_l = \setOf{ (\tableCell, \qCons); \forall i, j, k}  \cup \setOf{ (\header, \qCons); \forall i, k}  
      \end{equation} \\
            Collection of basic variables connected to option  $m$ & 
       \begin{equation}
              \mathcal{O}_m = \setOf{ (\tableCell, \option); \forall i, j, k}  \cup \setOf{ (\header, \option); \forall i, k}  
      \end{equation} \\
            Collection of basic variables in column $k$ of table $i$ connected to option  $m$: & 
       \begin{equation}
              \mathcal{M}_{i, k, m} = \setOf{ (\tableCell, \option); \forall j}  \cup \setOf{ (\header, \option)}  
      \end{equation} \\
      \hline 
    \end{tabular}
    }
  \caption{All the sets useful in definitions of the constraints in Table~\ref{table:constraints}. }
  \label{table:ilp-sets}
\end{table*}

\newpage

\begin{table*}
	\small 
	\renewcommand{\arraystretch}{0.0}
    \linespread{\linespacingInTable}\selectfont\centering
    \resizebox{\textwidth}{!}{
	\begin{tabular}{|L{9.5cm}L{7.5cm}|}
      \hline 
		If any cell in row $j$ of table $i$ is active, the row should be active.
        & 
      \begin{equation}
      	  \label{cons:rowIsActiveIfAnyCellInRowIsActive}
          \xOne{\rowVar}  \geq \xTwo{\tableCell}{e}, \forall (\tableCell, e) \in \mathcal{R}_{ij}, \forall i, j, k 
      \end{equation} \\
      	If the row $j$ of table $i$ is active, at least one cell in that row must be active as well. 
        & 
      \begin{equation}
          \sum_{(\tableCell, e) \in \mathcal{R}_{ij}} \xTwo{\tableCell}{e}  \geq  \xOne{\rowVar}, \forall i, j
      \end{equation} \\
      	 Column $j$ header should be active if any of the basic variables with one end in this column header are active. 
        & 
      \begin{equation}
          \xOne{\header}  \geq \xTwo{\header}{e}, \forall (\header, e) \in \mathcal{H}_{ik}, \forall i, k
      \end{equation} 
      \\
            	If the header of column $j$ variable is active, at least one basic variable with one end in the end in the header
        & 
      \begin{equation}
          \sum_{(\header, e) \in \mathcal{H}_{ik}} \xTwo{\header}{e}  \geq  \xOne{\header}, \forall i 
      \end{equation} \\
		Column $k$ is active if at least one of the basic variables with one end in this column are active. 
        & 
      \begin{equation}
          \xOne{\columnVar} \geq \xTwo{\tableCell}{e}, \forall (\tableCell, e) \in \mathcal{C}_{ik}, \forall i, k
      \end{equation} \\
            \hline
    \end{tabular}
    }
\end{table*}

\begin{table*}
	\small 
	\renewcommand{\arraystretch}{0.0}
 \linespread{\linespacingInTable}\selectfont\centering
    \resizebox{\textwidth}{!}{
	\begin{tabular}{|L{10.5cm}L{8cm}|}
          \hline 
                	If the column $k$ is active, at least one of the basic variables with one end in this column should be active. 
        & 
      \begin{equation}
          \sum_{(\tableCell, e) \in \mathcal{C}_{ik}} \xTwo{\tableCell}{e} \geq  \xOne{\header}, \forall i, k   
      \end{equation} \\
      	If a basic variable with one end in table $i$ is active, the table variable is active. 
        & 
      \begin{equation}
      		\label{cons:ifEdgeConnectedToTableIsActiveTableIsActive}
          \xTwo{\tableCell}{e} \geq \xOne{\tableVar}, \forall (\tableCell, e) \in \mathcal{T}_i, \forall i 
      \end{equation} \\
         If the table $i$ is active, at least one of the basic variables with one end in the table should be active.  
        & 
      \begin{equation}
      		\label{cons:ifTableIsActiveAtLeastAnEdgeIsConnctedToIt}
          \sum_{(t, e) \in \mathcal{T}_i} \xTwo{t}{e} \geq \xOne{\tableVar}, \forall i 
      \end{equation} \\
	  	If any of the basic variables with one end in option $\option$ are on, the option should be active as well.  
        & 
      \begin{equation}
      		\label{cons:optionShouldBeActiveIfAnyEdgeConnectedToItIsActive}
          \xOne{\option} \geq \xTwo{x}{\option}, \forall (e, \option) \in \mathcal{O}_m  
      \end{equation} \\
          If the question option $\option$ is active, there is at least one active basic element connected to it 
        & 
      \begin{equation}
      		\label{cons:atLeastOneActiveEdgeConnectedToOption}
          \sum_{(e, a) \in \mathcal{O}_m} \xTwo{x}{a} \geq  \xOne{\option}   
      \end{equation} \\
          If any of the basic variables with one end in the constituent $\qCons$, the constituent must be active.  
        & 
      \begin{equation}
      		\label{cons:ifEdgeConnectedToConsConsMustBeActive}
          \xOne{\qCons} \geq \xTwo{e}{\qCons}, \forall (e, \qCons) \in \mathcal{Q}_l  
      \end{equation} \\
          If the constituent $\qCons$ is active, at least one basic variable connected to it must be active. 
        & 
      \begin{equation}
      		\label{cons:ifQConsIsActiveAtLeastOneEdgeConnectedToIt}
          \sum_{(e, \qCons) \in \mathcal{Q}_l} \xTwo{e}{\qCons}  \geq  \xOne{\qCons}   
      \end{equation} \\
          Choose only a single option. 
        & 
      \begin{equation}
      	  \label{cons:OnlyASingleOption}
          \sum_{m}\xOne{\option} \leq 1, \quad  \sum_{m}\xOne{\option} \geq 1 
      \end{equation} \\
           There is an upper-bound on the number of active tables; this is to limit the solver and reduce the chance of using spurious tables.  
        & 
      \begin{equation}
      		\label{eq:MAXTABLESTOCHAIN}
          \sum_{i} \xOne{\tableVar} \leq \textsc{MaxTablesToChain}  
      \end{equation} \\
          The number of active rows in each table is upper-bounded. 
        & 
      \begin{equation}
          \sum_{j} \xOne{\rowVar} \leq \textsc{MaxRowsPerTable}, \forall i
      \end{equation} \\
          The number of active constituents in each question is lower-bounded. Clearly We need to use the question definition in order to answer a question.  
        & 
      \begin{equation}
          \sum_{l} \xOne{\qCons} \geq \textsc{MinActiveQCons}  
          \label{eq:MinActiveQCons}
      \end{equation} \\
          A cell is active if and only if the sum of coefficients of all external alignment to it is at least a minimum specified value 
        & 
        \begin{multline}
           \label{cons:aCellIsActiveIffTheSumOfExtAlignIsAtLeastSth}
          \sum_{(\tableCell, e) \in \mathcal{E}_{i, j, k}}  \xTwo{\tableCell}{e}  \geq \xOne{\tableCell} \\ 
          \times \textsc{MinActiveCellAggrAlignment}, \forall i,j, k 
		\end{multline} \\
          A title is active if and only if the sum of coefficients of all external alignment to it is at least a minimum specified value
        & 
        \begin{multline}
          \sum_{(\title, e) \in \mathcal{H}_{i, k}} \xTwo{\tableCell}{e} \geq \xOne{\tableCell} \\  \times \textsc{MinActiveTitleAggrAlignment}, \forall i, k
      	\end{multline} \\
         If a column is active, at least one of its cells must be active as well. & 
      \begin{equation}
          \sum_{j} \xOne{\tableCell} \geq \xOne{\columnVar}, \forall i, k
      \end{equation} \\
          At most a certain number of columns can be active for a single option 
 & 
      \begin{multline}
          \sum_{k} \xTwo{\columnVar}{\option}  \leq \textsc{MaxActiveChoiceColumn}, \\ \forall i, m
      \end{multline} \\
          If a column is active for a choice, the table is active too.  & 
      \begin{equation}
          	\xOne{\columnVar} \leq \xOne{\tableVar}, \forall i,k
      \end{equation} \\
          If a table is active for a choice, there must exist an active column for choice. & 
      \begin{equation}
          \xOne{\tableVar} \leq \sum_{k}  \xOne{\columnVar}, \forall i
      \end{equation} \\
          If a table is active for a choice, there must be some non-choice alignment.  & 
      \begin{equation}
          \xTwo{\tableVar}{\option} \leq \sum_{(e, e') \in \mathcal{N}_{i}} \xTwo{e}{e'}, \forall i, m
      \end{equation} \\
                 Answer should be present in at most a certain number of tables & 
      \begin{multline}
          \xTwo{\tableVar}{\option}  \leq \textsc{MaxActiveTableChoiceAlignmets}, \\ \forall i, m
      \end{multline} \\
          If a cell in a column, or its header is aligned with a question option, the column is active for question option as well.  & 
      \begin{multline}
          \xTwo{\tableCell}{\option}  \leq \xTwo{\columnVar}{\option}, \\ \forall i, k, m, \forall (\tableCell, \option) \in \mathcal{M}_{i, k, m}
      \end{multline} \\
          If a column is active for an option, there must exist an alignment to header or cell in the column. 	 & 
      \begin{equation}
          \xTwo{\columnVar}{\option} \leq \sum_{(\tableCell, \option) \in \mathcal{O}_{i, k, m}} \xTwo{\tableCell}{\option}, \forall i, m
      \end{equation} \\
      \hline 
    \end{tabular}
    }
\end{table*}

\begin{table*}
	\small 
	\renewcommand{\arraystretch}{0.0}
    \linespread{\linespacingInTable}\selectfont\centering
    \resizebox{\textwidth}{!}{
	\begin{tabular}{|L{10.0cm}L{8.5cm}|}
      \hline 
           At most a certain number of columns may be active for question option in a table. & 
      \begin{multline}
          \sum_{k} \xTwo{\columnVar}{\option} \leq \\ \textsc{MaxActiveChoiceColumnVars}, \forall i, m
      \end{multline} \\
          If a column is active for a choice, the table is active for an option as well. & 
      \begin{equation}
          \xTwo{\columnVar}{\option} \leq \xTwo{\tableVar}{\option}, \forall i, k, m
      \end{equation} \\
          If the table is active for an option, at least one column is active for a choice & 
      \begin{equation}
      \xTwo{\tableVar}{\option} \leq \sum_{k} \xTwo{\columnVar}{\option}, \forall i, m
      \end{equation} \\
      	 Create an auxiliary variable $\xOne{\text{whichTermIsActive}}$ with objective weight 1.5 and activate it, if there a ``which" term in the question. 
         & 
      \begin{multline}
          \sum_{l} 1\setOf{\qCons = \text{``which''} } \leq \xOne{\text{whichTermIsActive}}
      \end{multline} \\      
      Create an auxiliary variable $\xOne{\text{whichTermIsAligned}}$ with objective weight 1.5. Add a boost if at least one of the table cells/title aligning to the choice happens to have a good alignment ($\setOf{w(., .) > \textsc{MinAlignmentWhichTerm}}$) with the ``which'' terms, i.e. \textsc{WhichTermSpan} constituents after ``which''. 
         & 
      \begin{multline}
          \sum_i \sum_{(e_1, e_2) \in \mathcal{T}_i} \xTwo{e_1}{e_2} 
            \geq  \xOne{\text{whichTermIsAligned}}
      \end{multline} \\
      	 A question constituent may not align to more than a certain number of cells & 
      \begin{equation}
          \sum_{(e, \qCons) \in \mathcal{Q}_l } \xTwo{e}{\qCons} \leq \textsc{MaxAlignmentsPerQCons}
      \end{equation} \\
      	  Disallow aligning a cell to two question constituents if they are too far apart; in other words add the following constraint if the two constituents $\qCons$ and $\qConsPrime$ are more than \textsc{qConsCoalignMaxDist} apart from each other: & 
      \begin{equation}
          \xTwo{\tableCell}{\qCons} + \xTwo{\tableCell}{\qConsPrime} \leq 1,  \forall l, l', i, j, k
      \end{equation} \\
      	 For any two two question constraints that are not more than \textsc{qConsCoalignMaxDist} apart create an auxiliary binary variable $\xOne{\text{cellProximityBoost}}$ and set its weight in the objective function to be $1 / (l - l' + 1)$, where $l$ and $l'$ are the indices of the two question constituents. With this we boost objective score if a cell aligns to two question constituents that are within a few words of each other & 
      \begin{multline}
          \xOne{\text{cellProximityBoost}} \leq \xTwo{\tableCell}{\qCons}, \\ \xOne{\text{cellProximityBoost}} \leq  \xTwo{\tableCell}{\qConsPrime}, \forall i, j, k 
      \end{multline} \\
      If a relation match is active, both the columns for the relation must be active & 
      \begin{equation}
      	 \label{cons:IfRelationIsActiveBothColumnsMustBeActive}
         \xFour{\columnVar}{\columnVarPrime}{\qCons}{\qConsPrime} \leq \xOne{\columnVar}, \xFour{\columnVar}{\columnVarPrime}{\qCons}{\qConsPrime} \leq \xOne{\columnVarPrime}
      \end{equation} \\
       If a column is active, a relation match connecting to the column must be active & 
      \begin{equation}	\label{cons:ifColumnIsActiveARelationConnectingToItMustBeActive}
        \xOne{\columnVar} \leq \sum_{k'} (\xFour{\columnVar}{\columnVarPrime}{\qCons}{\qConsPrime} + \xFour{\columnVarPrime}{\columnVar}{\qCons}{\qConsPrime}), \forall k  
      \end{equation} \\
      	  If a relation match is active, the column cannot align to the question in an invalid position & 
      \begin{multline}
 \label{cons:IfRelationIsActiveTheColumnCannotAlignToQuestionInInvalidPosition}
     	\xFour{\columnVar}{\columnVarPrime}{\qCons}{\qConsPrime} \leq  1 - \xTwo{\tableCell}{\hat{\qCons}}, \\ \textrm{where } \hat{\qCons} \leq \qCons \textrm{ and } \tableCell \in \columnVar
      \end{multline} \\
      If a row is active, at least a certain number of its cells must be active & 
      \begin{equation}
          \sum_{k} \xOne{\tableCell} \geq \textsc{MinActiveCellsPerRow} \times \xOne{\rowVar}, \forall i, j
      \end{equation} \\
      If row is active, it must have non-choice alignments. & 
      \begin{equation}
          \xOne{\rowVar} \leq \sum_{(n, n') \in \mathcal{L}_{ij}} \xTwo{n}{n}
      \end{equation} \\
      If row is active, it must have non-question alignments & 
      \begin{equation}
          \xOne{\rowVar} \leq \sum_{(n, n') \in \mathcal{K}_{ij}} \xTwo{n}{n}
      \end{equation} \\
      If two rows of a table are active, the corresponding active cell variables across the two rows must match; in other words, the two rows must have identical activity signature & 
      \begin{multline}
          \xOne{\rowVar} + \xOne{\rowVarPrime} + \xOne{\tableCell}  \\ - \xOne{\tableCellPrime} \leq 2, \forall i, j, j', k, k' 
      \end{multline} \\
       If two rows are active, then at least one active column in which they differ (in tokenized form) must also be active; otherwise the two rows would be identical in the proof graph. & 
       \begin{equation}
         \sum_{\tableCell \neq \tableCellSameRow} \xOne{\columnVar} - \xOne{\rowVar} - \xOne{\rowVarPrime} \geq -1 
      \end{equation} \\
      If a table is active and another table is also active, at least one inter-table active variable must be active; & 
      \begin{equation}
          \xOne{\tableVar} + \xOne{\tableVarPrime} + \sum_{j, k, j', k'} \xTwo{\tableCell}{\tableCellPrimePrime}  \geq 1, \forall i, i' 
      \end{equation} \\
      \hline 
    \end{tabular}
set    }
    \caption{The set of all constraints used in our ILP formulation. The set of variables and are defined in Table~\ref{table:ilp-variables}. More intuition about constraints is included in Section 3. The sets used in the definition of the constraints are defined in Table~\ref{table:ilp-sets}.}
	\label{table:constraints}
\end{table*}

\makeatletter
\setlength{\@fptop}{0pt}
\makeatother

\clearpage
\newpage 

\section{Supplementary Details for Chapter 8}

We here provide detailed proofs of the formal results, followed by additional experiments. The following observation allows a simplification of the proofs, without loss of any generality.

\begin{remark}
\label{remark:folding}
Since our procedure doesn't treat similarity edges and meaning-to-symbol noise edges differently, we can `fold' $\varepsilon_-$ into $p_-$ and $p_+$ (by increasing edge probabilities). 
More generally, the results are identical whether one uses $p_+, p_-, \varepsilon_-$ or $p'_+, p'_-, \varepsilon'_-$, as long as: 
$$
\begin{cases}
p_+ \oplus \varepsilon_- = p'_+ \oplus  \varepsilon'_- \\ 
p_- \oplus \varepsilon_- = p'_- \oplus \varepsilon'_-
\end{cases}
$$ 
For any $p_+$ and $\varepsilon_-$, we can find a $p’_+$ such that $\varepsilon’_-$ = 0. Thus, w.l.o.g., in the following analysis we derive results only using $p_+$ and $p_-$ (i.e. assume $\varepsilon’_-$ = 0). Note that we  expand these terms to $p_+ \oplus \varepsilon_-$ and $p_- \oplus \varepsilon_-$ respectively in the final results.
\end{remark}

\subsection{Proofs: Possibility of Accurate Connectivity Reasoning}
\label{sec:supp:proofs}
In this section we provide the proofs of the additional lemmas necessary for proving the intermediate results. 
First we introduce a few useful lemmas, and then move on to the proof of Theorem~\ref{the:reasoning:possiblity:detailed}. 

We introduce the following lemmas which will be used in connectivity analysis of the clusters of the nodes $\oracle(m)$.
\begin{lemma}[Connectivity of a random graph~\citep{Gilbert59}]
\label{connectivity:probability}
Let $P_n$ denote the probability of the event that a random undirected graph $\renymodel$ ($p > 0.5$) is connected.
This probability can be lower-bounded as following: 
$$
P_n \geq 1 - \left[ q^{n-1} \left\lbrace (1 + q^{(n-2)/2} )^{n-1} - q^{(n-2)(n-1)/2} \right\rbrace + q^{n/2}\left\lbrace (1 + q^{(n-2)/2})^{n-1} - 1\right\rbrace \right],
$$
where $q = 1-p$. 
\end{lemma}

See \citet{Gilbert59} for a proof of this lemma. Since $q \in (0, 1)$, this implies that $P_n  \rightarrow 1$ as $n$ increases. The following lemma provides a simpler version of the above probability: 
\begin{corollary}[Connectivity of a random graph~\citep{Gilbert59}]
\label{cor:connectivity:likelihood}
The random-graph connectivity probability $P_n$ (Lemma~\ref{connectivity:probability}) can be lower-bounded as following: 
$$
P_n \geq 1 - 2e^3q^{n/2}
$$
\end{corollary}
\begin{proof}
We use the following inequality: 
$$ (1 + \frac{3}{n})^n \leq e^3 $$

Given that $q \leq 0.5, n\geq 1$, one can verify that $q^{(n-2)/2} \leq 3/n$. Combining this with the above inequality gives us,  
$(1 + q^{{n-2}/2})^{n-1} \leq e^3 $. 

With this, we bound the two terms within the two terms of the target inequality: 
$$
\begin{cases}
(1 + q^{(n-2)/2} )^{n-1} - q^{(n-2)(n-1)/2} \leq e^3  \\ 
(1 + q^{(n-2)/2})^{n-1} - 1 \leq e^3 
\end{cases}
$$

$$
\left[ q^{n-1} \left\lbrace (1 + q^{(n-2)/2} )^{n-1} - q^{(n-2)(n-1)/2} \right\rbrace + q^{n/2}\left\lbrace (1 + q^{(n-2)/2})^{n-1} - 1\right\rbrace \right] \leq  e^3 q^{n-1}  + e^3 q^{n/2} \leq  2e^3 q^{n/2}
$$
which concludes the proof. 

\end{proof}

We show a lower-bound on the probability of $s$ and $s'$ being connected given the connectivity of their counterpart nodes in the meaning graph. This lemma will be used in the proof of Theorem~\ref{the:reasoning:possiblity}:

\begin{lemma}[Lower bound]
\label{lem:LBthm1}
$\prob{  s \pathk{\tilde{d}} s' | m \pathk{d} m' } \geq \left(1 - 2 e^3 \varepsilon_+ ^{\lambda/2}\right)^{d+1} \cdot  \left(1 - (1-p_+)^{\lambda^2}\right)^{d}$.
\end{lemma}
\begin{proof}
We know that $m$ and $m'$ are connected through some intermediate nodes $m_1, m_2, \cdots, m_\ell$ ($\ell < d$). We show a lower-bound on having a path in the symbol-graph between $s$ and $s'$, through clusters of nodes $\oracle(m_1), \oracle(m_2), \cdots, \oracle(m_\ell)$. We decompose this into two events: 
$$
\begin{cases}
e_1[v] & \text{For a given meaning node } v \text{ its cluster in the symbol-graph, } \oracle(v) \text{ is connected. }\\
e_2[v, u] & \text{For any two connected nodes } (u, v) \text{ in the meaning graph, there is at least an edge} \\
 & \text{connecting their clusters } \oracle(u), \oracle(v) \text{ in the symbol-graph.}
\end{cases}
$$
The desired probability can then be refactored as: 
\begin{align*}
\prob{  s \pathk{\tilde{d}} s' | m \pathk{d} m' }
  & \geq   \prob{  \left(\bigcap_{v\in\{s,m_1,\dots,m_\ell,s'\}} e_1[v]\right)\cap \left(\bigcap_{(v,u)\in\{(s,m_1),\dots,(m_\ell,s')\}} e_2[v,u]\right)  } \\
  & \geq \prob{e_1}^{d+1} \cdot \prob{e_2}^d. 
\end{align*}

We split the two probabilities and identify lower bounds for each. Based on Corollary~\ref{cor:connectivity:likelihood},
$\prob{e_1} \geq 1 - 2e^3\varepsilon_+ ^{\lambda/2}$, and as a result $\prob{e_1}^{d+1} \geq  \left(1 - 2e^3 \varepsilon_+ ^{\lambda/2}\right)^{d+1}$. The probability of connectivity between pair of clusters is 
$\prob{e_2}  = 1 - (1-p_+)^{\lambda^2}$. Thus, similarly,  $\prob{e_2}^{d} \geq  \left(1 - (1-p_+)^{\lambda^2}\right)^{d}$. Combining these two, we obtain:
\begin{equation}
\label{eqn:lower-bound}
    \prob{  s \pathk{\tilde{d}} s' | m \pathk{d} m' } \geq \left(1 - 2 e^3 \varepsilon_+ ^{\lambda/2}\right)^{d+1} \cdot  \left(1 - (1-p_+)^{\lambda^2}\right)^{d}
\end{equation}
\end{proof}

The connectivity analysis of $G_S$ can be challenging since the graph is a non-homogeneous combination of positive and negative edges. 
For the sake of simplifying the probabilistic arguments, given symbol graph $G_S$, we introduce a non-unique simple graph $\tilde{G}_S$ as follows.
\begin{definition}
Consider a special partitioning of $V_G$ such that the $d$-neighbourhoods of $s$ and $s'$ form two of the partitions and the rest of the nodes are arbitrarily partitioned in a way that the diameter of each component does not exceed $\tilde{d}$.
\begin{itemize}
    \item The set of nodes $V_{\tilde{G}_S}$ of $\tilde{G}_S$ corresponds to the aforementioned partitions.
    
    \item There is an edge $(u,v)\in E_{\tilde{G}_S}$ if and only if at least one node-pair from the partitions of $V_G$ corresponding to $u$ and $v$, respectively, is connected in $E_{G_S}$.
\end{itemize}
\end{definition}

In the following lemma we give an upper-bound on the connectivity of neighboring nodes in $\tilde{G}_S$: 
\begin{lemma}
\label{lemma:graph:homogination2}
When $G_S$ is drawn at random, the probability that an edge connects two arbitrary nodes in $\tilde{G}_S$ is at most $(\lambda \ball(d))^2 p_-$.
\end{lemma}
\begin{proof}
Recall that a pair of nodes from $\tilde{G}_S$, say \((u,v)\), are connected when at least one pair of nodes from corresponding partitions in $G_S$ are connected. Each $d$-neighbourhood in the meaning graph has at most \(\ball(d)\) nodes. It implies that each partition in  $\tilde{G}_S$ has at most \(\lambda \ball(d)\) nodes. Therefore, between each pair of partitions, there are at most $(\lambda \ball(d))^2$ possible edges. By union bound, the probability of at least one edge being present between two partitions is at most $(\lambda \ball(d))^2 p_-$.
\end{proof}

Let $v_s, v_{s'}\in V_{\tilde{G}_S}$ be the nodes corresponding to the components containing $s$ and $s'$ respectively. The following lemma establishes a relation between connectivity of $s,s'\in V_{G_S}$ and the connectivity of $v_s, v_{s'}\in V_{\tilde{G}_S}$: 

\begin{lemma}
\label{lemma:graph:homogination}
\( \prob{ s \pathk{\tilde{d}} s' | m \npathk{} m' }\leq \prob{\text{There is a path from $v_s$ to $v_{s'}$ in $\tilde{G}_S$ with length } \tilde{d} }\). 
\end{lemma}

\begin{proof}
Let $L$ and  \(R\) be the events in the left hand side and right hand side respectively.  
Also for a permutation of nodes in \(G_S\), say \(p\), let \(F_p\) denote the event that all the edges of \(p\) are present, i.e., 
\(L=\cup F_p\).  
Similarly, for a permutation of nodes in \(\tilde{G}_S\), say \(q\), let \(H_q\) denote the event that all the edges of \(q\) are present.
Notice that \(F_p\subseteq H_q\) for \(q\subseteq p\), because if all the edges of \(p\) are present the edges of \(q\) will be present. Thus,
\[L=\bigcup_p F_p\subseteq \bigcup_p H_{p\cap E_{\tilde{G}_S}} =\bigcup_q H_q = R.\]
This implies that \(\prob{L}\leq \prob{R}\).
\end{proof}

{
\begin{lemma}[Upper bound]
\label{lem:UBthm1}
    If $(\lambda\ball(d))^2 p_- \leq \frac{1}{2 e  n}$, then 
    $\prob{ s \pathk{\leq \tilde{d}} s' \mid m \npathk{} m' }
      \leq 2en (\lambda\ball(d))^2 p_-.$
\end{lemma}

\begin{proof}

To identify the upper bound on $\prob{ s \pathk{\leq \tilde{d}} s' | m \npathk{} m' } $, recall the definition of \(\tilde{G}_S\), given an instance of $G_S$ 
(as outlined in Lemmas~\ref{lemma:graph:homogination2} and~\ref{lemma:graph:homogination}, for 
$\tilde{p} = (\lambda \ball(d))^2 p_-$).
Lemma~\ref{lemma:graph:homogination} relates the connectivity of $s$ and $s'$ to a connectivity event in \(\tilde{G}_S\), i.e., \( \prob{ s \pathk{\leq \tilde{d}} s' \mid m \npathk{} m' } \leq \prob{\text{there is a path from $v_s$ to $v_{s'}$ in $\tilde{G}_S$ with length } \tilde{d} }\), where $v_s, v_{s'}\in V_{\tilde{G}_S}$ are the nodes corresponding to the components containing $s$ and $s'$ respectively. Equivalently, in the following, we prove that the event $\text{dist}(v_s,v_{s'})\leq \tilde{d}$ happens with a small probability:

\[
\prob{ s \pathk{\leq \tilde{d}} s' } = 
\prob{ \bigvee_{\ell=1, \cdots, \tilde{d}} s \pathk{\ell} s' }
\leq \sum_{\ell\leq \tilde{d}} { n \choose \ell} \tilde{p}^\ell
\leq \sum_{\ell\leq \tilde{d}} (\frac{e n}{\ell})^\ell \tilde{p}^\ell
\]

\[
\leq \sum_{\ell\leq \tilde{d}} ({e n})^\ell \tilde{p}^\ell
\leq 
{e n\tilde{p}}\frac{
        ({e n\tilde{p}})^{\tilde{d}}-1
    }{
        {e n\tilde{p}}-1
    }
\leq \frac{e n\tilde{p}}{1-e n\tilde{p}}    
\leq 2 e n\tilde{p}. 
\]
where the final inequality uses the assumption that $\tilde{p}\leq \frac{1}{2 e n}$.

\end{proof}
}

Armed with the bounds in Lemmas~\ref{lem:LBthm1} and~\ref{lem:UBthm1}, we are ready to provide the main proof:
\begin{proof}[Proof of 
Theorem~\ref{the:reasoning:possiblity}]
Recall that the algorithm checks for connectivity between two given nodes $s$ and $s'$, i.e., 
$s \pathk{\leq \tilde{d}} s'$. With this observation, we aim to infer whether the two nodes in the meaning graph are connected ($m \pathk{\leq d} m'$) or not ($ m \npathk{} m'$). 
We prove the theorem by using lower and upper bound for these two probabilities, respectively:
{ 
\begin{align*}
\gamma 
  & = \prob{  s \pathk{\leq \tilde{d}} s' | m \pathk{d} m' } - \prob{ s \pathk{\leq \tilde{d}} s' | m \npathk{} m' } \\
  & \geq LB\left(\prob{  s \pathk{\leq \tilde{d}} s' | m \pathk{d} m' }\right) - UB\left(\prob{ s \pathk{\leq \tilde{d}} s' | m \npathk{} m' }\right) \\ 
  & \geq \left(1 - 2 e^3 \varepsilon_+ ^{\lambda/2}\right)^{d+1} \cdot  \left(1 - (1-p_+)^{\lambda^2}\right)^{d}- 2en (\lambda\ball(d))^2 p_-. 
\end{align*}
%
where the last two terms of the above inequality are based on the results of Lemmas~\ref{lem:LBthm1} and~\ref{lem:UBthm1}, with the assumption for the latter that $(\lambda\ball(d))^2 p_- \leq \frac{1}{2 e  n}$. To write this result in its general form we have to replace $p_+$ and $p_-$, with $p_+ \oplus \varepsilon_-$ and  $p_- \oplus \varepsilon_-$, respective (see Remark~\ref{remark:folding}). 

}

\end{proof}

\subsection{Proofs: Limitations of Connectivity Reasoning}
We provide the necessary lemmas and intuitions before proving the main theorem. 

A random graph is an instance sampled from a distribution over graphs. 
In the $\renymodel$ Erd\H{o}s-Renyi model, a graph is constructed in the following way:  
Each edge is included in the graph with probability $p$, independent of other edges. 
In such graphs, on average, the length of the path connecting any node-pair is short (logarithmic in the number of nodes). 
\begin{lemma}[Diameter of a random graph, Corollary 1 of \citep{ChungLu02}]\label{lemma:diameter}
If $n\cdot p = c > 1$ for some constant $c$, then almost-surely the diameter of $\renymodel$ is $\Theta(\log n)$. 
\end{lemma}

We use the above lemma to prove Theorem~\ref{th:multi-hop:impossiblity}. 
Note that the overall noise probably (i.e., $p$ in Lemma~\ref{lemma:diameter}) in our framework is $p_- \oplus \varepsilon_- $.  

\begin{proof}[Proof of Theorem~\ref{th:multi-hop:impossiblity}]
Note that the $|V_{G_S}|=\lambda\cdot n$.
By Lemma \ref{lemma:diameter}, the symbol graph has diameter $\Theta(\log \lambda n)$.
This means that for any pair of nodes $s,s'\in V_{G_S}$, we have $s\pathk{\Theta(\log \lambda n)}s'$.
Since $\tilde{d}\geq\lambda d\in \Omega (\log \lambda n) $, the multi-hop reasoning algorithm finds a path between $s$ and $s'$ in symbol graph and returns $\mathsf{connected}$ regardless of the connectivity of $m$ and $m'$.
\end{proof}

\subsection{Proofs: Limitations of General Reasoning}

The proof of the theorem follows after introducing necessary lemmas.

In the following lemma, we show that the spectral differences between the two symbol graphs in the locality of the target nodes are small. 
For ease of exposition, we define an intermediate notation, for a normalized version of the Laplacians: $\tilde{L} = L / \| L\|_2$ and $\tilde{L}' = L' / \| L'\|_2$.

\begin{lemma}
\label{lemma:cut-norm}
The norm-2 of the Laplacian matrix corresponding to the nodes participating in a cut, can be upper-bounded by the number of the edges participating in the cut (with a constant factor). 
\end{lemma}

\begin{proof}[Proof of Lemma~\ref{lemma:cut-norm}]
Using the definition of the Laplacian: 
$$
\|L_C\|_2 \leq \| A - D \|_2 \leq   \| A \|_2 + \| D \|_2
$$
where $A$ is the adjacency matrix and $D$ is a diagonal matrix with degrees on the diagonal. We bound the norms of the matrices based on size of the cut (i.e., number of the edges in the cut). For the adjacency matrix we use the Frobenius norm:  
$$
\| A \|_2 \leq  \| A \|_F = \sqrt{\sum_{ij} a_{ij}} = 2 \cdot |C|
$$
where $|C|$ denotes the number of edges in $C$.
To bound the matrix of degrees, we use the fact that norm-2 is equivalent to the biggest eigenvalue, which is the biggest diagonal element in a diagonal matrix: 
$$
\| D \|_2 = \sigma_{\max}(D) = \max_i deg(i) \leq |C|
$$
With this we have shown that: $\|L_C\|_2 \leq 3|C|$. 
\end{proof}

For sufficiently large values of $p$, $\renymodel$ is a connected graph, with a high probability. More formally: 
\begin{lemma}[Connectivity of random graphs]
\label{lemma:er:connectedness}
In a random graph $\renymodel$, for any $p$ bigger than ${{\tfrac {(1+\varepsilon )\ln n}{n}}}$, the graph will almost surely be connected.
\end{lemma}
The proof can be found in \citep{ErdosR60}. 

\begin{lemma}[Norm of the adjacency matrix in a random graph]
\label{lemma:adj:norm}
For a random graph $\renymodel$, let $L$ be the adjacency matrix of the graph. 
For any $\varepsilon > 0$: 
$$
\lim_{n \rightarrow +\infty} \mathbb{P} \left( \left| \|L\|_2   - \sqrt{2n\log n} \right| > \varepsilon \right)  \rightarrow 0
$$
\end{lemma}
\begin{proof}[Proof of Lemma~\ref{lemma:adj:norm}]
From Theorem 1 of~\citep{DingJiot10} we know that: 
$$
\frac{ \sigma_{\max}{(L)} }{ \sqrt{n \log n}}  \overset{P}{\rightarrow} \sqrt{2}
$$
where $\overset{P}{\rightarrow}$ denote \emph{convergence in probability}. 
And also notice that norm-2 of a matrix is basically the size of its biggest eigenvalue, which concludes our proof. 
\end{proof}

\begin{lemma}
\label{lemma:impossiblity:general:reasoning}
For any pair of meaning-graphs $G$ and $G'$ constructed according to Definition~\ref{cut:construction}, and,  
\begin{itemize}
    \item $d > \log n$, 
    \item $p_- \oplus \varepsilon_- \geq c \log n \big/ n$ for some constant $c$, 
    \item $\tilde{d}\geq \lambda d$, 
\end{itemize}
with $L$ and $L'$ being the Laplacian matrices corresponding to the $\tilde{d}$-neighborhoods of the corresponding nodes in the surface-graph; we have: 
$$
\frac{\|L-L'\|_2}{\|L\|_2} \leq \frac{ \sqrt{\lambda} \ball(1)}{ \sqrt{2n \log (n \lambda)} }, 
$$
with a high-probability.
\end{lemma}

\begin{proof}[Proof of Lemma~\ref{lemma:impossiblity:general:reasoning}]
In order to simplify the exposition, w.l.o.g. assume that $\varepsilon_- =0$ (see Remark~\ref{remark:folding}). 
Our goal is to find an upper-bound to the fraction $\frac{\|L-L'\|_2}{\|L\|_2}$. 
Note that the Laplacians contain only the local information, i.e., $\tilde{d}-$neighborhood. 
First we prove an upper bound on the nominator. 
By eliminating an edge in a meaning-graph, the probability of edge appearance in the symbol graph changes from $p_+$ to $p_-$. 
The effective result of removing edges in $C$ would appear as i.i.d. $\bern{p_+-p_-}$. 
Since by definition, $\ball(1)$ is an upper bound on the degree of meaning nodes, the size of minimum cut should also be upper bounded by \ball(1).
Therefore, the maximum size of the min-cut $C$ separating two nodes $m \pathk{d} m'$ is at most $\ball(1)$. 
To account for vertex replication in symbol-graph, the effect of cut would appear on at most $\lambda \ball(1)$ edges in the symbol graph. Therefore, we have$\|L-L'\|_2\leq \lambda \ball(1)$ using Lemma~\ref{lemma:cut-norm}.


As for the denominator, the size of the matrix $L$ is the same as the size of $\tilde{d}$-neighborhood in the symbol graph. 
We show that if $\tilde{d} > \log (\lambda n)$  the neighborhood almost-surely covers the whole graph. 
While the growth in the size of the $\tilde{d}$-neighborhood is a function of both $p_+$ and $p_-$, to keep the analysis simple, we underestimate the neighborhood size by replacing $p_+$ with $p_-$, i.e., the size of the $\tilde{d}$-neighborhood is lower-bounded by the size of a $\tilde{d}$-neighborhood in $\mathsf{G}(\lambda \cdot n, p_-)$. 

By Lemma~\ref{lemma:er:connectedness} the diameters of the symbol-graphs $G_S$ and $G_S'$ are both $\Theta(\log (\lambda n))$.  
Since $\tilde{d} \in \Omega(\log (\lambda n))$, $\tilde{d}$-neighborhood covers the whole graph for both $G_S$ and $G_S'$.

Next, we use Lemma~\ref{lemma:adj:norm} to state that $\|L\|_2$ converges to $\sqrt{2 \lambda n \log (\lambda n)}$, in probability. 

Combining numerator and denominator, we conclude that the fraction, for sufficiently large $n$, is upper-bounded by: 
$\frac{
\lambda\ball(1)
}{
\sqrt{2 \lambda n \log (\lambda n)}
},$
which can get arbitrarily small, for a big-enough choice of $n$.





\end{proof}

\begin{proof}[Proof of Lemma~\ref{lem:normLaplacianClose}]
We start by proving an upper bound on $\tilde{L}-\tilde{L'}$ in matrix inequality notation. Similar upper-bound holds for  $\tilde{L'}-\tilde{L}$ which concludes the theorem.
\begin{align*}
\tilde{L}-\tilde{L'} &= \frac{L}{\norm{{L}}}-\frac{L'}{\norm{L'}} \\ 
& \preceq \frac{L}{\norm{L}}-\frac{L'}{\norm{L-L'}+\norm{L}} \\ 
& =\frac{L\cdot  \norm{L-L'}}{\norm{L}^2}+\frac{L-L'}{\norm{L}}\\ 
& \preceq \frac{ \sqrt{\lambda} \ball(1)}{ \sqrt{2n \log (n \lambda)} }I+\frac{ \sqrt{\lambda} \ball(1)}{ \sqrt{2n \log (n \lambda)} }I.
\end{align*}
The last inequality is due to Lemma \ref{lemma:impossiblity:general:reasoning}.
By symmetry the same upper-bound holds for \(\tilde{L}'-\tilde{L}\preceq
2\frac{ \sqrt{\lambda} \ball(1)}{ \sqrt{2n \log (n \lambda)} }I\). This means that \(\norm{\tilde{L}-\tilde{L}'}\leq \frac{ 2\sqrt{\lambda} \ball(1)}{ \sqrt{2n \log (n \lambda)} }\). 
\end{proof}

\begin{lemma}
\label{lemma:compositon}
Suppose $f$ is an indicator function on an open set\footnote{\url{https://en.wikipedia.org/wiki/Indicator_function}}, 
it is always possible to write it as composition of two functions:
\begin{itemize}
   \item A continuous and Lipschitz function: $g: \mathbb{R}^{d} \rightarrow (0, 1), $
   \item A thresholding function: $H(x)= \mathbf{1} \{ x > 0.5 \}.$
\end{itemize}
such that: $ \forall x \in \mathbb{R}^d  :  \; \; \;  f(x) = h(g(x))$. \end{lemma}
\begin{proof}[Proof of Lemma~\ref{lemma:compositon}]
Without loss of generality, we assume that the threshold function is defined as $H(x) = \mathbf{1} \{ x > 0.5 \}$. One can verify that a similar proof follows for $H(x) = \mathbf{1} \{ x \geq 0.5 \}$. 
We use notation $f^{-1}(A)$ the set of pre-images of a function $f$, for the set of outputs $A$. 

First let's study the collection of inputs that result in output of $1$ in $f$ function. 
Since $f=h\circ g$, then $f^{-1}(\{1\})=g^{-1}(h^{-1}(\{1\}))=g^{-1}((0.5,1))$ and 
$f^{-1}(\{0\})=g^{-1}(h^{-1}(\{0\}))=g^{-1}((0,0.5))$.
Define $C_0$ and $C_1$, such that $C_i \triangleq f^{-1}(\{i\})$; note that  
since $g$ is continuous and $(0.5,1)$ is open $C_1$ is an open set (hence $C_1$ is closed). 
Let
$d:\mathbb R^n\to \mathbb R$ be defined by, 
$$
d(x) \triangleq \text{dist}(x,C_0)=\inf_{c\in C_0}\|x-c\|. 
$$
Since $C_0$ is closed, it follows $d(x)=0$ if and only if $x\in C_0$. Therefore, letting $$g(x)=\frac12 + \frac12\cdot \frac{d(x)}{1+d(x)},$$
then $g(x)=\frac12$ when $x\in C_0$, while $g(x)>\frac12$ when $x\not\in C_0$. This means that letting $h(x)=1$ when $x>\frac12$ and $h(x)=0$ when $x\le \frac12$, then $f=h\circ g$.
One can also verify that this construction is $1/2$-Lipschitz; this follows because $d(x)$ is $1$-Lipschitz, which can be proved using the triangle inequality

Hence the necessary condition to have such decomposition is $f^{-1}(\{1\})$ and $f^{-1}(\{0\})$ be open or closed. 
\end{proof}

\begin{proof}[Proof of Lemma~\ref{lemma:improbable:recovery}]

Note that $f$ maps a high dimensional continuous space to a discrete space. To simplify the argument about $f$,
we decompose it to two functions: a continuous function $g$ mapping matrices to $(0,1)$ and a threshold function $H$ (e.g. $0.5 + 0.5 \text{sgn}(.)$) which maps to one if $g$ is higher than a threshold and to zero otherwise. Without loss of generality we also normalize $g$ such that the gradient is less than one. Formally, 
$$
f = H \circ g, \text{ where } g: \mathbb{R}^{\dimm \times \dimm } \rightarrow (0, 1), \norm{ \nabla g \Bigr|_{\tilde{L}}}\leq 1.
$$
Lemma~\ref{lemma:compositon} gives a proof of existence for such decompositon, which depends on having open or closed pre-images. 

One can find a differentiable and Lipschitz function $g$ such that 
it intersects with the threshold specified by $H$, in the borders where $f$ changes values.  

With $g$ being Lipschitz, one can upper-bound the variations on the continuous function: 
$$
\norm{g(\tilde{L}) - g(\tilde{L'})} \leq M \norm{\tilde{L} - \tilde{L}'}. 
$$
According to Lemma~\ref{lem:normLaplacianClose}, $\norm{\tilde{L} - \tilde{L}'}$ is upper-bounded by a decreasing function in $n$. 

For uniform choices $(G,G',m,m') \sim \mathcal{G}$ the Laplacian pairs $(\tilde{L}, \tilde{L}')$ are randomly distributed in a high-dimensional space, and for big enough $n$, there are enough portion of the $(\tilde{L}, \tilde{L}')$  (to satisfy $1-\beta$ probability) that appear in the same side of the hyper-plane corresponding to the threshold function (i.e. $f(\tilde{L}) = f(\tilde{L}')$). 
\end{proof}


\subsection{Further experiments}
\begin{figure*}
    \centering
    \includegraphics[scale=0.35]{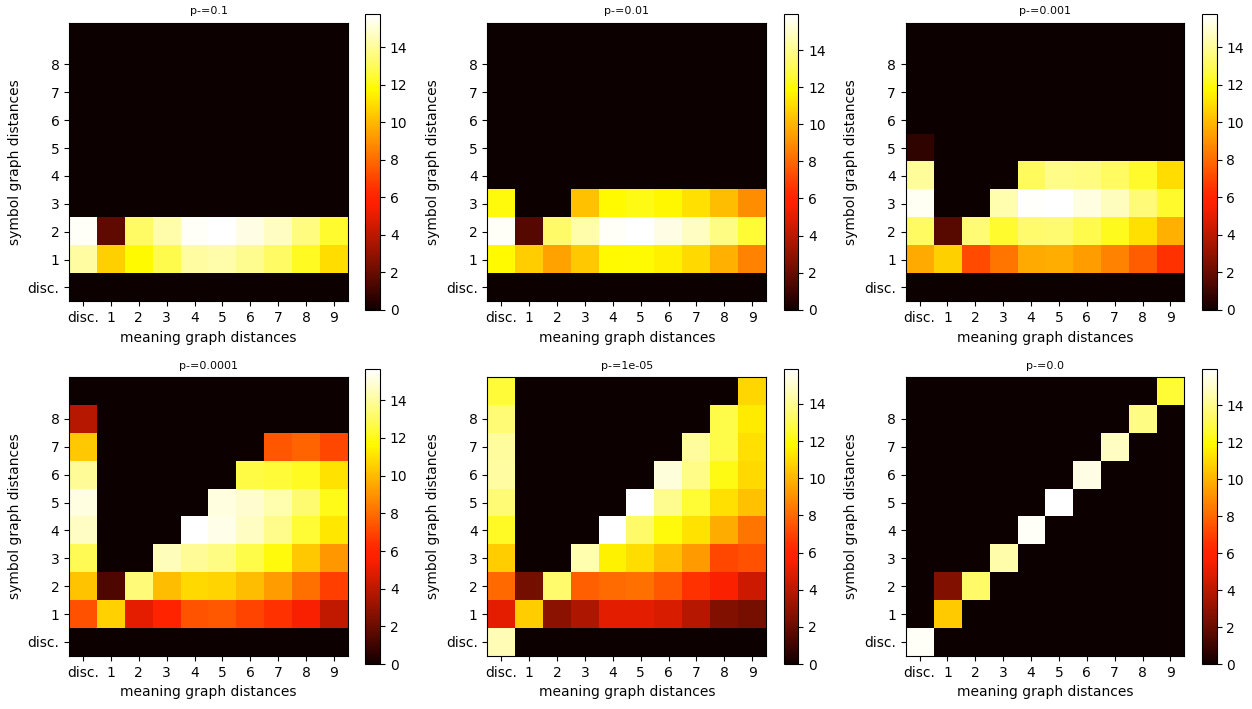}
    \caption{With varied values for $p_-$ a heat map representation of the distribution of the average distances of node-pairs in symbol graph based on the distances of their corresponding meaning nodes is presented. }
    \label{fig:exp:d:vs:d}
\end{figure*}
To evaluate the impact of the other noise parameters in the sampling process, we compare the average distances between nodes in the symbol graph for a given distance between the meaning graph nodes. In the Figure~\ref{fig:exp:d:vs:d}, we plot these graphs for decreasing values of $p_-$ (from top left to bottom right). With high $p_-$ (top left subplot), nodes in the symbol graph at distances lower than two, regardless of the distance of their corresponding node-pair in the meaning graph. As a result, any reasoning algorithm that relies on connectivity can not distinguish symbolic nodes that are connected in the meaning space from those that are not. As the $p_-$ is set to lower values (i.e. noise reduces), the distribution of distances get wider, and correlation of distance between the two graphs increases. In the bottom middle subplot, when $p_-$ has a very low value, we observe a significant correlation that can be reliably utilized by a reasoning algorithm.

\addtocontents{toc}{\protect\setcounter{tocdepth}{1}} 
\end{appendixf}

\singlespacing
\newenvironment{bibliof}{}{}
\titleformat{\chapter}[hang]{\large\center}{\thechapter}{0 pt}{}
\titlespacing*{\chapter}{0pt}{-25 pt}{6 pt} 
\begin{bibliof}

\renewcommand\bibname{BIBLIOGRAPHY}
\addcontentsline{toc}{chapter}{BIBLIOGRAPHY}
\bibliographystyle{abbrvnat}
\bibliography{ccg} 

\end{bibliof}

\end{document}